%% file: main.tex
\documentclass[10pt,twocolumn,letterpaper]{article}
\usepackage{float}
\usepackage{multirow}
\usepackage{tabularx}
\usepackage[pagenumbers]{cvpr}     % To produce the REVIEW version
\input{preamble}
\definecolor{cvprblue}{rgb}{0.21,0.49,0.74}
\usepackage[pagebackref,breaklinks,colorlinks,allcolors=cvprblue]{hyperref}
\def\confName{CVPR}
\def\confYear{2027}

\title{Video Harness: Evidence-Constrained Video Editing with Agentic Search}

\input{vh-authors}
\input{cuhk-title}
\AddToHookNext{cmd/@maketitle/after}{\par\vspace{8pt}}

\renewenvironment{abstract}
  {\par\centerline{\large\bfseries Abstract}\nobreak\vskip 8pt
   \fontsize{10}{11.2}\selectfont\noindent\itshape\ignorespaces}
  {\par}
\begin{document}
\input{fig/teaser}
\input{sec/0_abstract}    
\input{sec/1_intro}
\input{sec/2_related_work}

\input{sec/2_Method}

\input{sec/2_Experiment}

\input{sec/3_Conclusion}
\clearpage

\input{sec/llm_statement}

{
    \small
    \bibliographystyle{ieeenat_fullname}
    \bibliography{main}
}
\clearpage
\input{sec/fulu}

\clearpage
\input{sec/X_suppl}   

\end{document}

%% file: preamble.tex
\usepackage{cuted} %< for strip environment

\usepackage{currfile} %< for \currfilebase macro

\usepackage{caption} %< for "\captionof" commands (teaser)

%% file: vh-authors.tex
\author{%
  \begin{tabular}{@{}c@{\hspace{2em}}c@{}}
    Zhiteng Zheng & Xiangyu Yue\thanks{Corresponding author.} \\[-2pt] 
    {\small\texttt{eecs@link.cuhk.edu.hk}} &
    {\small\texttt{xyyue@ie.cuhk.edu.hk}} \\[1pt]
    \multicolumn{2}{c}{\small MMLab, CUHK}
  \end{tabular}%
}

%% file: fig/teaser.tex
\begin{strip}
  \centering
  \includegraphics[
  width=\textwidth,
  trim={0mm 13mm 0mm 18mm},
  clip
]{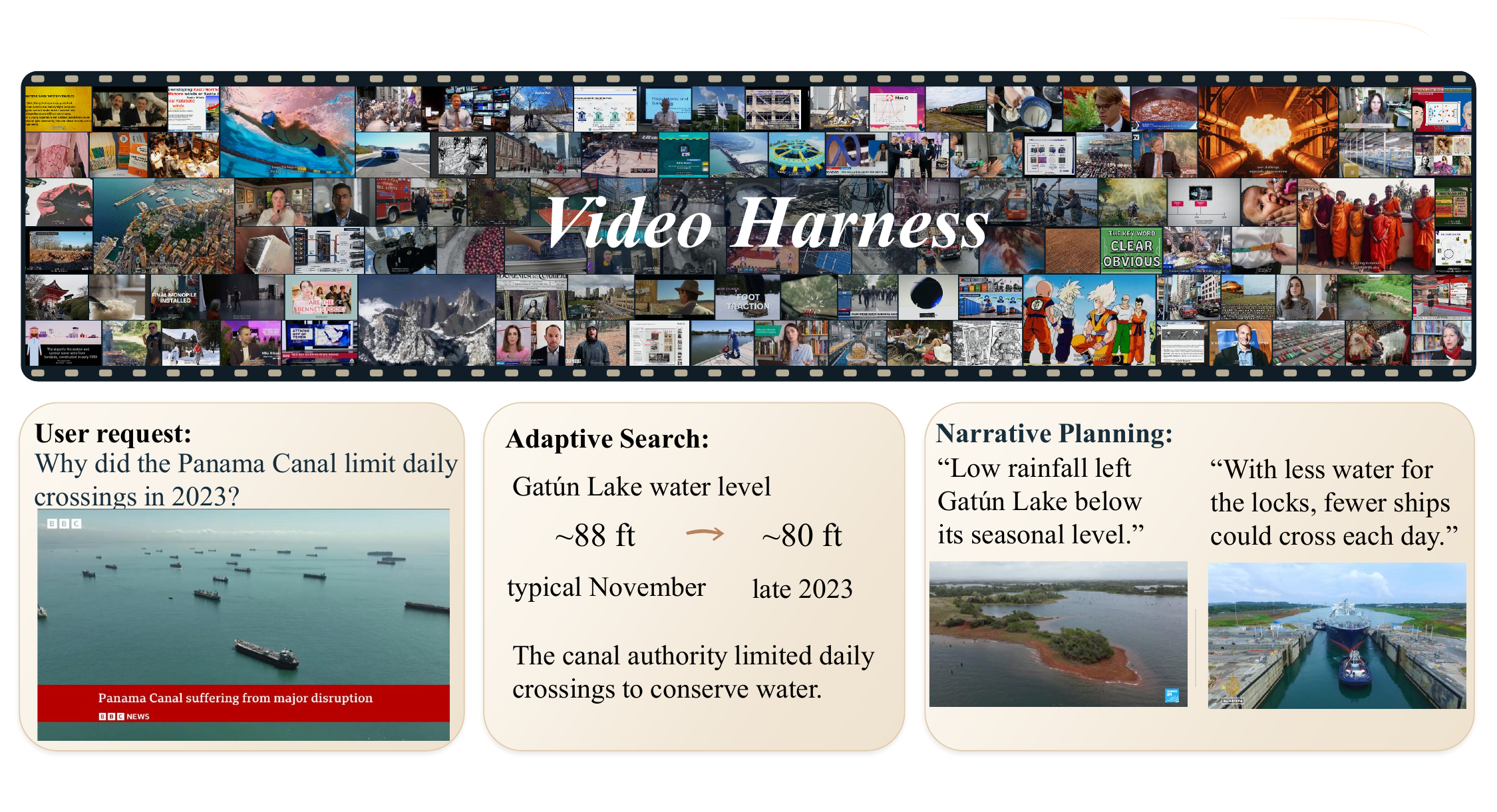}
  \captionsetup{hypcap=false}
  \captionof{figure}{\textbf{VideoHarness.}
    In the Panama Canal example, VideoHarness connects footage of waiting
    ships to water-level records and uses views of the shoreline and locks
    to explain why low rainfall led to fewer daily crossings.}
  \label{fig:videoharness-teaser}
\end{strip}

%% file: sec/0_abstract.tex
\begin{abstract}
An informative video needs both a well-informed narrative and footage that can carry it. We propose Video Harness, an agent framework that combines adaptive agentic search with narration-first planning. Its central idea is to establish what the video should communicate before selecting what it should show. Multi-step search supplies source-grounded knowledge when research offers a clear benefit, and the complete narration script \(N\) is written before footage acquisition. The script becomes a blueprint for global shot assignment, where visual support, measured narration timing, and footage-reuse constraints jointly determine the edit. Capacity and conflict diagnostics guide targeted footage retrieval
and targeted script revision to satisfy narrative and timing requirements. Evaluated against four systems on shared prompts, Video Harness achieves the highest overall automated score. Across prompts spanning five content-oriented prompt types, Video Harness outperforms Crayotter, the strongest baseline, by 18.24 points in mean prompt fulfillment. In the paired search-on/off comparisons, enabling search improves the mean score by 7.62 points, with a 29.05-point gain on news prompts.
\end{abstract}

%% file: sec/1_intro.tex
\section{Introduction}
\label{sec:intro}

Given a short natural-language request, the goal of explanatory video
editing is to combine real-world footage and narration into a coherent
account of a topic. This pairing can help viewers understand complex
processes. In a manufacturing video, for example, viewers can follow
a part through successive stages while the narration explains why one
stage must precede the next. The challenge is that a brief request
often specifies only the topic, and even footage that clearly shows
each step may leave the relationships between steps unexplained.

Language-driven video editing connects editorial intent to specific
choices about footage. B-Script uses a transcript to recommend where
to insert B-roll and what to search for~\cite{huber2019b},
while LAVE translates a user's editing objectives into actions on
their footage~\cite{10.1145/3640543.3645143}. Recent agents extend
this control to shot retrieval, sequence planning, and timeline
execution~\cite{zhou2026videoagent,li2026direct}.
For narrated videos, the relationship between script and footage
can run in either direction. VideoNarrator writes narration to fit
recorded scenes~\cite{yang2024synchronized}, and VideoNarrationAgent
retrieves background knowledge to explain what those scenes
show~\cite{wang2026howtonarrate}. REGen starts from a script and retrieves interview excerpts from
a source documentary to fill quotation slots in a
teaser~\cite{xu2026regen}. For open-ended editing requests,
Crayotter retrieves footage, checks its coverage of the brief,
and builds an editing plan from the resulting material
pool~\cite{yan2026crayotter}. An open-ended explanatory request
also leaves the factual content to the editor: which facts should
the narration convey, and what footage can support each one?

External research can help determine what an informative video should
say. ReAct interleaves reasoning with actions that gather
information~\cite{yao2022react}, and STORM uses questions from
multiple perspectives to research a topic and organize its findings
into an outline~\cite{shao2024assisting}. Visual generation
systems use retrieval in different ways: Gen-Searcher gathers facts
and reference images for image synthesis~\cite{feng2026gen}, while
FACT-V incorporates retrieved knowledge into text-to-video
prompts~\cite{lee2025rewind}. PresentAgent-2 carries research into
a complete video, combining retrieved text, images, and video with
narrated slides~\cite{wu2026presentagent}. KIVI-Bench highlights
why the visual choices matter: even when asked to explain a procedure,
video generators struggle to depict its operations
accurately~\cite{wang2026knowledge}. For editing with real footage, this
raises a concrete question: which source interval shows the action
or entity described in each passage of narration?

Connecting external research to video editing poses three challenges.
First, research needs vary across requests. A visual montage may
require little external information, whereas a report on recent
events depends on up-to-date sources.
Second, assessing factual support and selecting visual evidence
require different judgments. A report may establish that an event
occurred, while footage of its location may provide only context.
Third, shot selection must account for the timing of the full
narration~\cite{yang2024synchronized,li2026direct,yan2026crayotter}.
Two passages may compete for the same source interval, and the
usable footage for a passage may be too short to cover its narration.

We introduce \textbf{Video Harness}, an agent framework that integrates
\textbf{adaptive agentic search} with evidence-constrained video editing.
In evidence-constrained editing, we specify what each passage needs
to show and how footage may support it. A shot can be used for that
passage only if the system verifies that it meets these requirements.
Given a user request, a routing model determines whether external
research is beneficial. When needed, the search agent iteratively
retrieves evidence, checks key factual statements against the sources, and organizes the selected
information into the script.
This script serves as a shared blueprint for production, guiding
footage retrieval and defining the visual requirements of
individual narration units.

Video Harness then connects this narrative to the final video through
\textbf{visual verification} and \textbf{global shot assignment}.
Narrative-guided retrieval discovers source videos and candidate
shots, which are assessed for direct evidence, source or speaker attribution,
contextual illustration, or neutral bridge according to
each unit's support requirements.
Video Harness selects verified clips across narration units to match
their measured durations. Each clip stays within its source interval,
and the same footage is used only once.
When admissible material is insufficient, targeted footage retrieval or
targeted script revision addresses the gaps.
Together, these mechanisms align researched content with the
visual evidence and temporal structure of the finished video.

Our contributions are threefold:
\begin{itemize}
    \item We present \textbf{Video Harness}, an agent framework that connects
    adaptive agentic search, \textbf{narration-first planning}, and
    real-world footage editing for informative video creation.

    \item We use each passage's visual requirements and measured duration
to assign verified footage across the narration. When passages compete
for the same footage or lack enough material, the system identifies
the affected passages and guides further retrieval or script revision.

    \item We evaluate Video Harness across diverse video prompt types and paired search conditions, demonstrating its
    strengths in content-oriented prompts and quantifying the benefits
    of external factual research for video creation.
\end{itemize}

%% file: sec/2_related_work.tex
\section{Related Work}
\label{sec:related}

\subsection{Automated Video Editing.}
Automated video editing assembles existing footage into a complete
video by selecting shots, sequencing clips, and coordinating content.
LLM-based agents take videos as input and plan editing actions toward
specified objectives, such as following a given
script~\citep{10.1145/3640543.3645143,zhou2026videoagent}.
These objectives extend to music-guided
editing~\citep{zhao2026cutclawagentichourslongvideo,li2026direct}.
Narration-driven approaches further organize audiovisual content
around a storyline, synchronizing narration with input
videos~\citep{yang2024synchronized,hu2026mcsc,yan2026crayotter}.
Open-ended requests, such as introducing and comparing universities,
also require determining which footage to acquire and how it supports
the intended explanation.
Video Harness addresses this setting through narration-first planning:
the script guides external footage acquisition, visual
verification, and global shot assignment.

\subsection{Language-Driven Content Creation.}
Natural-language interfaces allow users to express creative goals
through textual instructions.
In the visual domain, prompts guide the creation of coherent story
images and comics~\citep{wang2025autostory,zhou2024storydiffusion},
long-form narratives and illustrated
guides~\citep{yang2025seed,cui2025emu3}, and content requiring
specialized knowledge or external
references~\citep{shalev2026imagerag,
he2026mindbrushintegratingagenticcognitive,feng2026gen}.
Video creation introduces a temporal hierarchy: prompts are expanded
into multi-scene plans and shot-level specifications that coordinate
character identity and
transitions~\citep{lin2024videodirectorgptconsistentmultiscenevideo,
long2024videostudio,zhao2025moviedreamer,zheng2024videogen}.
Agents for end-to-end production further coordinate narration, music,
and sound effects~\citep{wu2025automated,huang2026vimax,xu2025mm}.
Language-driven creation also extends to music and songs from
instructions and
lyrics~\citep{yu2023musicagent,liang2024bytecomposer,
ding2025songcomposer,lei2024songcreator,liu2025songgen,
yuan2026yue,gong2026ace}, as well as games and interactive
worlds~\citep{chen2023gamegpt,earle2025dreamgarden,
bruce2024genie,che2025gamegen,parkerholder2025genie3}.
Video Harness focuses on composing informative videos from acquired
real-world footage, using narrative requirements to guide material
selection and visual evidence to govern its use.

\subsection{LLM-Empowered Agents.}
Agentic systems connect language reasoning with external execution.
They interleave reasoning and action so that observations inform
subsequent decisions~\citep{yao2022react}, decompose requests across
models and tools~\citep{shen2023hugginggpt}, and coordinate multiple
agents through conversation protocols and structured
roles~\citep{wu2023autogen,hong2024metagpt}.
Feedback supports revision through verbal self-reflection stored
in memory~\citep{shinn2023reflexion}, replanning after execution
failures~\citep{fourney2024magentic}, and iterative refinement with
verification in visual
creation~\citep{yang2024idea2img,wang2024genartist}.
Video Harness derives actionable feedback from visual support
assessments and global assignment constraints, identifying unmet
narration requirements, insufficient footage duration, and competition
for source intervals.
These diagnoses guide targeted footage retrieval and targeted script revision, linking agent decisions to the narrative and
temporal requirements of the final video.

%% file: sec/2_method.tex
\section{Method}
\label{sec:method}

% Allow one full-width float at the top of a regular two-column page.
\setcounter{dbltopnumber}{1}

\subsection{Overview}

\begin{figure*}[t]
    \centering
    \captionsetup{position=bottom,skip=2pt}
    \includegraphics[width=\textwidth]{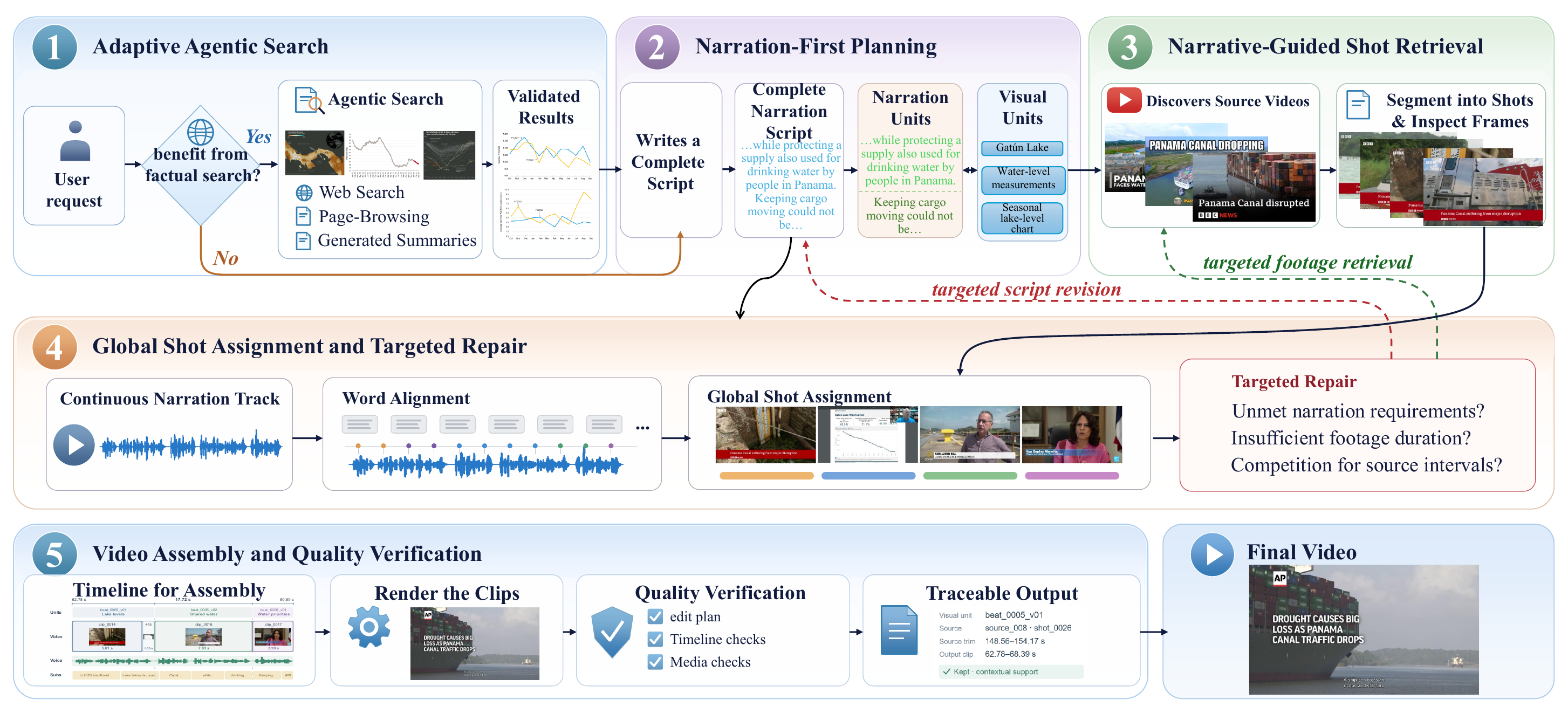}
    \caption{\textbf{Overview of Video Harness.}
    The script determines what footage to retrieve, and the synthesized
    narration determines how long each passage lasts. Global assignment
    uses both to build the video timeline.}
    \label{fig:method}
    \vspace{-2pt}
\end{figure*}

\textbf{Video Harness} is an agent framework that turns a natural-language
instruction \(q\) into a narrated, subtitled video using retrieved
footage in Fig.~\ref{fig:method}. It first writes a complete script \(N\),
drawing on web search when the request requires additional factual
support. The script guides both shot retrieval and timing. For each
passage, the system specifies visual requirements and verifies
candidate shots against them. It synthesizes the full narration to
measure how long each passage lasts. Video Harness assigns verified shots across visual units to match the measured narration durations. If the available footage cannot cover a passage, the system retrieves more footage or revises that passage. The resulting timeline guides assembly, subtitling, and verification.

\subsection{Adaptive Search-Augmented Narrative Planning}

Before drafting the script, a routing model assesses whether the request would
benefit from factual search. It considers the instruction, the date
of execution, and gaps in recency, precision, explanatory detail, or
source-dependent comparisons. It classifies the benefit as definite,
uncertain, or absent. Only a definite benefit triggers search, for
which the routing model specifies the information gaps, initial queries,
scope, and any reporting cutoff. Users can disable factual search
independently of YouTube footage retrieval.

Video Harness adapts Gen-Searcher's web-search and page-browsing
implementations. A language model refines queries, reads sources, and
decides when to stop. Each search entry links a factual statement to a source
identifier and a supporting quotation. Original passages are stored
separately from generated summaries. Statement admission requires a
matching passage, corroboration across source domains, and date checks
for time-sensitive requests. Failed searches trigger bounded retries
or a search for alternative sources. At budget exhaustion, the system
returns validated results or reports failure.

The writer uses the instruction and any validated statements selected for
the script. Selection preserves statement identifiers and citations, giving
priority to required facts, qualifications, and explanatory relationships
even when they are difficult to illustrate. A topic-level check of
YouTube metadata provides an initial estimate of footage availability.
With or without search, the writer produces the script
before the system acquires footage.

\subsection{From Narration to Visual Units}

Video Harness divides \(N\) into ordered narration units, each with its
text, character span, narrative function (e.g., a fact or transition),
and visual requirements. It then splits these into visual units
\(\mathcal B=(b_1,\ldots,b_H)\) so that several shots can illustrate
one passage of continuous narration. Each \(b_h\) refers to a script
span and specifies the entities and actions to show, preferred scenes,
prohibited content, eligible sources, and permitted forms of visual
support. These specifications guide retrieval. For example, a factual
statement may permit contextual footage, while a passage requiring
direct depiction of an action needs a shot that shows it.

\subsection{Narrative-Guided Shot Retrieval and Verification}

\begin{figure*}[t]
    \centering
    \captionsetup{position=bottom,skip=0pt}
    \includegraphics[width=\linewidth,trim={0pt 6cm 0pt 0pt},clip]{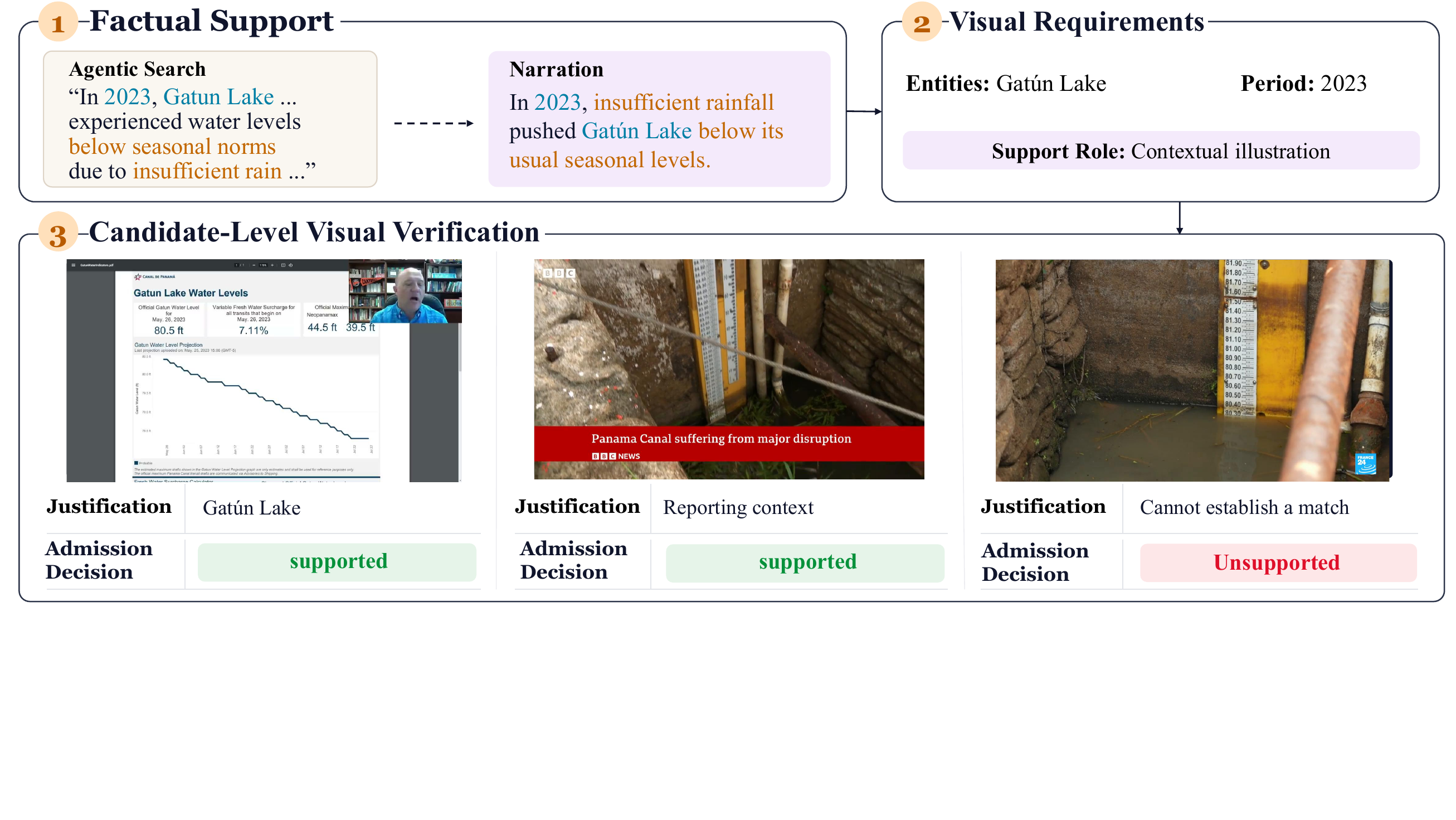}
    \caption{\textbf{Candidate-level visual verification.}
Agentic search supports the narration that low rainfall left
Gat\'{u}n Lake below its usual seasonal level in 2023.
The verifier admits the dated chart (left) and the gauge (center)
as contextual illustrations. The chart names the lake, and the
accompanying report links the gauge to it. The verifier rejects
the gauge on the right because it cannot establish the same link.}
    \label{fig:evidence-support}
    \vspace{-0.2cm}
\end{figure*}

Video Harness discovers source videos for the full script, then
retrieves shots for each visual unit. Source queries cover the
objects, processes, and scenes in the narrative. Videos selected by
metadata are downloaded, segmented into shots, and transcribed. A
vision-language model inspects frames near each shot's start, middle,
and end to describe visible objects, actions, scenes, and text. A shot
catalog stores these descriptions, transcripts, and quality metrics
alongside source identifiers and time intervals.

Shot retrieval uses the catalog independently of the source videos'
search-engine ranks. The system processes the catalog in batches and
uses compact narrative summaries to route each batch to relevant units.
For each unit \(b_h\), a language
model reranks candidates jointly using their metadata, transcripts, and
the unit's visual requirements. It rates each candidate \(j\) for
semantic relevance, metadata support, narrative fit, duration fit, and
redundancy. The resulting metadata score is:
\begin{equation}
\begin{aligned}
L_{hj}={}&w_{\mathrm{sem}}s_{\mathrm{sem}}
+w_{\mathrm{meta}}s_{\mathrm{meta}}
+w_{\mathrm{nar}}s_{\mathrm{nar}}\\
&+w_{\mathrm{dur}}s_{\mathrm{dur}}
+w_{\mathrm{dup}}(1-r_{\mathrm{dup}}),
\end{aligned}
\end{equation}
where the weights
\((w_{\mathrm{sem}},w_{\mathrm{meta}},w_{\mathrm{nar}},
w_{\mathrm{dur}},w_{\mathrm{dup}})\) are set to
\((0.28,0.24,0.24,0.14,0.10)\).

A vision-language model verifies representative frames from each
shortlisted shot and returns a visual-support score \(s_{\mathrm{vis}}\),
an admission decision, and a support role \(m_{hj}\). The four roles are
\emph{direct evidence}, \emph{source or speaker attribution},
\emph{contextual illustration}, and \emph{neutral bridge}. Contextual illustrations provide visual context for a factual statement. Neutral bridges are eligible only for permitted transitions.

Admission requires a permitted role, sufficient narrative-fit and
visual-support scores, and acceptable shot quality. The verifier rejects
unsupported, contradictory, or entity-conflicting shots and records
identity as unknown when it cannot establish a match.
Figure~\ref{fig:evidence-support} shows how reporting context affects
admission for a passage on Gat\'{u}n Lake. Admitted shots are ranked by:
\begin{equation}
S_{hj}=\min\left\{1,\,
\alpha L_{hj}+(1-\alpha)s_{\mathrm{vis}}
+\beta(m_{hj})\right\},
\end{equation}
where \(\alpha=0.55\). The role bonus \(\beta(m_{hj})\) is \(0.05\) for
direct evidence, \(0.025\) for attribution, and zero for the remaining
roles. The resulting pool \(\mathcal C_h\) records each admitted shot's
source interval, compatibility score, support role, and justification.
Only shots in \(C_h\) can be assigned to visual unit \(b_h\) in the
final timeline.
Rejection reasons are stored separately. If too few shots qualify, the
system expands retrieval within the catalog or across analyzed sources,
subject to a retrieval budget.

\subsection{Global Shot Assignment and Targeted Repair}

For each visual unit \(b_h\), global assignment needs both the admitted
shots in \(\mathcal C_h\) and the duration of the corresponding
narration. Video Harness measures this duration by synthesizing the full
script as one audio track and aligning the words to timestamps. For
English narration, Whisper-based alignment gives the start and end
times of each unit. Its interval and duration are:
\begin{equation}
\mathcal T_h=[\tau_h^{\mathrm{start}},\tau_h^{\mathrm{end}}],
\qquad
d_h=\tau_h^{\mathrm{end}}-\tau_h^{\mathrm{start}}.
\end{equation}
The system checks that the alignment covers the text, that timestamps
are monotonic, and that the audio supports the estimated boundaries.

Different visual units may need the same source interval. Selecting
the best shot for one unit can therefore leave another without coverage.
We use a mixed-integer linear program (MILP) to assign source intervals
jointly across units. One or more clips cover the measured narration duration of each unit.
The assignment avoids reusing source intervals, and each selected clip
retains the representative frame used for visual verification.
Appendix~\ref{app:global-assignment} gives the objective, constraints,
and solver details.

Local capacity estimates flag potential shortages and guide
candidate expansion, additional footage retrieval, or targeted
script revision. These heuristic flags can trigger recovery
before the global solver is invoked.
Separately, when the global MILP confirms infeasibility,
a diagnostic optimization identifies a set of units whose
removal restores feasibility. A second test introduces
hypothetical independent footage to estimate the additional
duration and number of distinct shots required.
The resulting report links shortages to script spans,
competing shots, and sources.

The controller can expand retrieval within the catalog, search other
analyzed sources, or acquire more footage. New shots undergo the same
verification before reassignment. Repairs may also update source
routes or visual-support roles after validation. When a passage needs
revision, only the writer edits its text. The system retains unaffected
text and regenerates or validates the affected speech and timing before
reassignment. Retrieval and revision have separate budgets.
Appendix~\ref{app:capacity-feedback} details the procedure.

% \begin{figure*}[t]
%     \centering
%     \captionsetup{position=bottom,skip=-2pt}
%     \includegraphics[width=\textwidth,trim={0 3mm 0 0},clip]{sec/fig4.pdf}
%     \caption{\textbf{Global shot assignment.}
%     The MILP selects source intervals from the admitted pools and assigns
%     their durations and positions jointly across units.
%     Clips A and B together cover the first visual unit
%     ($5.61\,\mathrm{s} + 1.00\,\mathrm{s} = 6.61\,\mathrm{s}$),
%     while clips C and D cover the second and third visual units,
%     respectively.}
%     \label{fig:global-allocation-repair}
% \end{figure*}

\subsection{Video Assembly and Quality Verification}

A feasible assignment gives each visual unit an ordered list of clips
\(\mathcal A_h=\{(v_{hk},u_{hk},\ell_{hk})\}_{k=1}^{K_h}\), where the
three entries specify the source video, start time in that video, and
clip duration. Video Harness extracts and standardizes the clips, places
them in narrative order, and fills each unit's measured interval on the
continuous narration track. It then derives subtitle timing from the
word alignment and renders the clips with the narration and subtitles.

Verification proceeds at three levels. For the edit plan, the system
checks schemas, script spans, and unit coverage, and compares each
selected clip with its saved assessment of support role, eligible
sources, and prohibited content. Timeline checks enforce source bounds,
non-reuse, duration coverage, and a gap-free, non-overlapping sequence.
They flag continuity risks separately from hard-constraint violations.
Media checks inspect the rendered streams, decoding, durations, and
subtitle timing, and compare suspected black frames or freezes with
the source clips.

Citations link factual statements in the narration to their sources. Shot
identifiers, source intervals, assessments, and assignments link the
rendered clips to their origins. The system also records model and
prompt identifiers and artifact hashes. A failed check triggers a
bounded retry, additional retrieval, targeted script revision, or a
correction to the edit structure or rendered media. Checkpoints are
reused only when input, model, prompt, and artifact identities match,
and failed outputs remain available for inspection. Completion
requires all planning, semantic, technical, and integrated checks to pass.

%% file: sec/2_experiment.tex
\section{Experiments}
\label{sec:experiments}

\begin{figure}[htbp]
    \centering
    \captionsetup{position=bottom,skip=2pt}
    \includegraphics[width=0.87\columnwidth]{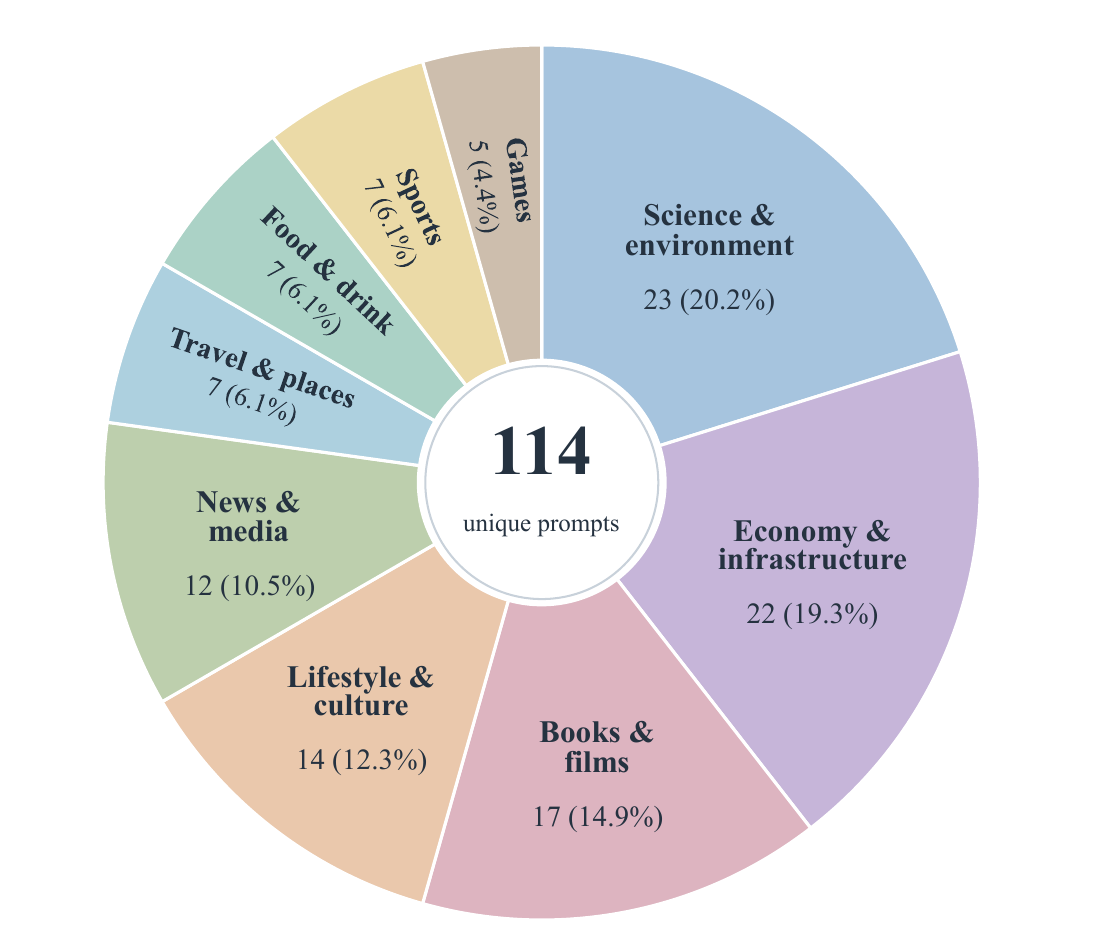}
    \caption{\textbf{Evaluation prompts by content topic.}
    Counts and percentages give the distribution of the prompts
    across nine content topics.}
    \label{fig:prompt-topic-distribution}
    \vspace{-5pt}
\end{figure}

\paragraph{Prompts and baselines.}
We evaluate 114 prompts spanning nine content topics in
Fig.~\ref{fig:prompt-topic-distribution} and seven prompt types in
Table~\ref{tab:topic-category-results}. Content topics describe the
subject of a video, while prompt types describe what the user asks the
video to accomplish. The three cross-system evaluations use the 558 available
videos from the 570 system--prompt pairs. 

Table~\ref{tab:generation-models} lists the
configured model for each generation module.
Table~\ref{tab:generation-budgets} in the appendix reports the
search, footage retrieval, and retry budgets in our experiments.

\begin{table}[tb]
    \centering
    \caption{\textbf{Generation models by module.}
    These are generation models, distinct from the evaluation judges.}
    \label{tab:generation-models}
    \footnotesize
    \setlength{\tabcolsep}{2pt}
    \renewcommand{\arraystretch}{1.05}
    \begin{tabular*}{\columnwidth}{@{\extracolsep{\fill}}ll@{}}
        \toprule
        \textbf{Module} & \textbf{Model} \\
        \midrule
        Routing model
            & GPT-6-astra \\
        Agentic search
            & GPT-6-astra\\
        Narrative planning
            & GPT-6-astra \\
        Source-level analysis
            & GPT-5.6-sol \\
        Metadata ranking
            & GPT-5.6-sol \\
        Candidate-level visual verification
            & GPT-5.6-sol \\
        Narration synthesis
            & MiniMax Speech-2.8-hd \\
        Targeted script revision
            & GPT-6-astra \\
        Semantic checks
            & GPT-5.6-sol \\
        \bottomrule
        \vspace{-18pt}
    \end{tabular*}
\end{table}

\subsection{Overall Video Quality}
\label{sec:overall-video-quality}

We independently score all videos with GPT-6.1-sol using the five
criteria of Crayotter~\cite{yan2026crayotter}. The judge rates theme
alignment, content richness, narrative coherence, editing smoothness,
and visual quality on an integer scale from 1 to 5. It receives the
original prompt, an independent automatic speech recognition
 transcript of the final soundtrack, and up to 56 frames. We sample
32 frames uniformly and four around each of up to six detected cuts,
omitting unavailable frames. We average the five ratings within each completed video.
Missing outputs receive zero in all five criteria.
We compute each system's score over all prompt-level scores.

\begin{table}[tb]
    \centering
    \caption{\textbf{Overall video quality.}
    Completed videos are rated on a 1--5 scale.}
    \label{tab:overall-video-quality}
    \footnotesize
    \setlength{\tabcolsep}{1.2pt}
    \resizebox{\columnwidth}{!}{%
        \begin{tabularx}{\columnwidth}{@{}l*{6}{>{\centering\arraybackslash}X}@{}}
            \toprule
            \textbf{Method}
                & \textbf{\boldmath Theme$\uparrow$}
                & \textbf{\boldmath Content$\uparrow$}
                & \textbf{\boldmath Story$\uparrow$}
                & \textbf{\boldmath Editing$\uparrow$}
                & \textbf{\boldmath Visual$\uparrow$}
                & \textbf{\boldmath Overall$\uparrow$} \\
            \midrule
            Crayotter
                & 3.02 & 3.34 & 3.08 & 2.64 & 2.29 & 2.87 \\
            ChatCut
                & 2.44 & 2.50 & 2.96 & 2.57 & \textbf{3.24} & 2.74 \\
            CapCut-Mate
                & 2.75 & 2.97 & 3.59 & 2.95 & 3.04 & 3.06 \\
            CutClaw
                & 2.03 & 2.67 & 2.14 & 2.93 & 2.95 & 2.54 \\
            \textbf{Video Harness}
                & \textbf{3.85} & \textbf{4.01} & \textbf{3.96}
                & \textbf{3.15} & 2.98 & \textbf{3.59} \\
            \bottomrule
        \end{tabularx}%
    }
    \vspace{-8pt}
\end{table}

Video Harness ranks first overall at 3.59, followed by
CapCut-Mate, Crayotter, ChatCut, and
CutClaw in Table~\ref{tab:overall-video-quality}.
Its largest margins over the strongest baseline on each criterion
are in theme alignment and content richness.
Video Harness also leads in narrative coherence and editing smoothness.
ChatCut achieves the highest visual quality, which scores 3.24,
compared with 2.98 for Video Harness.

To complement the automated evaluation, three human evaluators rated
the videos generated for 20 selected prompts. They used the same five
criteria as the automated evaluation: theme alignment, content richness,
narrative coherence, editing smoothness, and visual quality.
As shown in Table~\ref{tab:mixed-quality} in the appendix, VideoHarness
received the highest mean overall score of 4.01. Crayotter and
CapCut-Mate tied for the highest baseline score at 3.20.

\subsection{Audiovisual Quality and Visual Evidence}
\label{sec:dimension-specific-evaluation}

We evaluate seven dimensions of audiovisual quality on a 1--5 scale in
Table~\ref{tab:custom-quality} and assess visual evidence at the
level of individual factual statements in Table~\ref{tab:custom-evidence}.
GPT-6.1-sol scores narration content, visual presentation, semantic
and temporal alignment between narration and footage, local editorial
continuity, and global narrative coherence. Gemini 3.8 Flash rates
audio production from the final soundtrack. We reuse the independent
ASR transcripts. In Table~\ref{tab:custom-quality}, \emph{Continuity}
denotes local editorial continuity, and \emph{Coherence}
denotes global narrative coherence.

\textbf{Narration and editing.}
Video Harness has the highest average score across the seven dimensions at 4.29, compared with 4.08 for CapCut-Mate and 3.61 for ChatCut. Video Harness scores highest in three areas, with 4.82 for narration content, 4.19 for continuity between nearby shots, and 4.96 for coherence across the full video. CapCut-Mate has the highest score for audio production at 4.96.

\begin{table*}[t]
    \centering
    \caption{\textbf{Audiovisual quality by dimension.}
    Completed videos are rated on a 1--5 scale; Score equally weights the seven dimension means.}
    \label{tab:custom-quality}
    \small
    \setlength{\tabcolsep}{3pt}
        \begin{tabular*}{\textwidth}{@{\extracolsep{\fill}}lcccccccc@{}}
        \toprule
        \multirow{2}{*}{Method}
            & \multicolumn{3}{c}{\textbf{Narration and presentation}}
            & \multicolumn{2}{c}{\textbf{Audiovisual alignment}}
            & \multicolumn{2}{c}{\textbf{Editing and narrative}}
            & \multirow{2}{*}{Score\,${\uparrow}$} \\
            & Narration\,${\uparrow}$   & Audio\,${\uparrow}$  & Visual\,${\uparrow}$
            & Semantic\,${\uparrow}$    & Timing\,${\uparrow}$
            & Continuity\,${\uparrow}$  & Coherence\,${\uparrow}$ & \\
        \midrule
        Crayotter
            & 2.56 & 2.67 & 2.93 & 3.35 & 4.44
            & 3.99 & 3.74 & 3.38 \\
        ChatCut
            & 4.12 & 3.89 & \textbf{3.46} & 1.99 & 4.11
            & 3.69 & 4.00 & 3.61 \\
        CapCut-Mate
            & 4.43 & \textbf{4.96} & 3.31 & 2.32 & \textbf{4.71}
            & 4.00 & 4.84 & 4.08 \\
        CutClaw
            & 3.83 & 3.74 & 3.26 & \textbf{4.04} & 4.59
            & 4.02 & 2.61 & 3.73 \\
        \textbf{Video Harness}
            & \textbf{4.82} & 4.75 & 3.24 & 3.80 & 4.31
            & \textbf{4.19} & \textbf{4.96} & \textbf{4.29} \\
        \bottomrule
    \end{tabular*}
    \vspace{-4pt}
\end{table*}

\textbf{Audiovisual timing.}
To assess whether footage appears when the corresponding content is
spoken, we sample complete clauses near one-third and two-thirds of
cumulative speech duration, with two seconds of context on either
side. When word timestamps are unavailable, we use fixed windows at
one-third and two-thirds of the video duration. Gemini 3.8 Flash
examines the footage and audio on a shared timeline. We localize
matched speech using available ASR word timings before GPT-6.1-sol
assigns a score. A video receives a timing score only when a correspondence
can be assessed. CapCut-Mate achieves the highest reported timing score at 4.71.
ChatCut gives a score of 4.11.
Video Harness scores 4.31 on all prompts.

\textbf{Visual evidence.}
GPT-6.1-sol compares the original visually testable statements
with the footage in sampled narration windows for each video.
We classify their visual alignment as Direct, Contextual, or
Insufficient, using the definitions in
Table~\ref{tab:custom-evidence}.
Percentages pool statements across these videos.
For Video Harness, 33.78\% receive Direct, 63.79\% receive Contextual,
and 2.44\% receive Insufficient.

\begin{table}[tb]
    \centering
    \caption{\textbf{Visual evidence matching} (\%).
    \emph{Statements} denotes the number of original visually testable
    statements from completed videos.
    Direct and Contextual denote direct depiction and relevant
    illustrative footage; Insufficient denotes insufficient visual
    alignment, including unrelated footage or material conflicts.
    Percentages pool statements across the original sampled narration
    windows. Missing outputs and videos without eligible statements
    do not contribute.}
    \label{tab:custom-evidence}
    \footnotesize
    \setlength{\tabcolsep}{1.2pt}
    \begin{tabularx}{\columnwidth}{@{}l*{4}{>{\centering\arraybackslash}X}@{}}
          \toprule
        \textbf{Method}
            
            & \textbf{\boldmath Direct$\uparrow$}
            & \textbf{\boldmath Contextual$\uparrow$}
            & \textbf{\boldmath Insufficient$\downarrow$} \\
        \midrule
        Crayotter    & 27.15 & 57.73 & 15.13 \\
        ChatCut      & 5.78  & 46.77 & 47.45 \\
        CapCut-Mate  & 9.04  & 55.62 & 35.34 \\
        CutClaw      & 32.61 & 43.48 & 23.91 \\
        \textbf{Video Harness}
            & \textbf{33.78} & \textbf{63.79} & 2.44 \\
        \bottomrule
    \end{tabularx}
    \vspace{-14pt}
\end{table}

\subsection{Performance by Prompt Type}
\label{sec:prompt-type-comparison}

For each prompt, we fix a shared evaluation specification before
scoring. GPT-6.1-sol compares each video's ASR transcript and
temporally ordered frames against this specification using six
prompt-fulfillment metrics on a 0--100 scale.
\emph{Content Coverage} measures how much of the requested content the video
covers with supporting evidence.
\emph{Function} measures coverage of the required editorial functions.
\emph{Narrative} measures the fraction of specified pairwise narrative
relations that the video satisfies. \emph{Allocation} compares the
reference share assigned to each content unit by the evaluation
specification with the fraction of the video's total duration
covered by footage that visually matches that unit.
\emph{Evidence}
measures the fraction of evaluated atomic statements with supporting
evidence. \emph{Visual Match} measures the fraction of the video's
own duration covered by footage that both matches the narration
semantically and serves the specified purpose. We average the resulting prompt-level scores and give the six metric means equal weight, both within each prompt type and
across the full set.

\textbf{Prompt fulfillment.}
Video Harness leads on all six metrics, averaging 51.85 compared with
34.68 for Crayotter in Table~\ref{tab:topic-dimension-results}.
Video Harness has its biggest lead in fulfilling narrative dependencies. It scores 62.75, compared with 36.80 for CapCut-Mate, the next-best system on this measure.

\textbf{Differences across prompts.}
Table~\ref{tab:topic-category-results} shows that Video Harness ranks
first in six of the seven prompt types. On the 105 prompts that focus
on organizing information, Video Harness averages 52.33 and Crayotter
averages 34.09. The gap is smaller on the nine visual-experience and
montage prompts, where Video Harness averages 46.35 and Crayotter
averages 42.06. Crayotter leads on visual experience with 46.12,
followed by Video Harness at 40.91. On montage and highlights,
Video Harness scores 50.70 and Crayotter scores 38.57.

The overall ranking stays the same on the prompts completed by all systems. Video Harness leads with 51.79, followed by Crayotter
with 34.52, CapCut-Mate with 21.56, ChatCut with 18.79, and CutClaw
with 16.18.

\begin{table}[tb]
    \centering
    \caption{\textbf{Prompt fulfillment by prompt type} on a 0--100 scale.
    Cray., Chat., Cap., and Cut.\ denote Crayotter, ChatCut,
    CapCut-Mate, and CutClaw.}
    \label{tab:topic-taxonomy}
    \label{tab:topic-category-results}
    \footnotesize
    \setlength{\tabcolsep}{1.2pt}
    \begin{tabular*}{\columnwidth}{@{\extracolsep{\fill}}lrrrrr@{}}
        \toprule
        \textbf{Prompt type} & \textbf{Cray.} & \textbf{Chat.}
            & \textbf{Cap.} & \textbf{Cut.} & \textbf{Ours} \\
        \midrule
        Process explanation$\uparrow$
            & 38.55 & 25.09 & 24.99 & 27.38 & \textbf{57.11} \\
        Role comparison$\uparrow$
            & 33.65 & 18.96 & 24.30 & 13.84 & \textbf{54.32} \\
        Historical evolution$\uparrow$
            & 26.70 & 16.88 & 26.65 & 13.50 & \textbf{50.07} \\
        Mechanism explanation$\uparrow$
            & 37.68 & 15.81 & 24.17 & 15.72 & \textbf{57.16} \\
        Recommendations/rankings$\uparrow$
            & 30.26 & 12.29 & 11.98 & 12.94 & \textbf{43.06} \\
        Visual experience$\uparrow$
            & \textbf{46.12} & 20.15 & 39.24 & 16.44 & 40.91 \\
        Montage/highlights$\uparrow$
            & 38.57 & 16.11 & 15.20 & 10.19 & \textbf{50.70} \\
        All prompts$\uparrow$
            & 34.68 & 16.81 & 21.73 & 15.67 & \textbf{51.85} \\
        \bottomrule
    \end{tabular*}
    \vspace{-5pt}
\end{table}
\begin{table}[tb]
    \centering
    \caption{\textbf{Prompt fulfillment by metric} on a 0--100 scale. Method
    abbreviations follow Table~\ref{tab:topic-category-results}.}
    \label{tab:topic-dimension-results}
    \footnotesize
    \setlength{\tabcolsep}{2pt}
    \begin{tabular*}{\columnwidth}{@{\extracolsep{\fill}}lrrrrr@{}}
                \toprule
        \textbf{Metric} & \textbf{Cray.} & \textbf{Chat.}
            & \textbf{Cap.} & \textbf{Cut.} & \textbf{Ours} \\
        \midrule
        Content Coverage$\uparrow$ & 10.78 &  1.20 &  3.71 &  2.37 & \textbf{29.66} \\
        Function$\uparrow$         & 49.78 & 44.74 & 52.63 &  9.87 & \textbf{69.96} \\
        Narrative$\uparrow$        & 36.68 & 36.02 & 36.80 &  1.40 & \textbf{62.75} \\
        Allocation$\uparrow$       & 37.52 &  5.34 & 15.06 & 24.60 & \textbf{51.73} \\
        Evidence$\uparrow$         & 22.60 &  1.59 &  2.91 & 22.62 & \textbf{32.48} \\
        Visual Match$\uparrow$     & 50.73 & 11.96 & 19.25 & 33.15 & \textbf{64.54} \\
        \textbf{Score}$\uparrow$            & 34.68 & 16.81 & 21.73 & 15.67 & \textbf{51.85} \\
        \bottomrule
    \end{tabular*}
    \vspace{-18pt}
\end{table}

\begin{table*}[htbp]
\centering
\footnotesize
\setlength{\tabcolsep}{1.2pt}
\renewcommand{\arraystretch}{1.05}
\caption{Component ablations on matched completed prompts. Scores use a 1--5 scale; Score equally weights the seven
unrounded dimension means. Full denotes the complete-system control within each ablation.
The assignment comparison uses paired re-renderings, whereas
the feedback comparison uses the original Full outputs.}
\label{tab:ablation-quality}
\begin{tabularx}{\textwidth}{@{}l*{8}{>{\centering\arraybackslash}X}@{}}
\toprule
\textbf{Variant}
    & \textbf{\boldmath Narration$\uparrow$}
    & \textbf{\boldmath Audio$\uparrow$}
    & \textbf{\boldmath Visual$\uparrow$}
    & \textbf{\boldmath Semantic$\uparrow$}
    & \textbf{\boldmath Timing$\uparrow$}
    & \textbf{\boldmath Continuity$\uparrow$}
    & \textbf{\boldmath Coherence$\uparrow$}
    & \textbf{\boldmath Score$\uparrow$} \tabularnewline
\midrule
\multicolumn{9}{@{}l}{\textit{Global shot assignment}} \tabularnewline
Greedy & 3.65 & 4.91 & 3.17 & 3.76 & 4.43 & 4.13 & 4.91 & 4.14 \tabularnewline
Full   & 3.74 & 4.87 & 3.22 & 3.74 & 4.33 & 4.14 & 4.96 & 4.14 \tabularnewline
% \multicolumn{9}{@{}l}{\textit{Visual verification: 40 matched pairs}} \tabularnewline
% w/o verification & 4.78 & 4.88 & 3.05 & 3.88 & 4.33 & 4.23 & 4.90 & 4.29 \tabularnewline
% Full             & 4.95 & 4.70 & 3.23 & 3.75 & 4.29 & 4.18 & 4.93 & 4.29 \tabularnewline
% \midrule
\multicolumn{9}{@{}l}{\textit{Gap-aware feedback}} \tabularnewline
w/o gap details & 5.00 & 4.76 & 3.28 & 3.79 & 4.38 & 4.22 & 4.93 & 4.34 \tabularnewline
Full            & 5.00 & 4.69 & 3.28 & 3.78 & 4.28 & 4.25 & 4.90 & 4.31 \tabularnewline
\bottomrule
\end{tabularx}
\vspace{-2pt}
\end{table*}

\subsection{Ablation Studies}
\label{sec:ablation}

We select 40 prompts by category and full-system performance
and use the evaluation in
Sec.~\ref{sec:dimension-specific-evaluation}.
We report completion over all prompts.
Within each ablation, both variants' quality scores are
averaged over exactly the same prompts completed by both;
missing outputs are excluded.
Table~\ref{tab:ablation-quality} reports audiovisual quality
on matched completed prompts. Under strict verification and allocation requirements, the ablated
variants may fail to find a valid global assignment due to insufficient
visual support, footage shortages, or allocation conflicts, and thus
fail to produce a video. These failures are reflected in completion
counts but excluded from matched quality averages. Full completes 100\% of the prompts. Greedy completes 57.5\%, and the version without gap details completes 72.5\%.
Thus, despite similar quality among matched completions, the main
observed benefit of these components is improved completion.

\textbf{Global shot assignment.}
With the same timing, candidates, and constraints, MILP completes 100\% of the prompts. Sequential greedy selection without backtracking completes 57.5\% (Table~\ref{tab:ablation-outcomes}). On the prompts completed by both variants, the mean scores for Full
and sequential greedy selection without backtracking both round to 4.14.
\begin{table}[htbp]
    \centering
    \footnotesize
    \setlength{\tabcolsep}{1.5pt}
    \renewcommand{\arraystretch}{1.05}
    \caption{Completion over all prompts and statement-pooled
    Visual evidence matching (\%) on matched completed prompts. Category definitions follow
    Table~\ref{tab:custom-evidence}.}
    \label{tab:ablation-outcomes}
    \begin{tabularx}{\columnwidth}{@{}l*{4}{>{\centering\arraybackslash}X}@{}}
                \toprule
        \textbf{Variant}
            & \textbf{\boldmath Completed$\uparrow$}
            & \textbf{\boldmath Direct$\uparrow$}
            & \textbf{Contextual$\uparrow$}
            & \textbf{\boldmath Insufficient$\downarrow$} \\
        \midrule
        \multicolumn{5}{@{}l}{\textit{Global shot assignment}} \\
        Greedy & 57.5\% & 30.34 & 62.92 & 6.74 \\
        Full   & 100\% & 32.18 & 66.09 & 1.72 \\
        % \multicolumn{5}{@{}l}{\textit{Visual verification}} \\
        % w/o verification & 40/40 & 27.05 & 70.52 & 2.43 \\
        % Full             & 40/40 & 31.36 & 66.27 & 2.37 \\
        % \midrule
        \multicolumn{5}{@{}l}{\textit{Gap-aware feedback}} \\
        w/o gap details & 72.5\% & 30.42 & 67.30 & 2.28 \\
        Full            & 100\% & 32.95 & 65.13 & 1.92 \\
        \bottomrule
    \end{tabularx}
    \vspace{-5pt}
\end{table}
% \textbf{Candidate-level visual verification.}
% We turn off visual verification and keep source-level analysis and metadata ranking. Both versions complete all 40 prompts. With visual verification, the visual score is 3.23 and the supported-statement rate is 26.92\%. Without it, the visual score is 3.05 and the supported-statement rate is 24.62\%.

\textbf{Gap-aware feedback.}
We remove gap diagnostics, duration hints, and prior explanations while
retaining retrieval and script revision. Full produces 100\% outputs, while the variant without gap details produces 72.5\%. Restricting both variants to the same prompts completed
by both, Full achieves a mean audiovisual quality score of
4.31, compared with 4.34 for the variant without gap details, $\Delta=-0.03$, which is computed by
subtracting the variant's score from the Full score

\textbf{Agentic search.}
We compare Search ON and Search OFF on a separate cohort
of 40 deliberately selected, relatively complex prompts.
In the ON condition, the router described in Sec.~3.2 decides
whether to run factual search before drafting the narration.

On this cohort, factual search was triggered and executed
for all prompts in the ON condition.

In the OFF condition,
factual search is disabled for all prompts. Both conditions retain
the same footage retrieval procedure.
GPT-6.1-sol evaluates anonymized pairs using
shared prompt criteria, supplemental references, and an information
cutoff defined by the earlier completion time of each pair.

% \begin{table}[t]
%     \centering
%     \footnotesize
%     \setlength{\tabcolsep}{-4pt}
%     \renewcommand{\arraystretch}{1.0}
%     \caption{Weights for the paired Search ON/OFF comparison.}
%     \label{tab:search-evaluation-weights}
%     \begin{tabular}{@{}lr@{}}
%         \toprule
%         Criterion & Weight \\
%         \midrule
%         Prompt completion                      & 10\% \\
%         Factual accuracy                       & 10\% \\
%         Key-fact traceability                  & 20\% \\
%         Data adequacy                          & 15\% \\
%         Freshness                              & 15\% \\
%         Narrative clarity                      &  5\% \\
%         Visual support                         & 17\% \\
%         Audio, subtitles, and playback quality &  8\% \\
%         Total                                  & 100\% \\
%         \bottomrule
%     \end{tabular}
%     \vspace{-14pt}
% \end{table}

\begin{table}[htbp]
\centering
\footnotesize
\setlength{\tabcolsep}{2pt}
\renewcommand{\arraystretch}{1.05}
\captionsetup{position=top,skip=8pt}
\caption{Paired results for Search ON and Search OFF.
Win, Tie, and Loss give the percentage of pairs in each group
that Search ON wins, ties, or loses.
Score differences of at most 0.5 points in either direction count as ties. $\Delta$ is the mean paired score difference, computed as Search ON minus Search OFF.}
\label{tab:search-ablation}
\begin{tabularx}{\columnwidth}{@{}l*{6}{>{\centering\arraybackslash}X}@{}}
\toprule
\textbf{Group}
    & \textbf{\boldmath ON$\uparrow$}
    & \textbf{\boldmath OFF$\uparrow$}
    & {\boldmath$\Delta\uparrow$}
    & \textbf{Win$\uparrow$} & \textbf{Tie} & \textbf{Loss$\downarrow$} \\
\midrule
All prompts   & 75.52 & 67.91 &  +7.62 &  70.0\% & 7.5\% & 22.5\% \\
2026 news     & 71.92 & 42.87 & +29.05 & 100.0\% & 0.0\% &  0.0\% \\
Other prompts & 76.29 & 73.22 &  +3.07 &  63.6\% & 9.1\% & 27.3\% \\
\bottomrule
\end{tabularx}
\vspace{-2pt}
\end{table}

As shown in Table~\ref{tab:search-ablation}, search ON scores 75.52 and search OFF scores 67.91.
The largest gain occurs on the news prompts, where ON wins every pair.
The remaining prompts improve by 3.07 points.

%% file: sec/3_Conclusion.tex
\section{Conclusion}
\label{sec:conclusion}

We presented Video Harness, an agent framework for video creation that
integrates adaptive agentic search, narration-first planning, and
evidence-constrained editing. 
For each narration passage, the system specifies what the footage
needs to show and whether contextual footage is sufficient or a
direct view of the action is required.
A complete narration guides footage
retrieval, while visual support and measured narration timing govern
global shot assignment. Targeted footage retrieval and targeted script revision coordinate narrative requirements with available material.
Video Harness achieves the highest scores among the evaluated systems
for narration content and global narrative coherence. Search ON improves the mean composite score by
7.62 points over Search OFF. The gain is largest on the news
prompts, where Search ON wins every comparison by an average margin
of 29.05 points. Together, these components connect
research, narration, visual evidence, and editing in a unified workflow
for informative video creation.

%% file: sec/llm_statement.tex
\section*{LLM Usage Statement}
We used ChatGPT to help write code and revise the manuscript.
We reviewed and tested the AI-assisted code and checked the manuscript
and reported results. We designed the method and evaluation protocols,
drew the conclusions, and take full responsibility for the final paper.

%% file: sec/fulu.tex
\appendix

\section{Global Shot Assignment}
\label{app:global-assignment}

For each visual unit \(b_h\), we select verified footage to match its
narration duration \(d_h\). We choose the source intervals, start times,
and clip durations jointly across units.

\subsection{Candidate Interval Construction}

For each visual unit \(b_h\), we construct candidate intervals from
its pool of admitted shots \(\mathcal{C}_h\). Each shot is a candidate
interval on its own. Consecutive shots from the same source also form
a candidate interval when their boundaries meet. Let \(I_h\) denote
the set of all such intervals and \(J_{hr}\) the shots in interval \(r\).

We clip each interval to the source bounds and trim \(0.12\) seconds
from each outer edge. Given the resulting bounds \([a_{hr}, b_{hr}]\),
the usable duration is:
\begin{equation}
    c_{hr} = \min\{d_{\max},\, b_{hr}-a_{hr}\},
\end{equation}
where \(d_{\max}=8\) seconds. We discard intervals shorter than
\(d_{\min}=1\) second. Each interval takes the lowest compatibility
score and the worst rank among its shots:
\begin{equation}
    s_{hr} = \min_{j\in J_{hr}} S_{hj},
    \qquad
    \rho_{hr} = \max_{j\in J_{hr}} \mathrm{rank}_{hj}.
\end{equation}

\subsection{Objective and Duration Constraints}

For interval \(r\), \(y_{hr}\in\{0,1\}\) indicates selection,
\(\ell_{hr}\) is the allocated duration, and \(u_{hr}\) is the start
time in the source video. We maximize duration-weighted compatibility
with small penalties for the number of intervals and their ranks:
\begin{equation}
    \max \sum_h \sum_{r\in I_h}
    \bigl[s_{hr}\ell_{hr}-(\epsilon_0+\epsilon_1\rho_{hr})y_{hr}\bigr],
\end{equation}
where \(\epsilon_0=0.01\) and \(\epsilon_1=10^{-5}\).

The clips for each unit must match its narration duration exactly.
Each selected clip must also satisfy the duration limits and stay
within its source bounds:
\begin{align}
    \sum_{r\in I_h}\ell_{hr} &= d_h
        && \forall h, \\
    d_{\min}y_{hr} \leq \ell_{hr} &\leq c_{hr}y_{hr}
        && \forall h,r, \\
    a_{hr} \leq u_{hr}, \quad u_{hr}+\ell_{hr} &\leq b_{hr}
        && \forall h,r.
\end{align}

Each selected clip must retain the frame used to verify each member
shot \(j\in J_{hr}\). If the frame starts at \(t^{\mathrm{ver}}_{hj}\)
and lasts \(\delta_{hj}\), we impose:
\begin{equation}
\begin{aligned}
    u_{hr} &\leq b_{hr}+(t^{\mathrm{ver}}_{hj}-b_{hr})y_{hr}, \\
    u_{hr}+\ell_{hr}
        &\geq a_{hr}+(t^{\mathrm{ver}}_{hj}+\delta_{hj}-a_{hr})y_{hr}.
\end{aligned}
\end{equation}

\subsection{Shot Ownership and Source Constraints}

Selecting an interval reserves all its member shots. With binary
ownership variables \(o_{hj}\), each source shot can belong to at
most one selected interval across all units:
\begin{equation}
    o_{hj} = \sum_{\substack{r\in I_h\\j\in J_{hr}}} y_{hr},
    \qquad
    \sum_h o_{hj} \leq 1.
\end{equation}
Ownership applies to the whole shot even when only part of it is used.
For intervals from the same source that could still overlap, binary
ordering variables and big-\(M\) constraints keep the clips separate.
We also enforce any specified source counts for individual units or
groups of units, and any minimum durations for required sources.

\subsection{Solver and Assignment Reconstruction}

We solve the assignment with HiGHS through SciPy's MILP interface.
Within each source, we compact the selected intervals while preserving
their order, safe bounds, and verification frames. We then split them
at shot boundaries to record their source and shot identifiers and
assemble the clips in narration order.

Confirmed infeasibility triggers the diagnostic procedure in the
next section. If the solver returns no accepted solution and
infeasibility is unconfirmed, the controller may try a deterministic
fallback that checks for source overlap and shot reuse.

\section{Capacity Feedback and Controlled Revision}
\label{app:capacity-feedback}

Video Harness uses local capacity estimates and global assignment
diagnostics to guide targeted footage retrieval and targeted script
revision.

\subsection{Local Capacity Screening}

Before global assignment, we estimate the footage available to each
visual unit \(b_h\). Let \(\mathcal{G}_h\) contain the maximal runs
of adjacent admitted shots from the same source. We compute each
run's usable duration \(c_{hg}\) using the boundary margins and
duration cap defined in Appendix~\ref{app:global-assignment}.
The estimated capacity is:
\begin{equation}
    \widehat{C}_h =
    \sum_{g\in\mathcal{G}_h}
    \mathbb{I}[c_{hg} \geq d_{\min}]\,c_{hg}.
\end{equation}
For a measured narration duration \(d_h\), the estimated shortfall is:
\begin{equation}
    \Delta_h = \max(0, d_h-\widehat{C}_h).
\end{equation}

We also check whether enough sources are available and whether the
required clips fit within \(d_h\) at their minimum duration. These
checks flag potential shortages: they consider maximal runs and count
shared footage separately for each unit. A potential shortage can
trigger candidate expansion, additional footage retrieval, or targeted
script revision before global assignment.

\subsection{Global Conflict Diagnosis}

If local checks pass but the solver reports infeasibility, we solve
the same constraints again with a zero objective and presolve disabled.
If both runs report infeasibility, we diagnose the conflict. An
inconclusive result, such as a time limit, may instead trigger fallback
allocation. Any gaps in that allocation are recorded separately.

To identify the affected units, we introduce a binary variable
\(\xi_h\) for each visual unit and replace its duration constraint with:
\begin{equation}
    \sum_{r\in I_h}\ell_{hr} + d_h\xi_h = d_h.
\end{equation}
Setting \(\xi_h=1\) removes the unit's duration requirement and
relaxes its minimum source count. Shot ownership, source bounds,
and global source requirements still apply. The objective penalizes
removed units, with smaller terms for duration and position.

For each unit selected by an optimal diagnostic solution, we restore
its full narration requirement and add hypothetical footage from an
independent source under the same clip-duration limits. A feasible
assignment gives one estimate of how much additional footage and how
many distinct shots would restore coverage. We use shot ownership and
overlapping candidate intervals from the same source to identify units
that compete for footage.

The conflict report links the affected script spans to competing shots
and sources. It records the estimated footage needs, local capacity
estimates, and solver and validation outcomes. If diagnosis or
validation is inconclusive, the controller records the case as
unresolved.

\subsection{Targeted Footage Retrieval}

The controller expands the affected candidate pools within the shot
catalog using \(K=10,15,20\), then checks other analyzed sources for
footage matching the required entities. For a verified global conflict
with acquisition budget remaining, it retrieves additional footage.
The search query uses the affected narration, visual requirements,
competing shots and sources, and estimated duration shortfall.
New shots undergo the same semantic and visual checks before
reassignment.

\subsection{Controlled Revision}

For a local shortage, or a verified conflict after the footage
acquisition budget is exhausted, an approved repair may update source
routes, visual-support roles, or the affected narration.

When a passage needs revision, the controller maps the affected visual
unit to its narration unit. Only the writer edits the passage, using
the hash of the parent text to identify the version to revise.
Unaffected text is preserved, and factual changes are checked against
verified evidence. After a wording change, the system updates the
affected units and regenerates speech and timing before candidate
validation and reassignment. Visual-only changes can reuse the
existing narration audio.

Footage acquisition and script revision have separate budgets. The
controller records repair decisions and retains the diagnostics for
cases that remain unresolved when a budget is exhausted.

\section{Search, Footage, and Retry Budgets}
\label{app:generation-budgets}
Table~\ref{tab:generation-budgets} reports the
search, footage retrieval, and retry budgets used for the
latest production batch.
\begin{table}[!htbp]
    \centering
    \caption{\textbf{Video Harness generation budgets.}
    Values describe configured limits, not actual consumption.}
    \label{tab:generation-budgets}
    \footnotesize
    \setlength{\tabcolsep}{3pt}
    \renewcommand{\arraystretch}{1.05}
    \begin{tabular*}{\columnwidth}{
        @{\extracolsep{\fill}}
        p{0.57\columnwidth}
        p{0.37\columnwidth}@{}}
        \toprule
        \textbf{Configuration} & \textbf{Setting} \\
        \midrule
        \multicolumn{2}{@{}l}{\textit{Agentic search}} \\
        Maximum controller turns & 18 \\
        Maximum search queries & 24 \\
        Maximum documents read & 24 \\
        \midrule
        \multicolumn{2}{@{}l}{\textit{Footage retrieval and analysis}} \\
        Initial source-video acquisition cap & 12 videos \\
        Search results per query & 8 \\
        Adaptive shot budget per source & 24--240 shots \\
        Semantic-retrieval budget per source & 250 calls \\
        \midrule
        \multicolumn{2}{@{}l}{\textit{Targeted repair and retries}} \\
        Maximum material-feedback rounds & 5 \\
        Semantic-backend attempts per request
            & 2 primary + 2 fallback \\
        \bottomrule
    \end{tabular*}
    \vspace{-0.5cm}
\end{table}

\section{Search Evaluation Weights}
\label{app:search-evaluation-weights}

\begin{table}[H]
    \centering
    \footnotesize
    \setlength{\tabcolsep}{4pt}
    \renewcommand{\arraystretch}{1.05}
    \caption{Weights for the paired Search ON/OFF comparison.}
    \label{tab:search-evaluation-weights}

    \begin{tabular*}{\linewidth}{@{\extracolsep{\fill}}lr@{}}
        \toprule
        \textbf{Criterion} & \textbf{Weight} \\
        \midrule
        Prompt completion                      & 10\% \\
        Factual accuracy                       & 10\% \\
        Key-fact traceability                  & 20\% \\
        Data adequacy                          & 15\% \\
        Freshness                              & 15\% \\
        Narrative clarity                      &  5\% \\
        Visual support                         & 17\% \\
        Audio, subtitles, and playback quality &  8\% \\
        Total                                  & 100\% \\
        \bottomrule
    \end{tabular*}
    \vspace{-0.3cm}
\end{table}

\section{Human Evaluation of Video Quality}
\begin{table}[H]
    \centering
    \footnotesize
    \setlength{\tabcolsep}{2pt}
    \renewcommand{\arraystretch}{1.15}
    \caption{\textbf{Overall video quality as rated by human evaluators.}
H1, H2, and H3 denote the three evaluators.}
    \label{tab:mixed-quality}
    \begin{tabular*}{\columnwidth}{@{\extracolsep{\fill}}lcccc@{}}
        \toprule
        \textbf{Method} & \textbf{H1} & \textbf{H2}
                        & \textbf{H3} & \textbf{Mean} \\
        \midrule
        Crayotter
            & $3.55 \pm 0.59$ & $3.30 \pm 0.98$
            & $2.75 \pm 0.80$ & $3.20 \pm 0.62$ \\
        ChatCut
            & $3.52 \pm 0.59$ & $3.12 \pm 0.91$
            & $2.83 \pm 0.83$ & $3.16 \pm 0.59$ \\
        CapCut-Mate
            & $3.48 \pm 0.65$ & $3.40 \pm 0.75$
            & $2.74 \pm 0.88$ & $3.20 \pm 0.54$ \\
        CutClaw
            & $2.09 \pm 0.67$ & $2.76 \pm 0.83$
            & $2.07 \pm 0.85$ & $2.31 \pm 0.57$ \\
        \textbf{VideoHarness}
            & $\mathbf{4.57 \pm 0.21}$ & $\mathbf{3.97 \pm 0.67}$
            & $\mathbf{3.50 \pm 0.87}$ & $\mathbf{4.01 \pm 0.47}$ \\
        \bottomrule
    \end{tabular*}
    \vspace{-4pt}
\end{table}

%% file: sec/X_suppl.tex
% Requires graphicx and caption (the CVPR template loads caption).
% Replace the old supplementary title command AND the entire gallery block.
% Do not retain \maketitlesupplementary before this code.
% Keep the same 120 JPG files beside the main .tex file.

\newcommand{\GalleryFrame}[2]{%
  \begin{minipage}[t]{0.23\linewidth}
    \vspace{0pt}%
    \includegraphics[width=\linewidth]{#1}\par
    \vspace{2pt}%
    {\centering\fontsize{7}{8.5}\selectfont\ttfamily #2\par}%
  \end{minipage}%
}

\newcommand{\GalleryExample}[4]{%
  \par\noindent\makebox[\linewidth][c]{%
  \begin{minipage}{0.86\linewidth}
    \noindent #3\par
    \captionsetup{skip=3pt,hypcap=false}%
    \captionof{figure}{#1. Duration: #2.}%
    \label{#4}%
  \end{minipage}%
  }\par\addvspace{12pt}%
}

\clearpage
\onecolumn
\raggedbottom
\setcounter{page}{1}
\setlength{\parindent}{0pt}
\setlength{\parskip}{0pt}

\noindent\begin{minipage}{\textwidth}
\centering
\begin{center}
  {\Large\bfseries\thetitle\par}
  \vspace{0.5em}
  {\Large Supplementary Material\par}
\end{center}
\vspace{4pt}
\begin{minipage}{0.86\linewidth}
\section{Examples}
\label{sec:qualitative-video-gallery}
\raggedright\small
Figures~\ref{fig:video-microchip_manufacturing}--\ref{fig:video-vaccines_and_eradication}
present 30 selected VideoHarness outputs, with one figure per video.
Given the 200\,MB limit on supplementary material, we include one full-length video, \textit{How Microchips Are Made}, alongside the gallery.

\par
\end{minipage}\par\vspace{10pt}

\GalleryExample{How Microchips Are Made}{3:50}{%
  \GalleryFrame{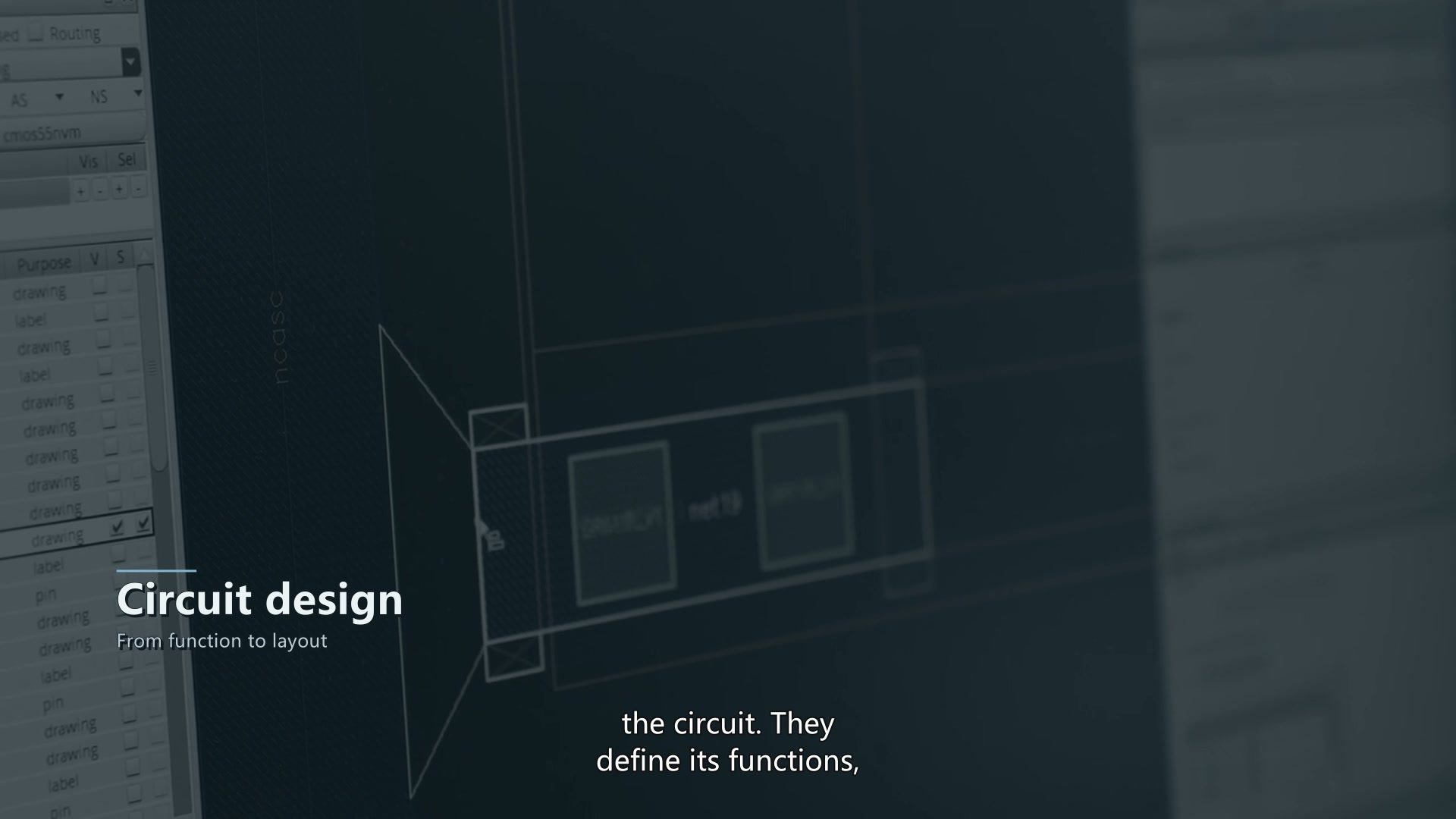}{00:24.0}\hfill%
  \GalleryFrame{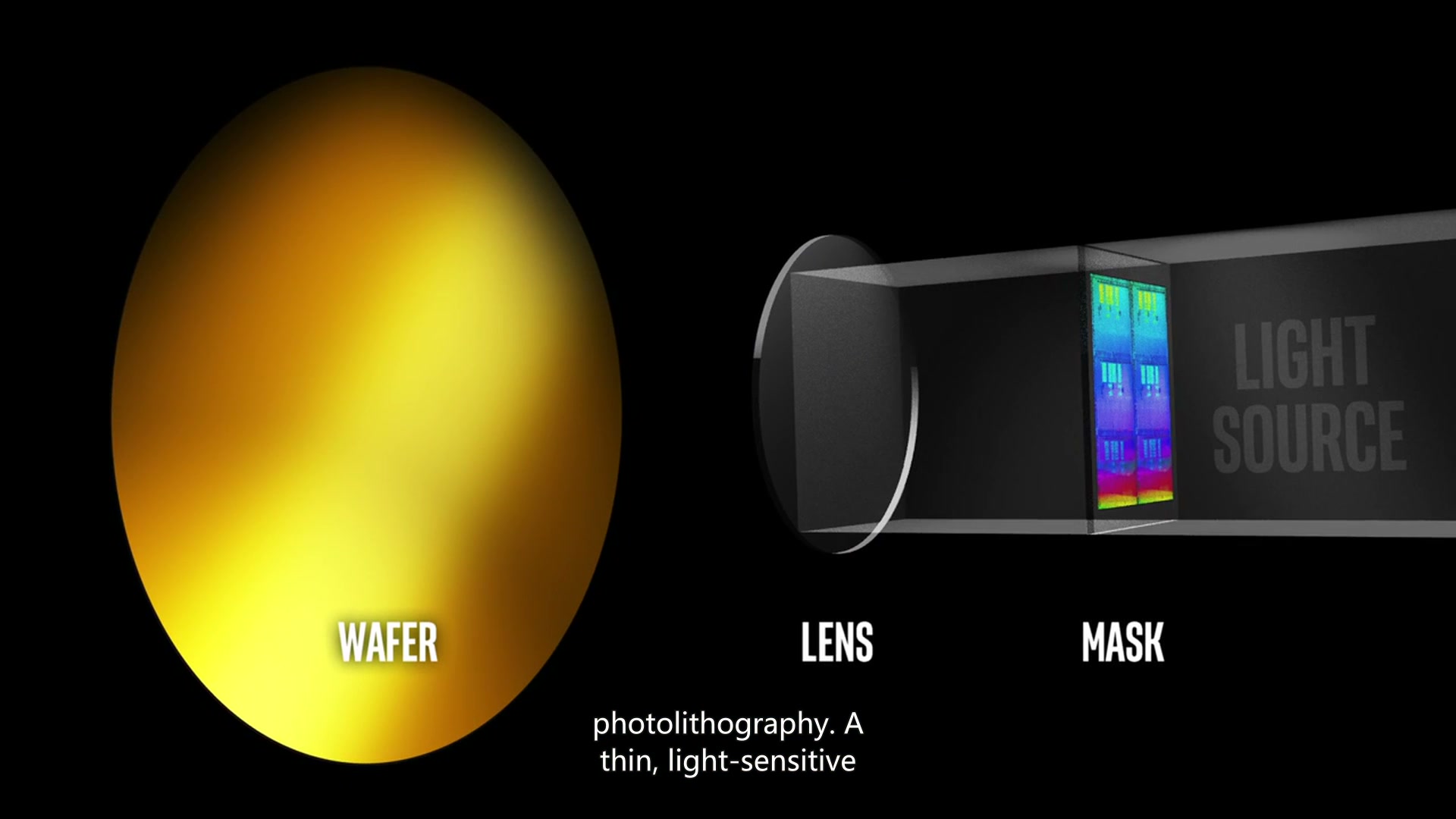}{01:21.5}\hfill%
  \GalleryFrame{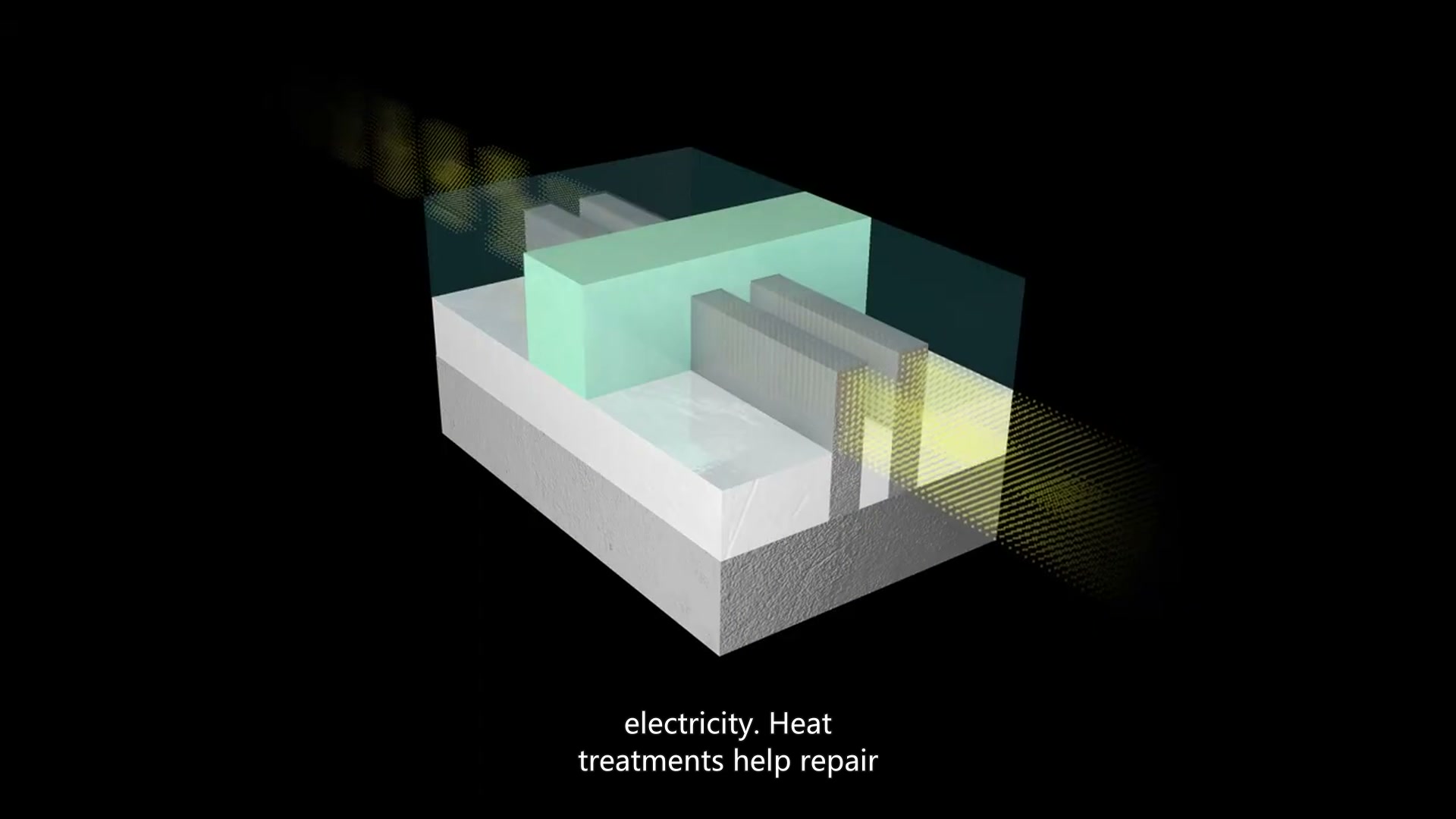}{01:59.9}\hfill%
  \GalleryFrame{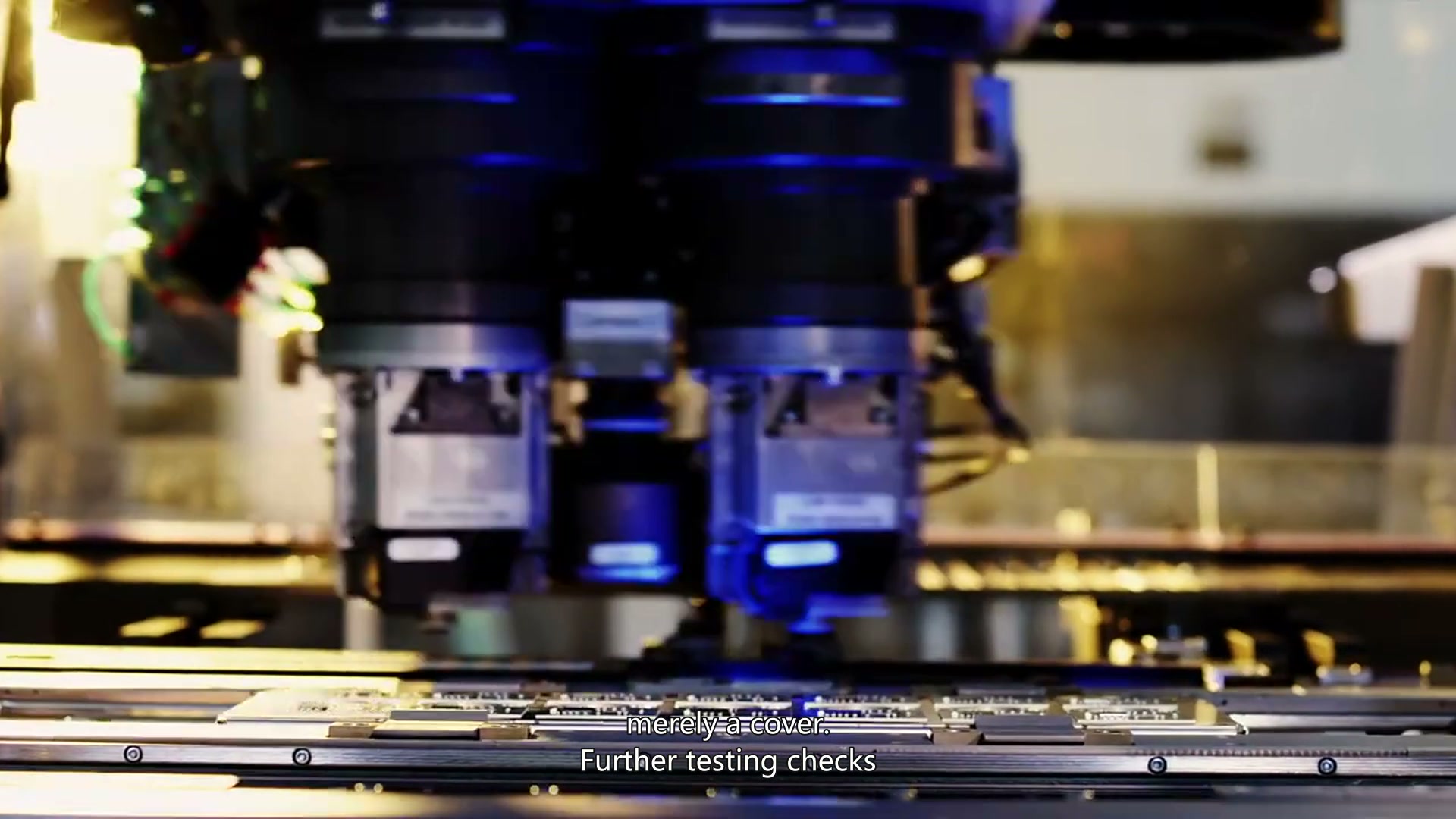}{03:26.3}%
}{fig:video-microchip_manufacturing}
\end{minipage}
\par\addvspace{12pt}

\GalleryExample{Why AI Is Straining Data-Center Power Supplies}{3:42}{%
  \GalleryFrame{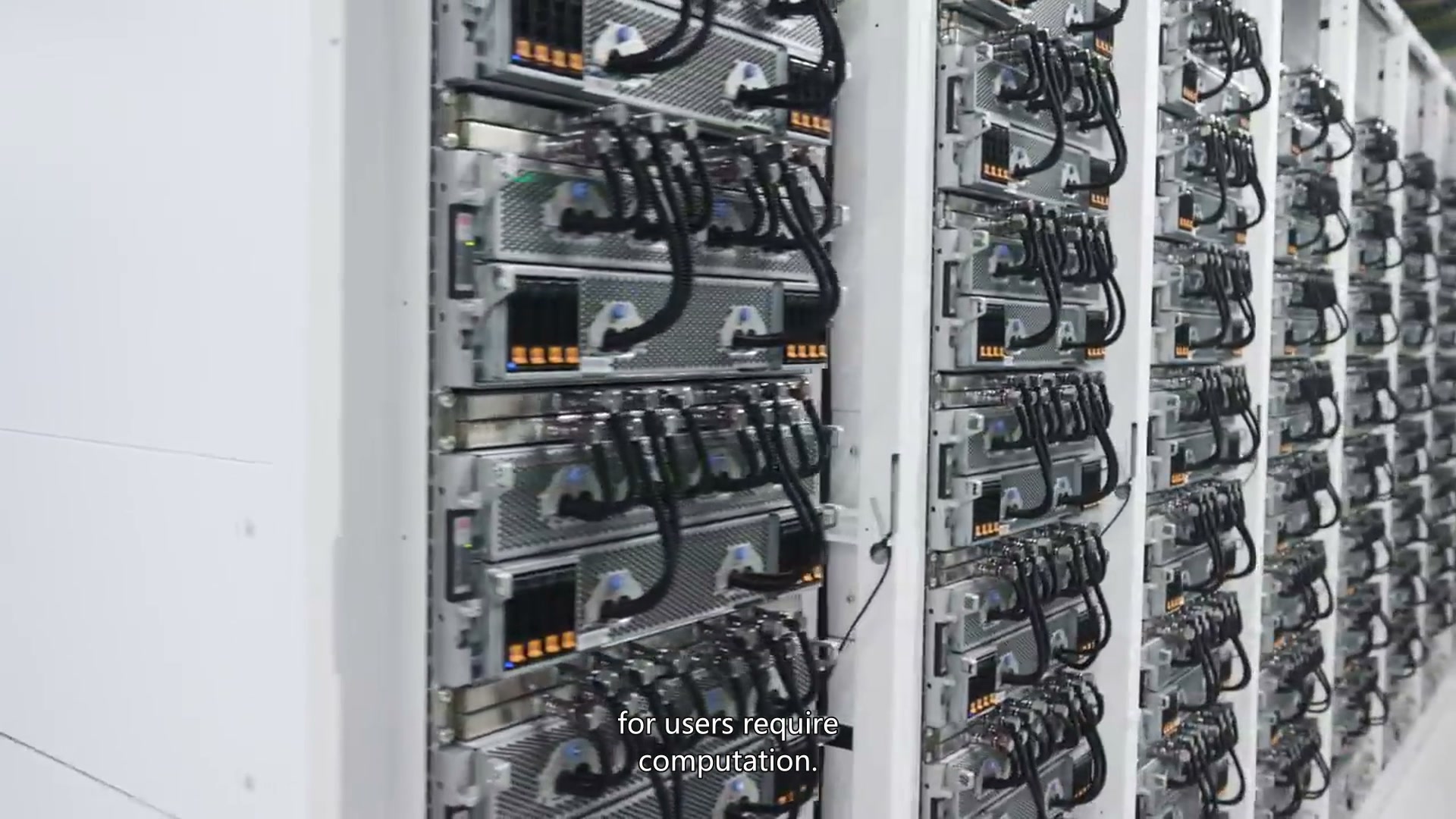}{00:32.4}\hfill%
  \GalleryFrame{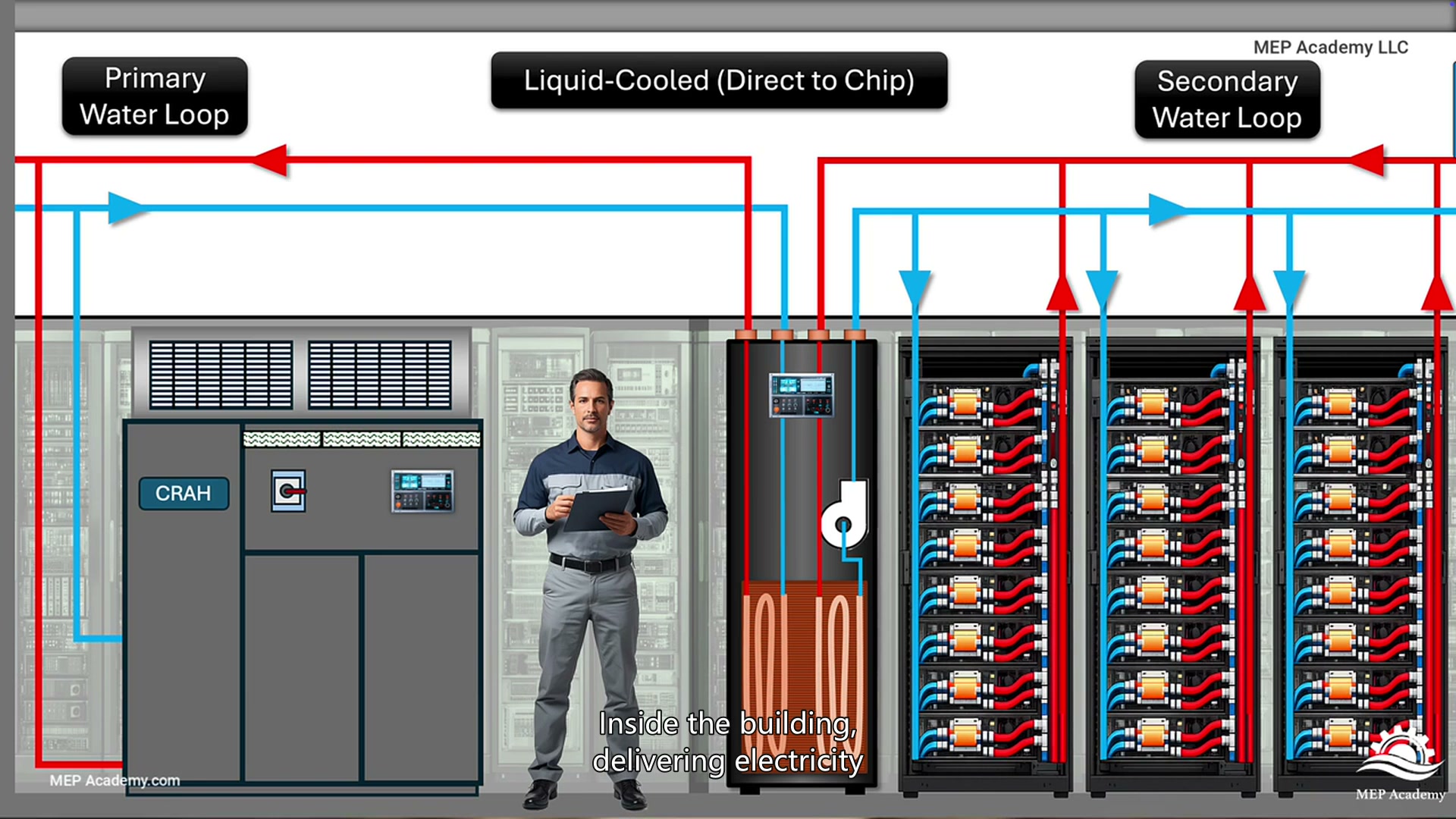}{01:28.0}\hfill%
  \GalleryFrame{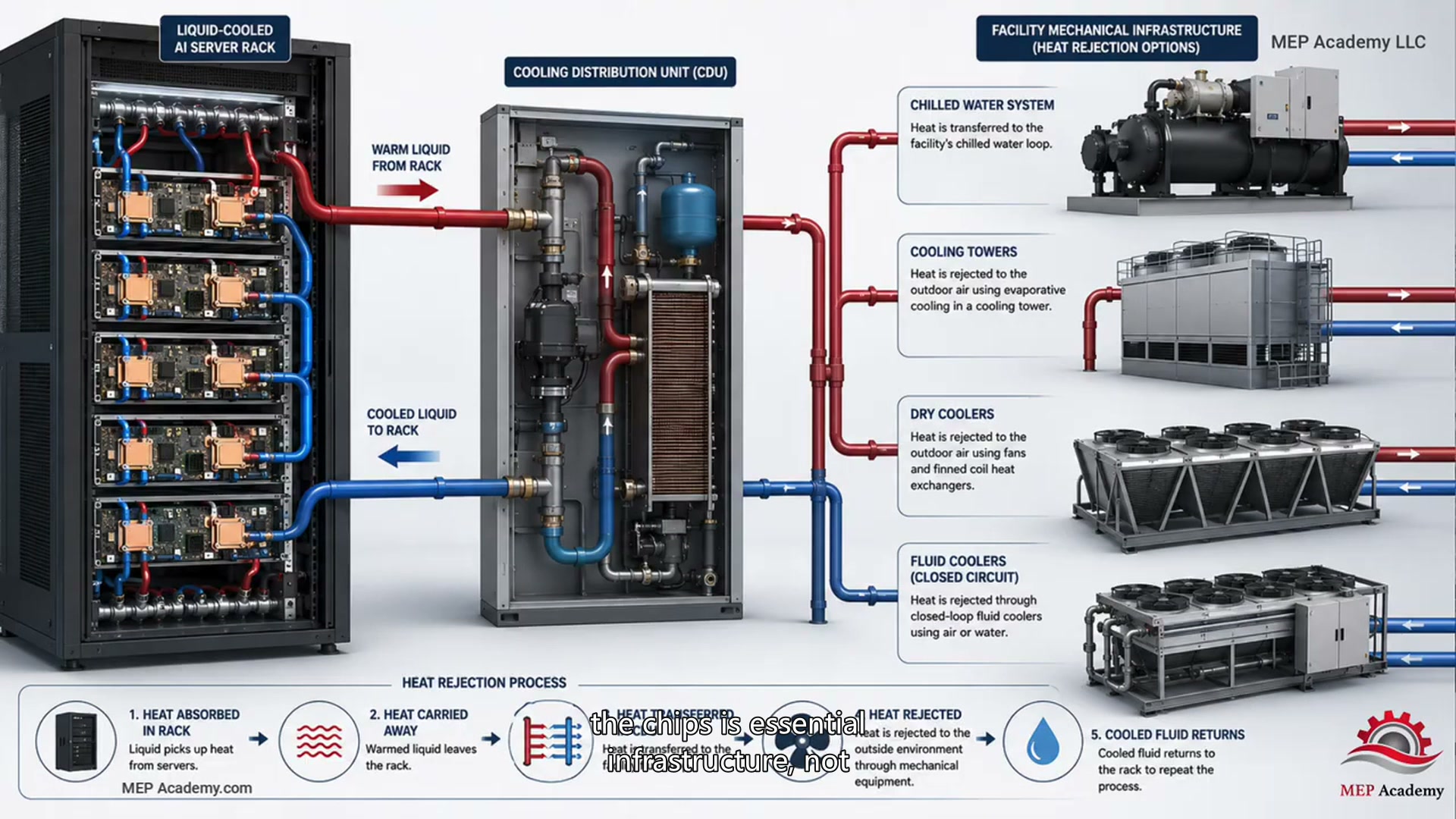}{02:23.5}\hfill%
  \GalleryFrame{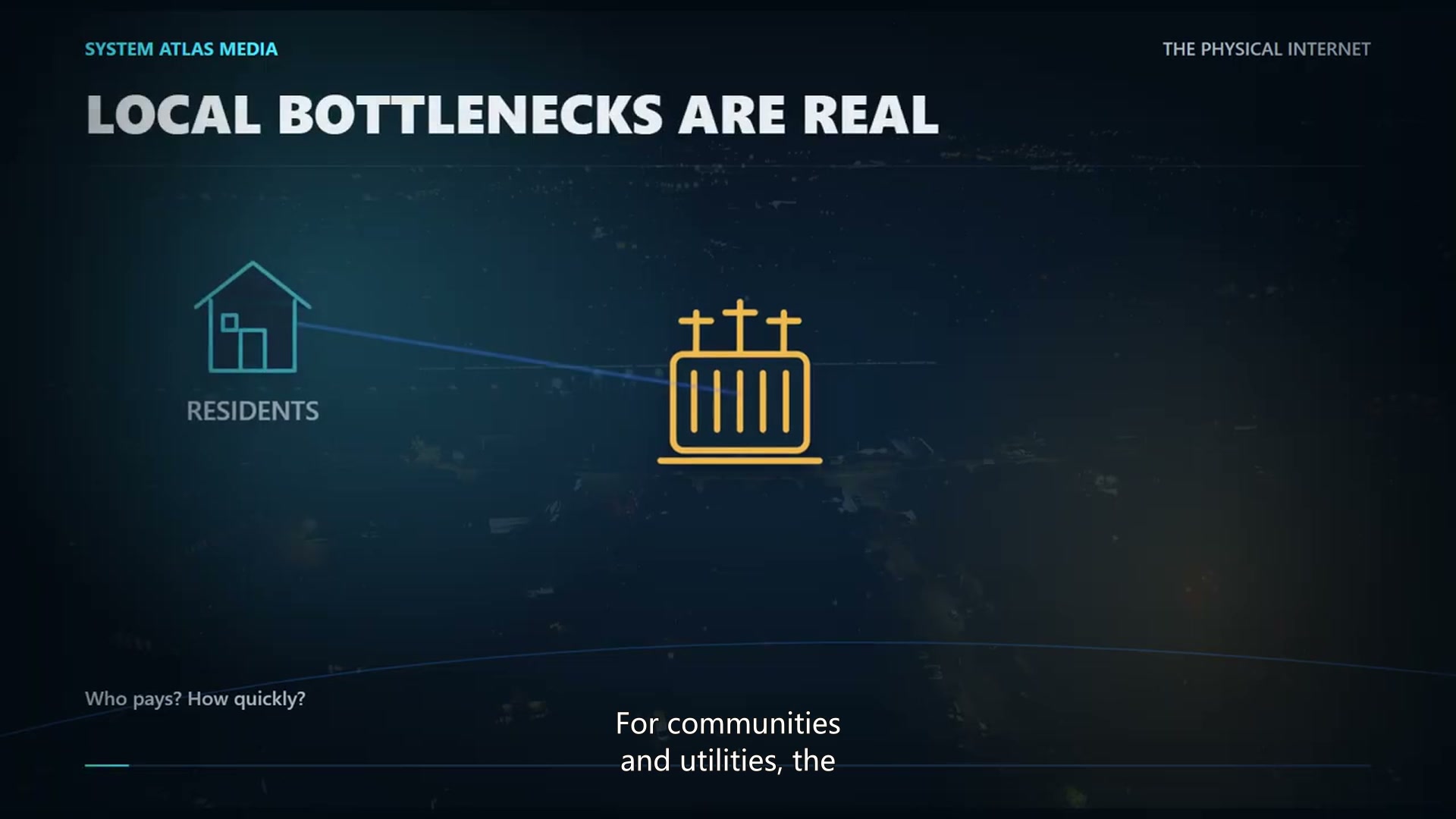}{03:09.9}%
}{fig:video-ai_data_center_power}
\GalleryExample{Three Days in New York City}{2:53}{%
  \GalleryFrame{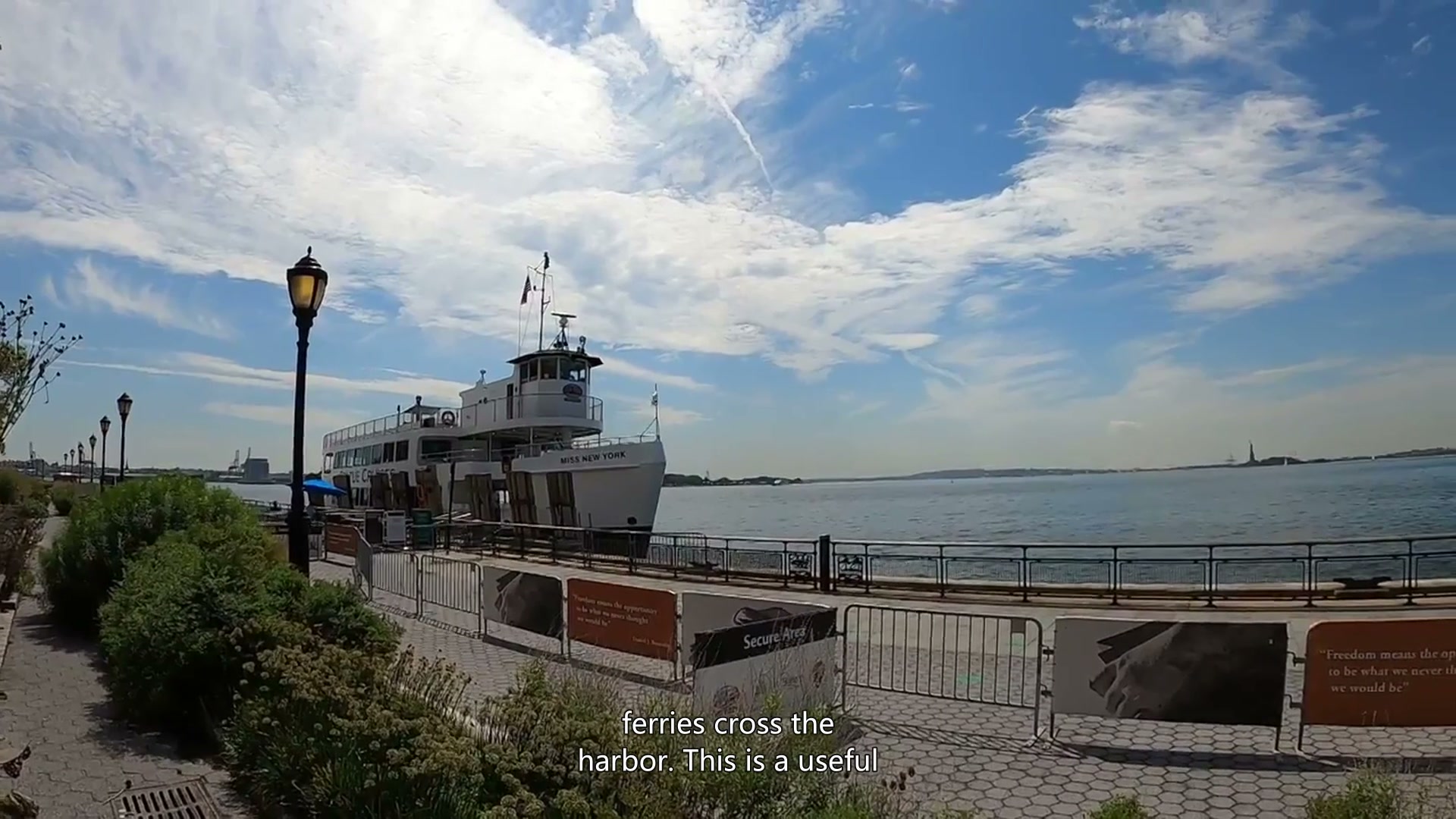}{00:25.2}\hfill%
  \GalleryFrame{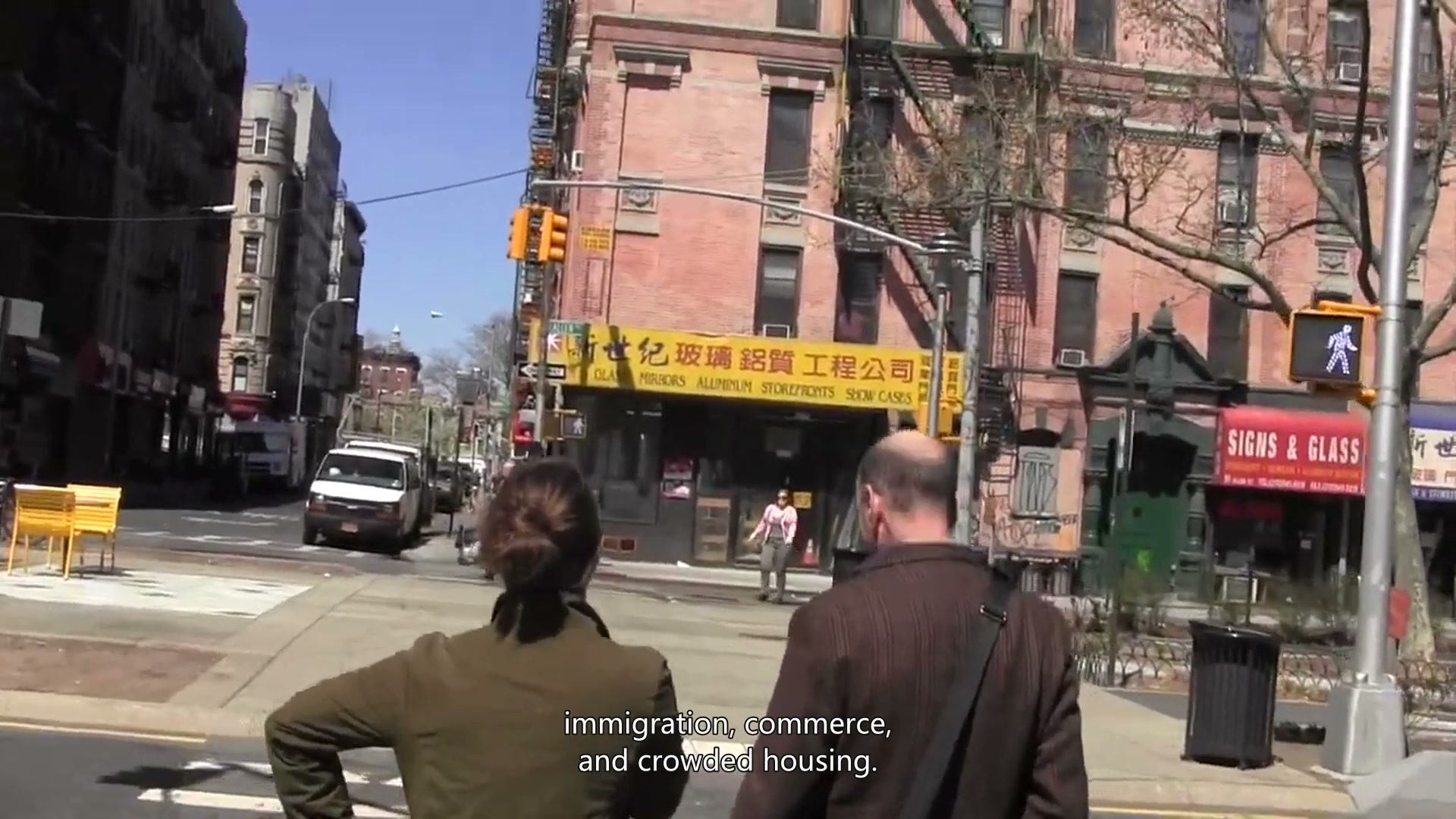}{00:54.1}\hfill%
  \GalleryFrame{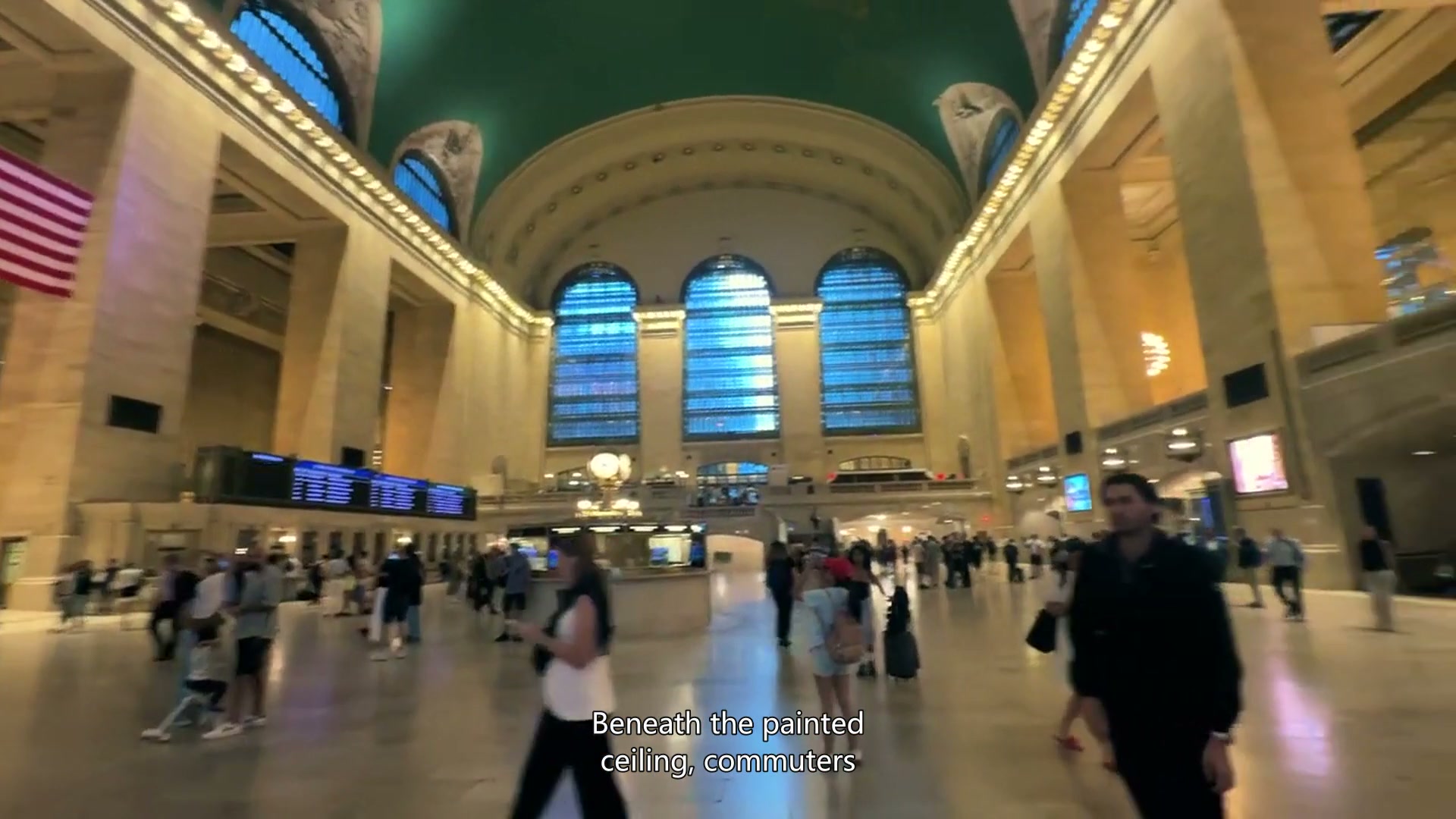}{01:22.9}\hfill%
  \GalleryFrame{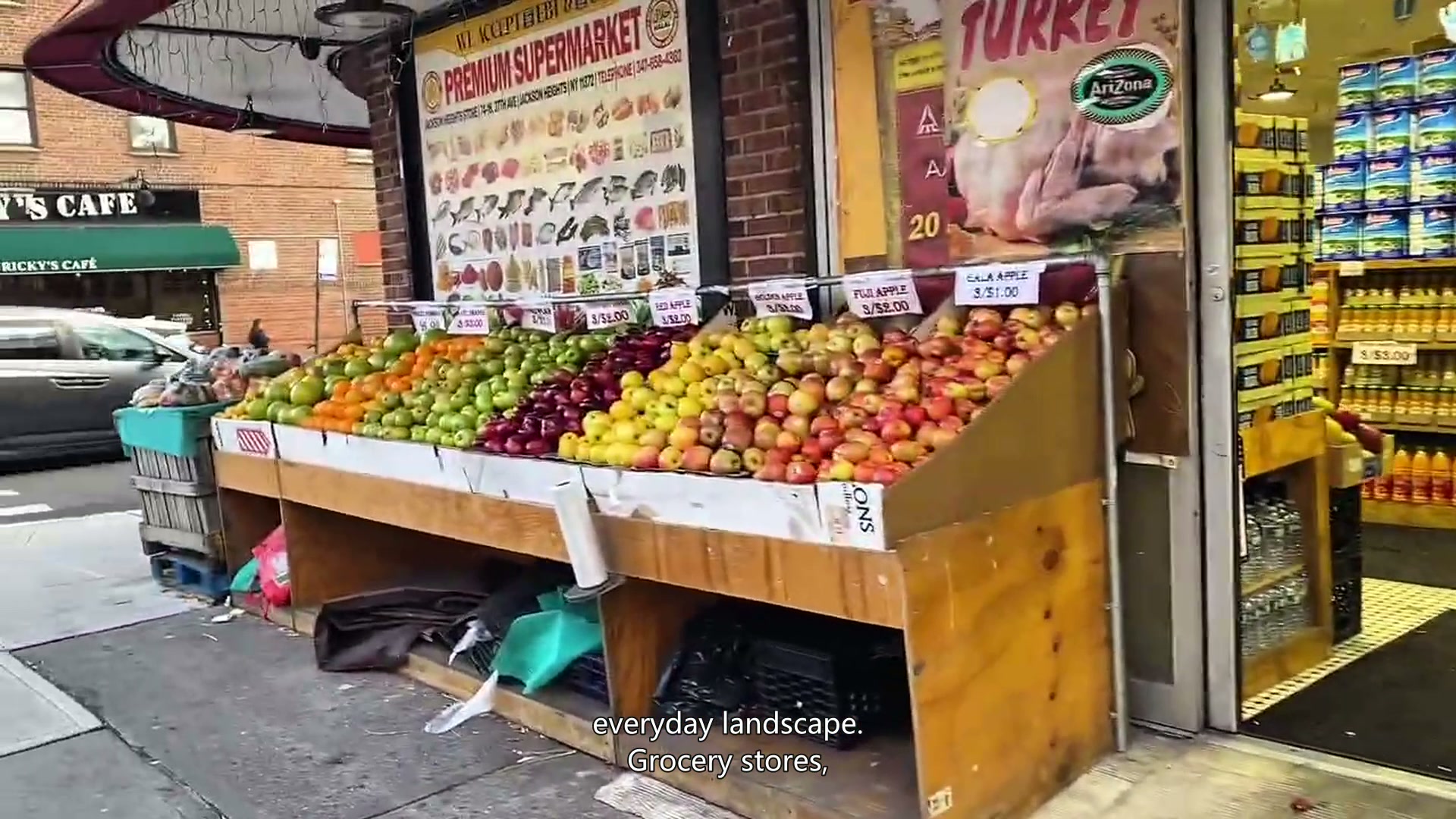}{02:20.6}%
}{fig:video-new_york_three_days}
\GalleryExample{Replacing a Bicycle Chain, Step by Step}{3:16}{%
  \GalleryFrame{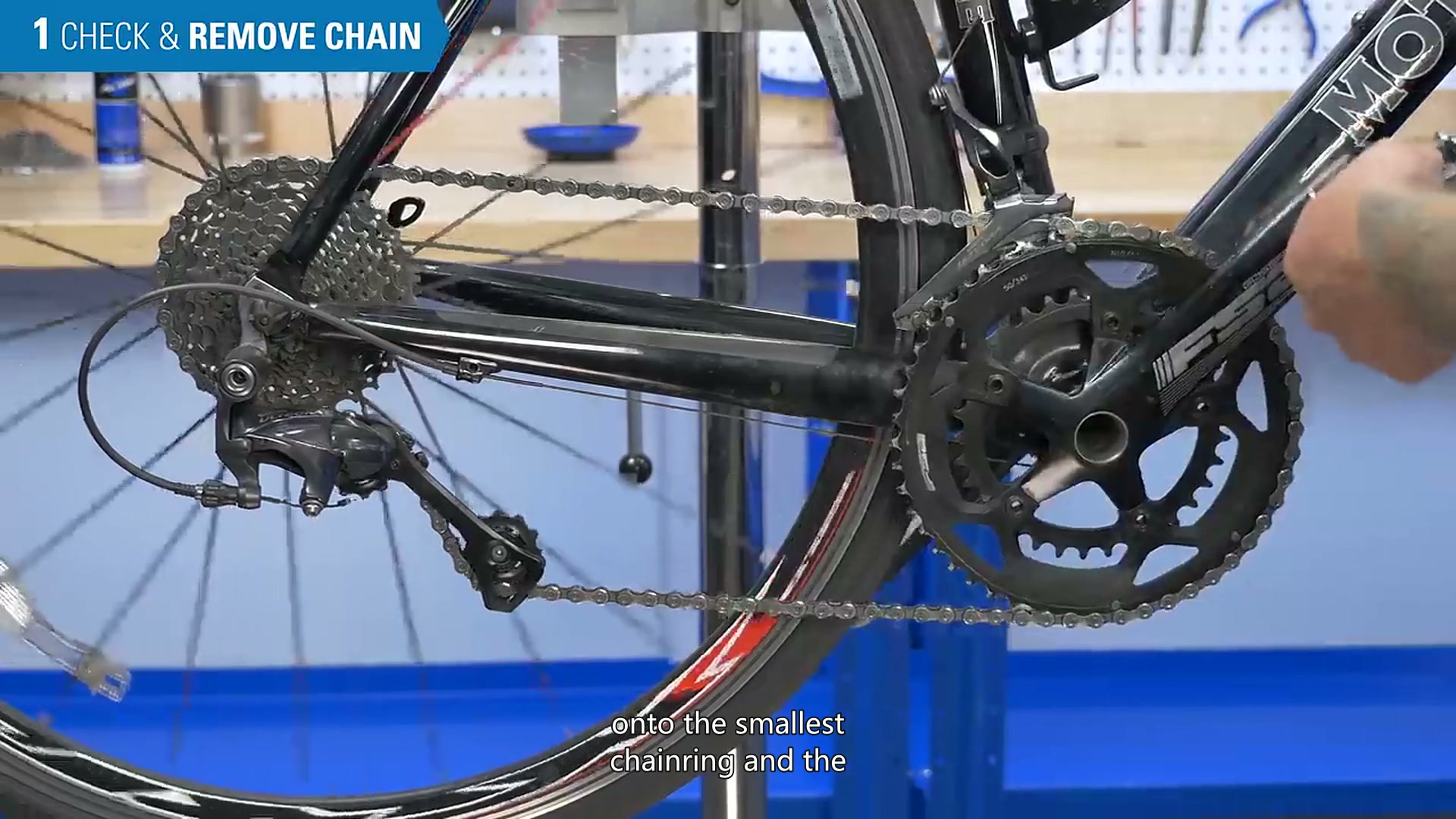}{00:28.6}\hfill%
  \GalleryFrame{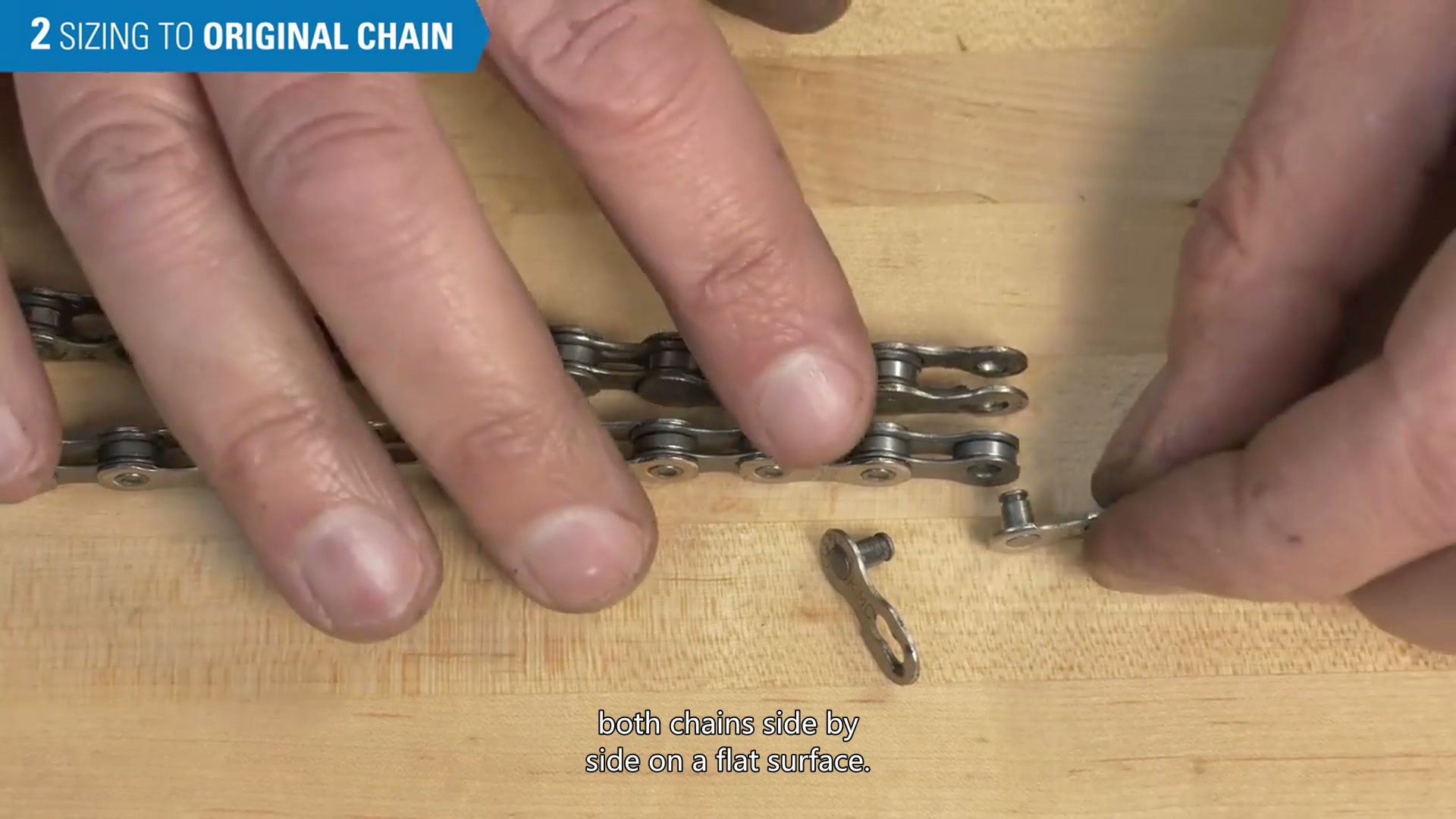}{01:17.6}\hfill%
  \GalleryFrame{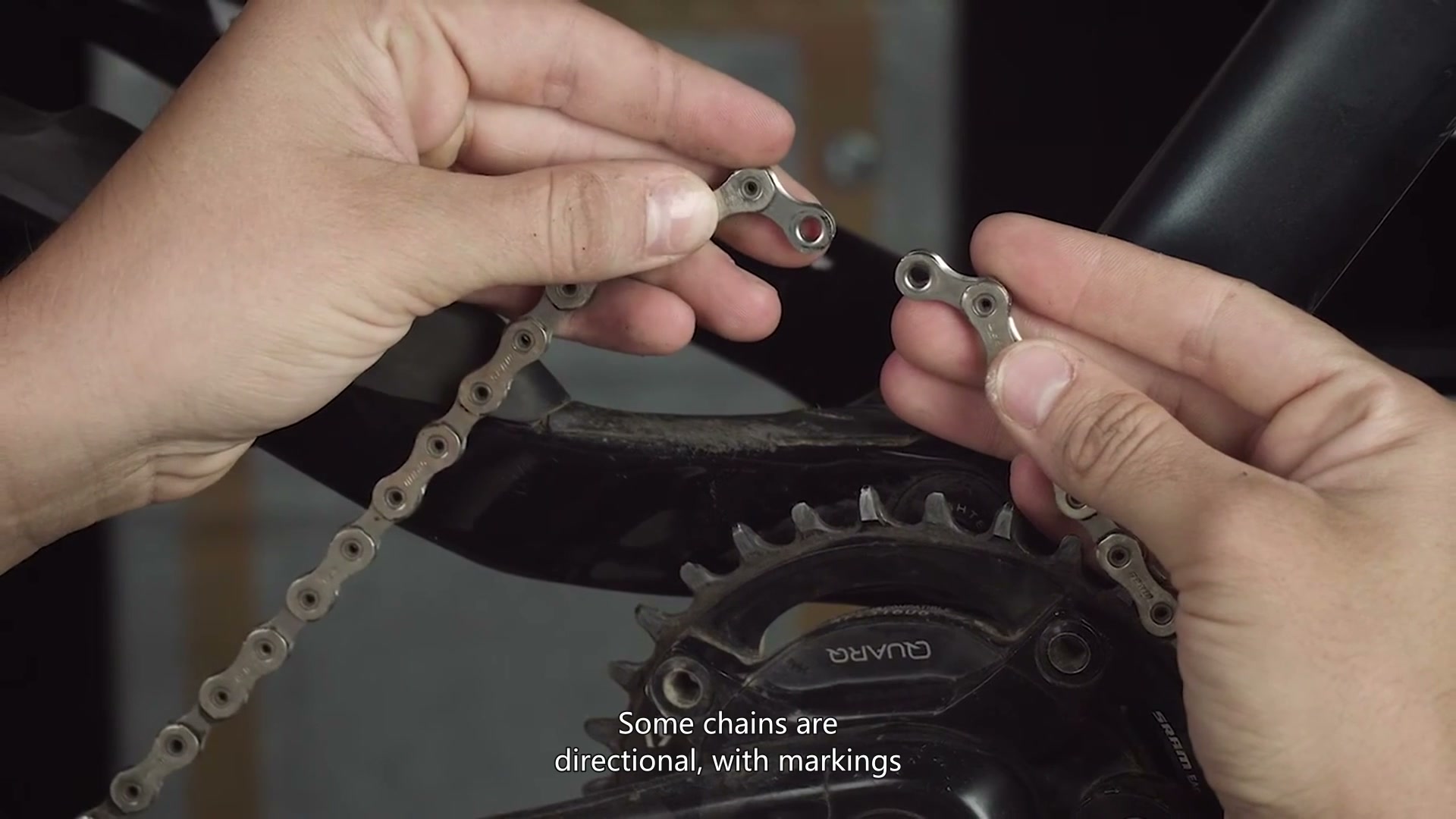}{01:58.4}\hfill%
  \GalleryFrame{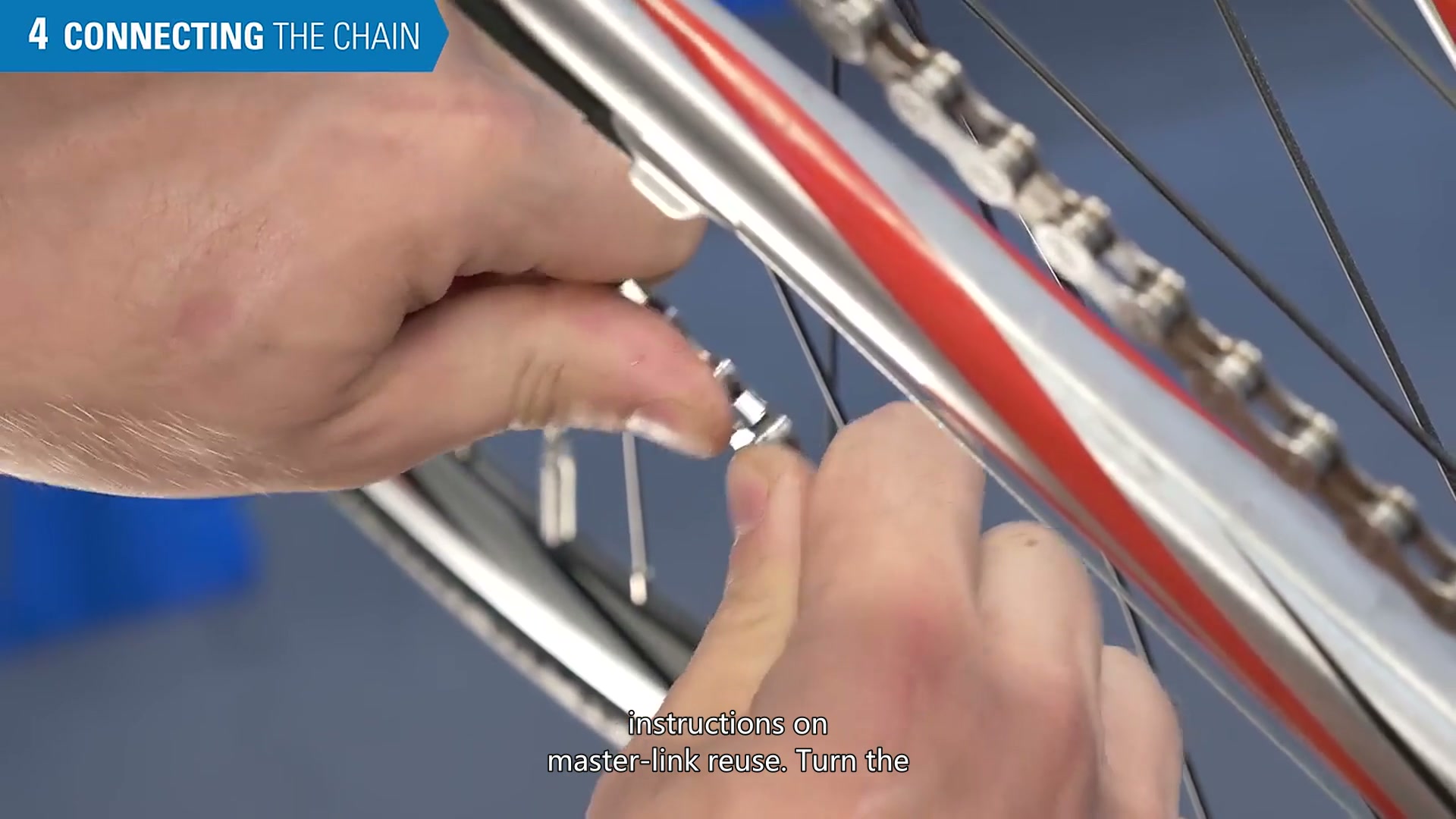}{02:39.3}%
}{fig:video-bicycle_chain_replacement}
\GalleryExample{What Does VAR Actually Do at the World Cup?}{3:36}{%
  \GalleryFrame{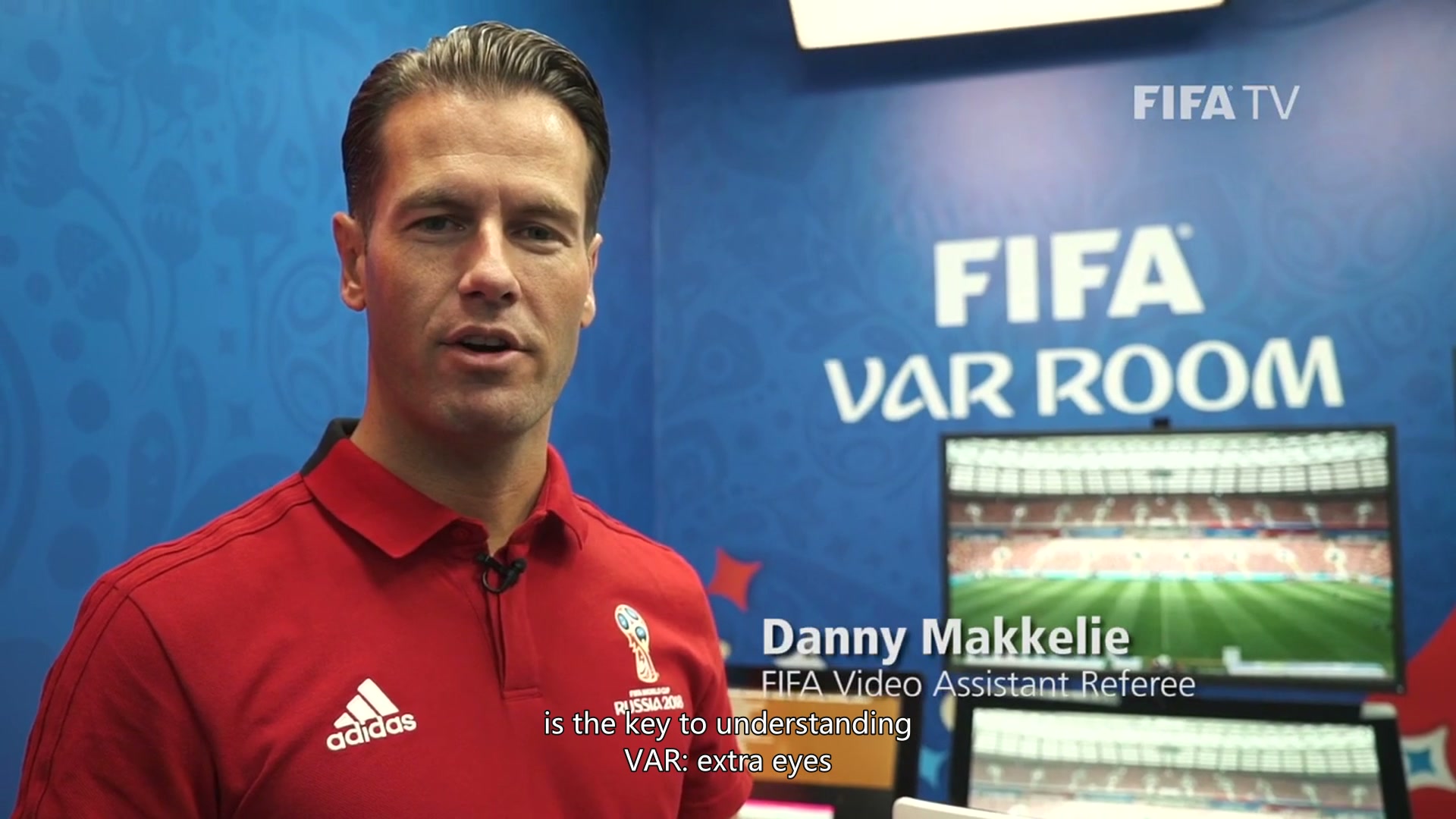}{00:22.5}\hfill%
  \GalleryFrame{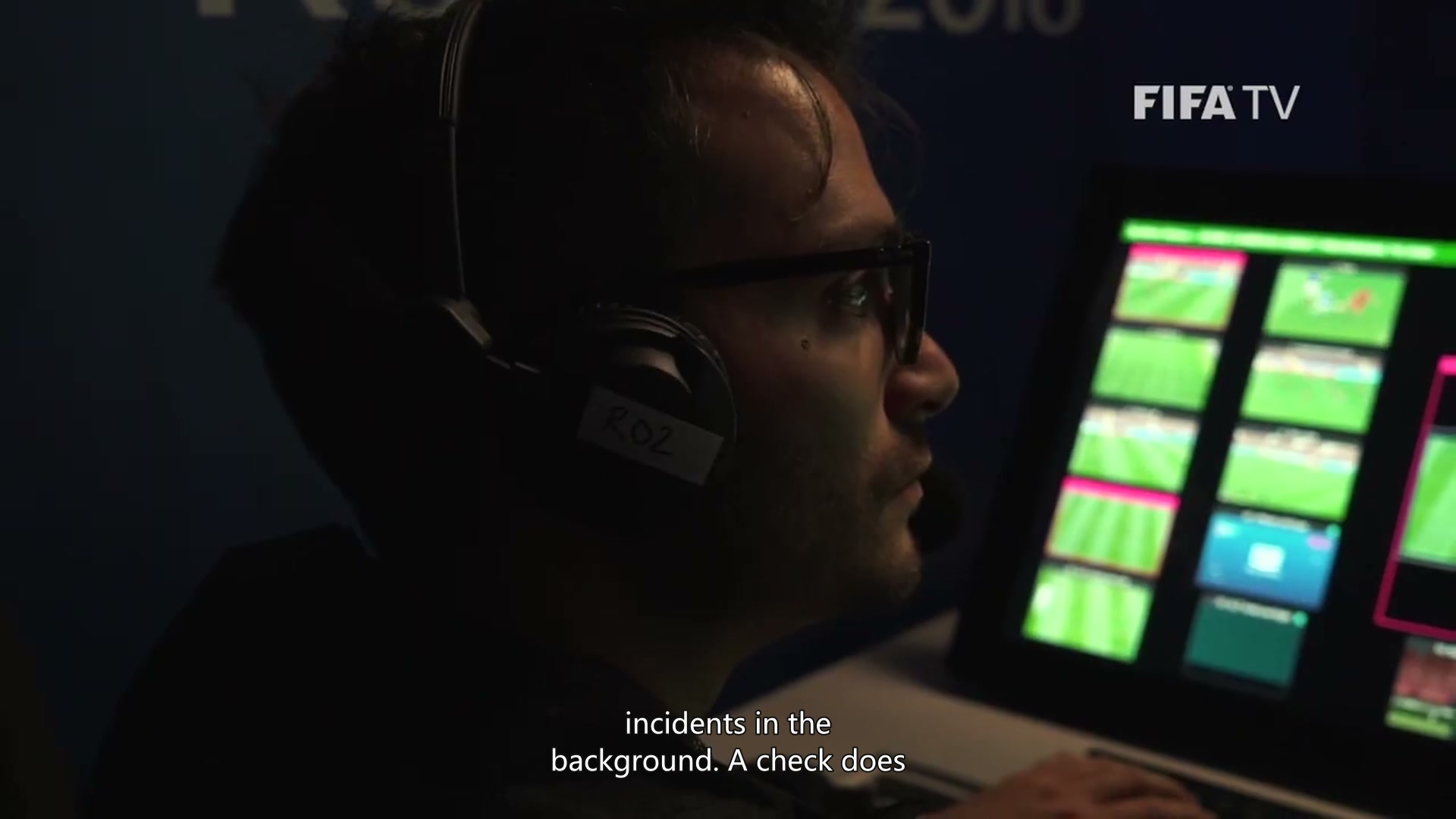}{01:16.6}\hfill%
  \GalleryFrame{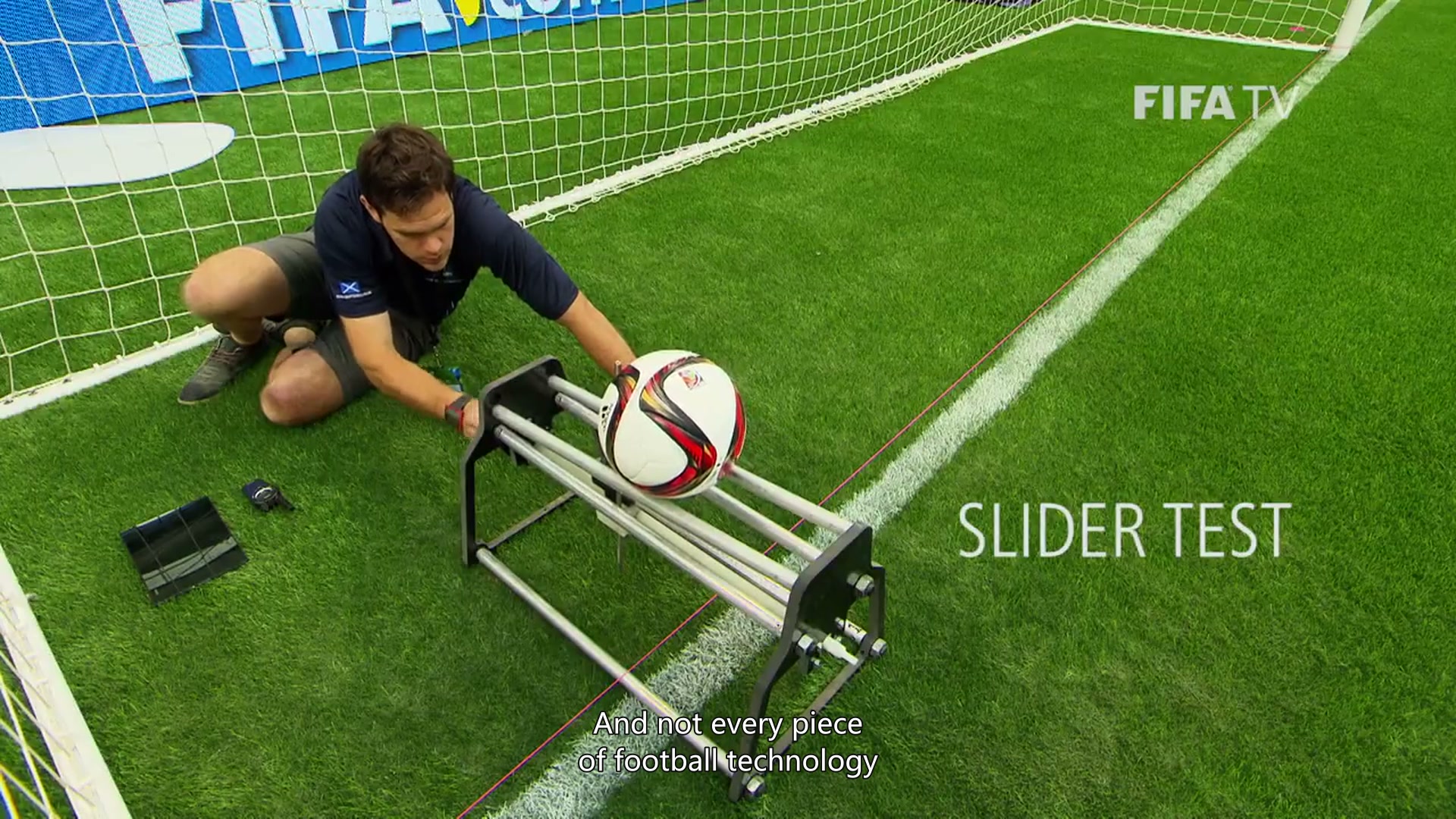}{02:10.7}\hfill%
  \GalleryFrame{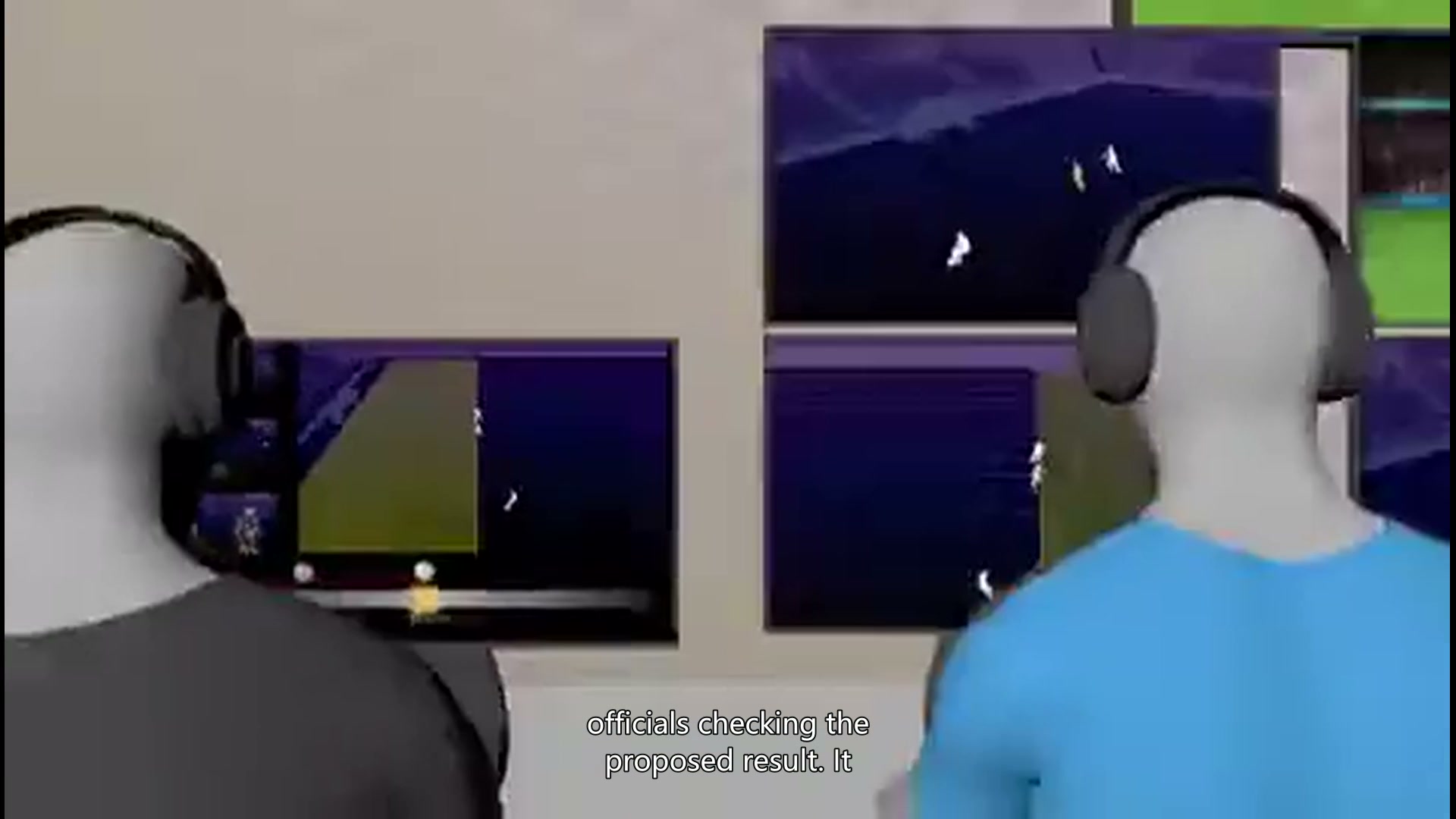}{03:04.8}%
}{fig:video-world_cup_var}
\GalleryExample{Comparing Video Game Genres}{3:23}{%
  \GalleryFrame{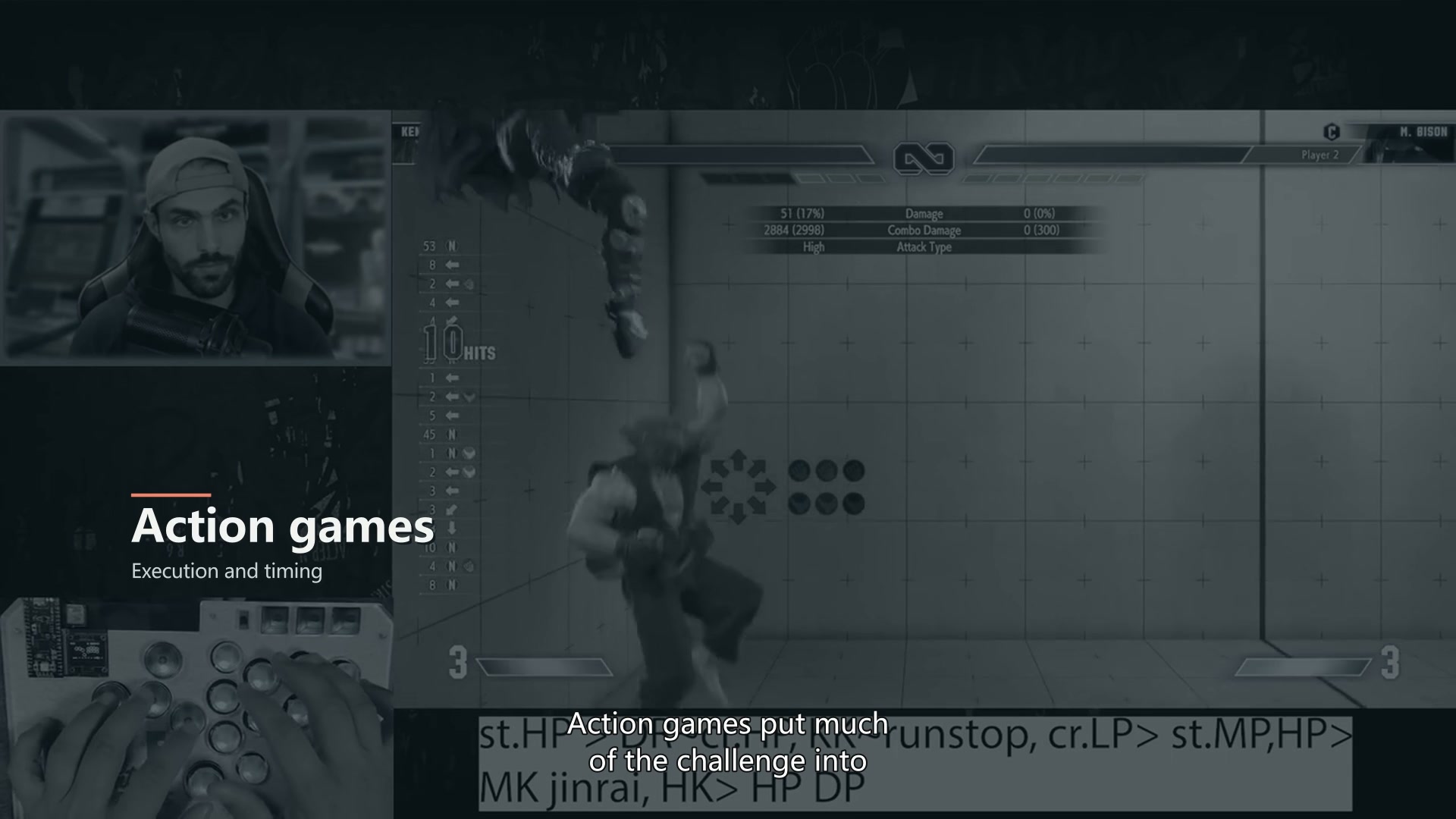}{00:21.2}\hfill%
  \GalleryFrame{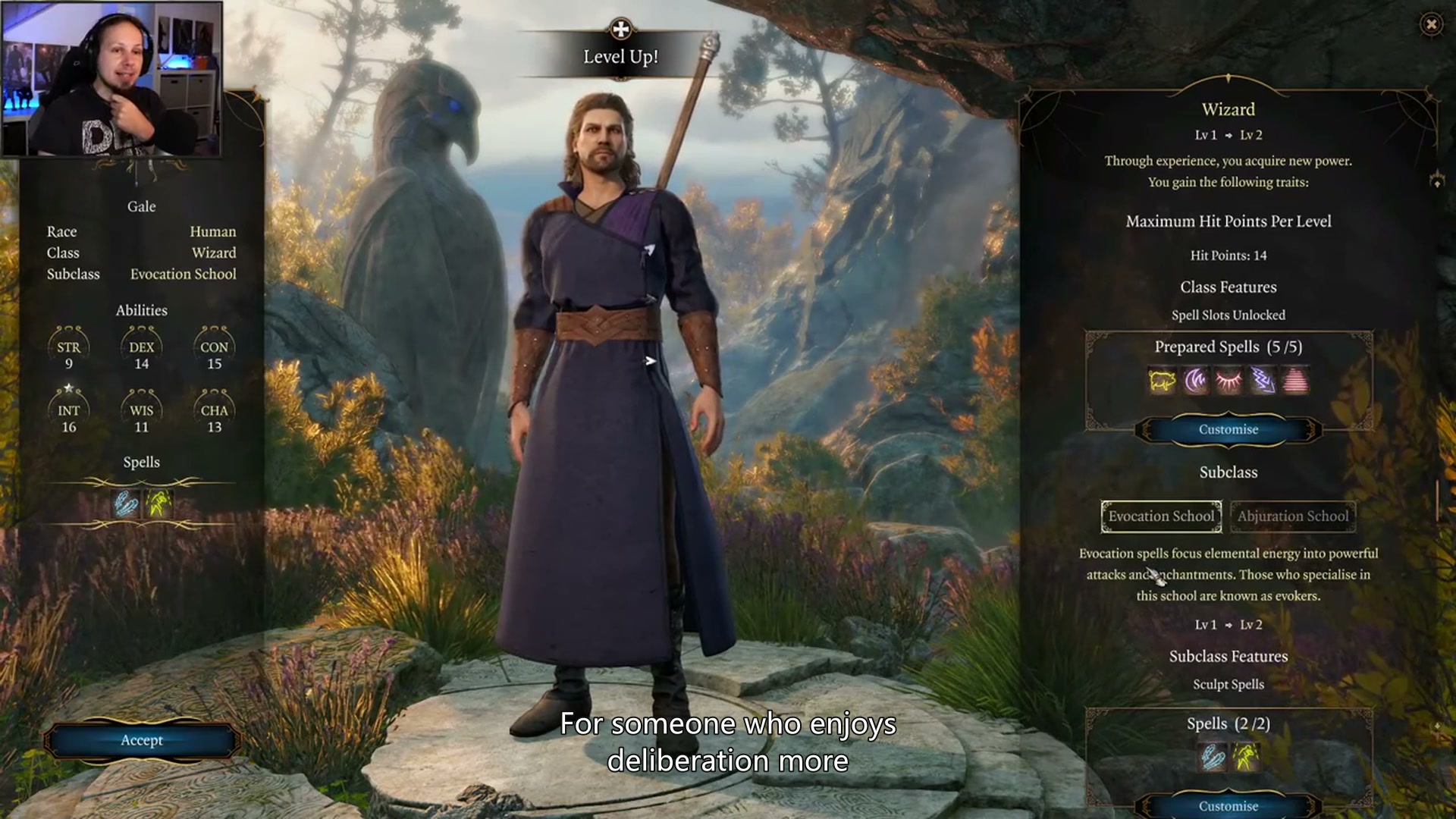}{01:12.1}\hfill%
  \GalleryFrame{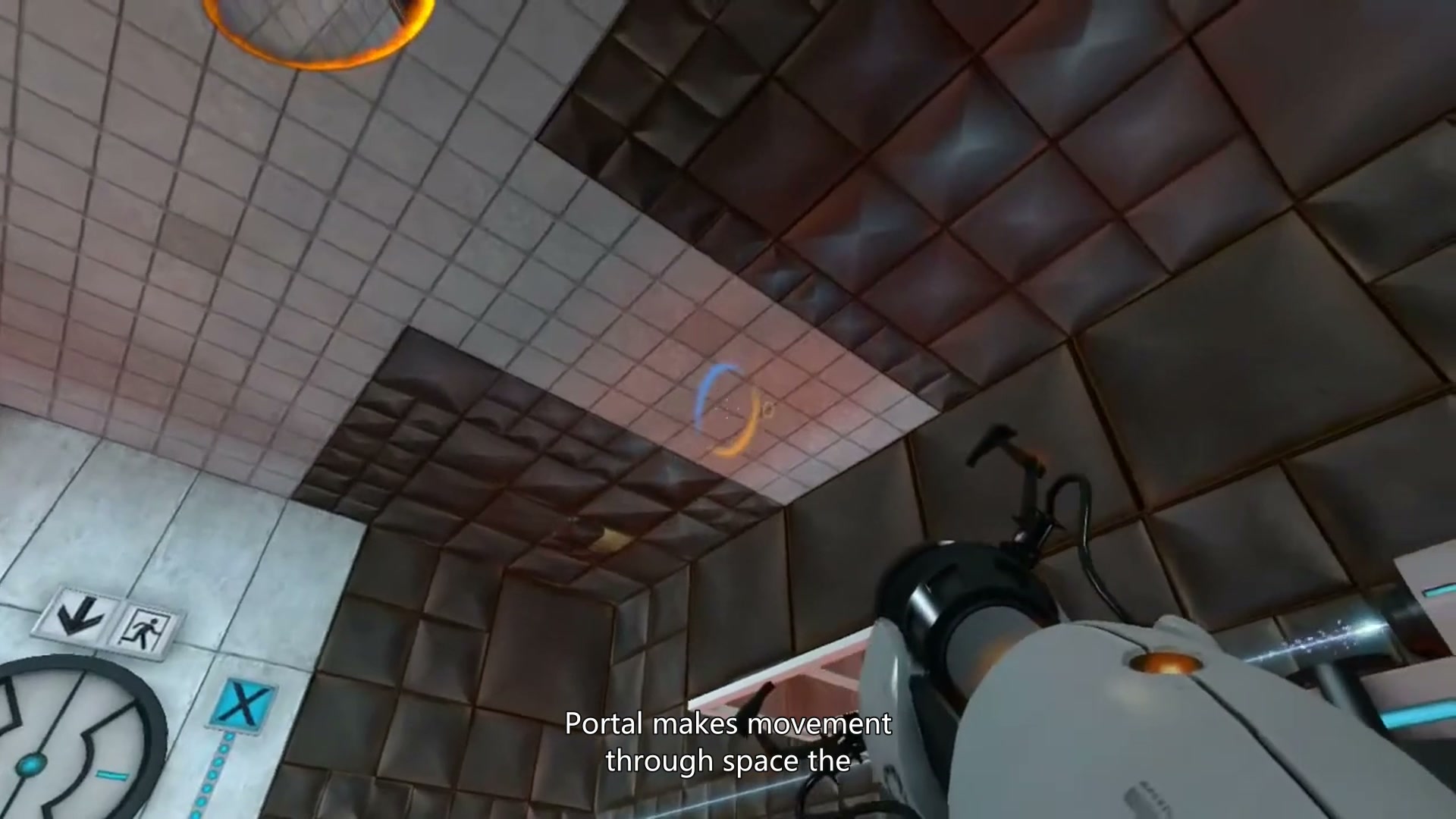}{02:03.0}\hfill%
  \GalleryFrame{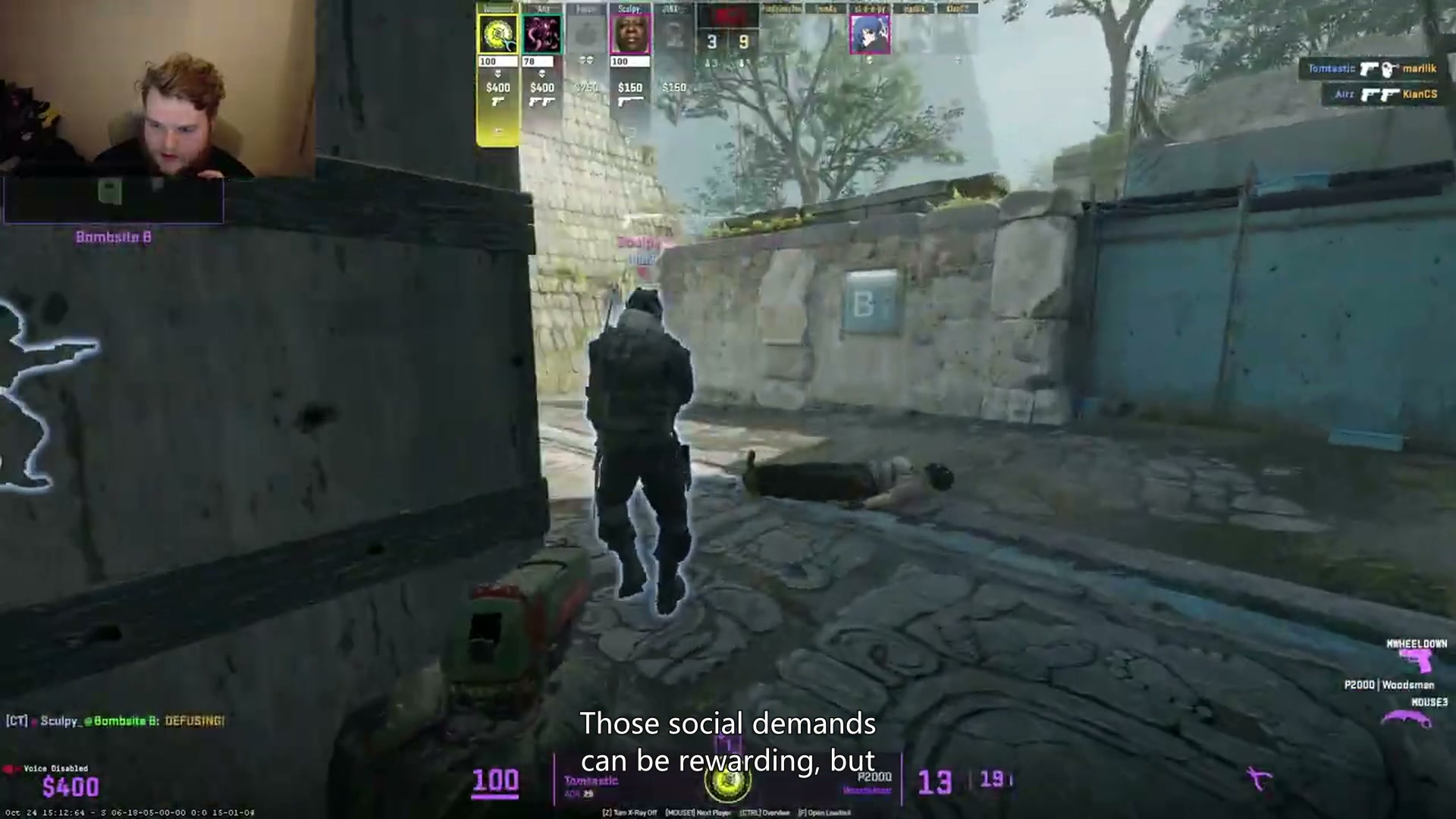}{02:53.9}%
}{fig:video-video_game_genres}
\GalleryExample{How Mountains Sustain Downstream Water Supplies}{2:58}{%
  \GalleryFrame{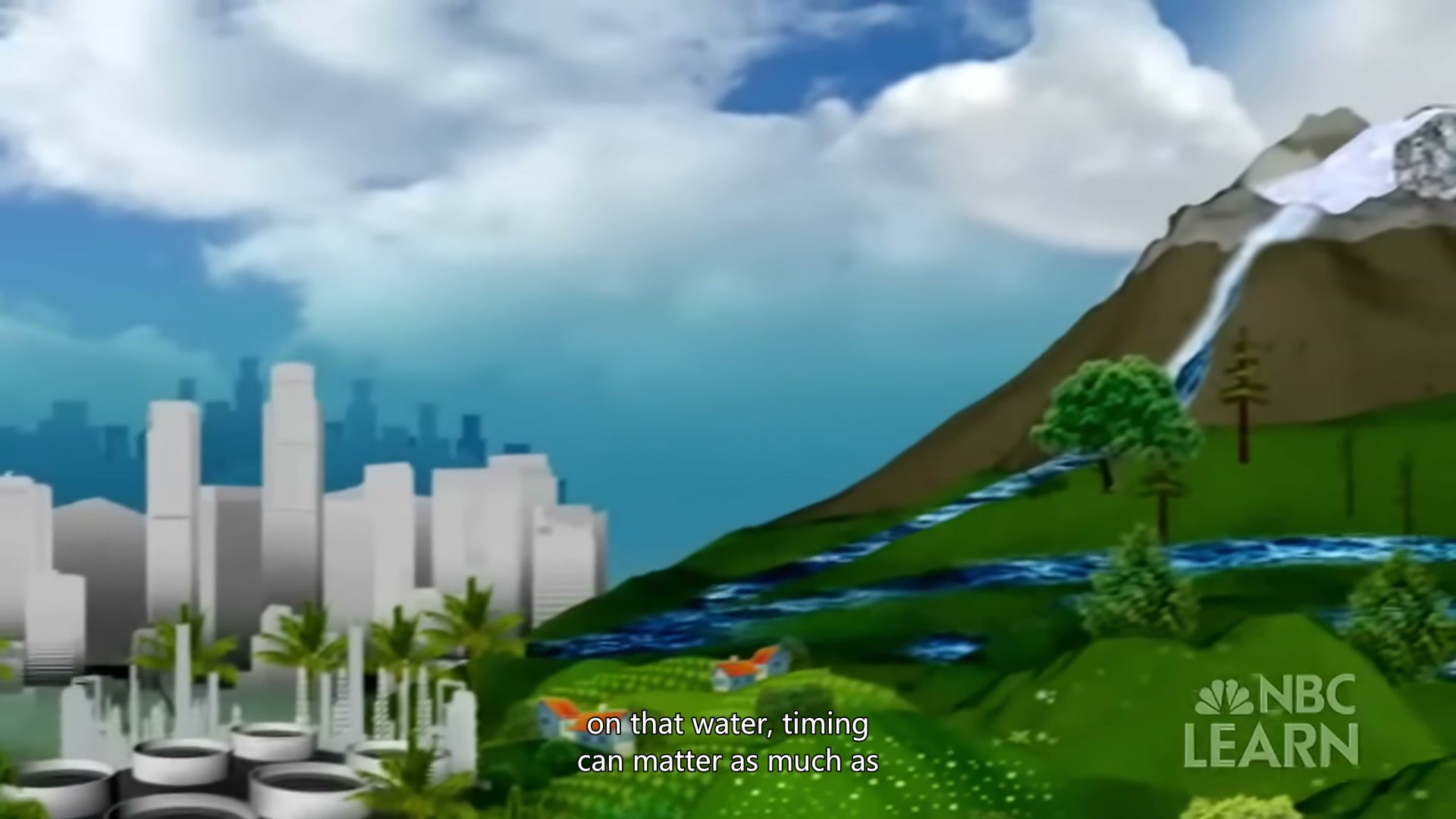}{00:18.6}\hfill%
  \GalleryFrame{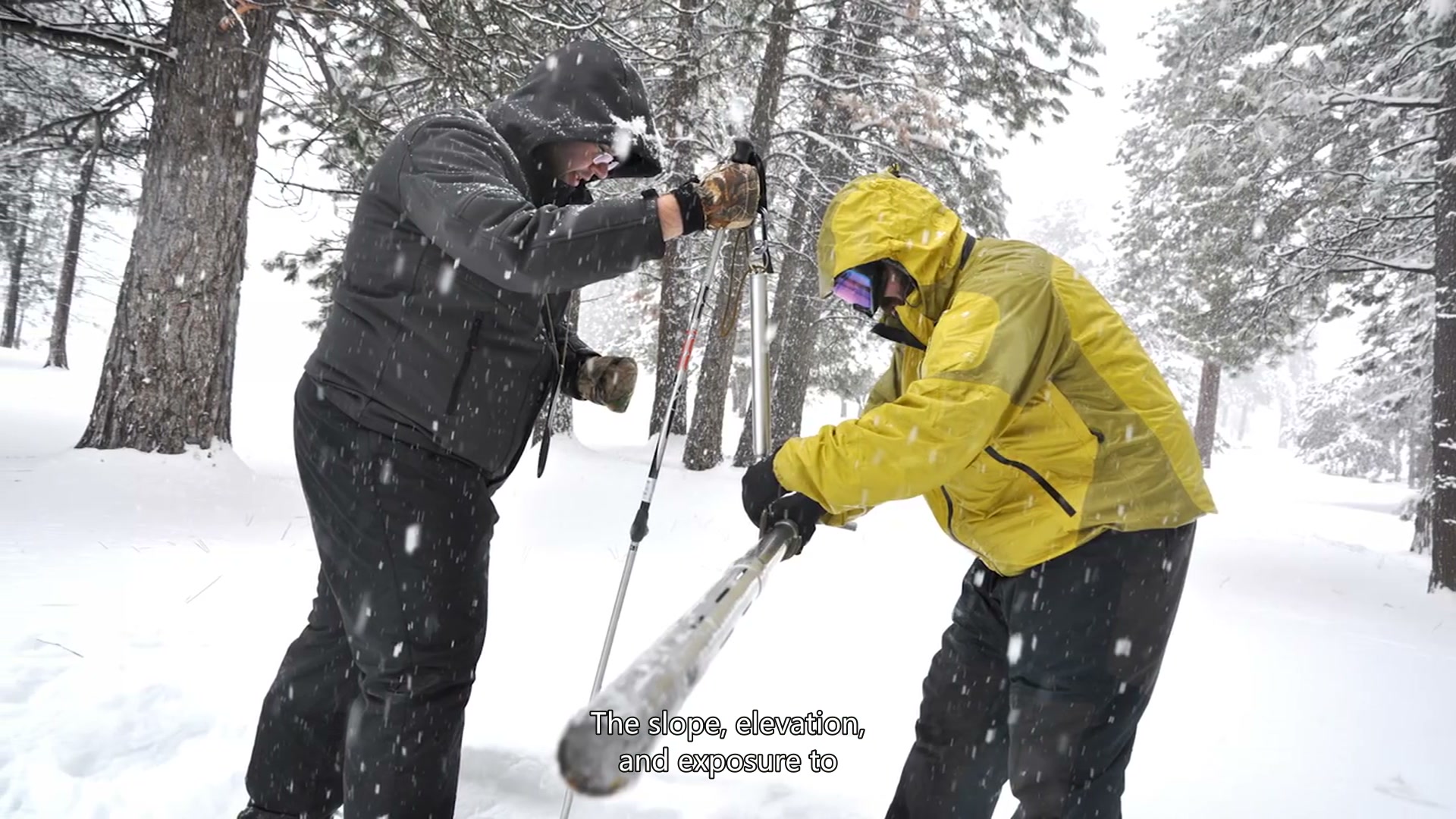}{01:03.4}\hfill%
  \GalleryFrame{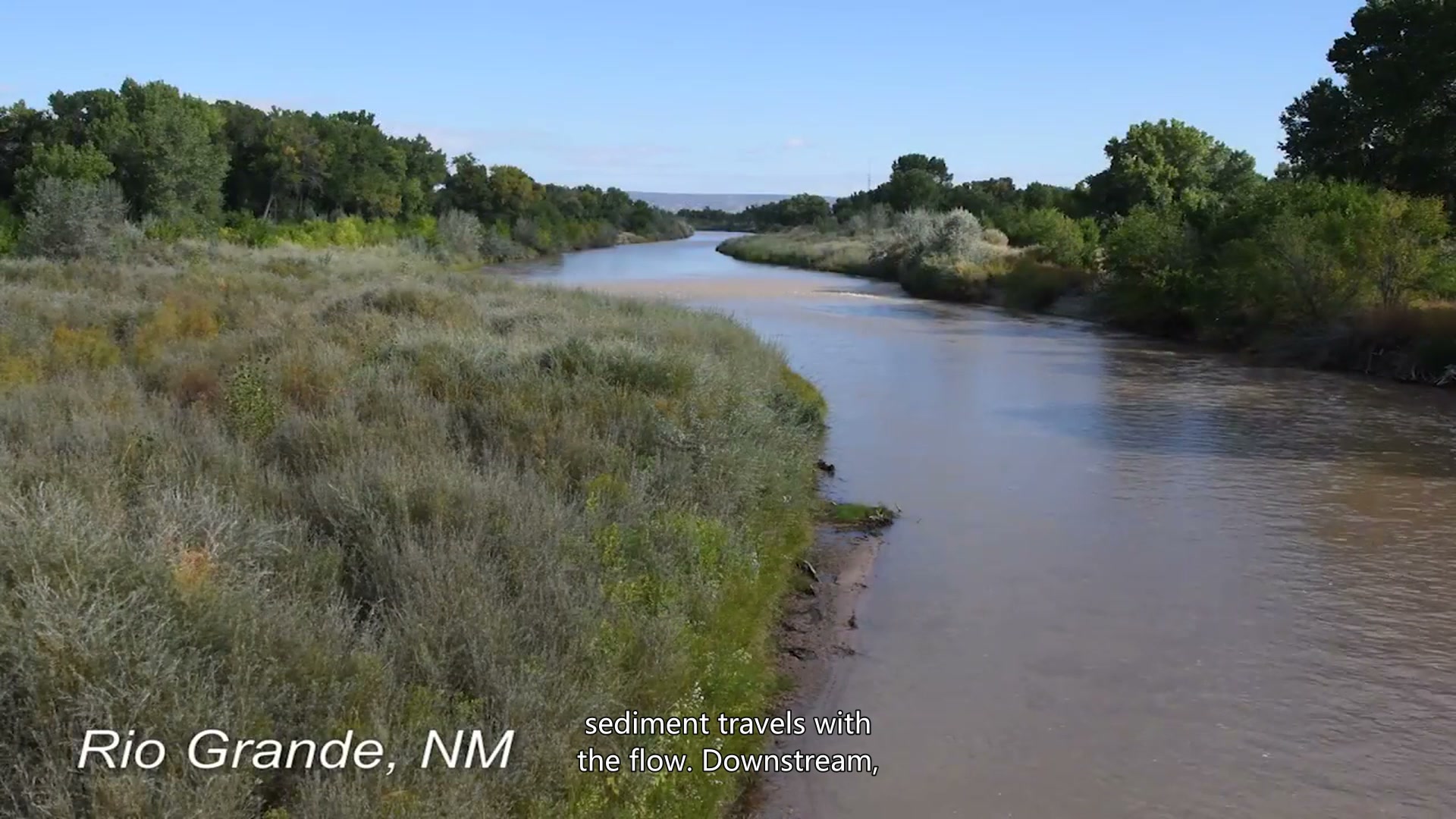}{01:48.1}\hfill%
  \GalleryFrame{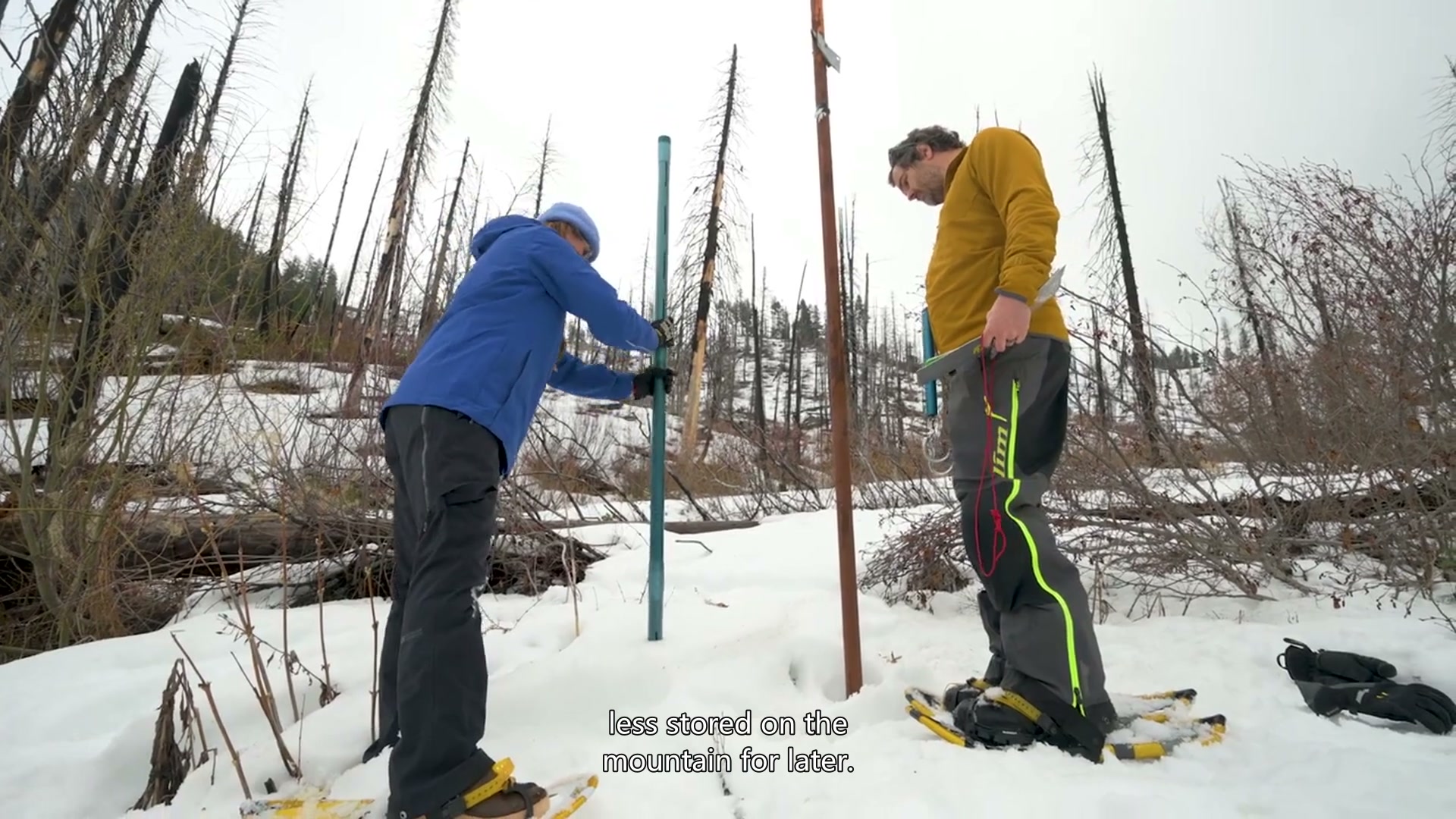}{02:32.8}%
}{fig:video-mountains_and_water_supply}
\GalleryExample{How the Netherlands Protects Low-Lying Land from Flooding}{3:23}{%
  \GalleryFrame{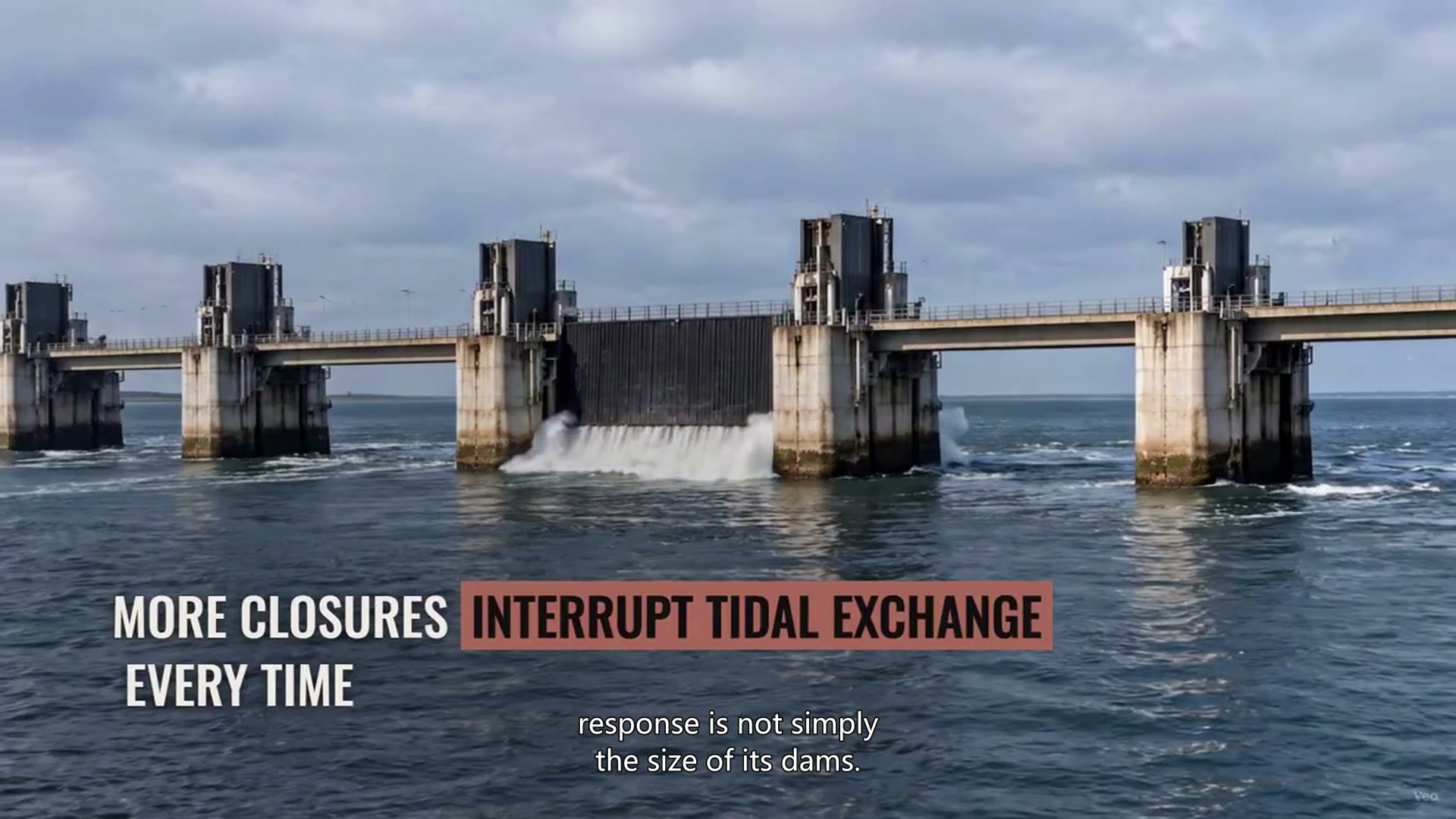}{00:21.2}\hfill%
  \GalleryFrame{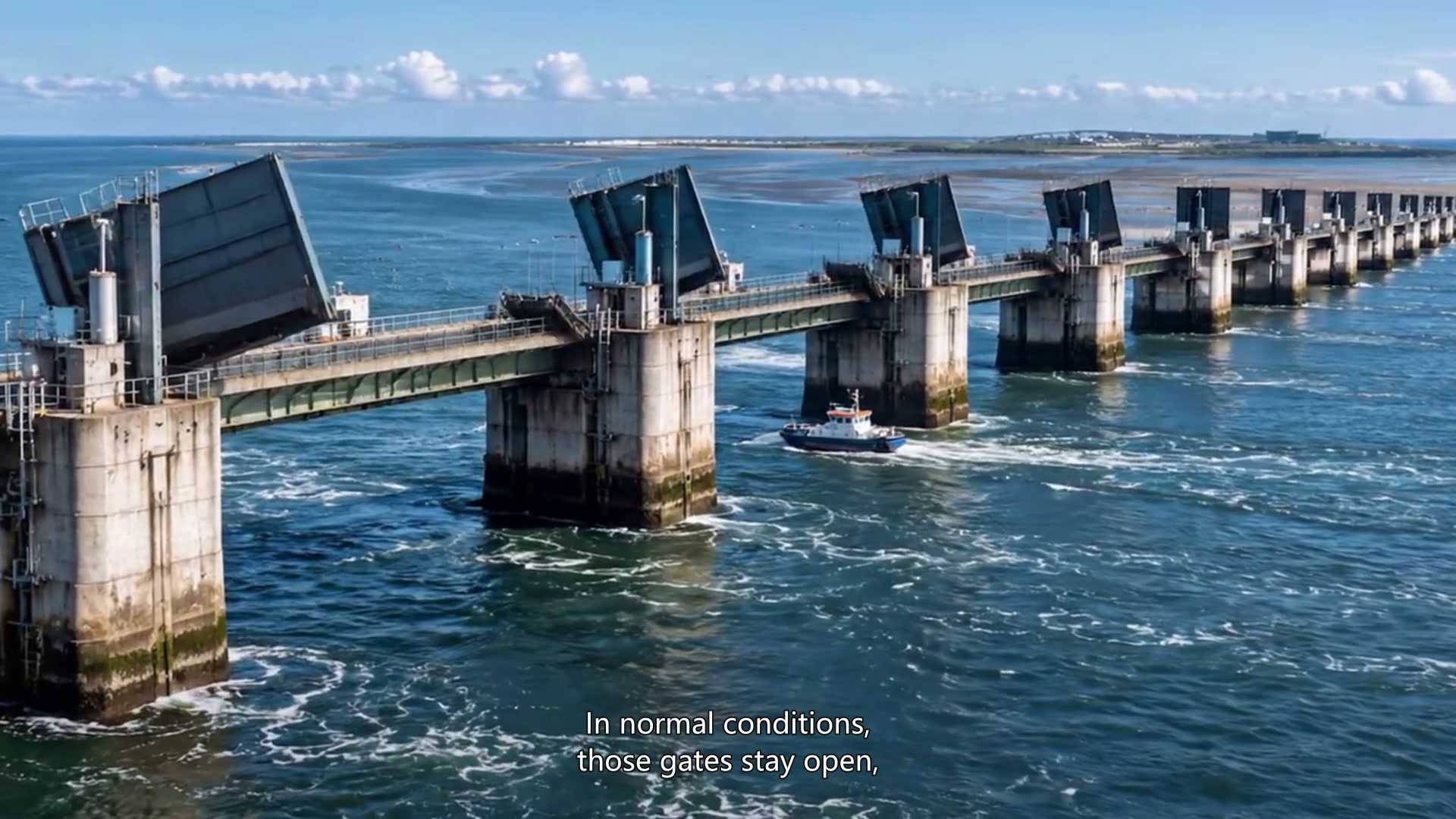}{01:12.2}\hfill%
  \GalleryFrame{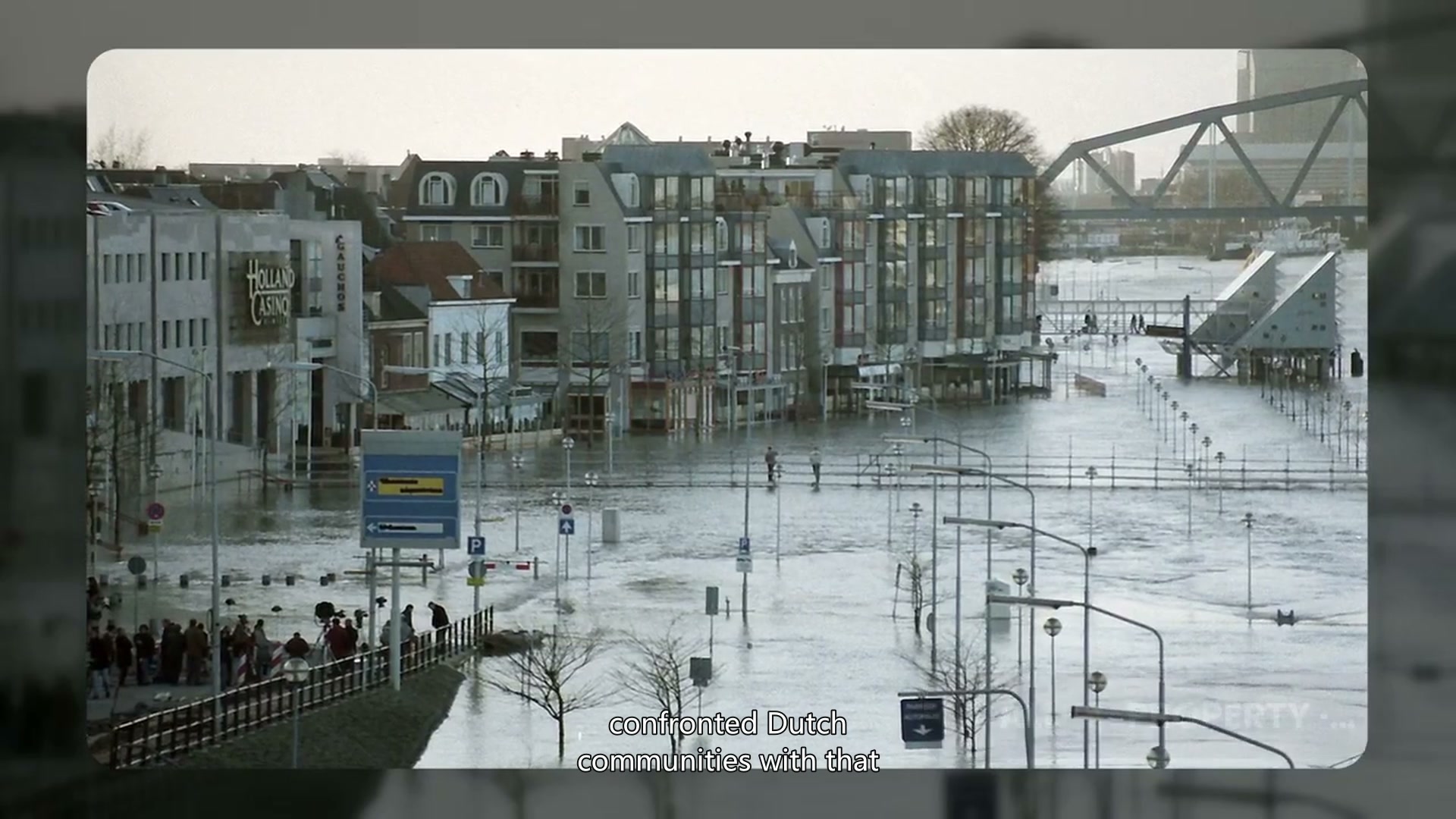}{02:03.1}\hfill%
  \GalleryFrame{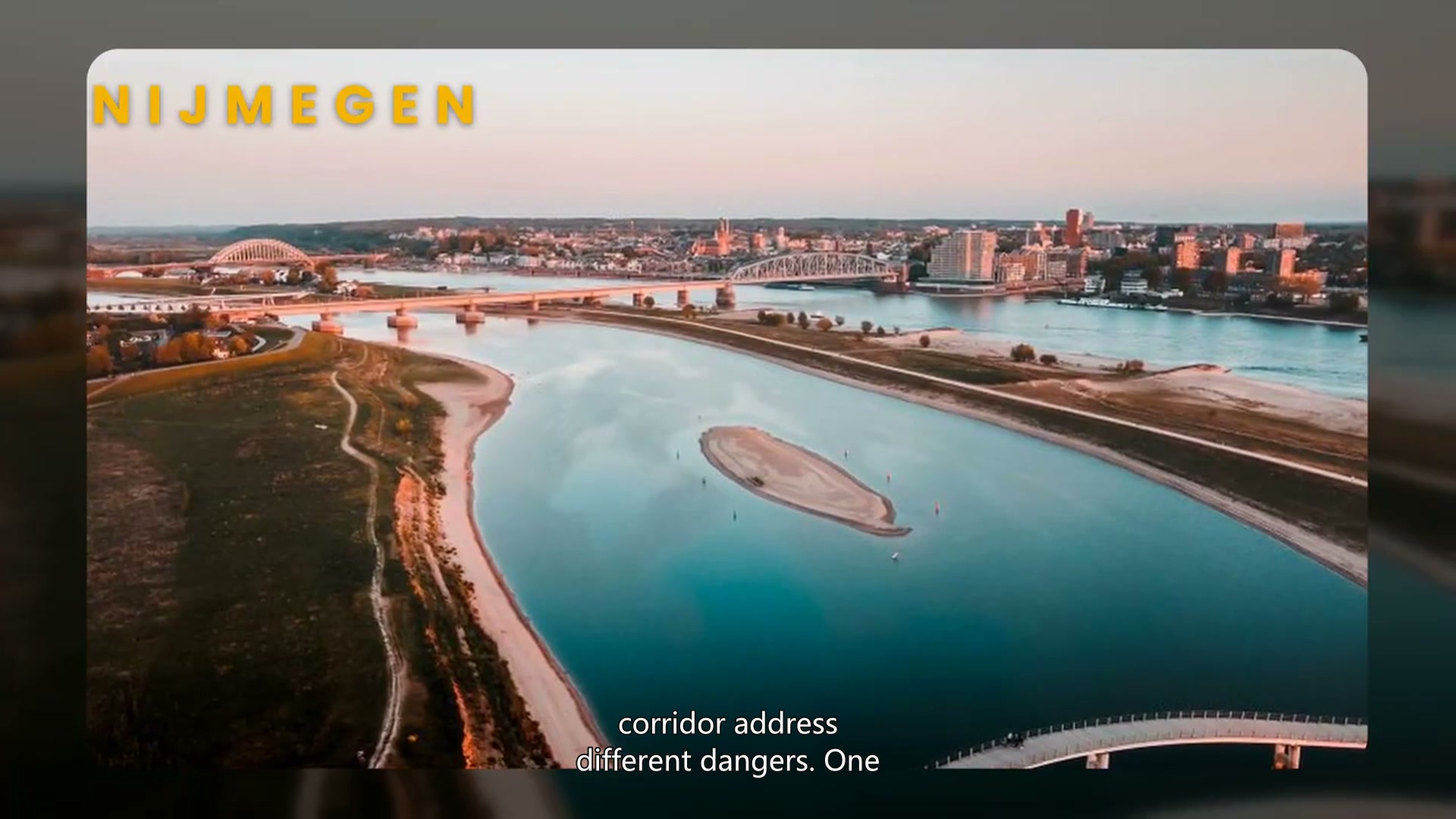}{02:54.0}%
}{fig:video-netherlands_flood_protection}
\GalleryExample{How Instant Noodles Became a Global Food}{3:33}{%
  \GalleryFrame{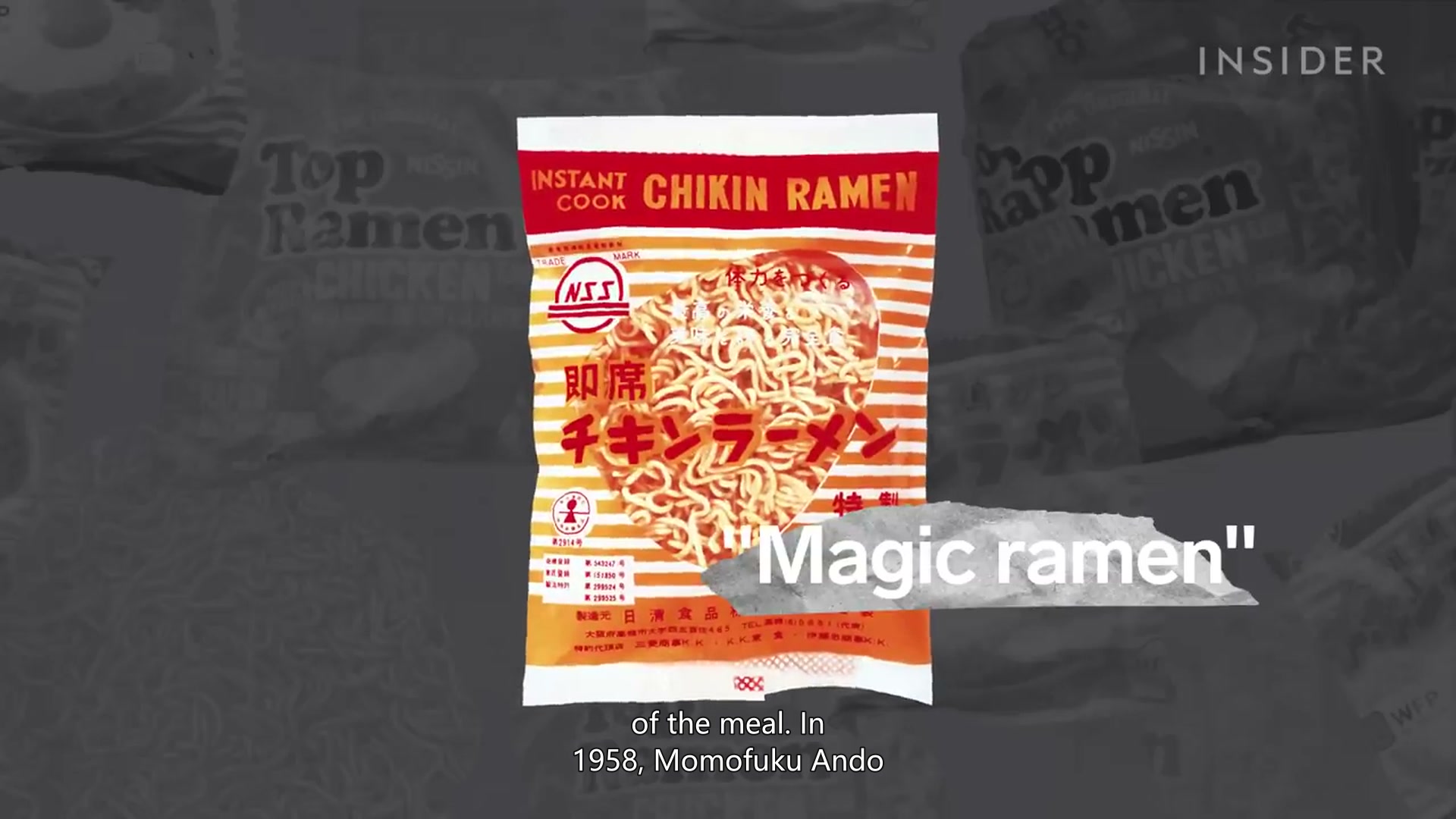}{00:22.2}\hfill%
  \GalleryFrame{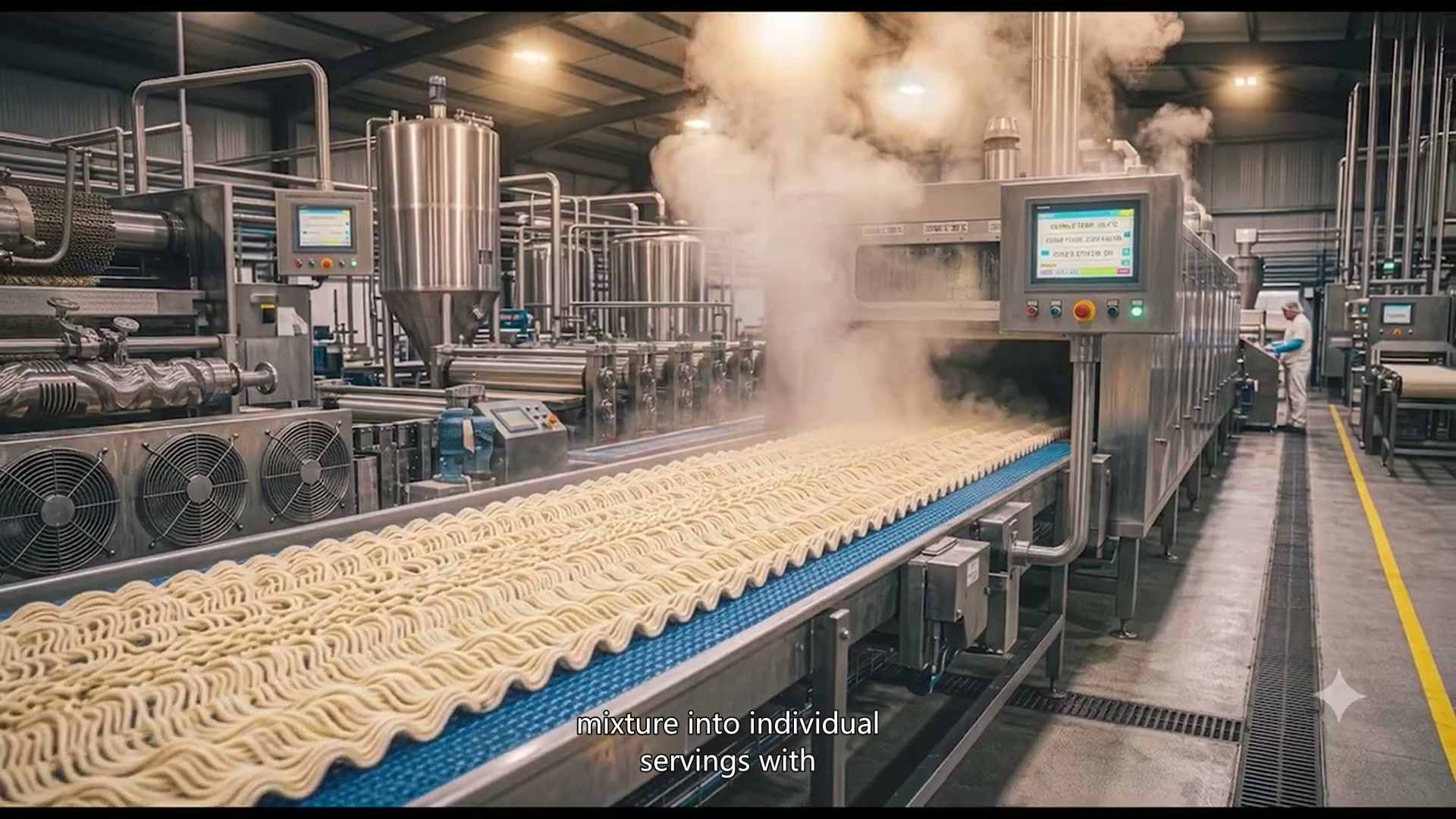}{01:15.6}\hfill%
  \GalleryFrame{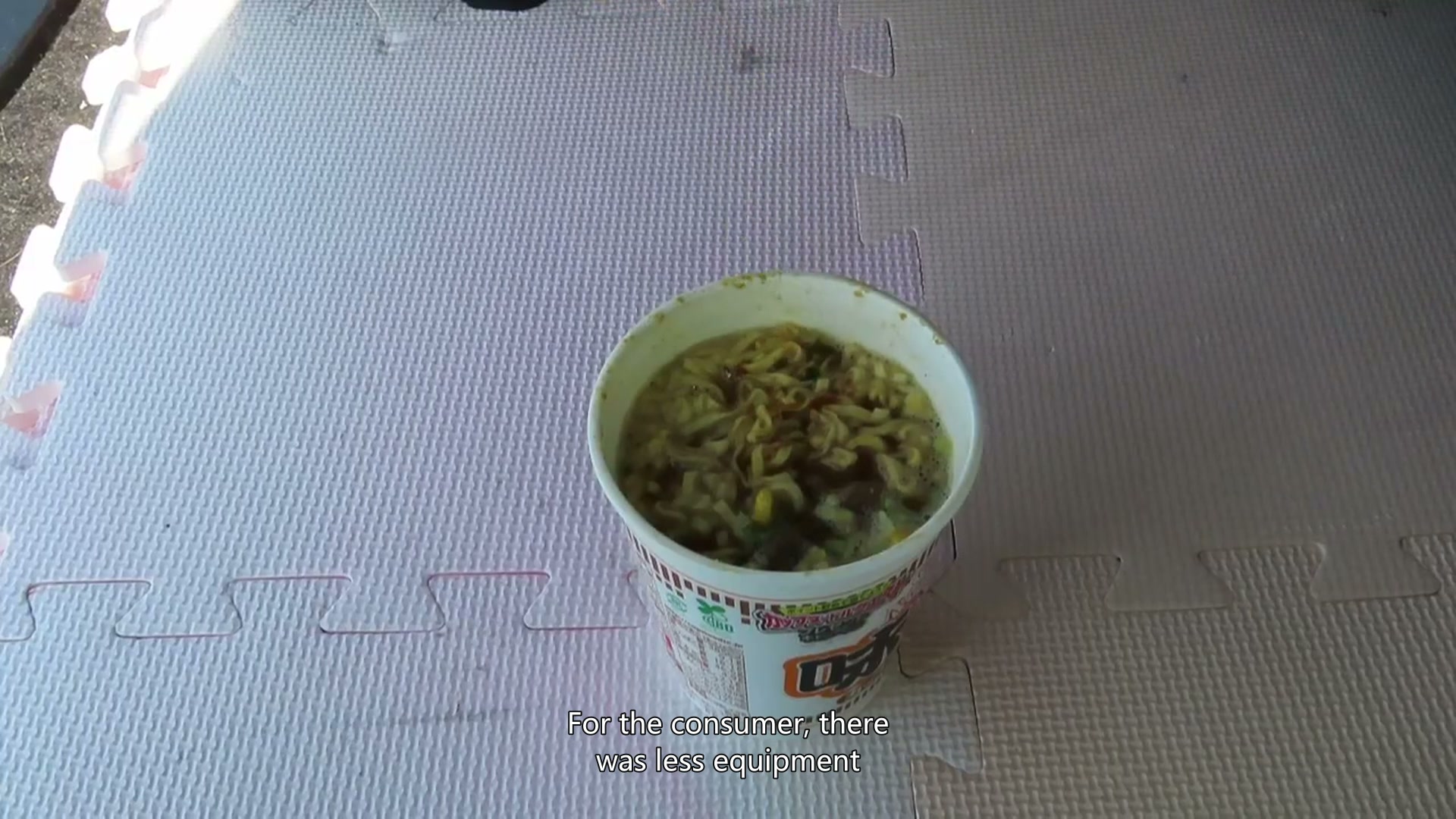}{02:09.0}\hfill%
  \GalleryFrame{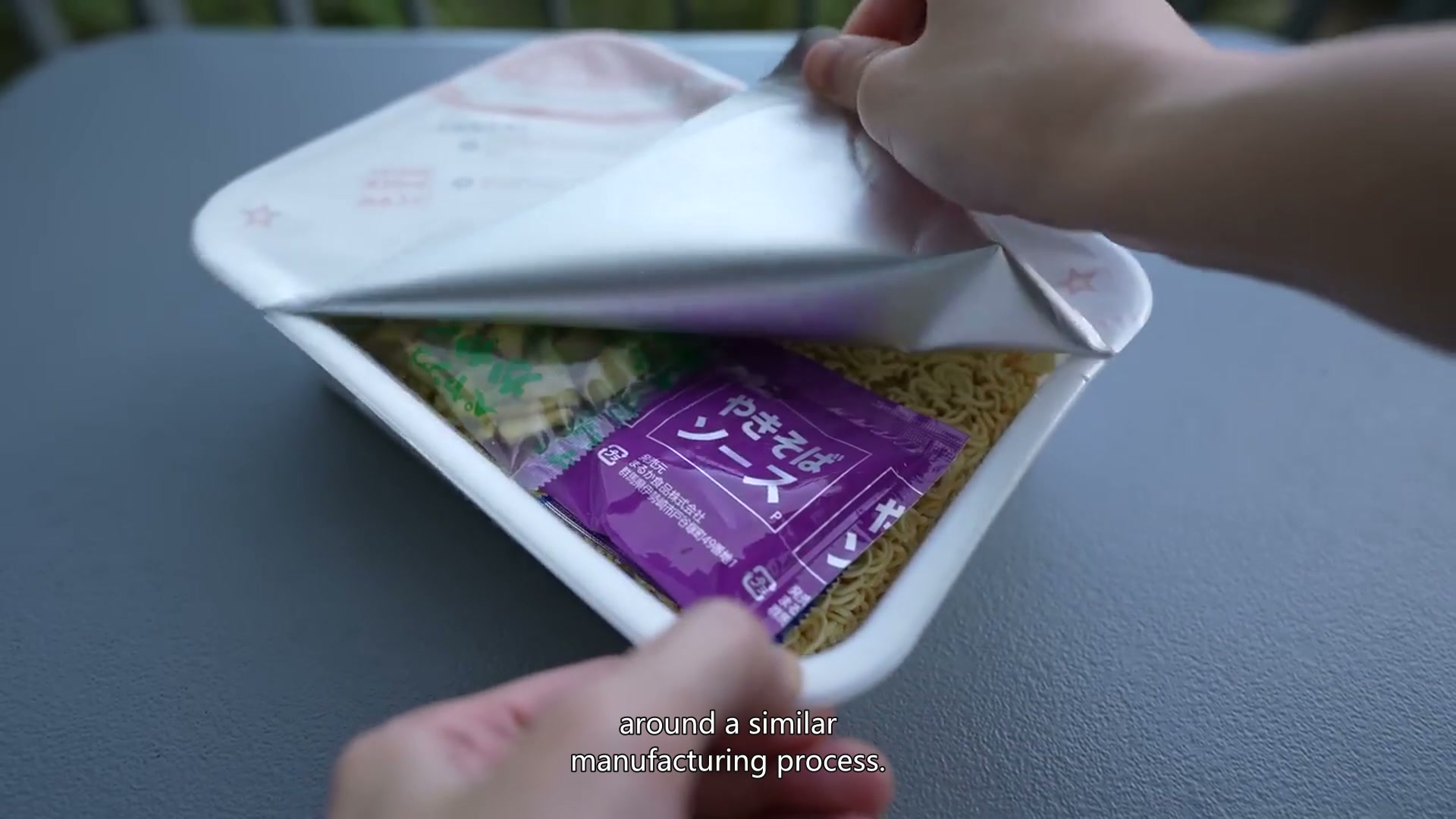}{03:02.4}%
}{fig:video-instant_noodle_industry}
\GalleryExample{A Guide to Dressing for Winter Travel}{3:15}{%
  \GalleryFrame{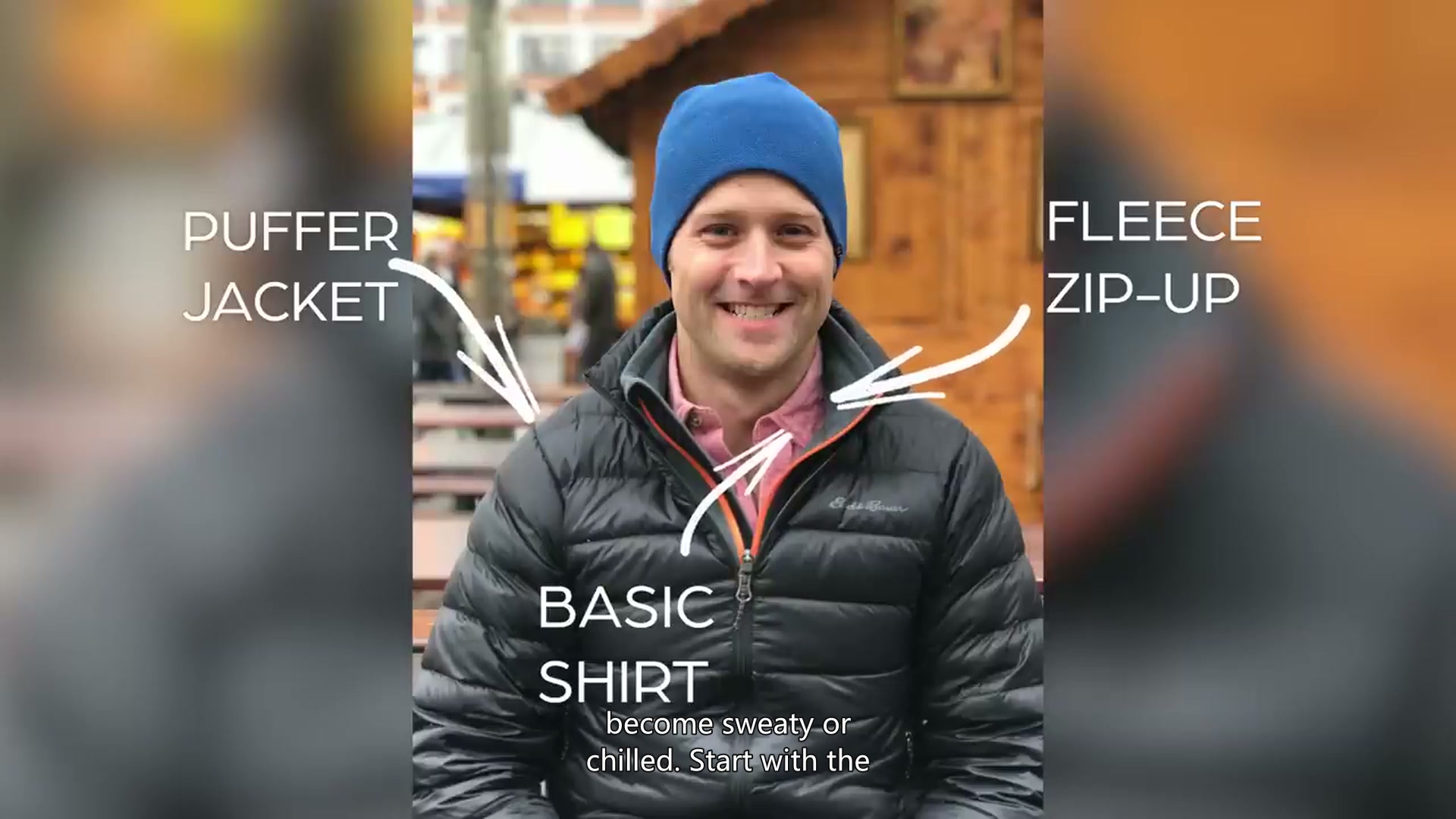}{00:20.3}\hfill%
  \GalleryFrame{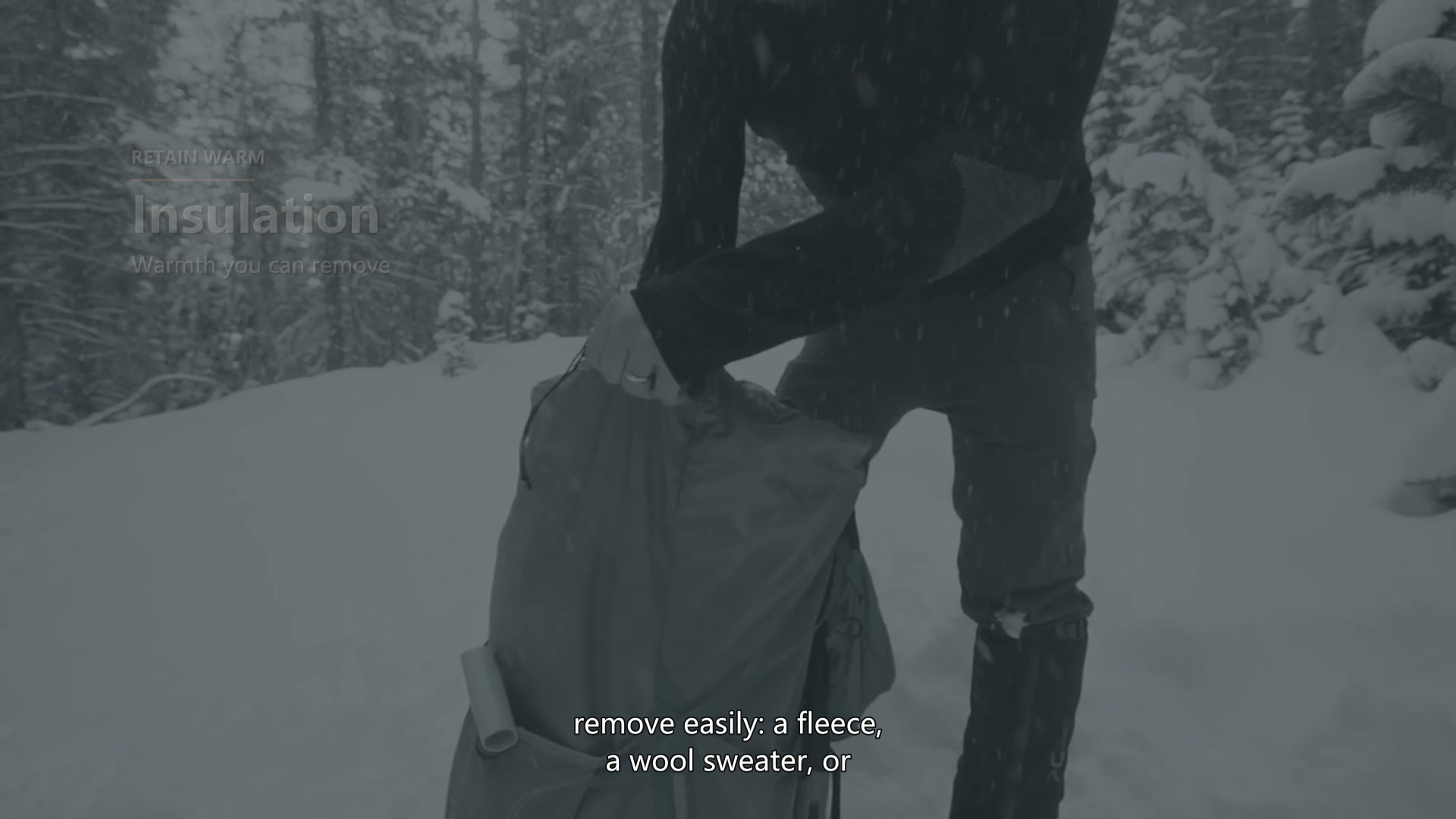}{01:09.2}\hfill%
  \GalleryFrame{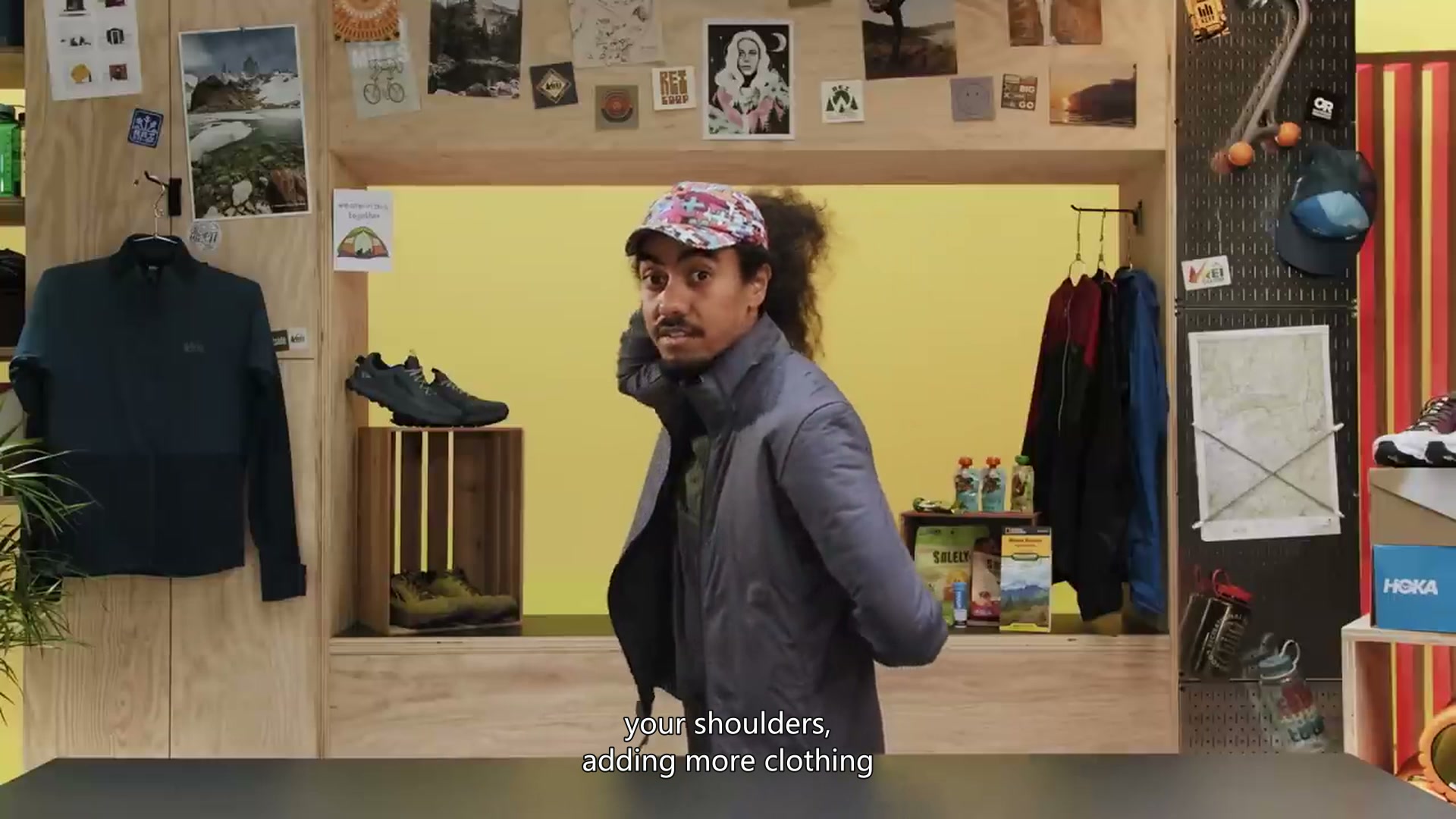}{01:58.0}\hfill%
  \GalleryFrame{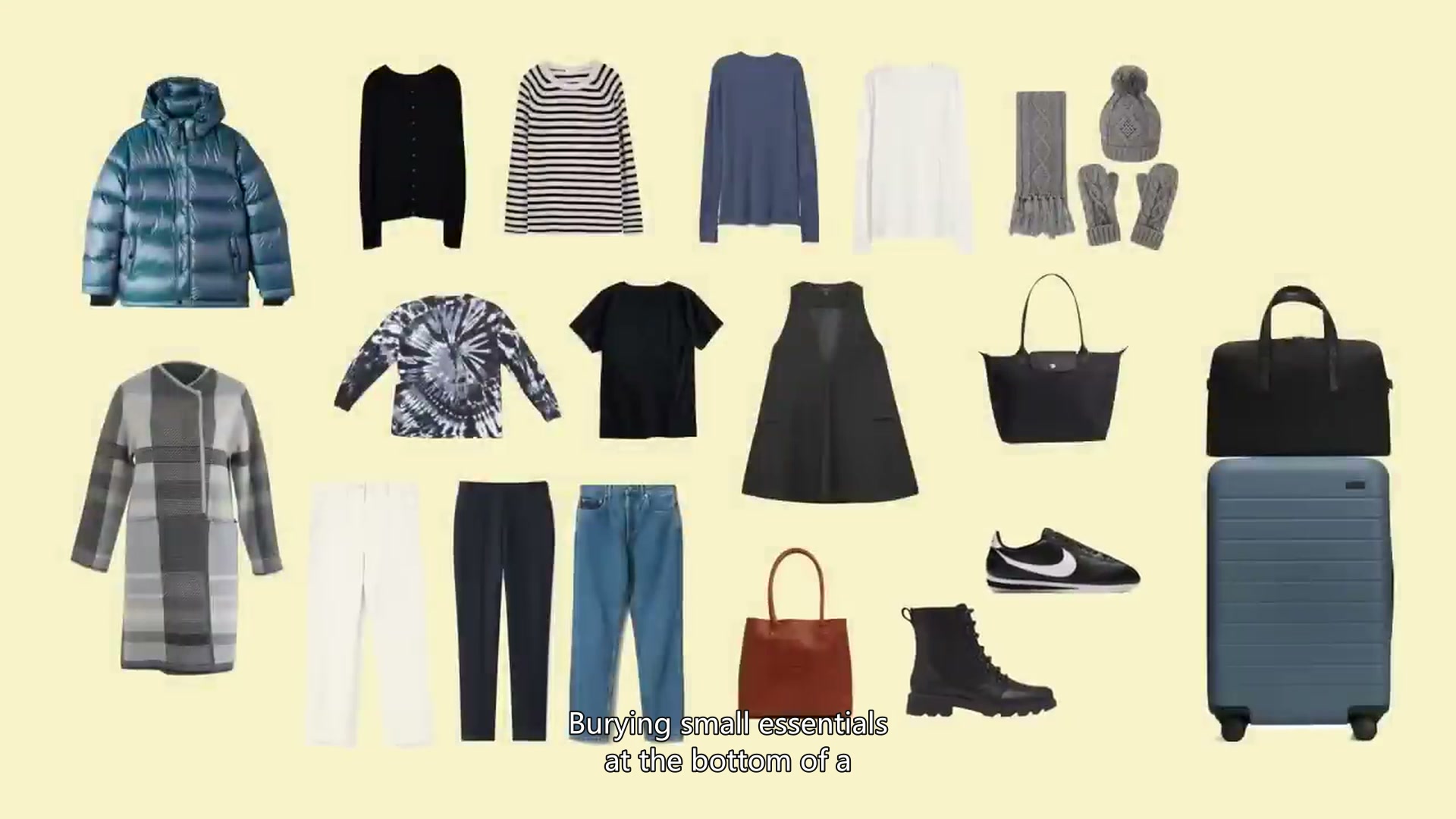}{02:46.8}%
}{fig:video-winter_travel_clothing}
\GalleryExample{Why Are Humans Returning to the Moon?}{3:23}{%
  \GalleryFrame{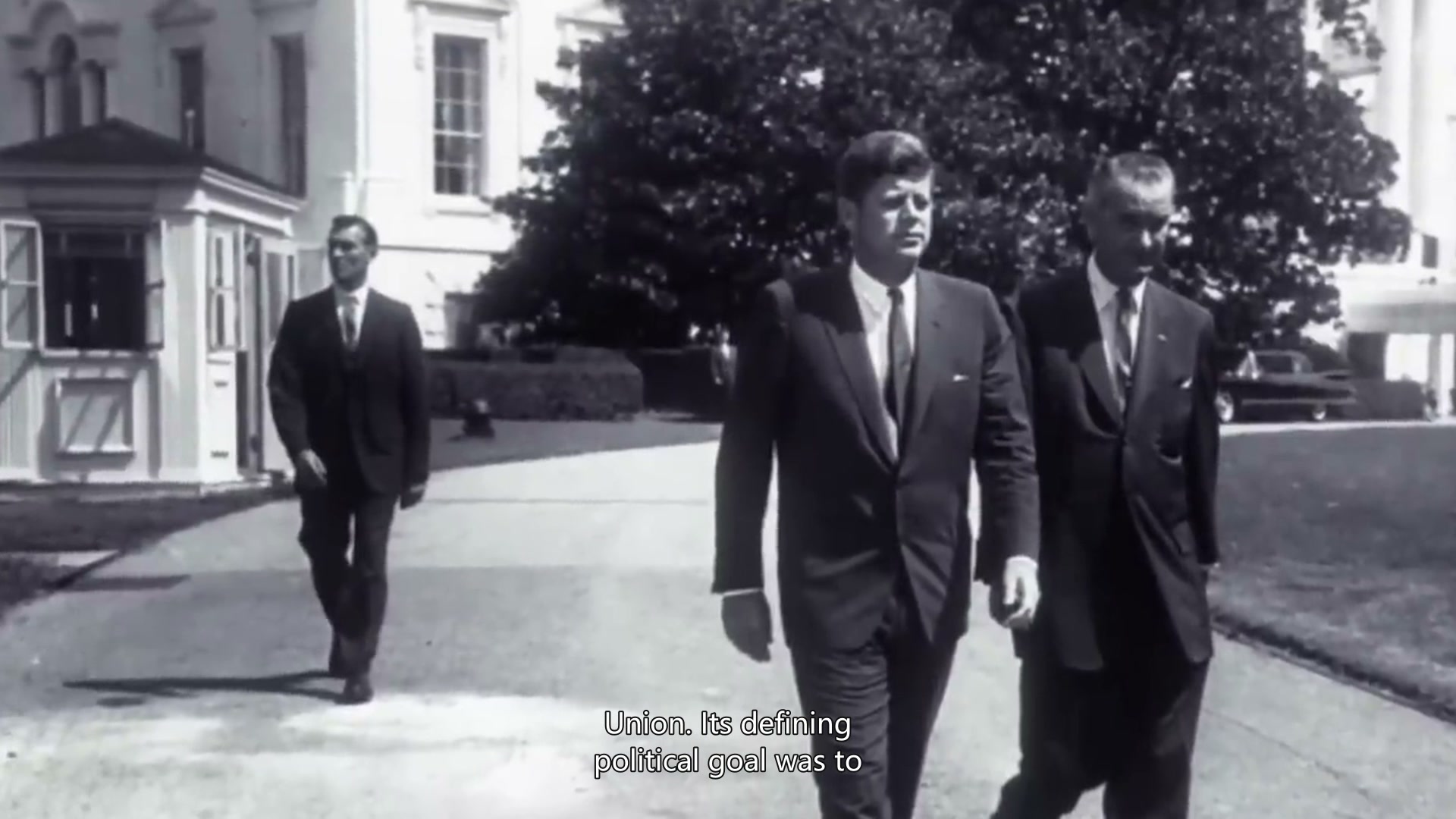}{00:21.2}\hfill%
  \GalleryFrame{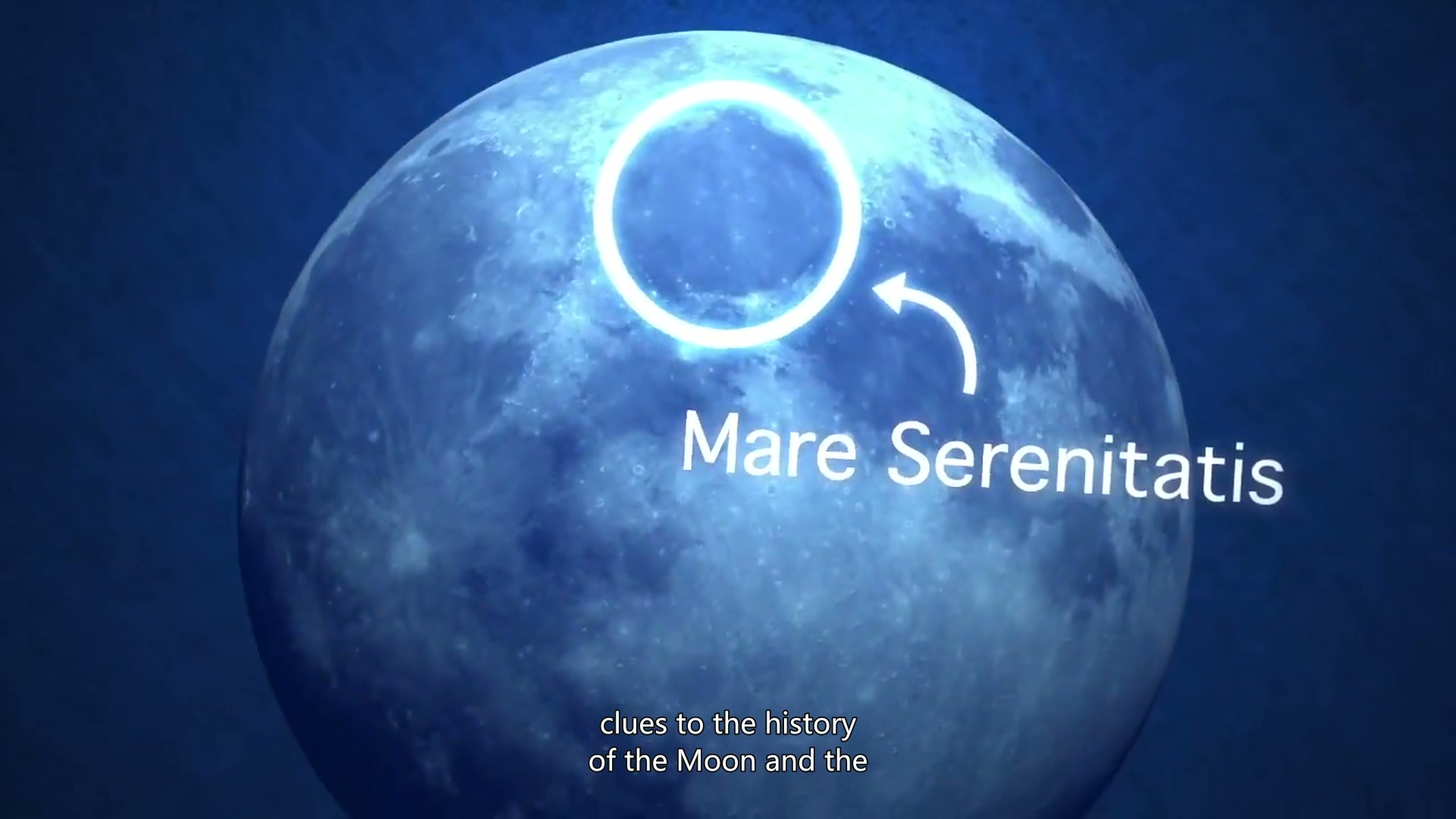}{01:12.0}\hfill%
  \GalleryFrame{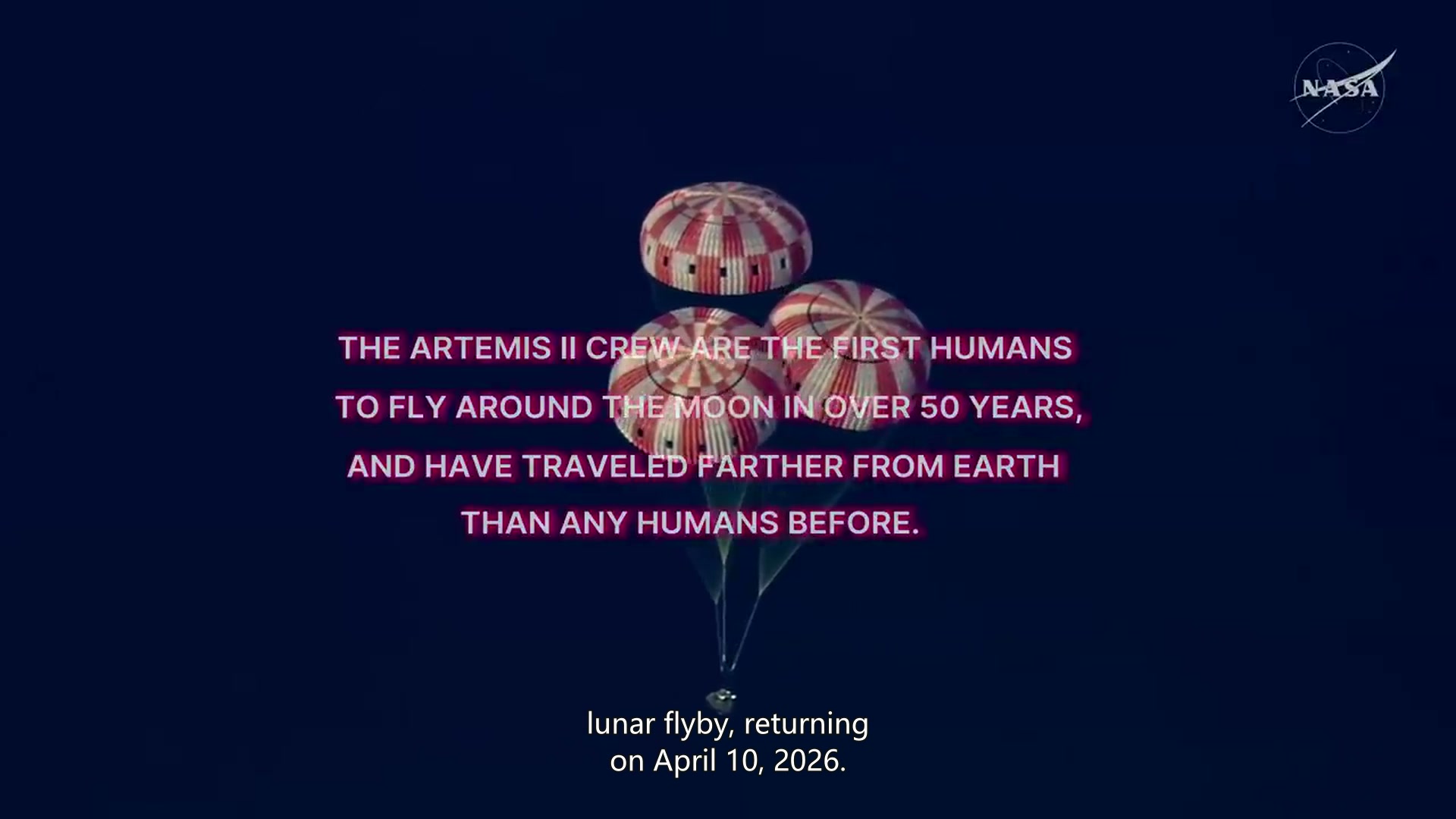}{02:02.8}\hfill%
  \GalleryFrame{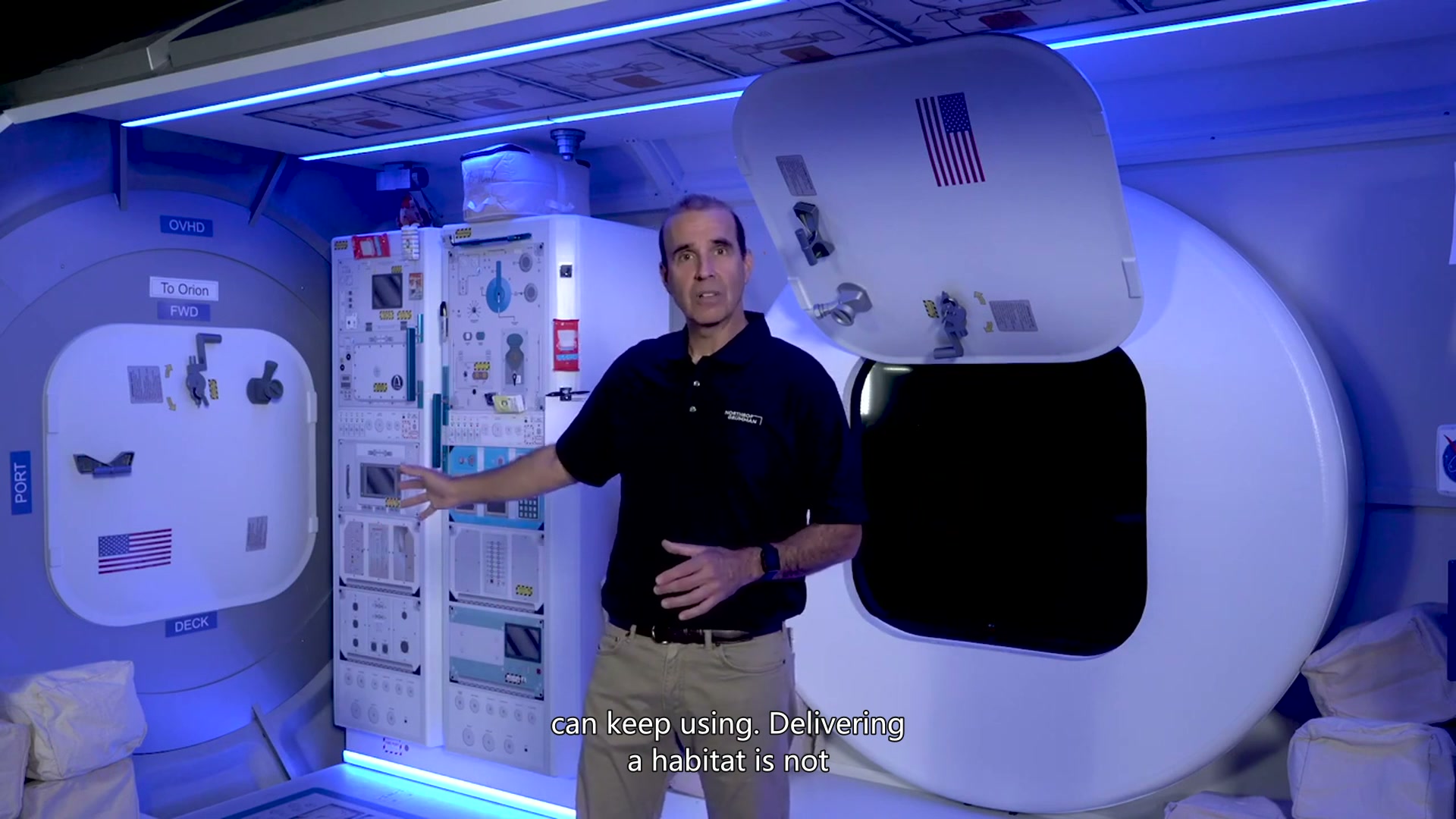}{02:53.6}%
}{fig:video-returning_to_the_moon}
\GalleryExample{Coffee: From Farm to Cup}{3:38}{%
  \GalleryFrame{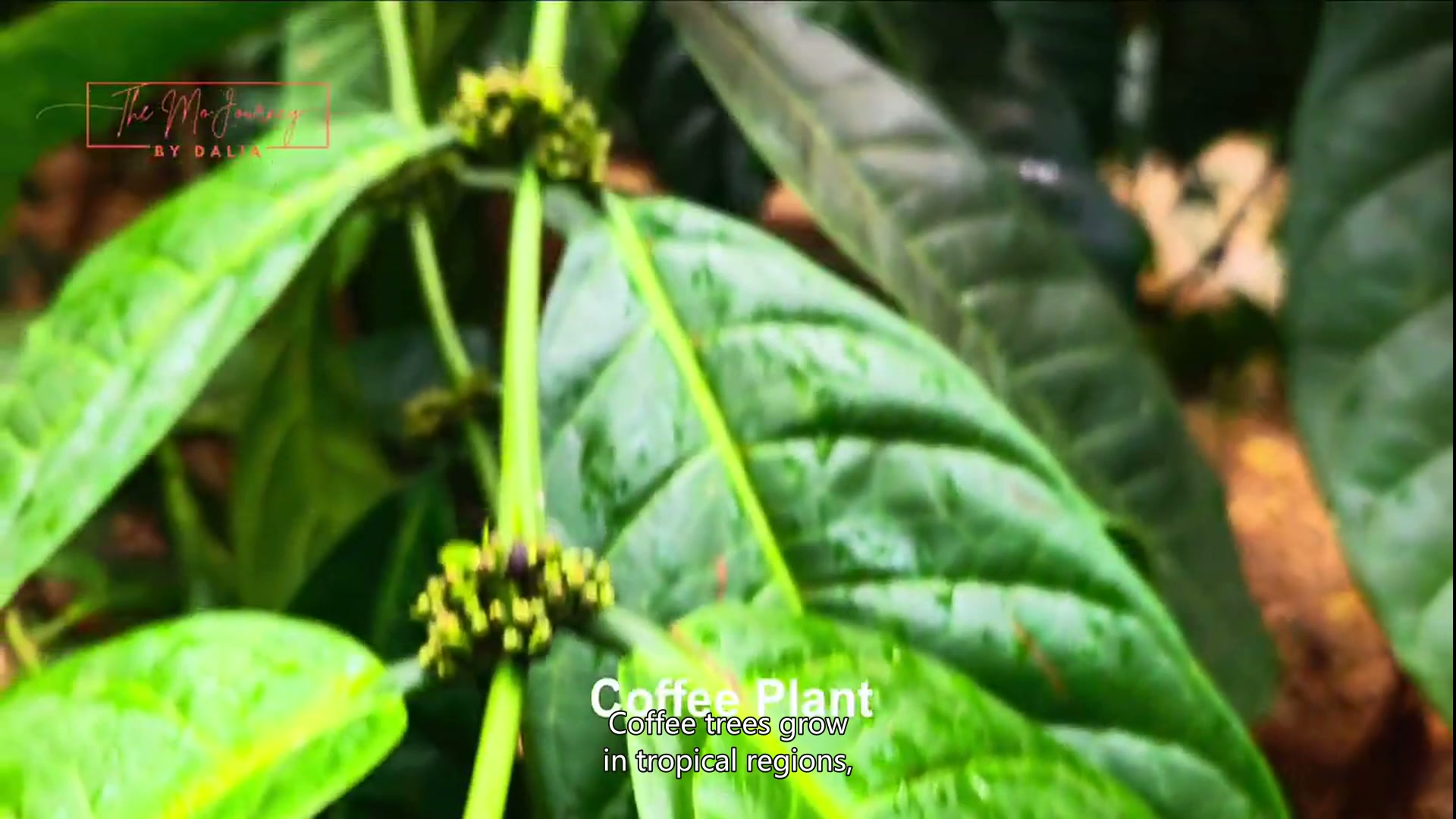}{00:22.7}\hfill%
  \GalleryFrame{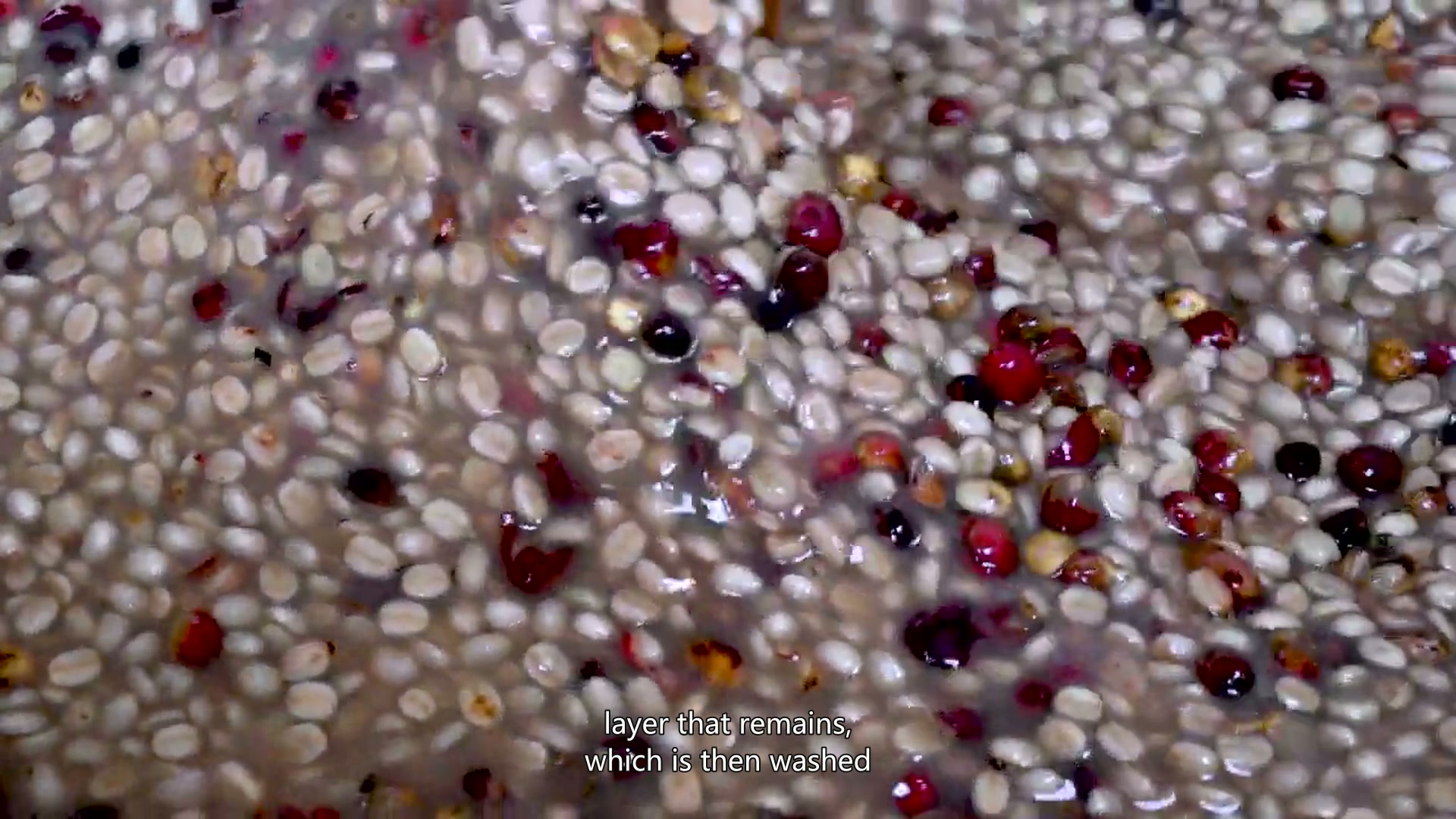}{01:17.3}\hfill%
  \GalleryFrame{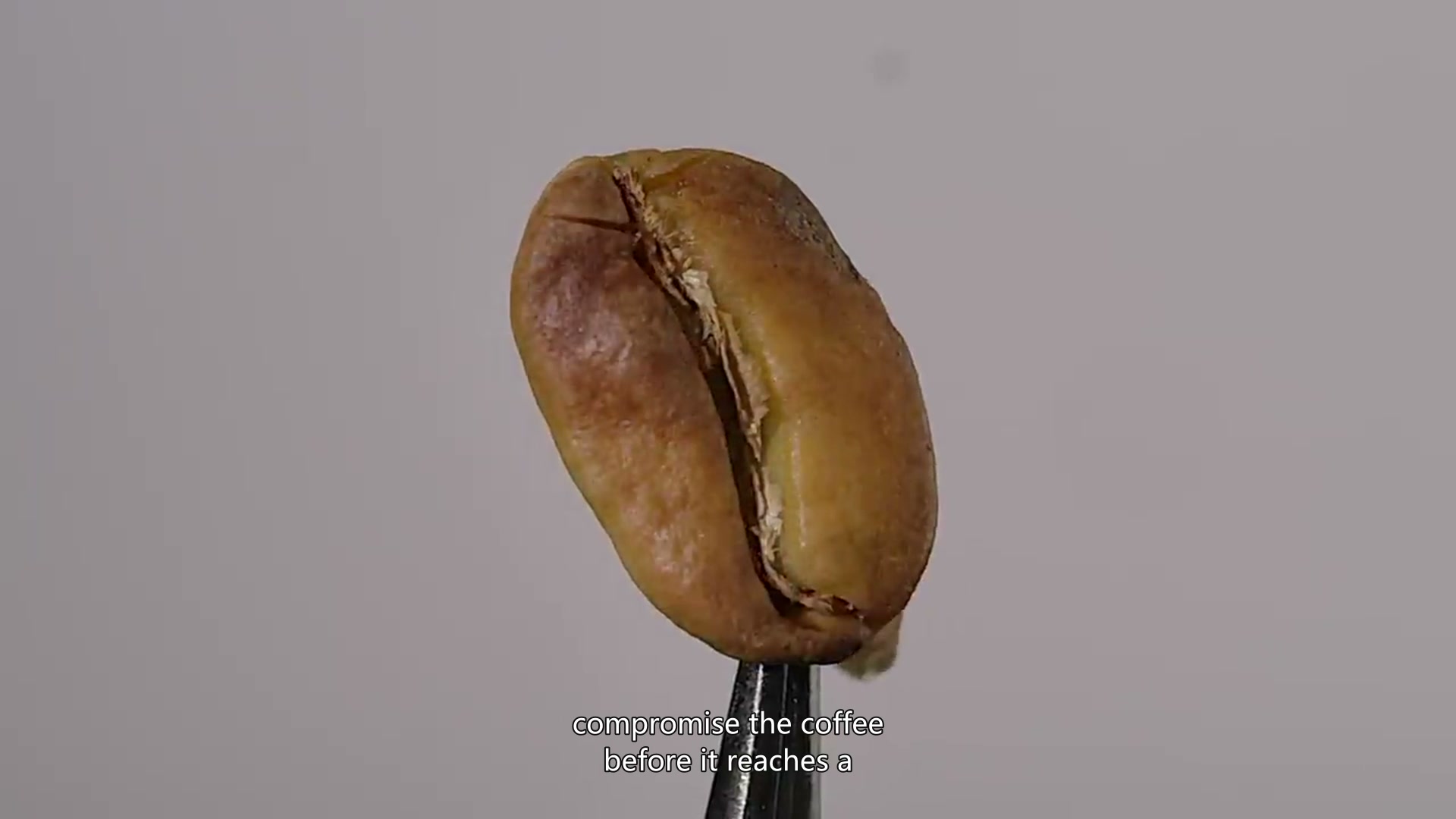}{02:12.0}\hfill%
  \GalleryFrame{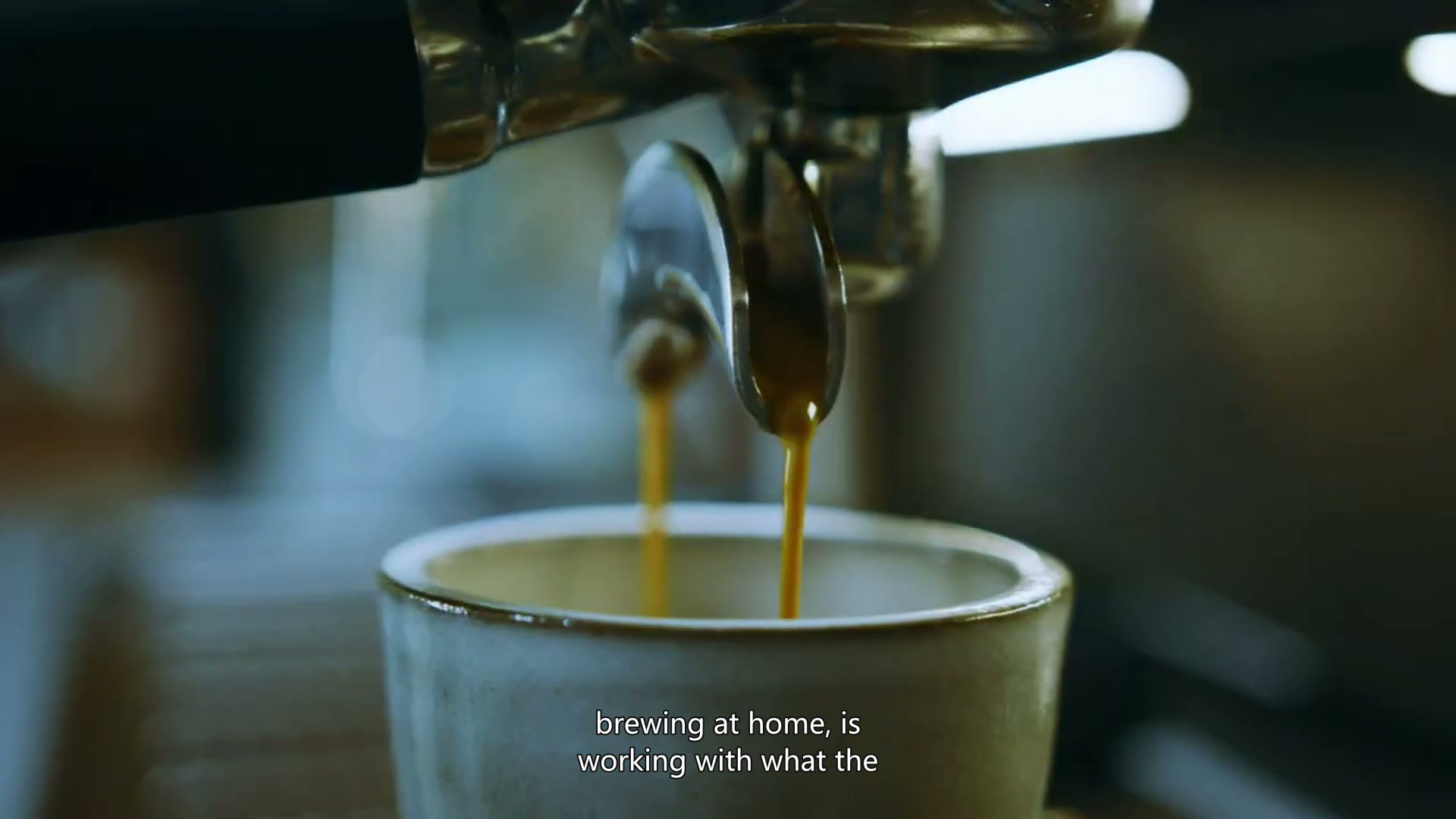}{03:06.6}%
}{fig:video-coffee_farm_to_cup}
\GalleryExample{How Printing Transformed the Spread of Knowledge}{3:30}{%
  \GalleryFrame{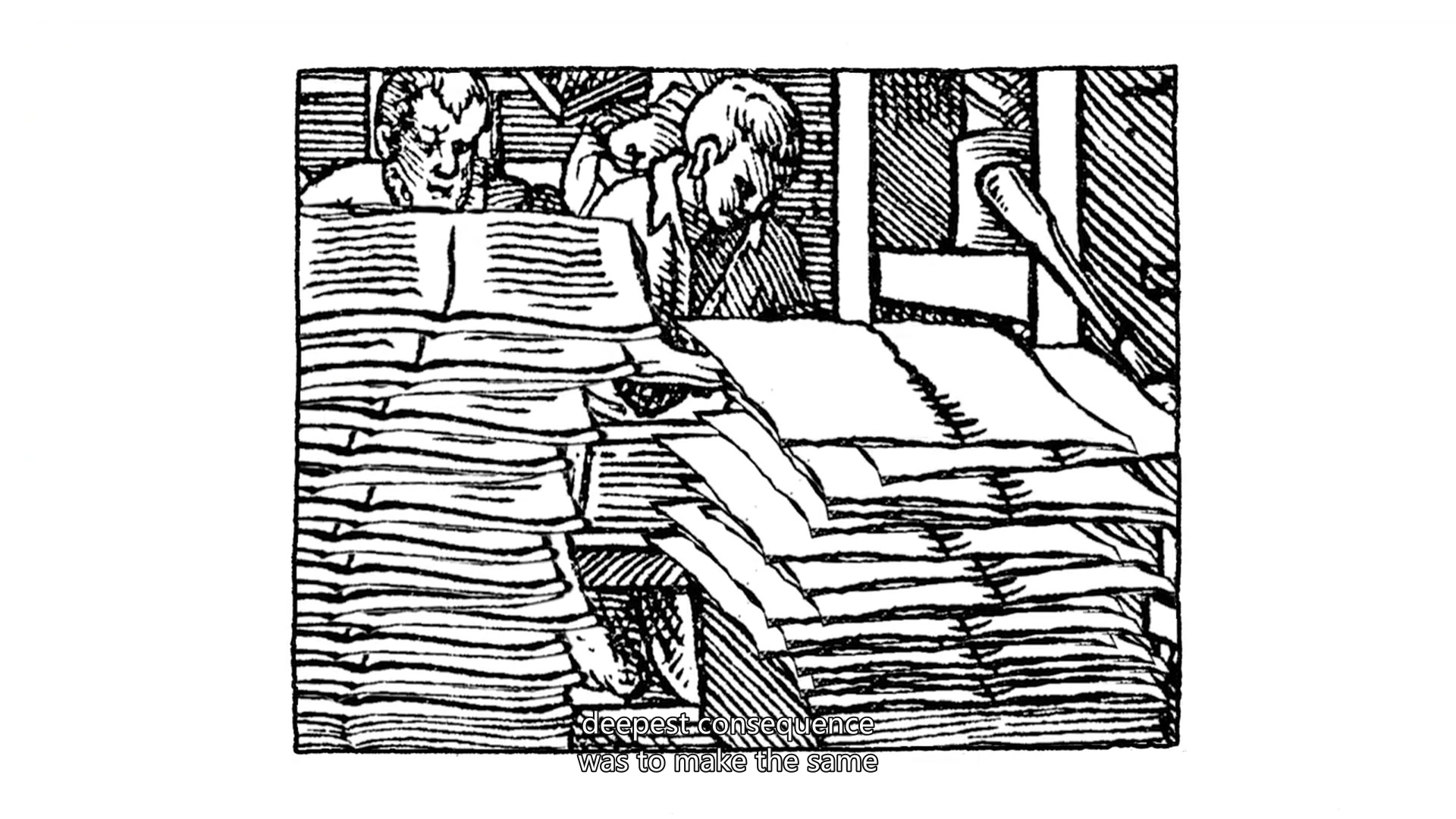}{00:21.9}\hfill%
  \GalleryFrame{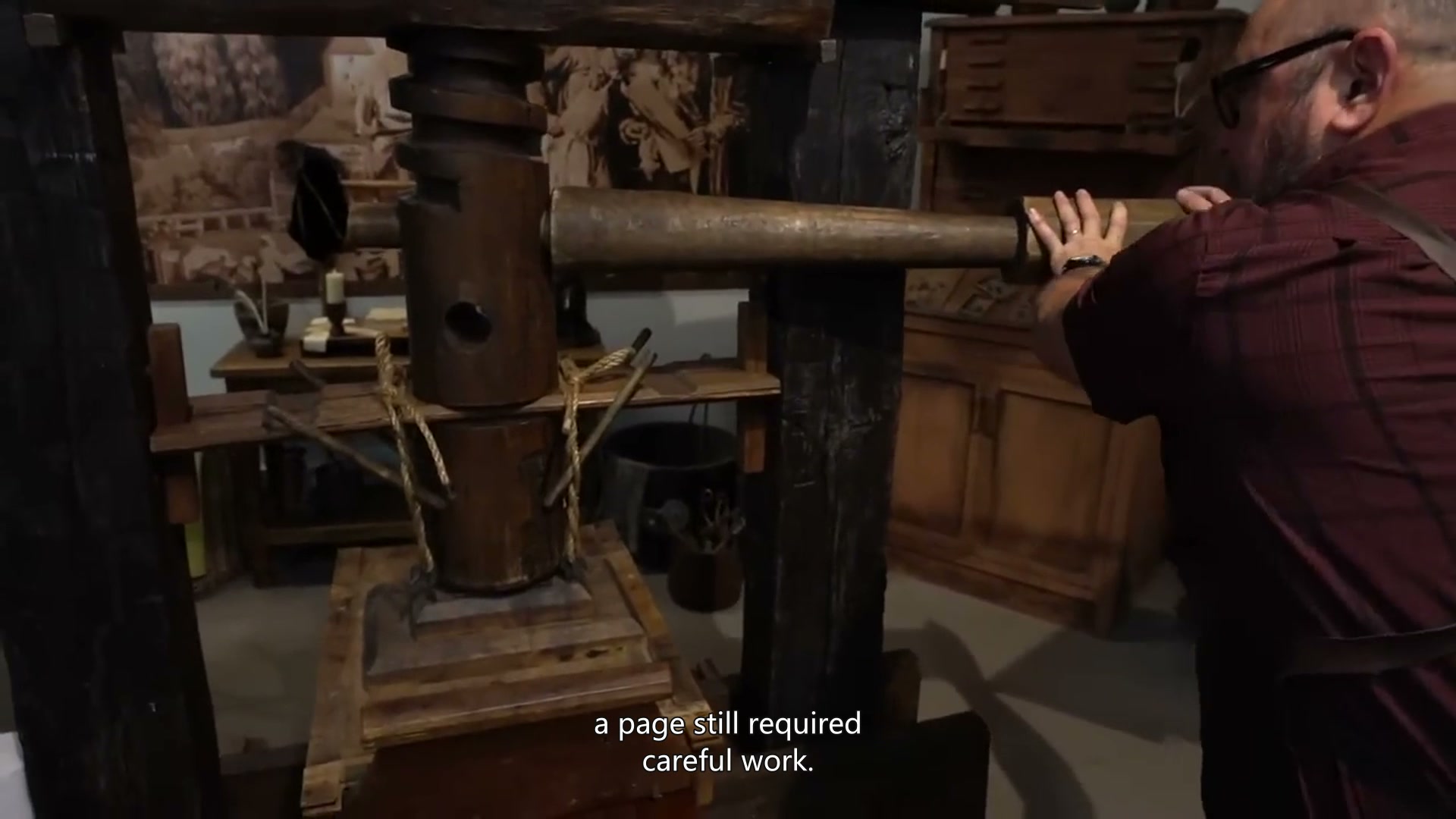}{01:14.6}\hfill%
  \GalleryFrame{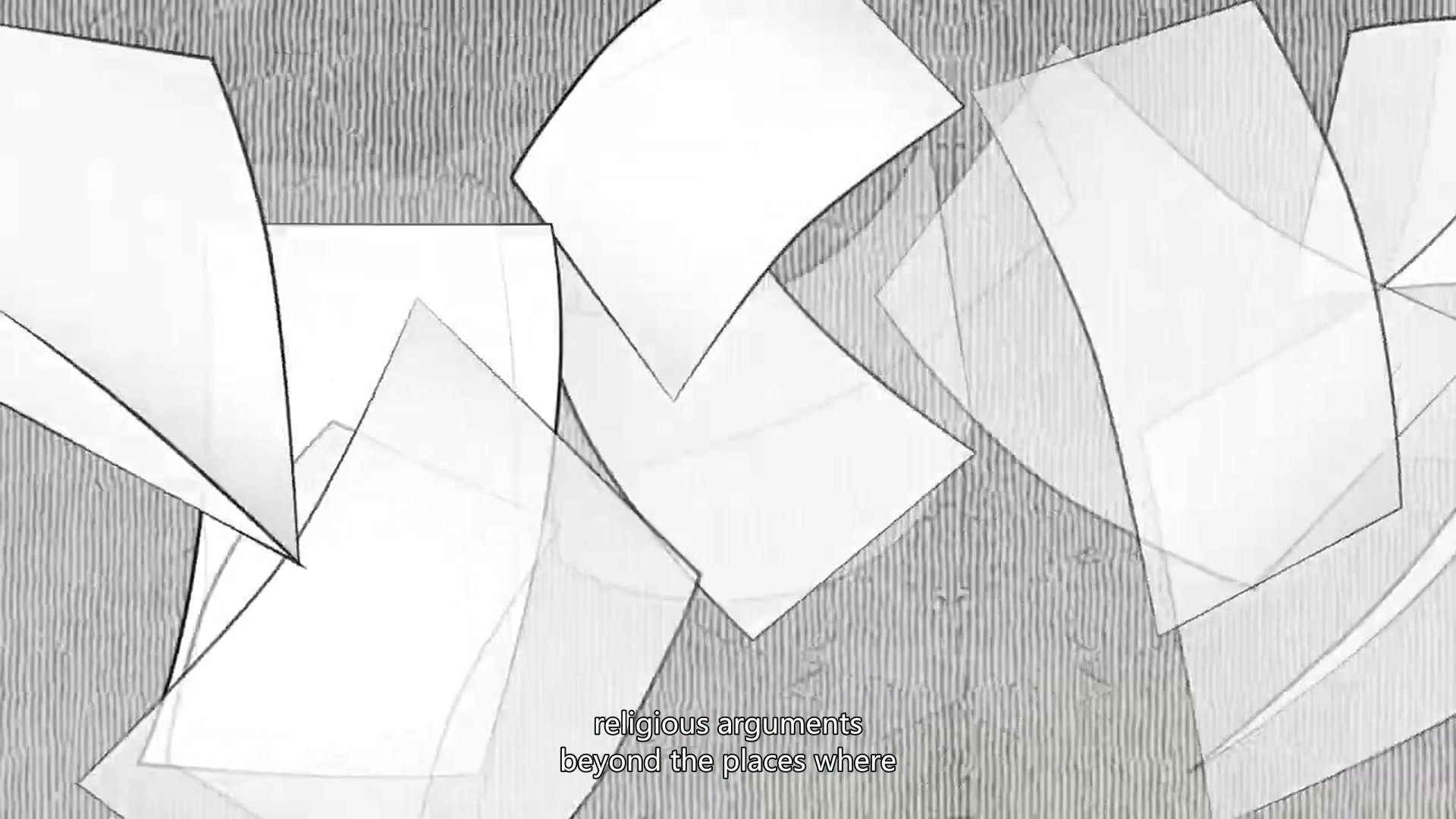}{02:07.3}\hfill%
  \GalleryFrame{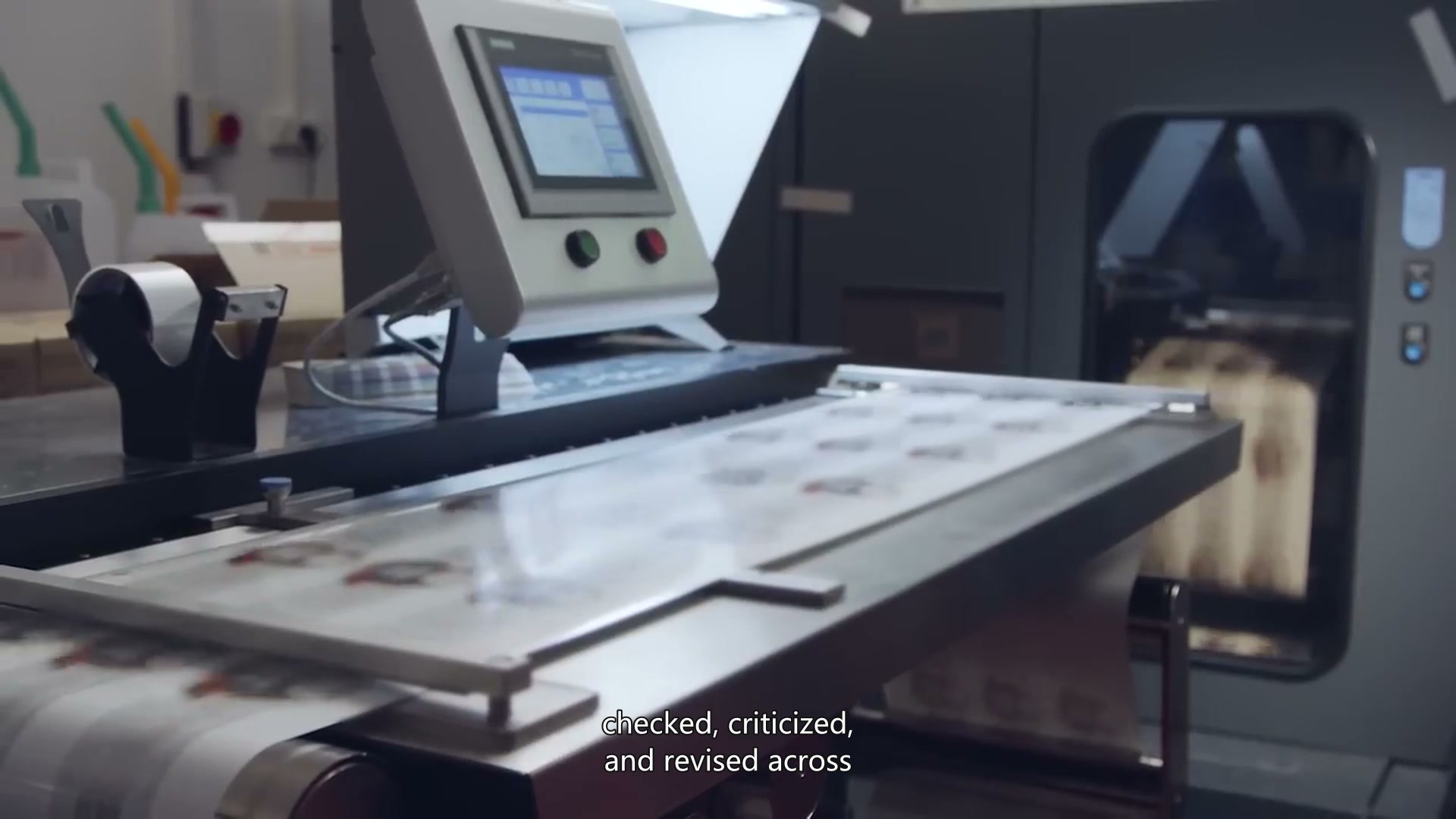}{02:59.9}%
}{fig:video-printing_and_knowledge}
\GalleryExample{How Libraries Preserve and Organize Knowledge}{3:38}{%
  \GalleryFrame{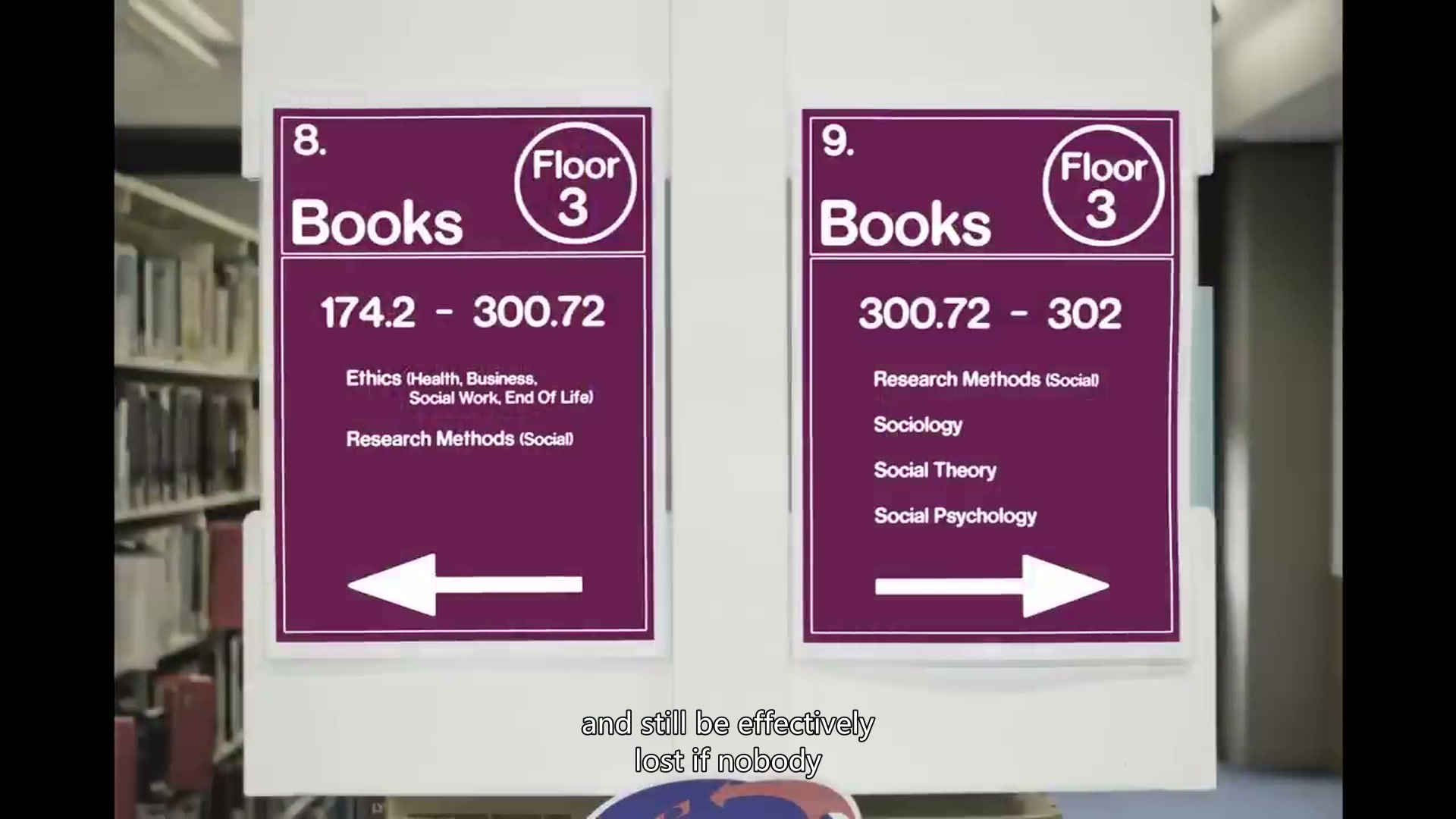}{00:04.5}\hfill%
  \GalleryFrame{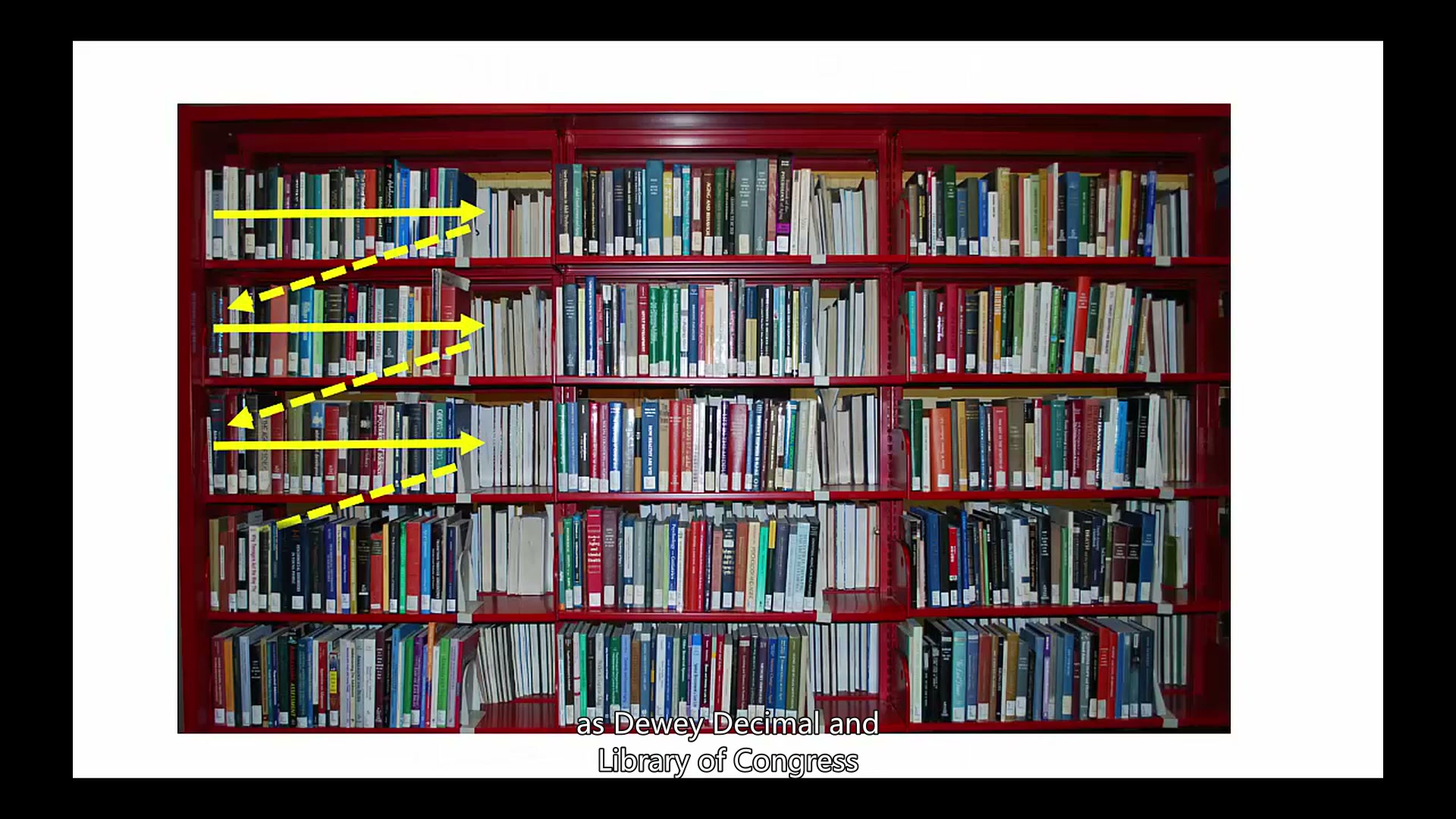}{01:17.4}\hfill%
  \GalleryFrame{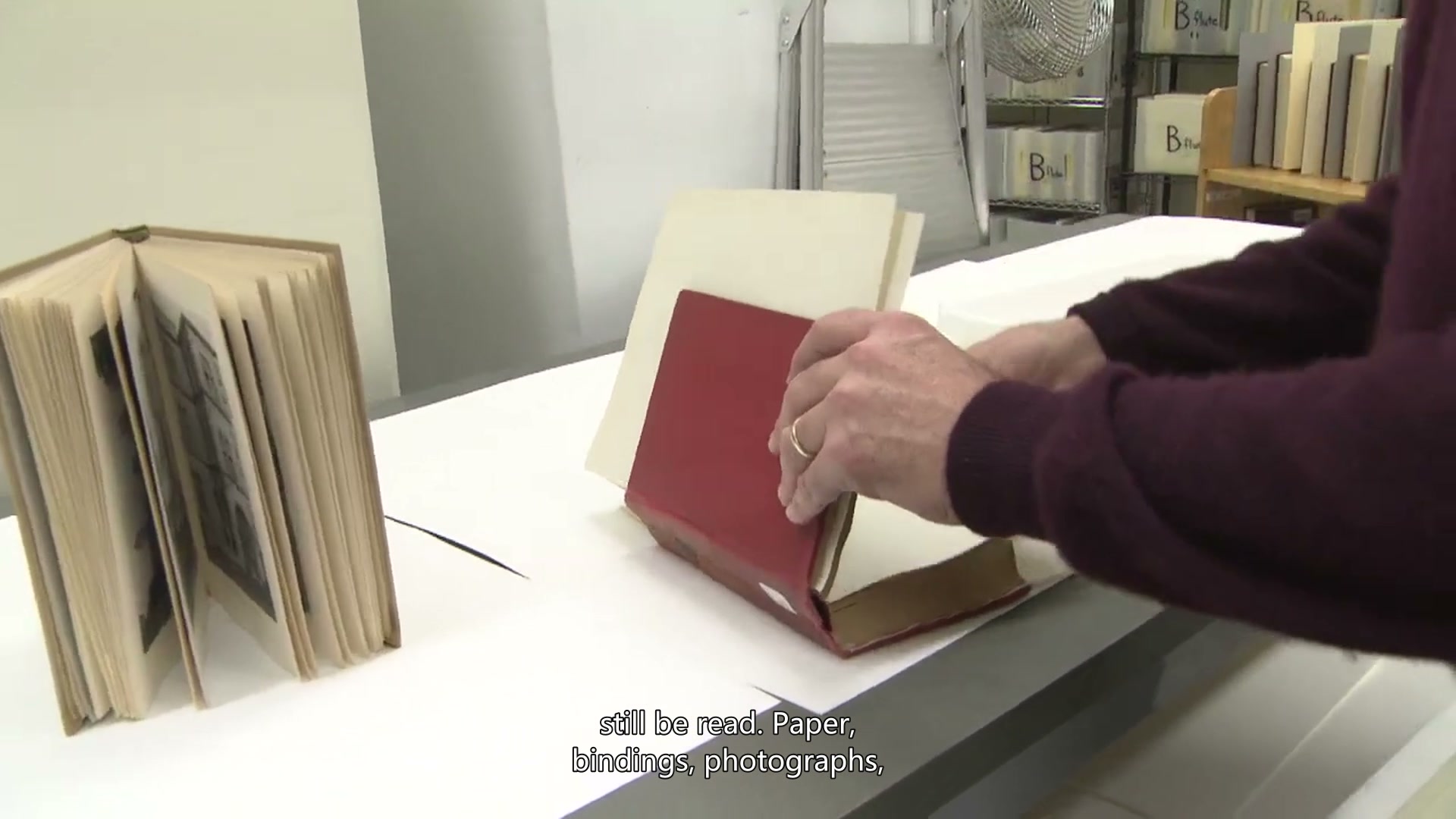}{01:35.6}\hfill%
  \GalleryFrame{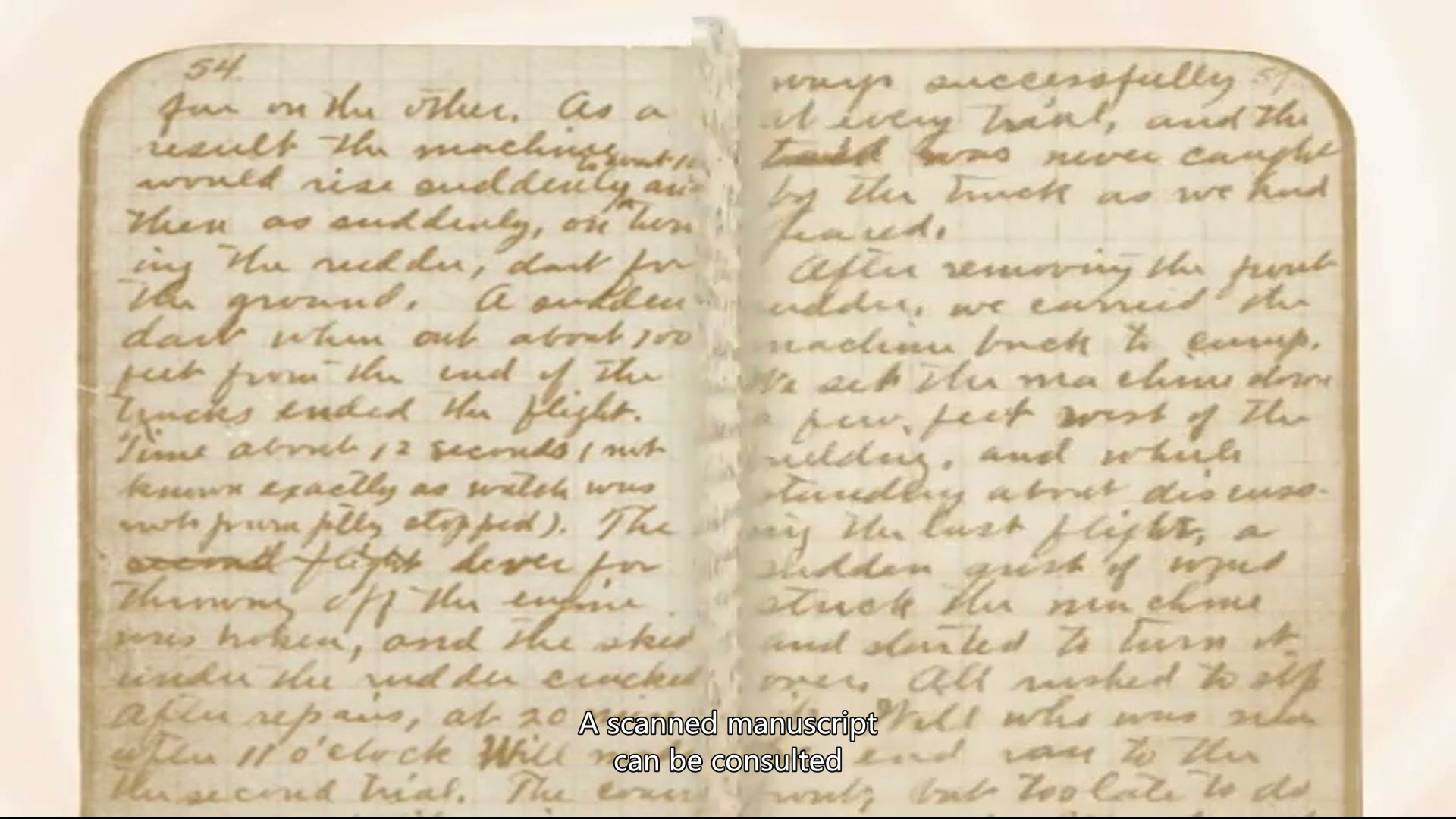}{02:12.1}%
}{fig:video-libraries_and_knowledge}
\GalleryExample{Why Classic Film Posters Are Instantly Recognizable}{3:22}{%
  \GalleryFrame{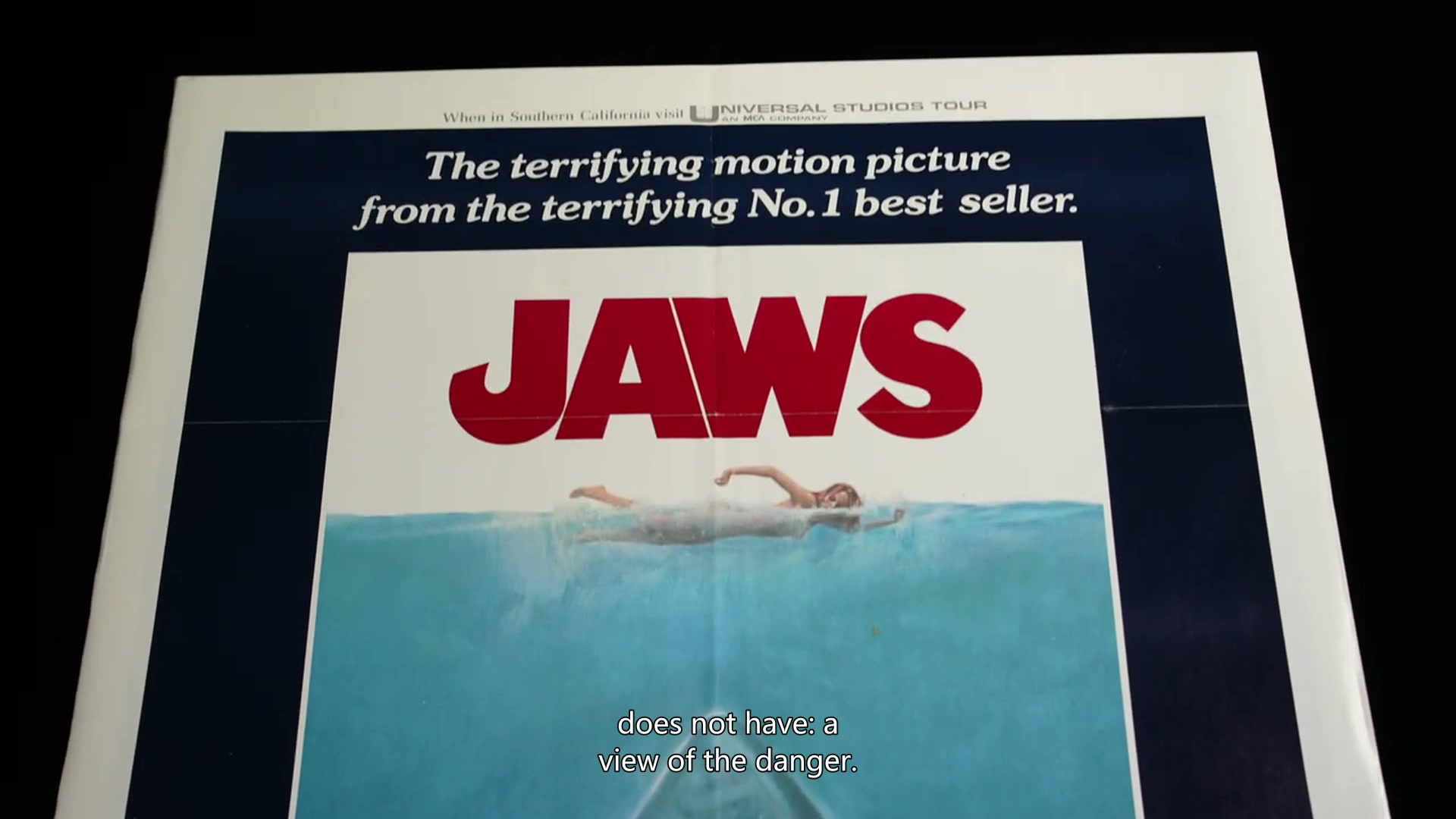}{00:12.6}\hfill%
  \GalleryFrame{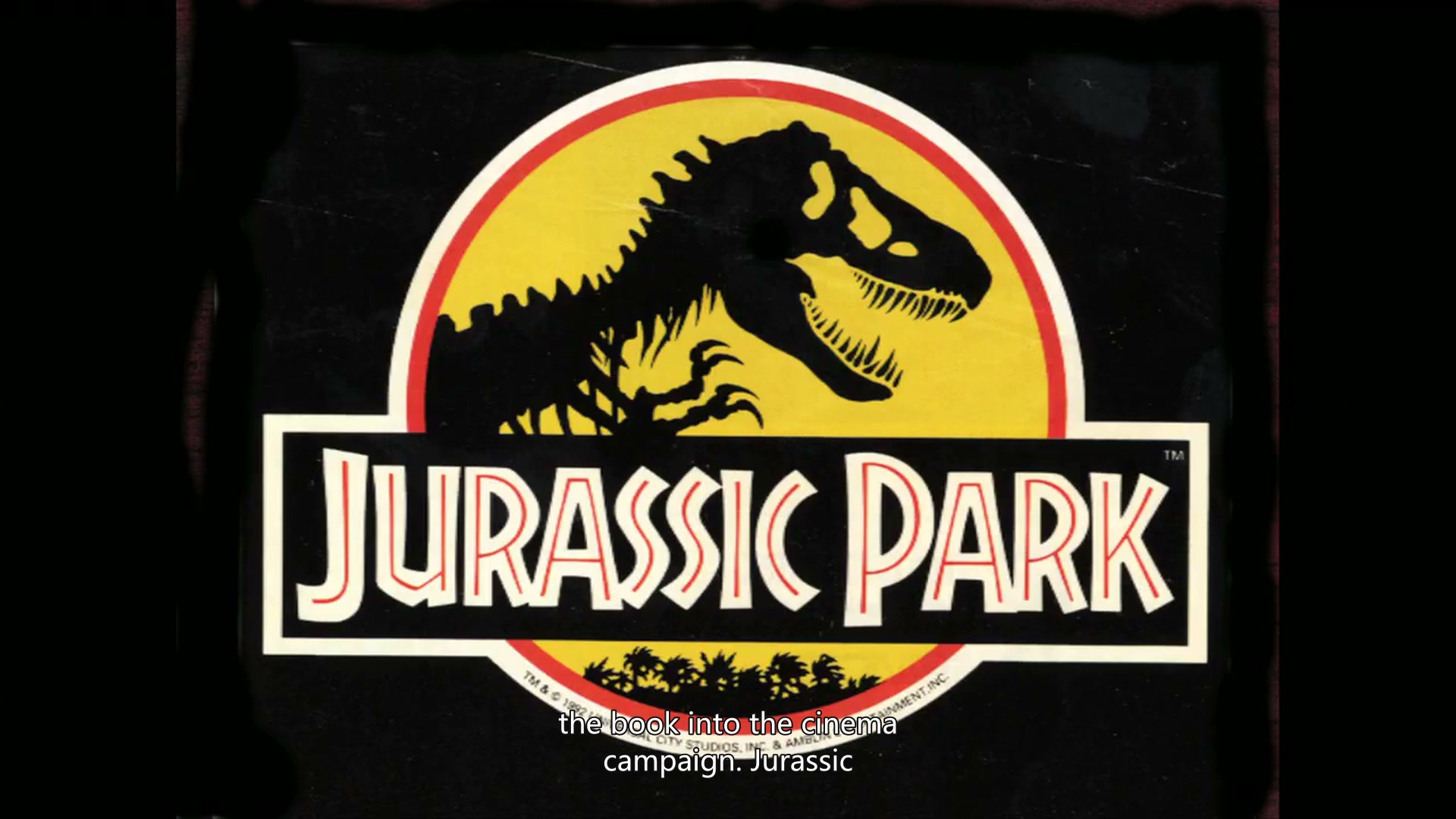}{01:20.1}\hfill%
  \GalleryFrame{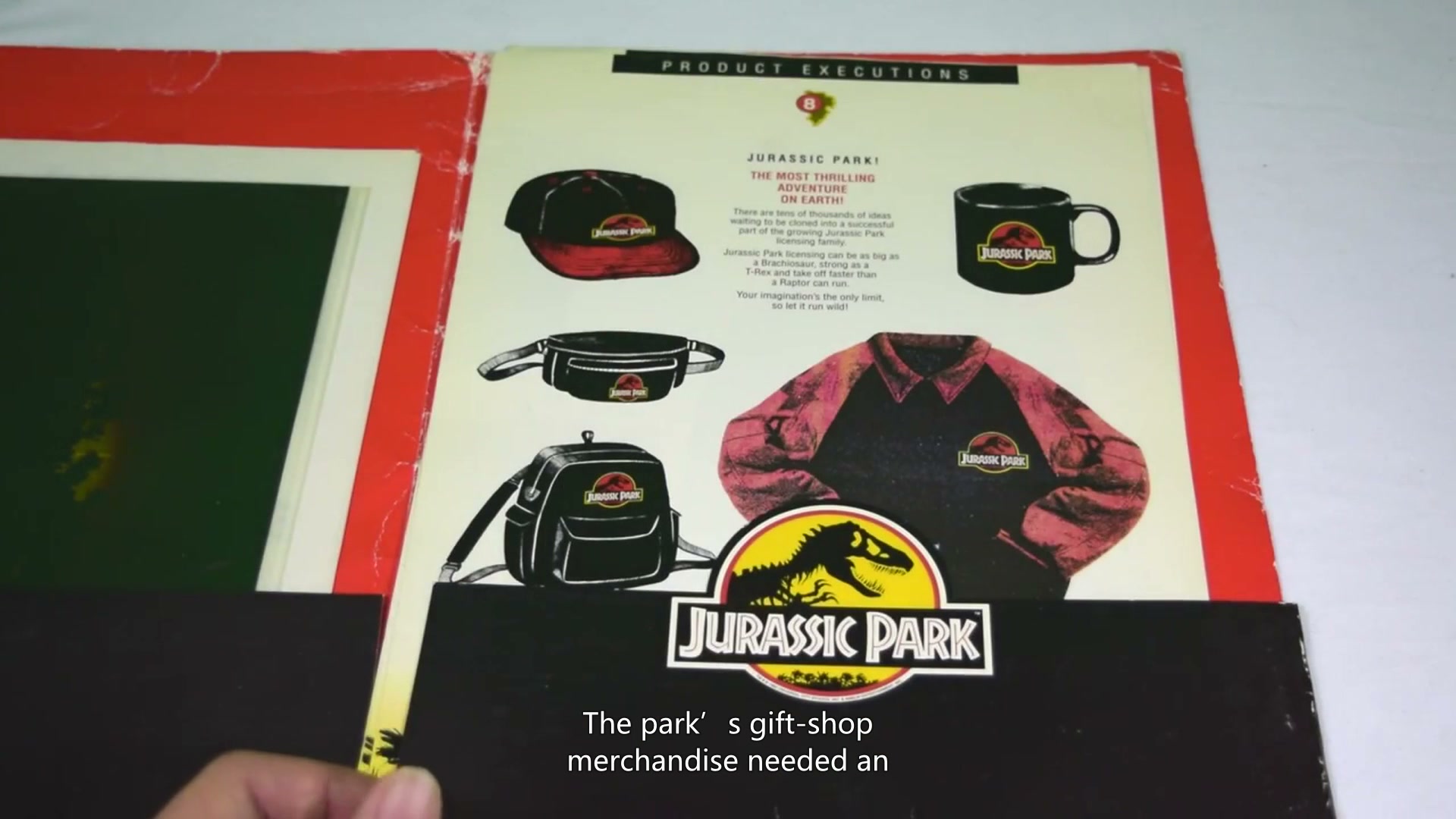}{02:02.3}\hfill%
  \GalleryFrame{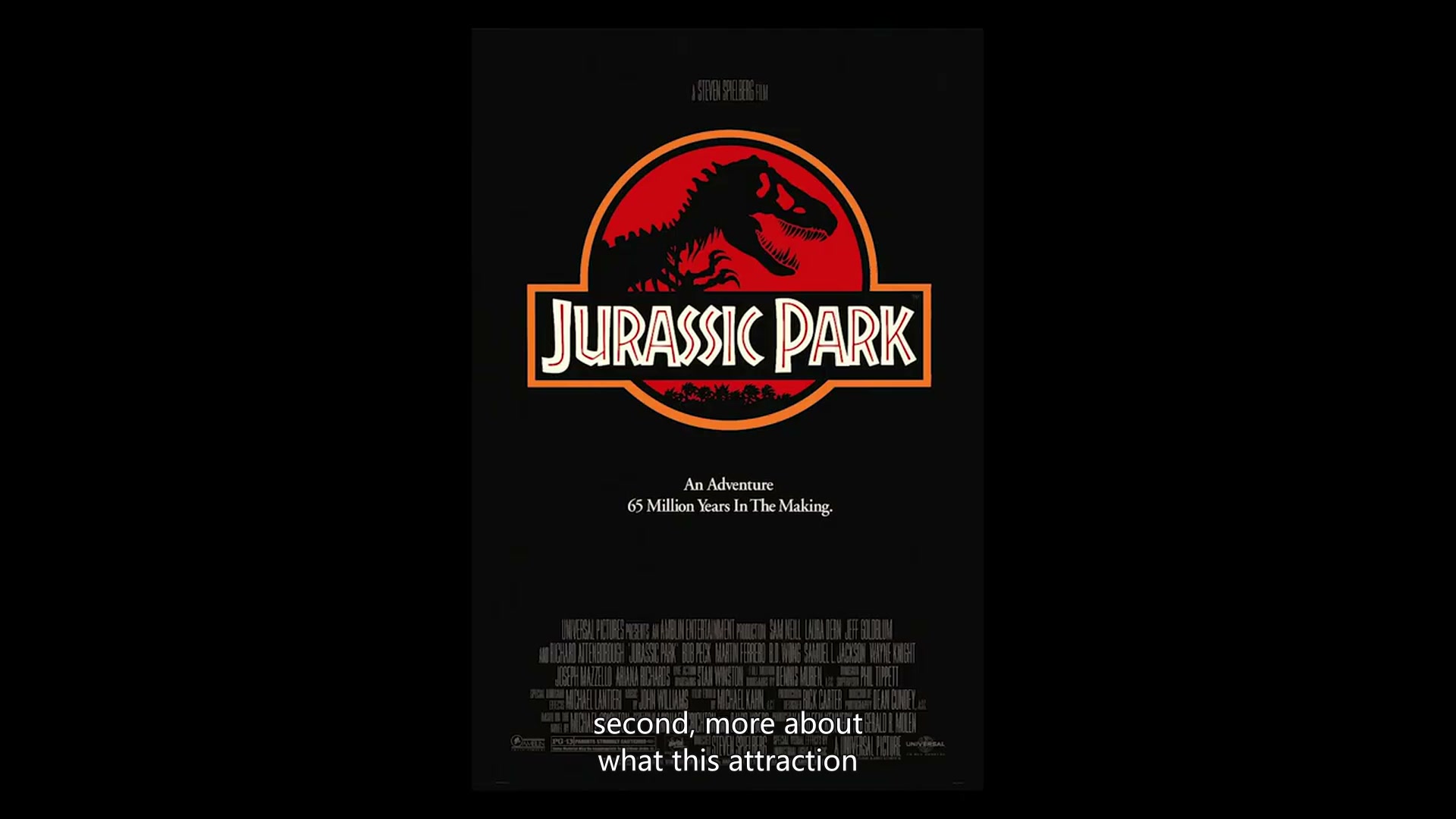}{02:36.1}%
}{fig:video-classic_film_posters}
\GalleryExample{The Titanic: Historical Facts and Movie Portrayals}{3:34}{%
  \GalleryFrame{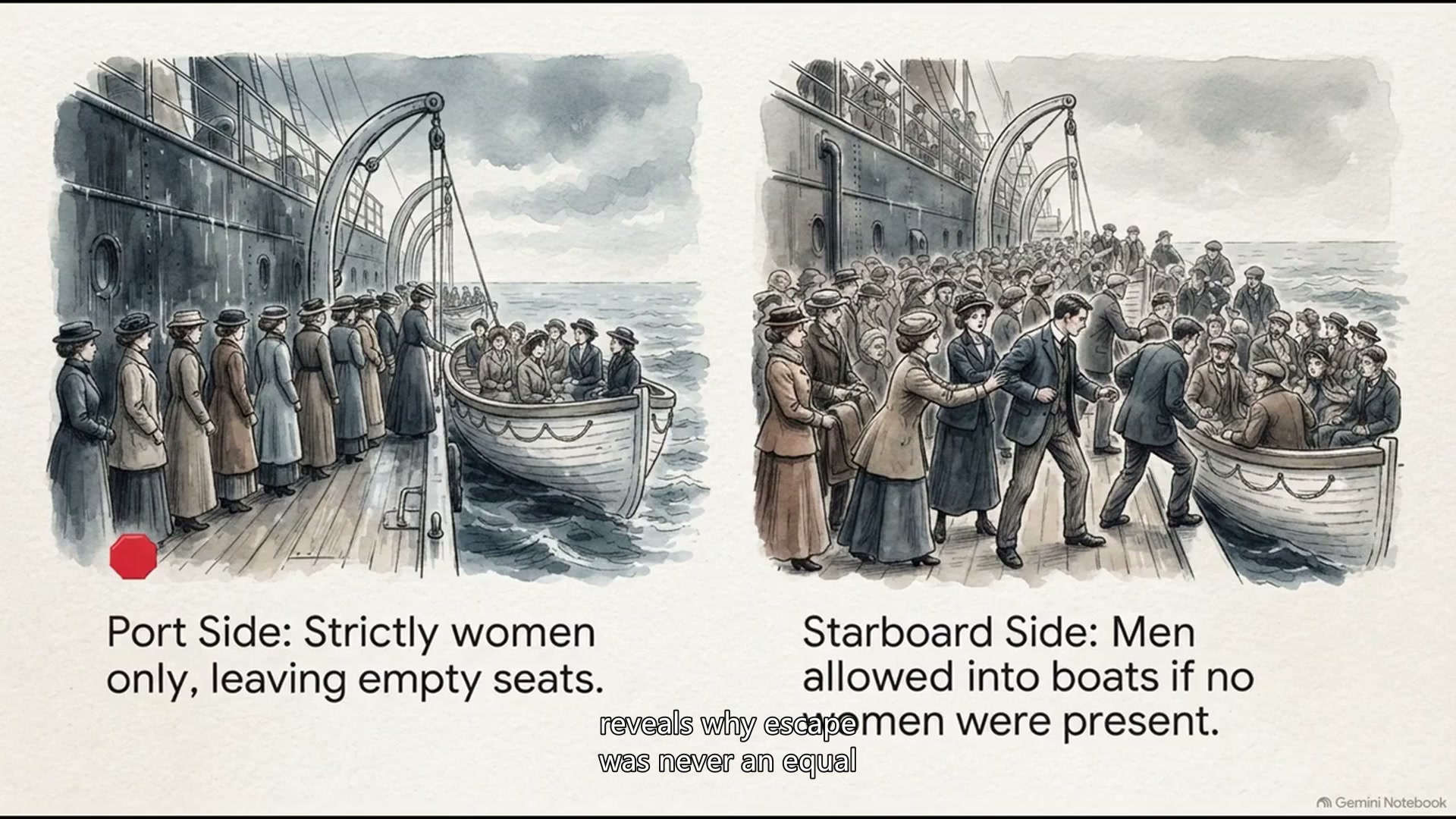}{00:22.4}\hfill%
  \GalleryFrame{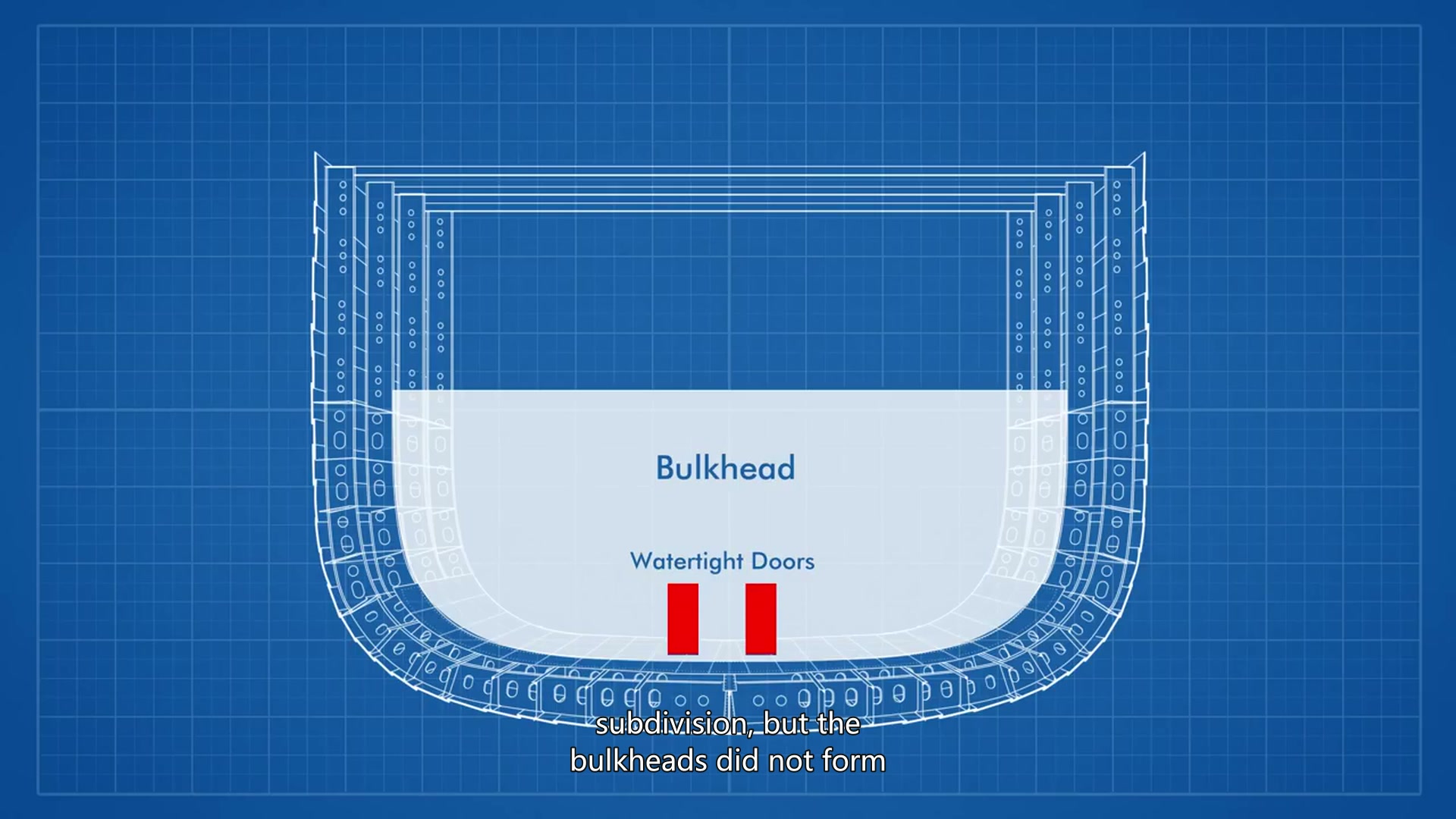}{01:16.1}\hfill%
  \GalleryFrame{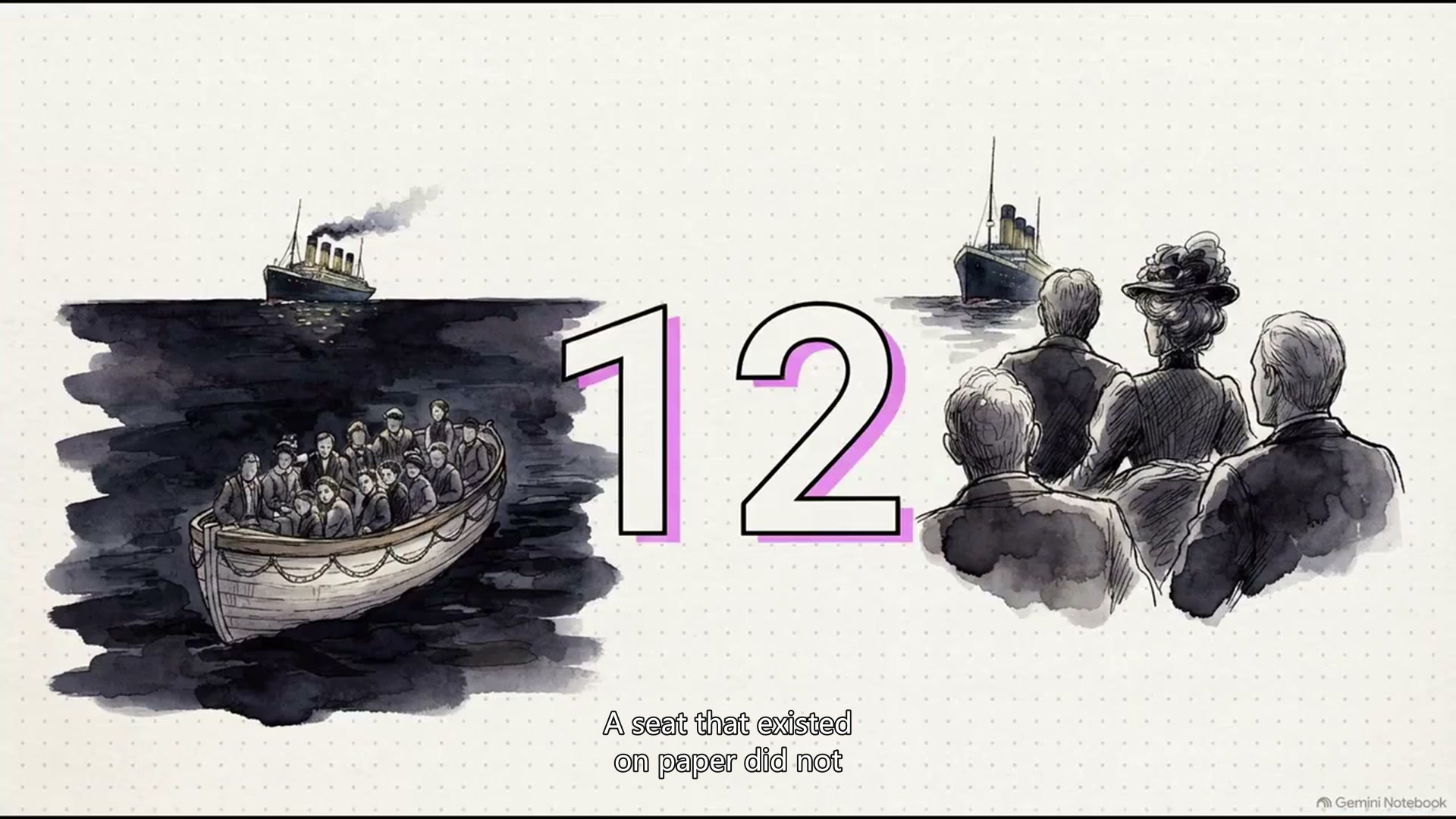}{02:09.8}\hfill%
  \GalleryFrame{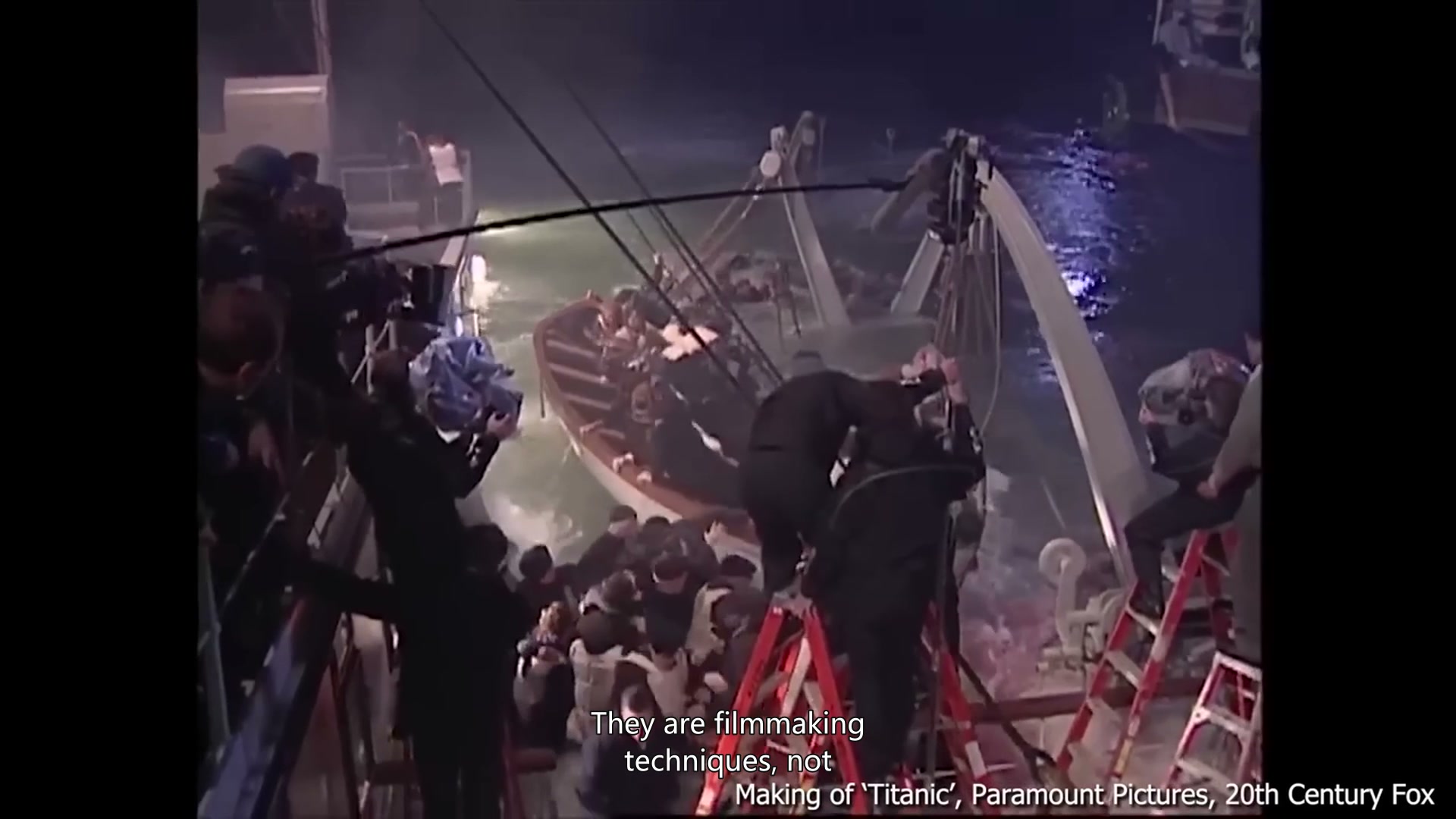}{03:03.5}%
}{fig:video-titanic_history_and_film}
\GalleryExample{Comparing Climate Across Asia}{3:39}{%
  \GalleryFrame{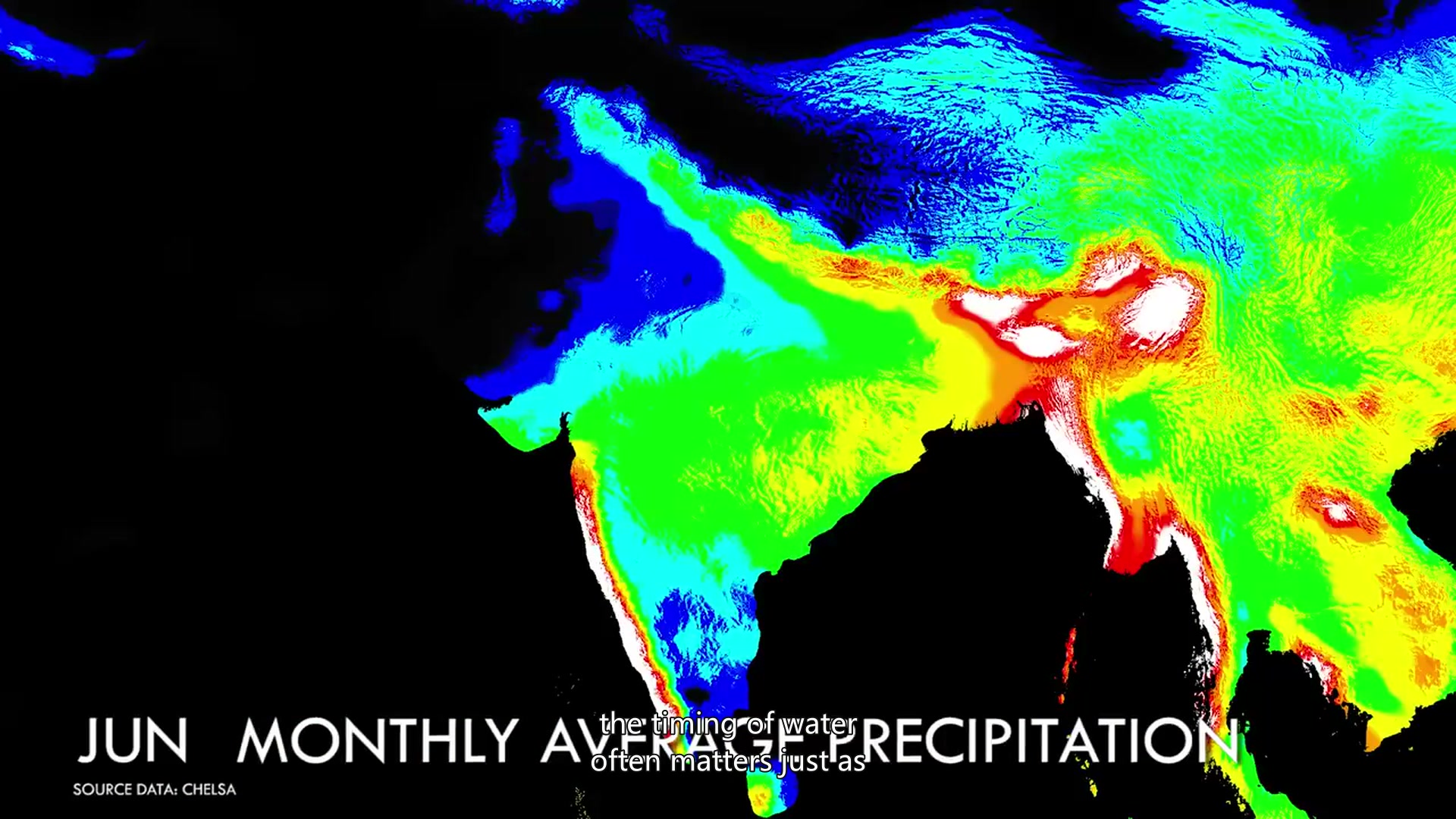}{00:22.8}\hfill%
  \GalleryFrame{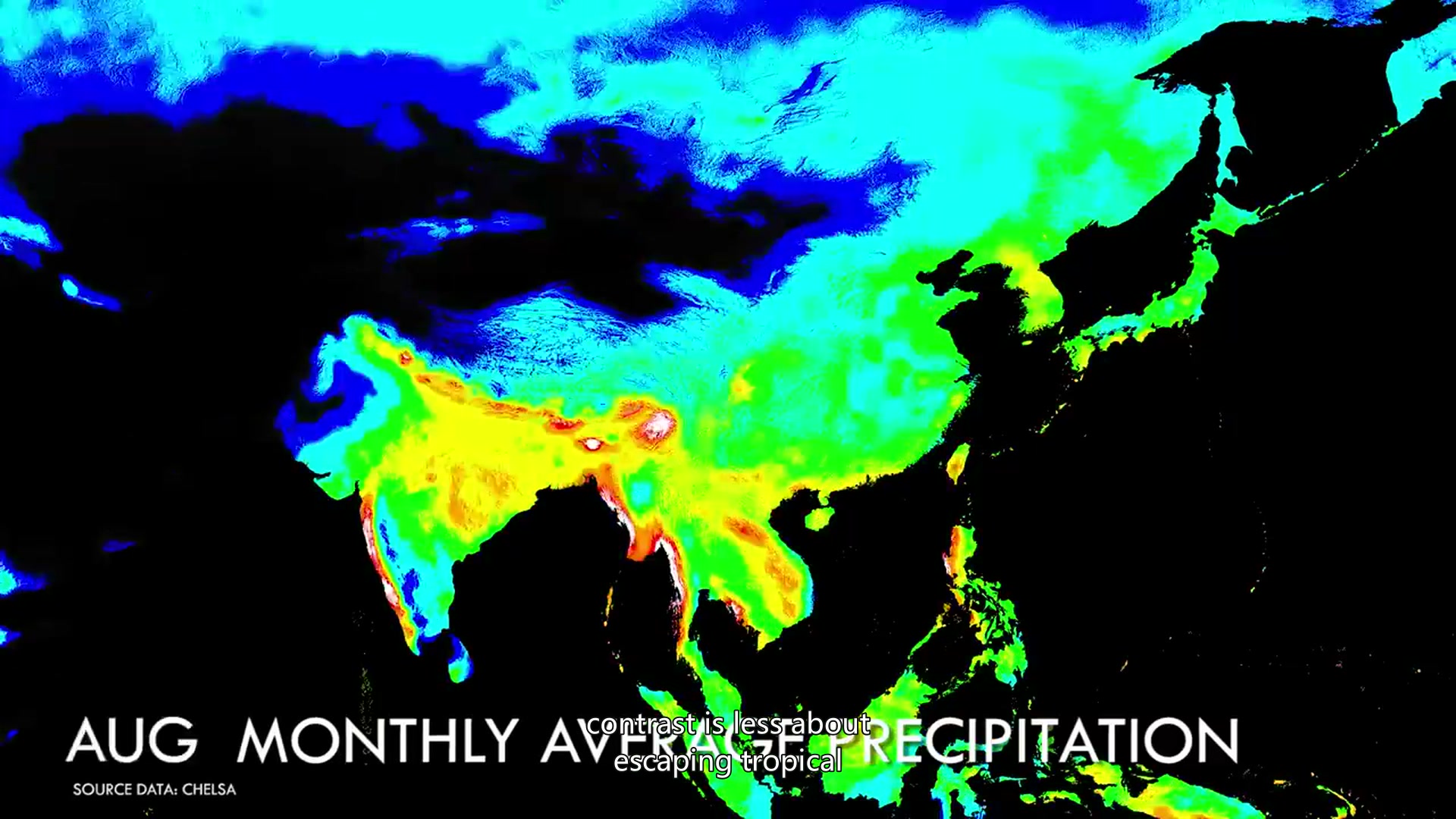}{01:17.5}\hfill%
  \GalleryFrame{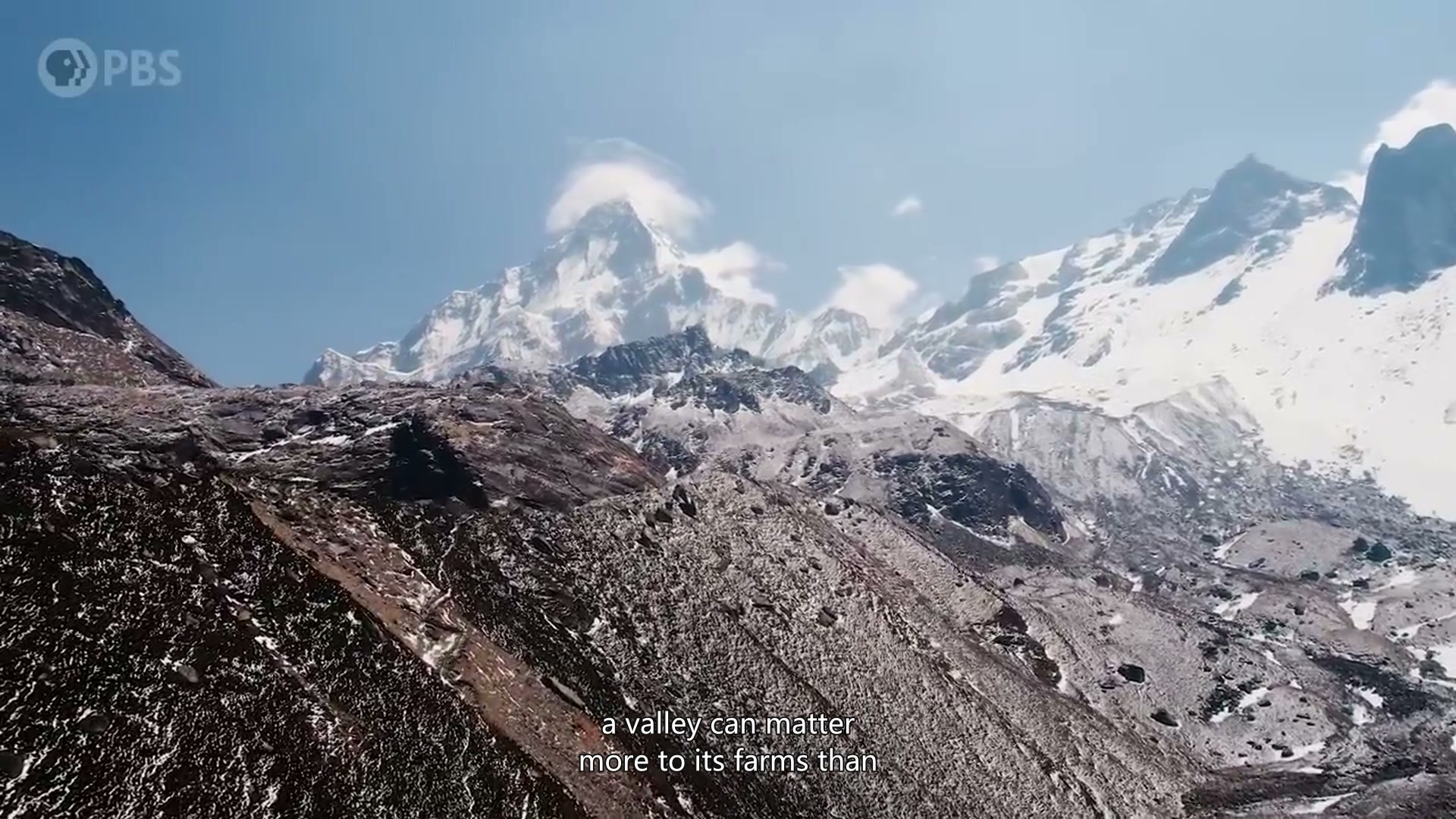}{02:12.3}\hfill%
  \GalleryFrame{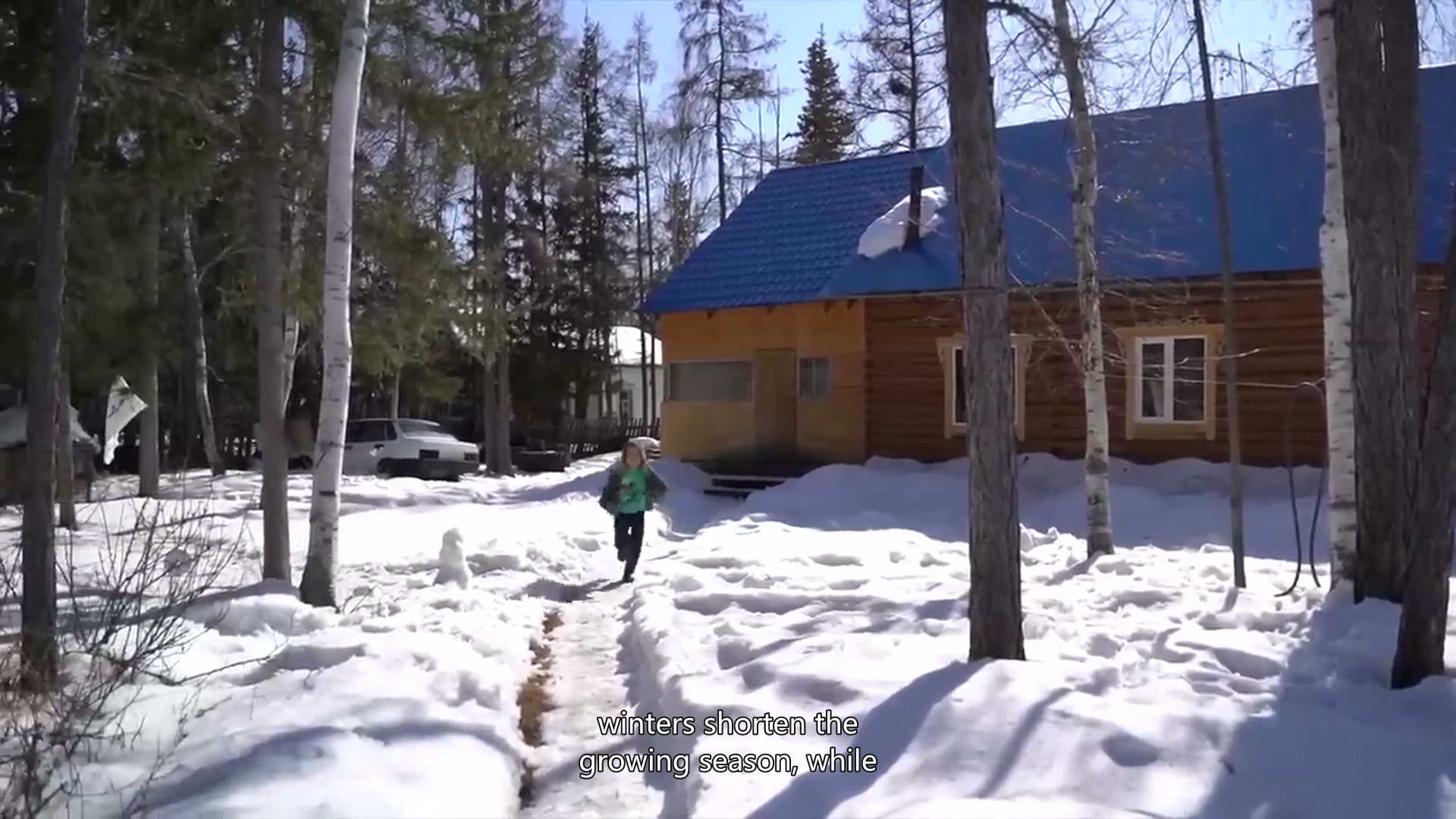}{03:07.0}%
}{fig:video-asian_climate_comparison}
\GalleryExample{Why Global Supply Chains Develop Bottlenecks}{3:43}{%
  \GalleryFrame{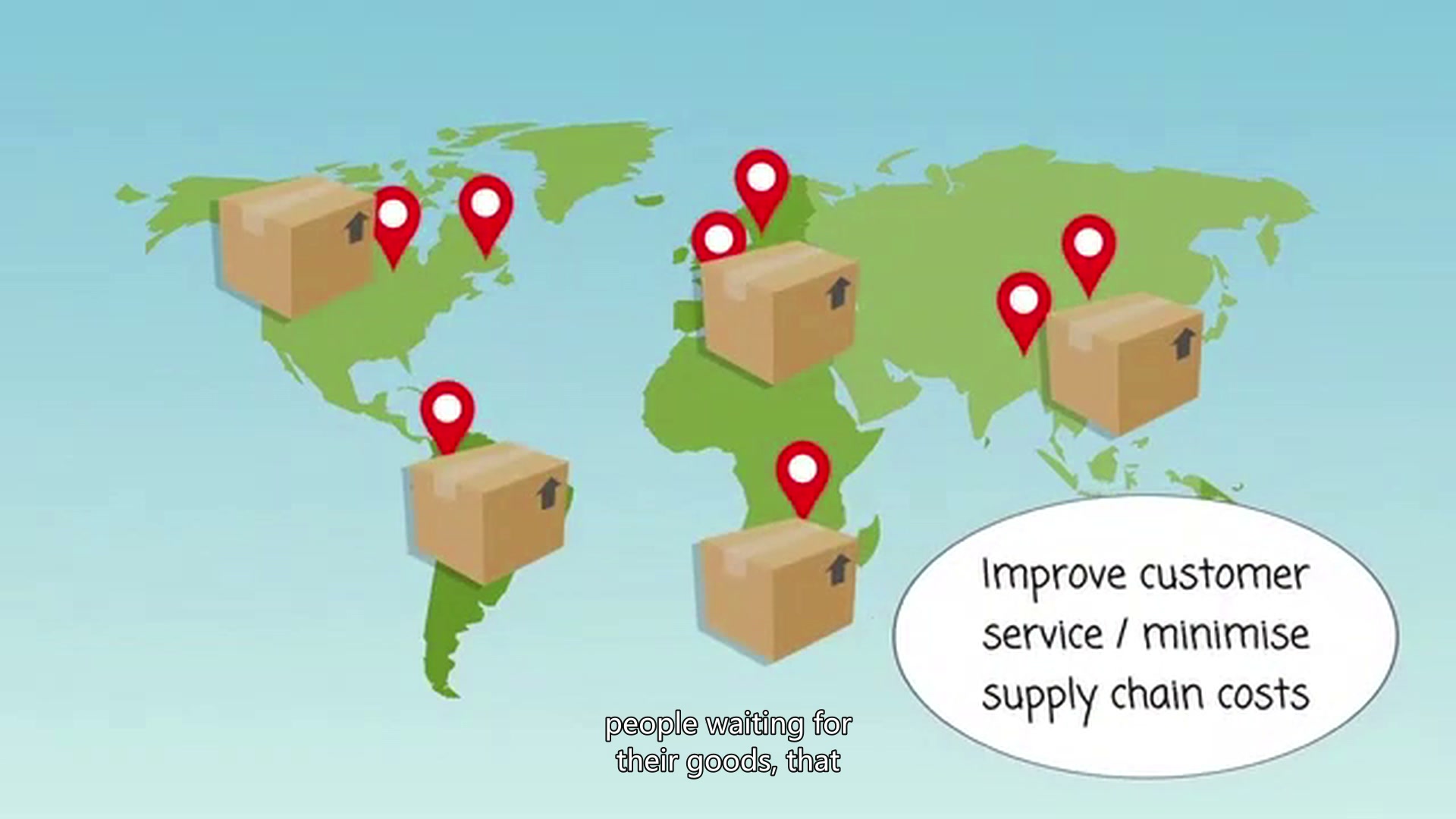}{00:23.2}\hfill%
  \GalleryFrame{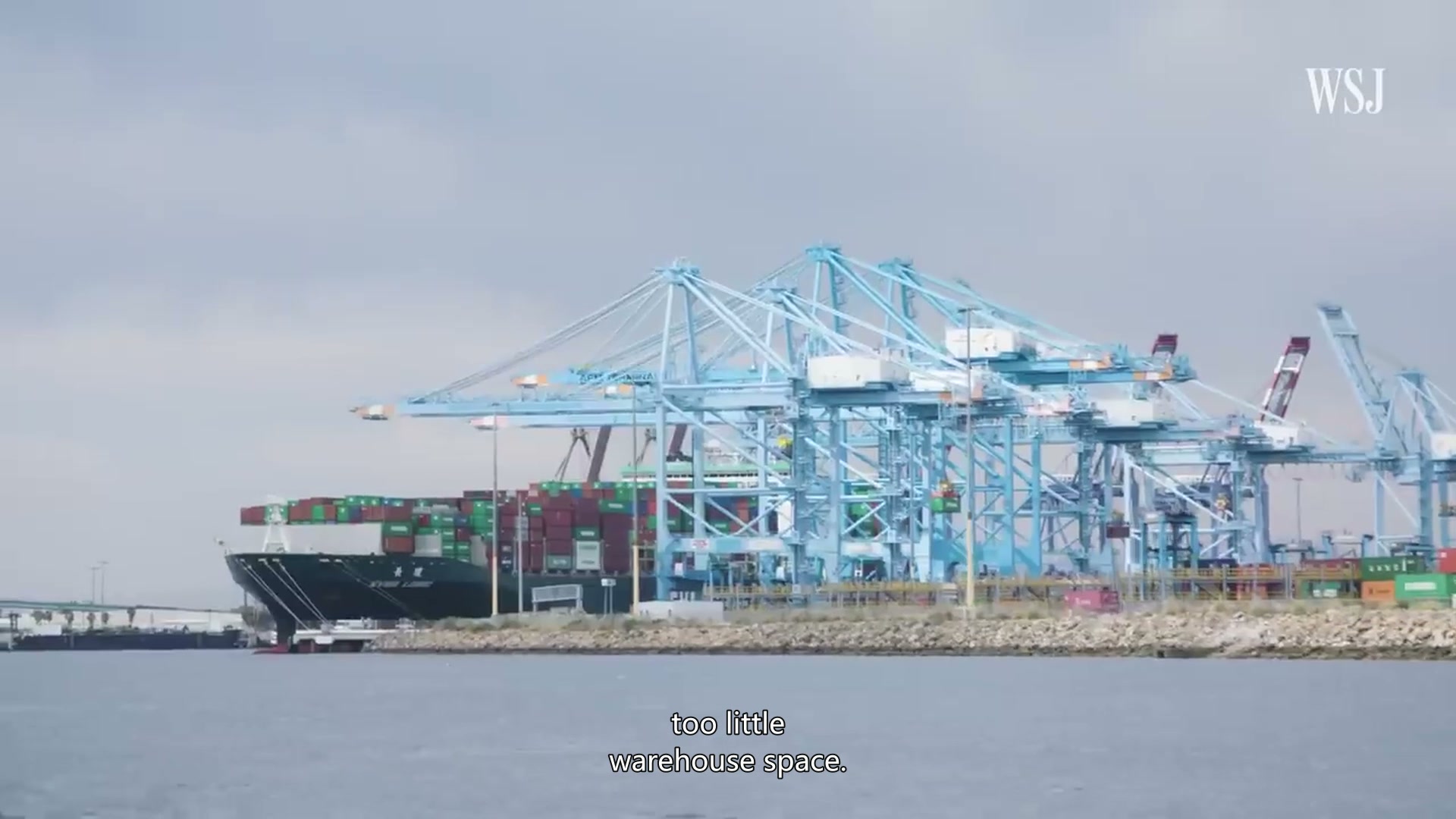}{01:19.0}\hfill%
  \GalleryFrame{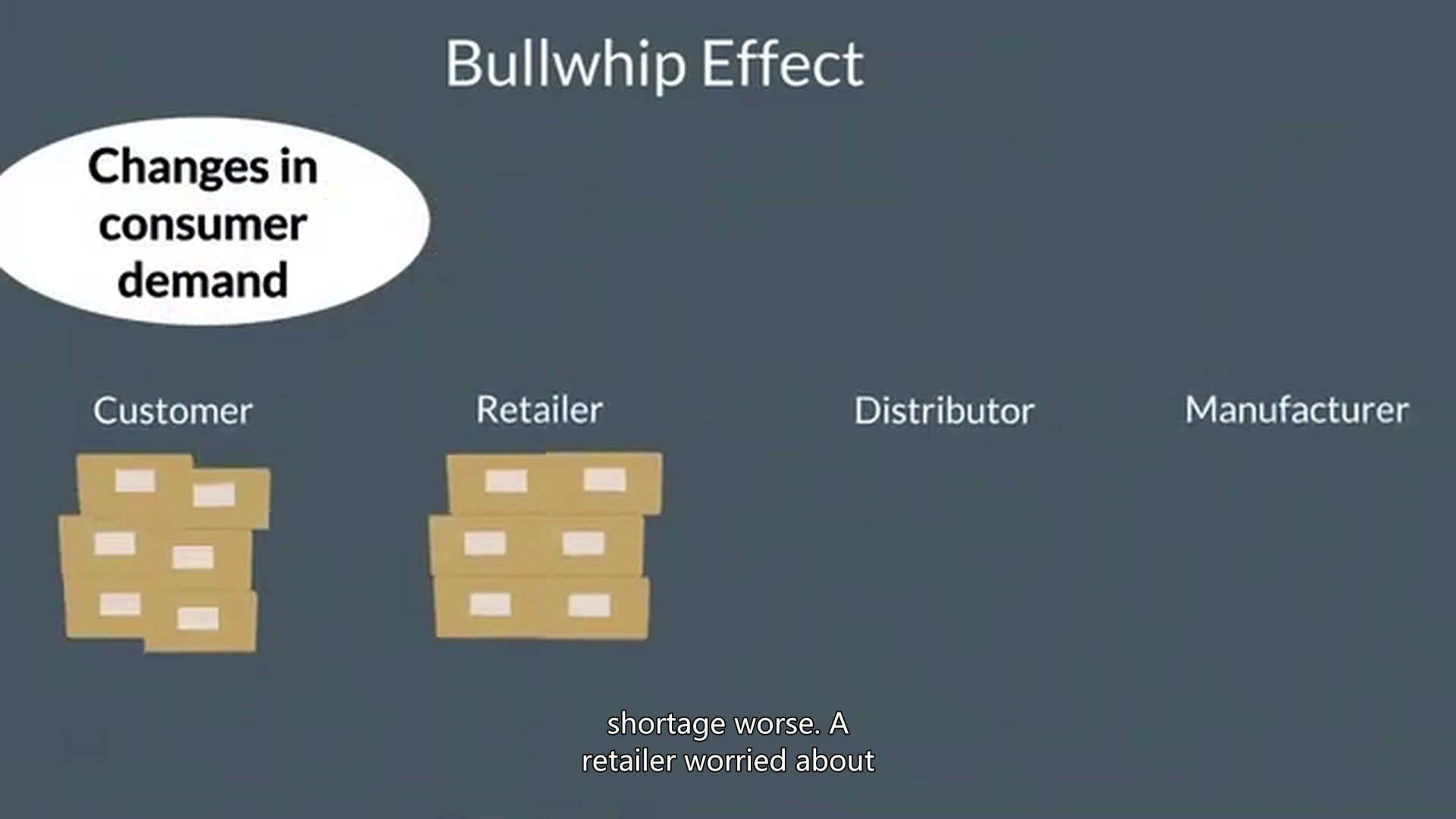}{02:14.8}\hfill%
  \GalleryFrame{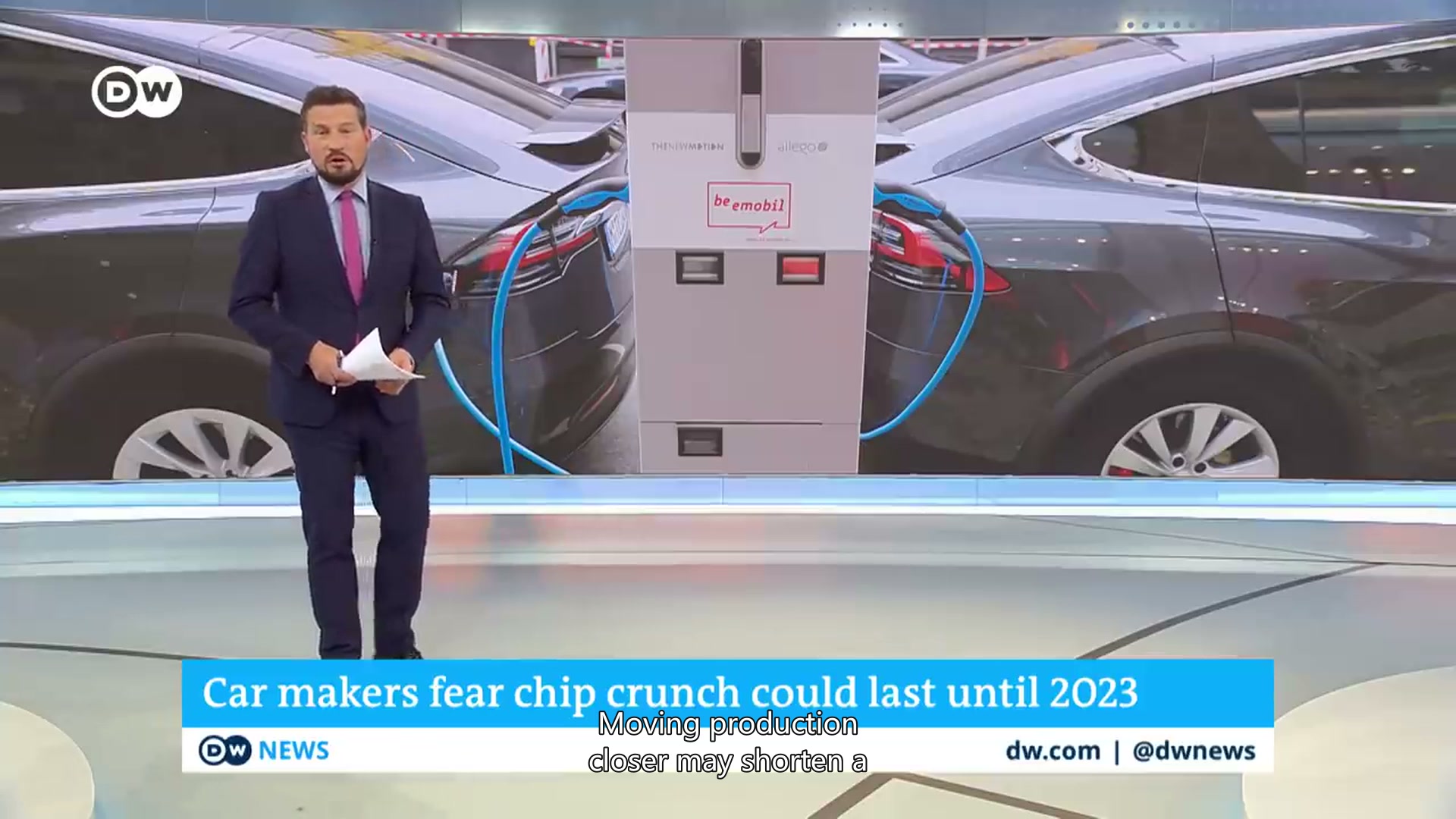}{03:10.6}%
}{fig:video-supply_chain_bottlenecks}
\GalleryExample{E-Waste: Collection, Sorting, Dismantling, and Metal Recovery}{3:31}{%
  \GalleryFrame{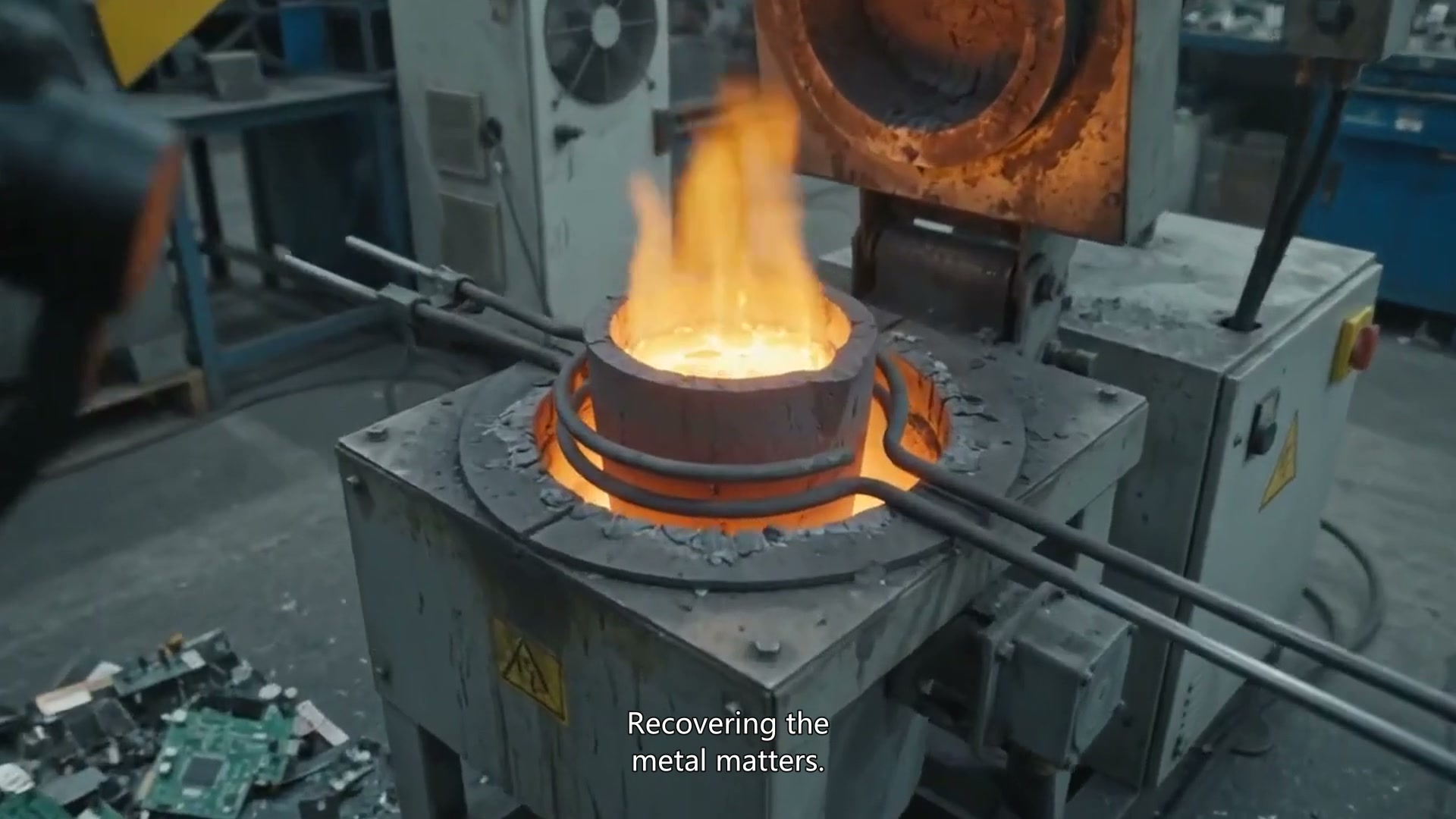}{00:22.1}\hfill%
  \GalleryFrame{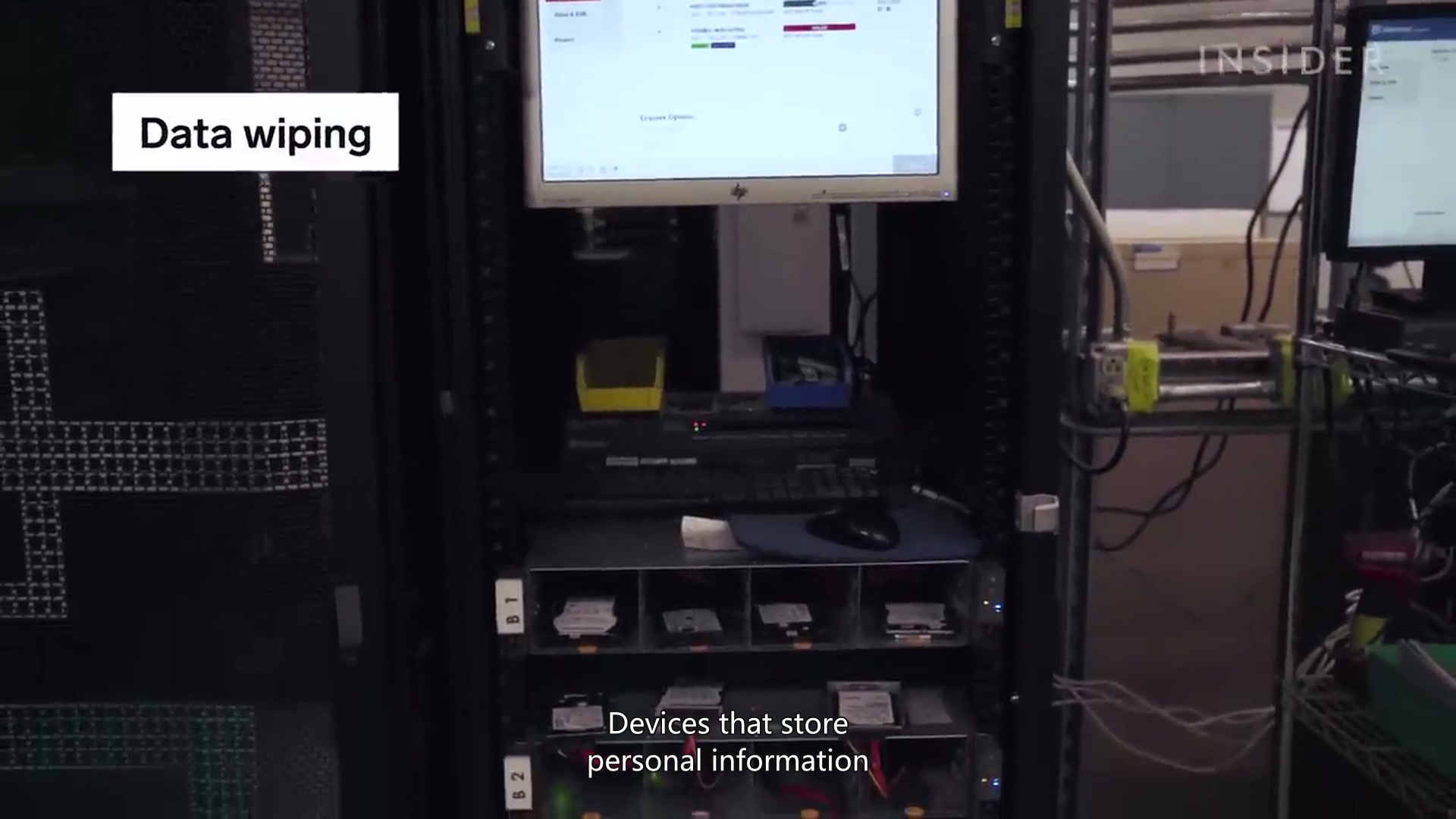}{01:15.0}\hfill%
  \GalleryFrame{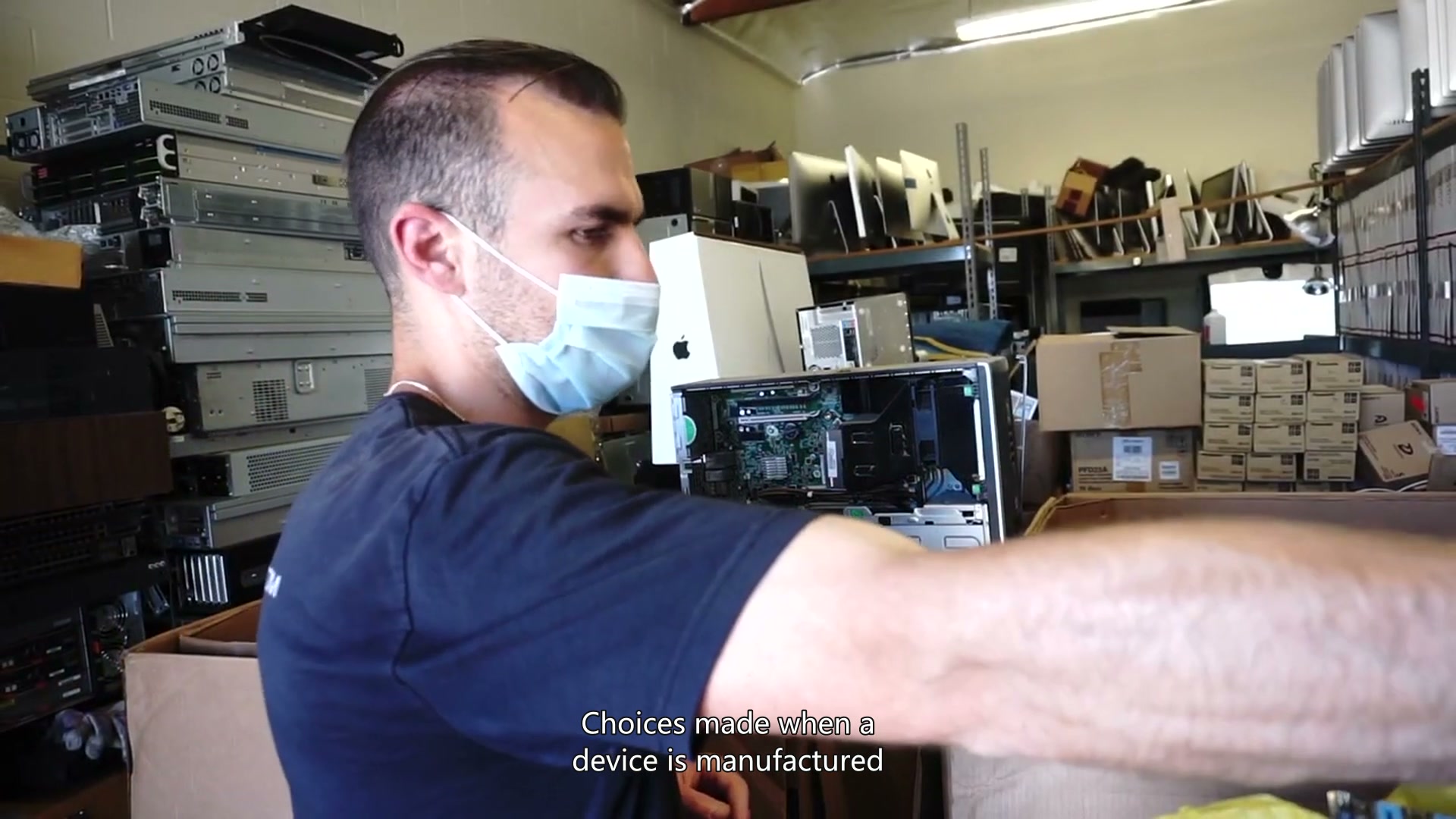}{02:08.0}\hfill%
  \GalleryFrame{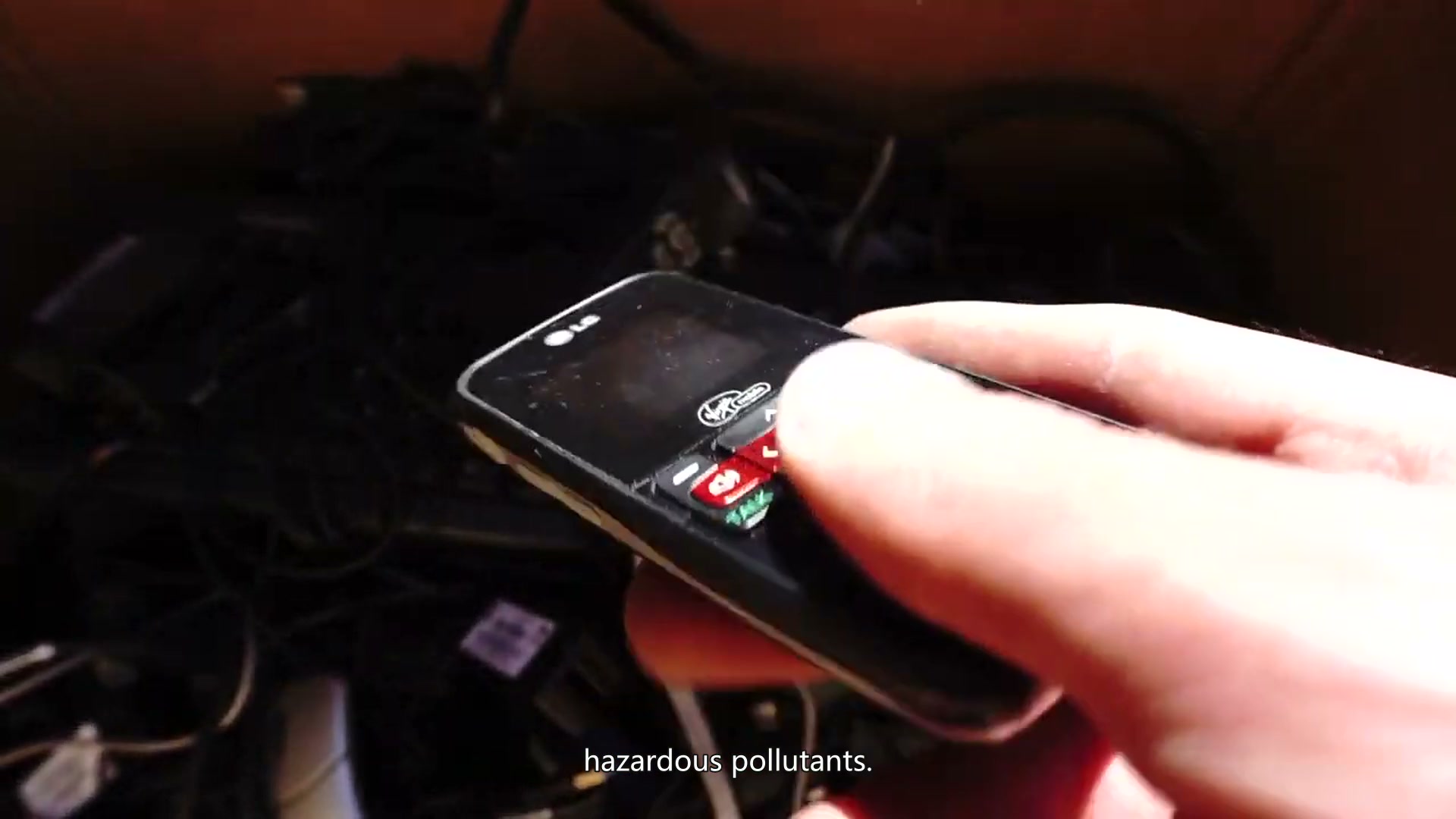}{03:01.0}%
}{fig:video-electronic_waste_recycling}
\GalleryExample{How the Internet Became Everyday Infrastructure}{3:53}{%
  \GalleryFrame{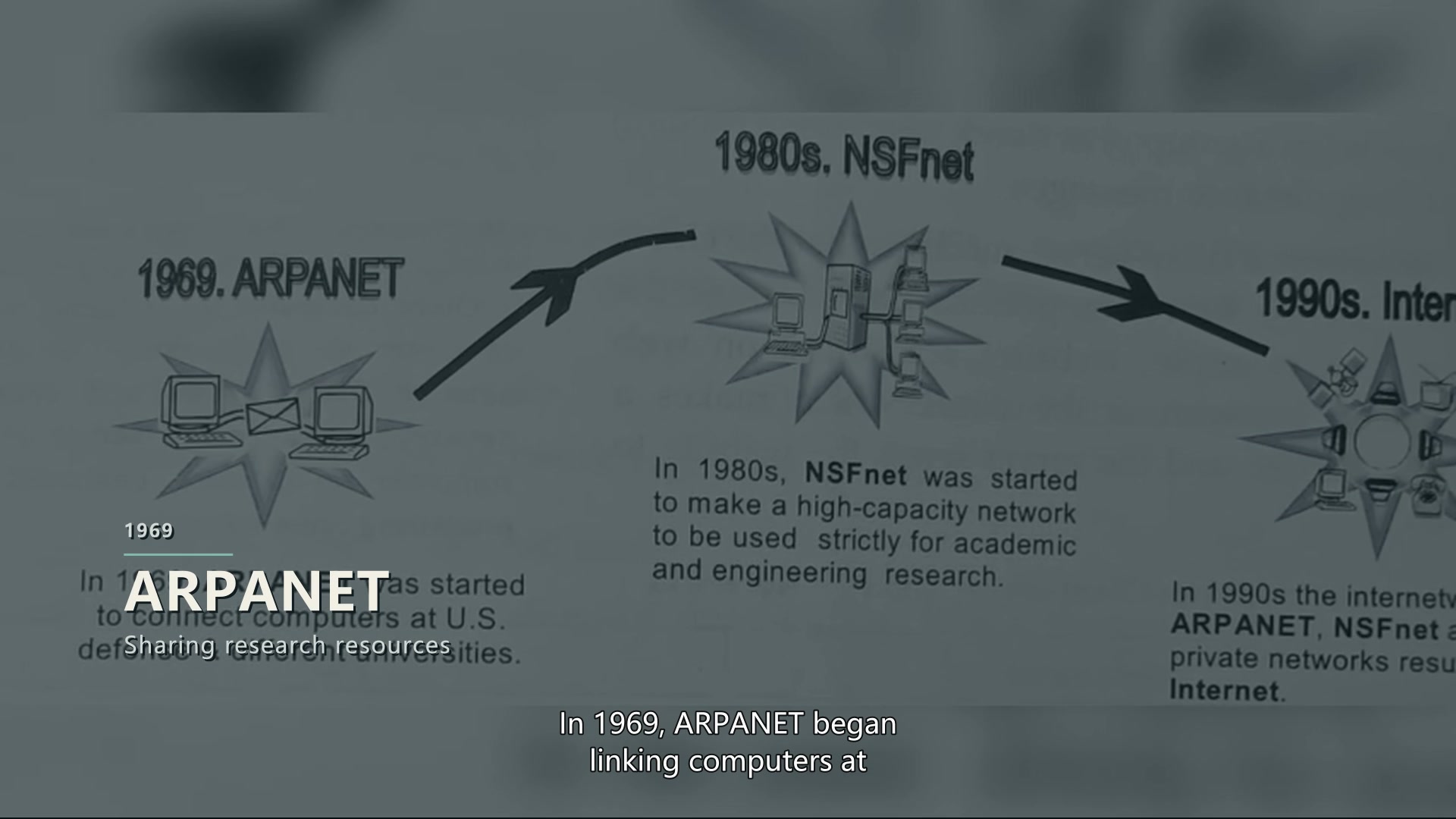}{00:24.4}\hfill%
  \GalleryFrame{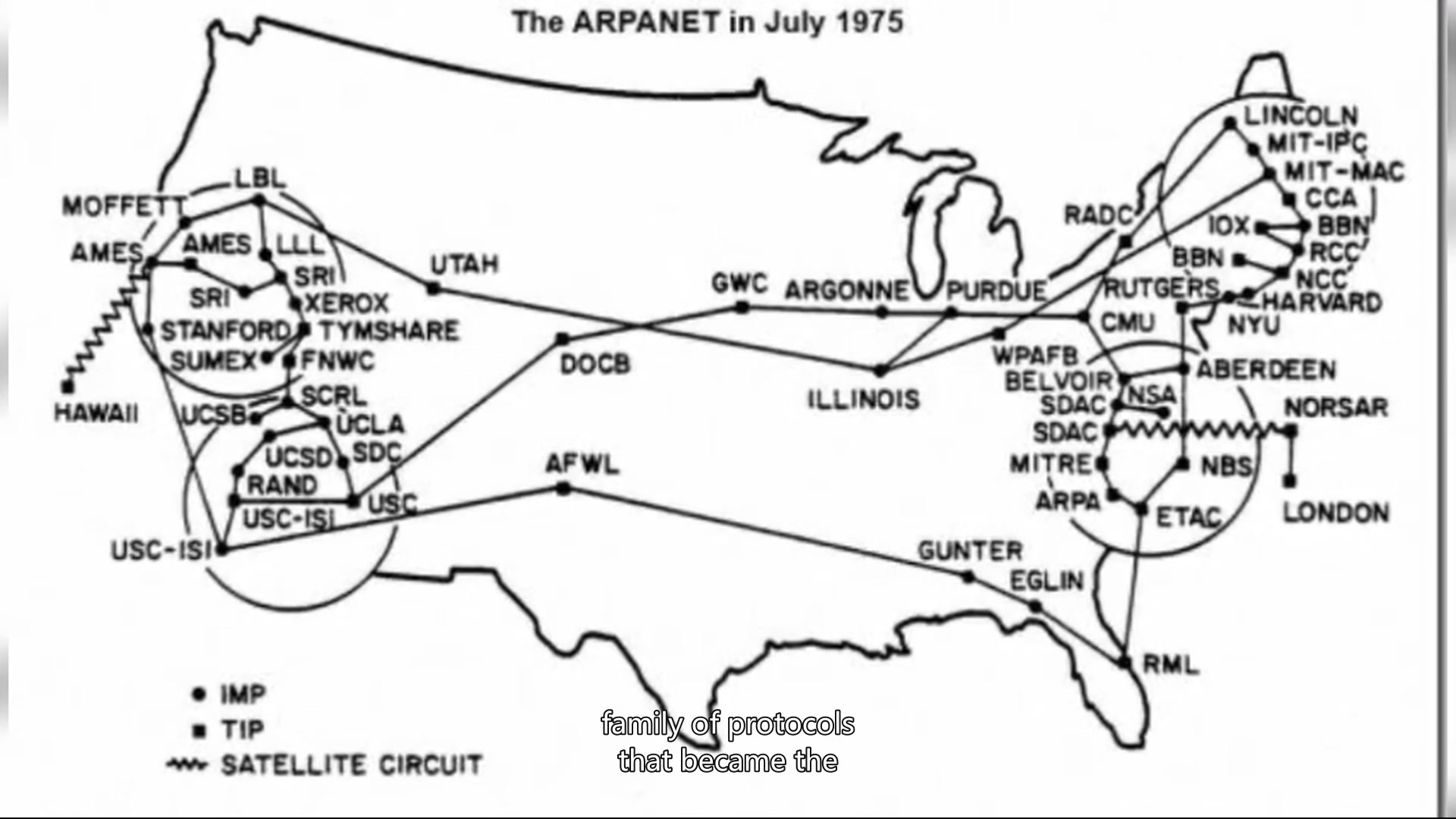}{01:22.8}\hfill%
  \GalleryFrame{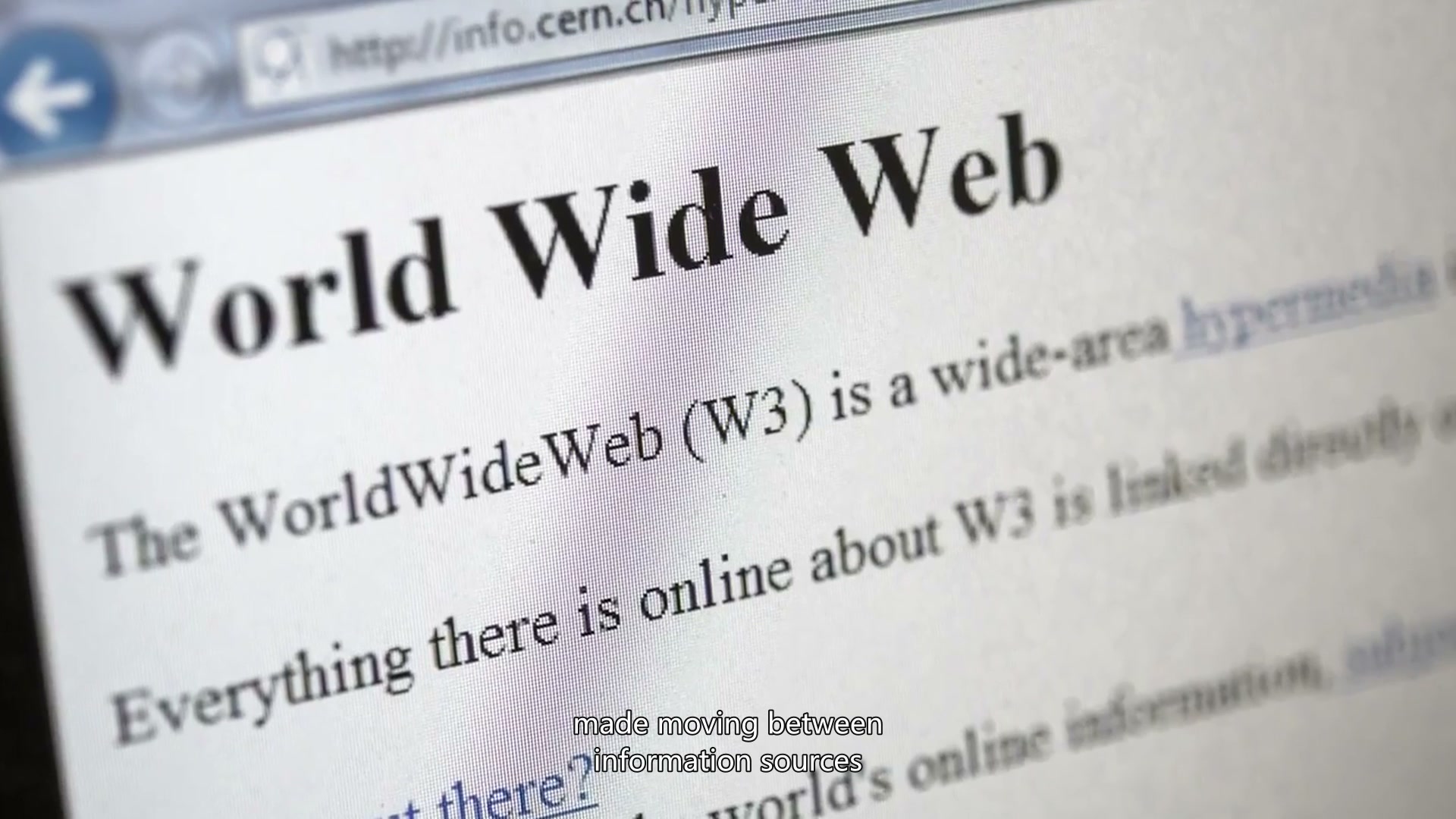}{02:21.3}\hfill%
  \GalleryFrame{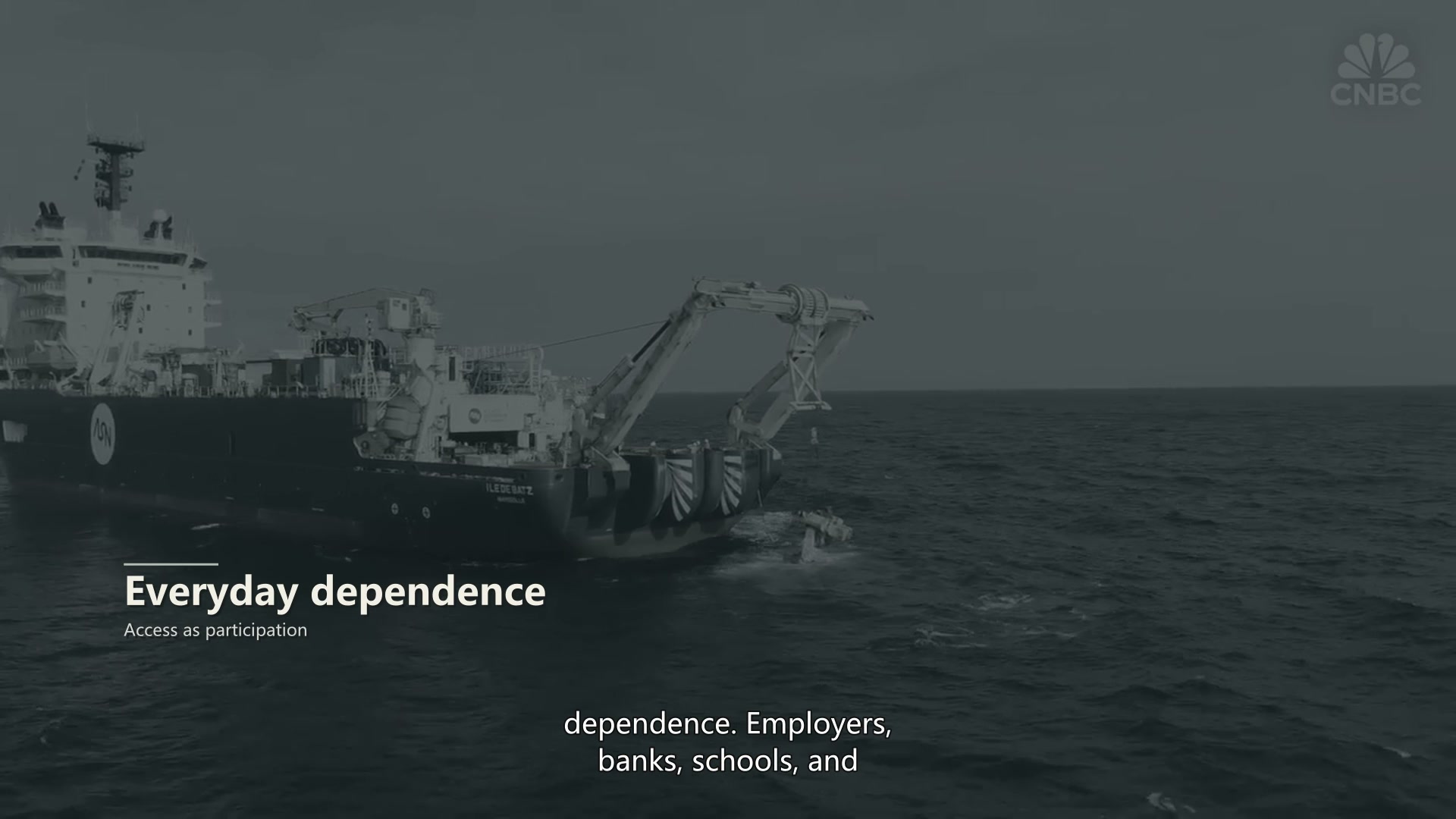}{03:19.8}%
}{fig:video-history_of_the_internet}
\GalleryExample{How Super Mario Teaches Players Through Play}{3:16}{%
  \GalleryFrame{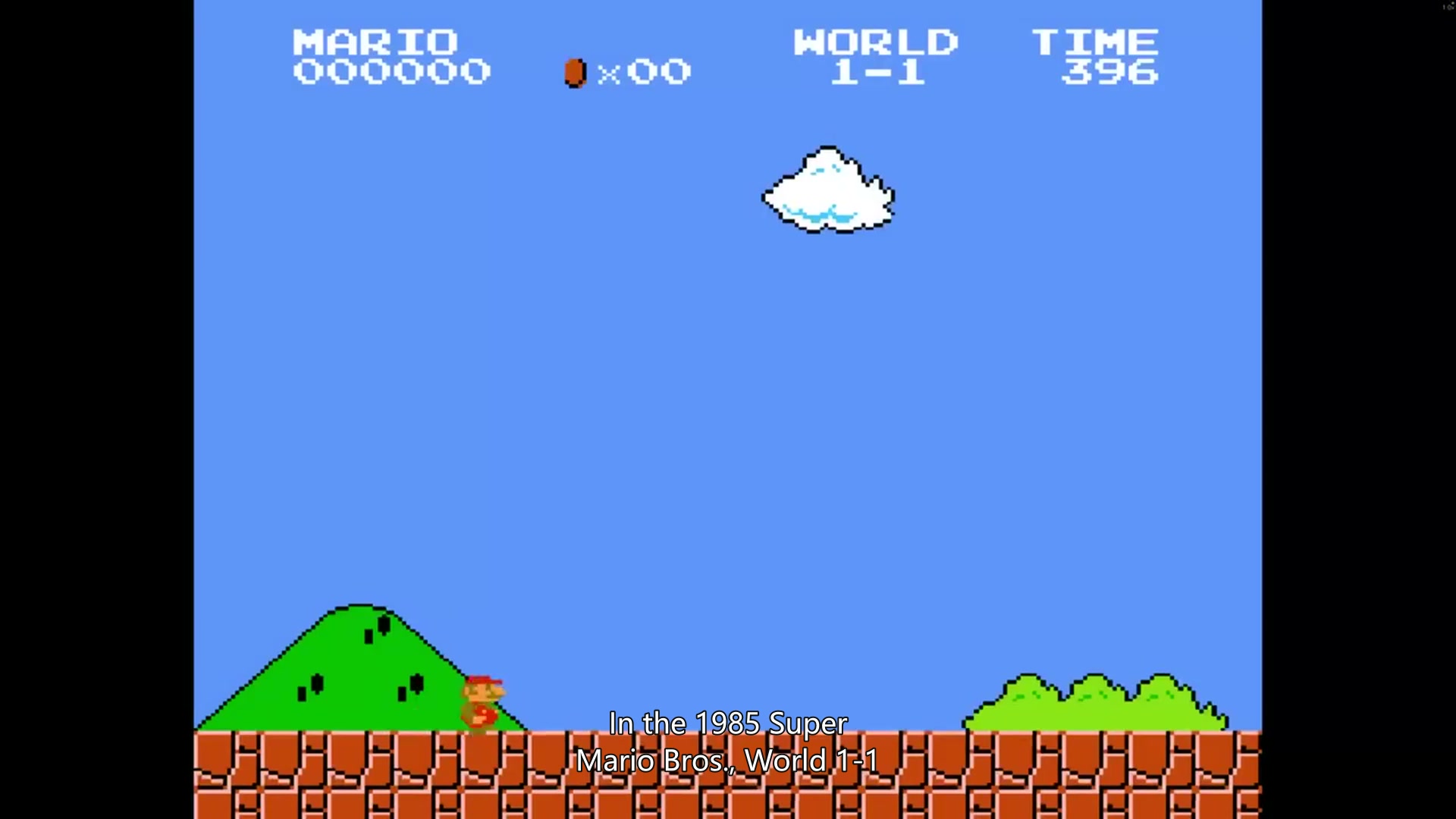}{00:20.4}\hfill%
  \GalleryFrame{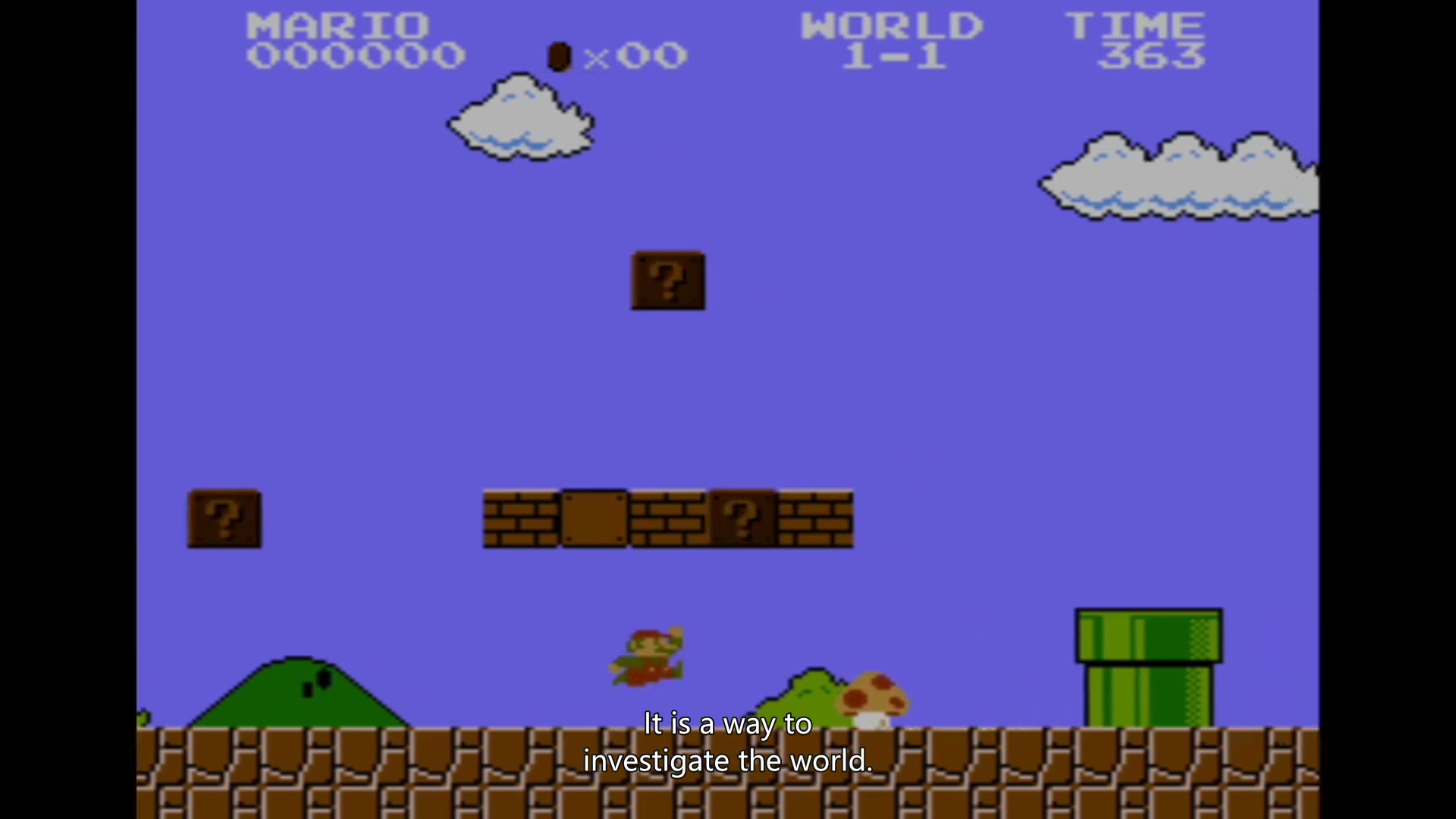}{01:09.5}\hfill%
  \GalleryFrame{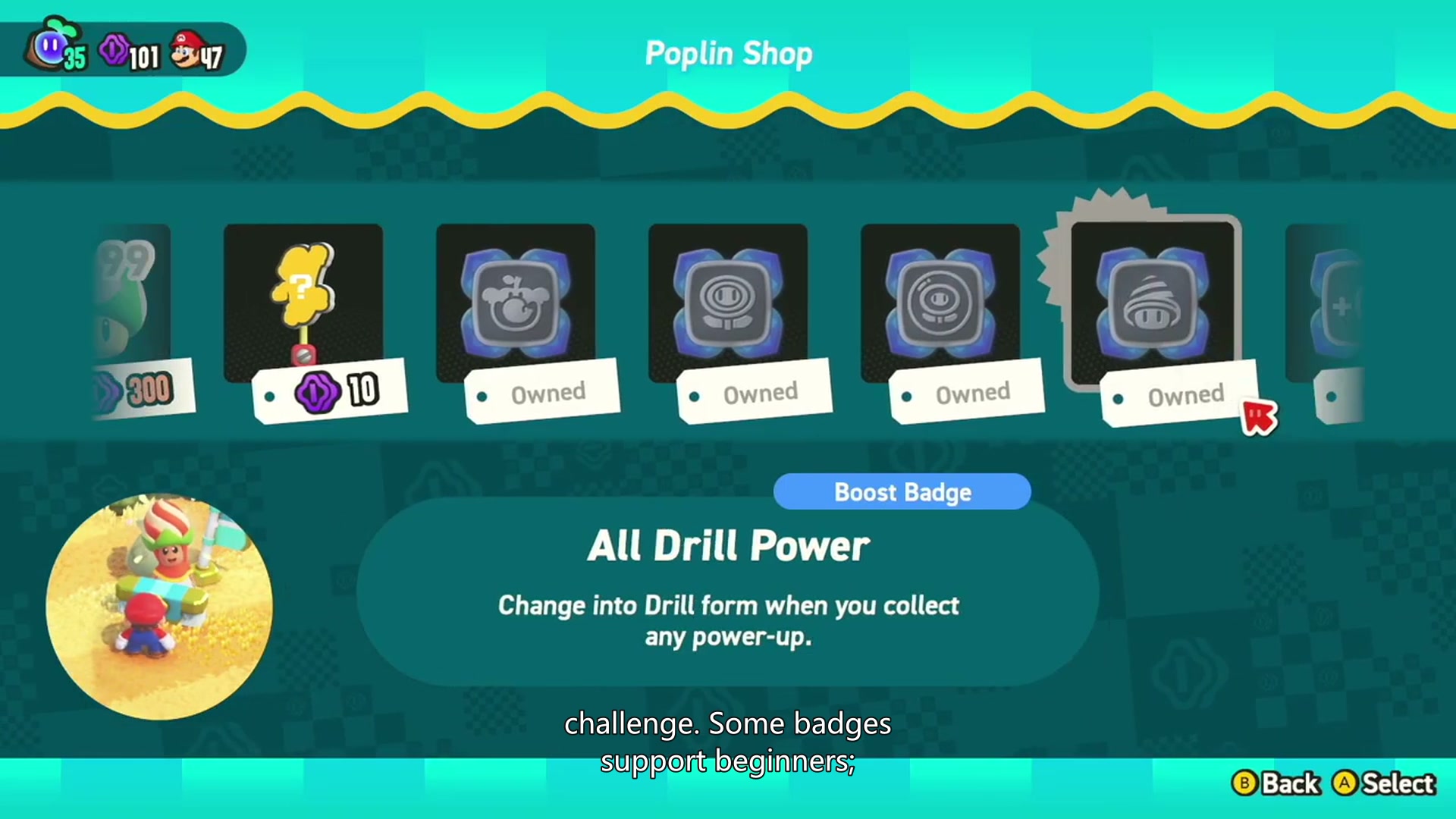}{01:58.6}\hfill%
  \GalleryFrame{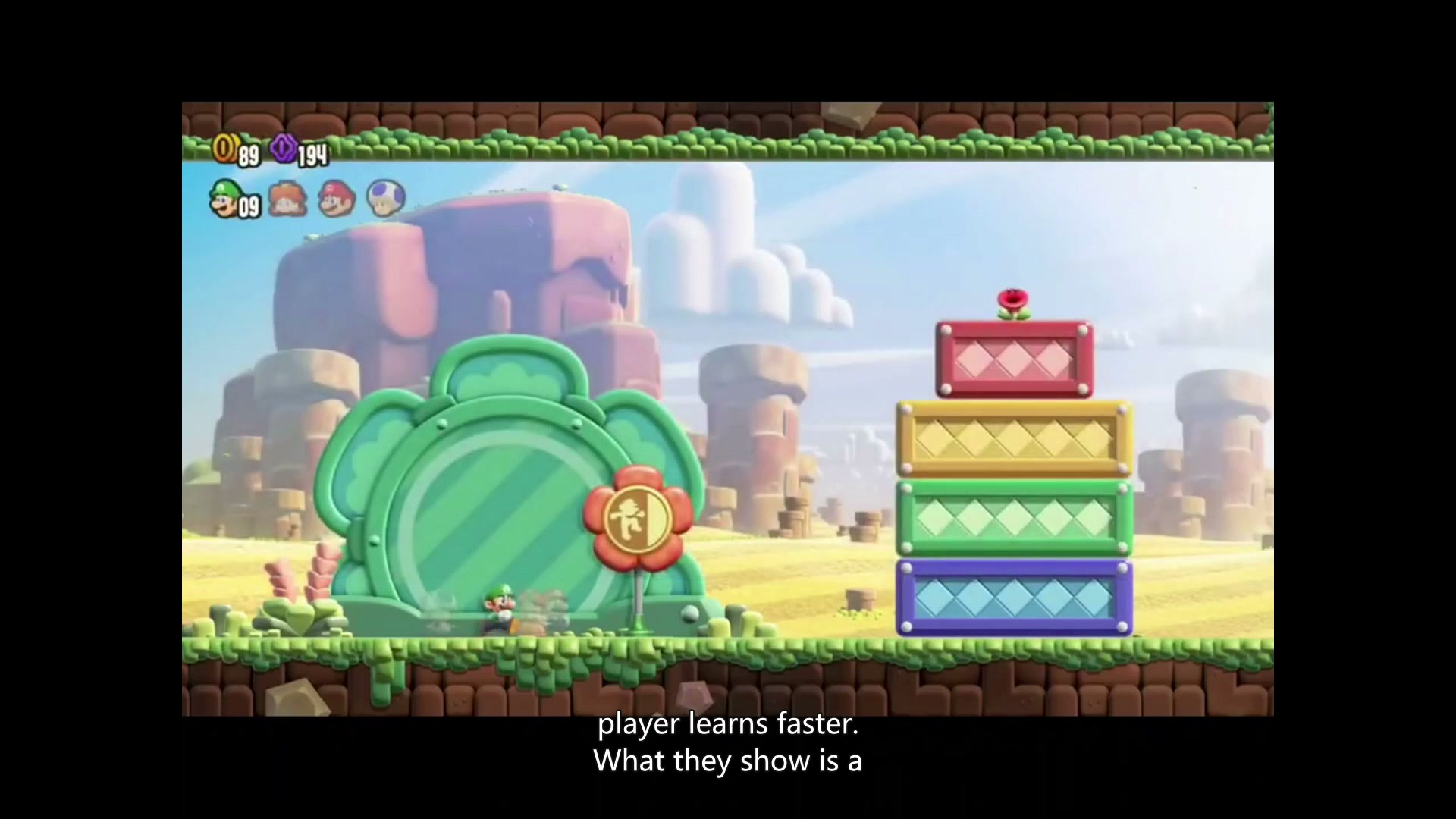}{02:47.7}%
}{fig:video-super_mario_game_design}
\GalleryExample{Conventional and Renewable Energy: Trade-Offs and Choices}{3:48}{%
  \GalleryFrame{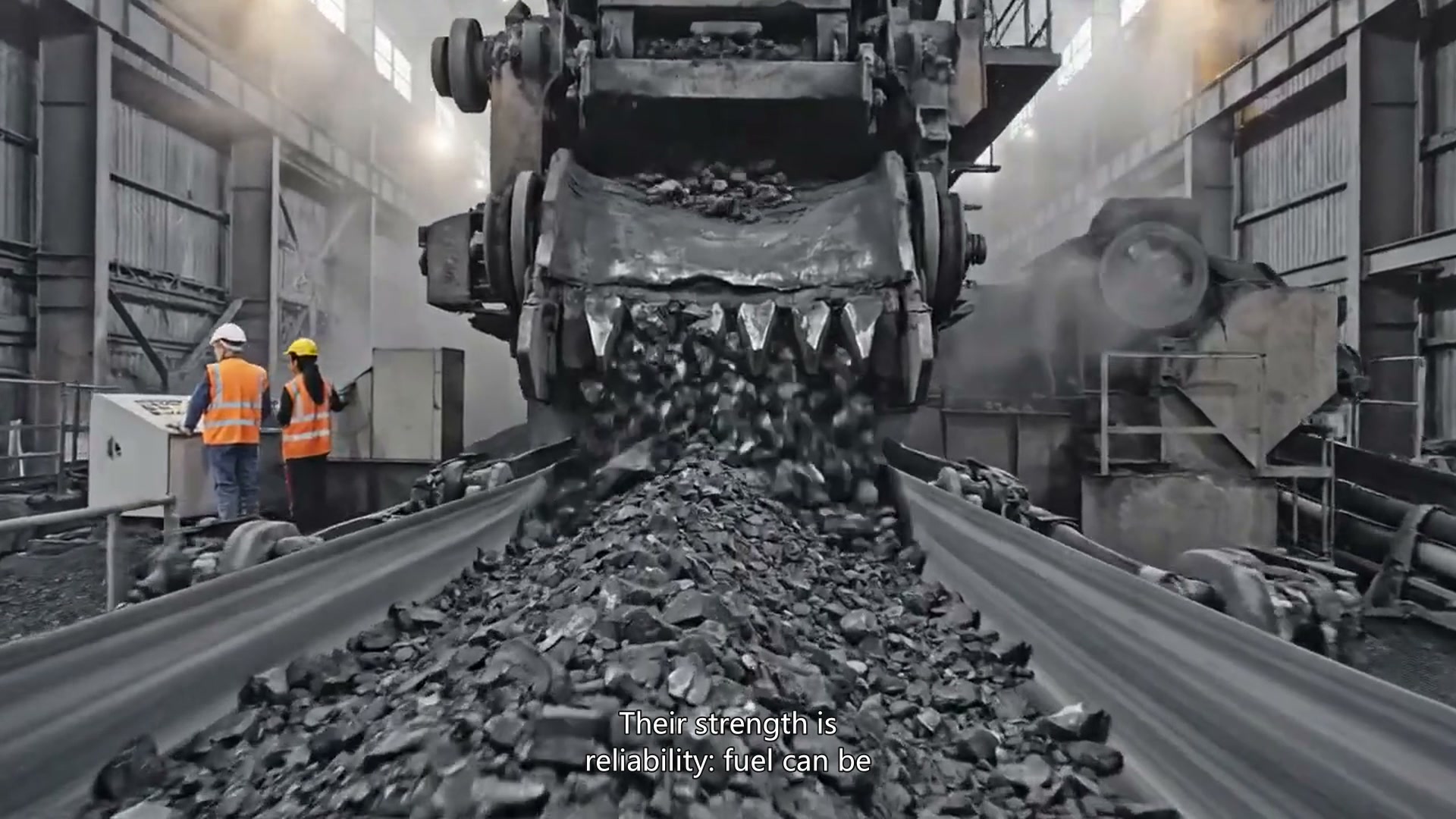}{00:23.8}\hfill%
  \GalleryFrame{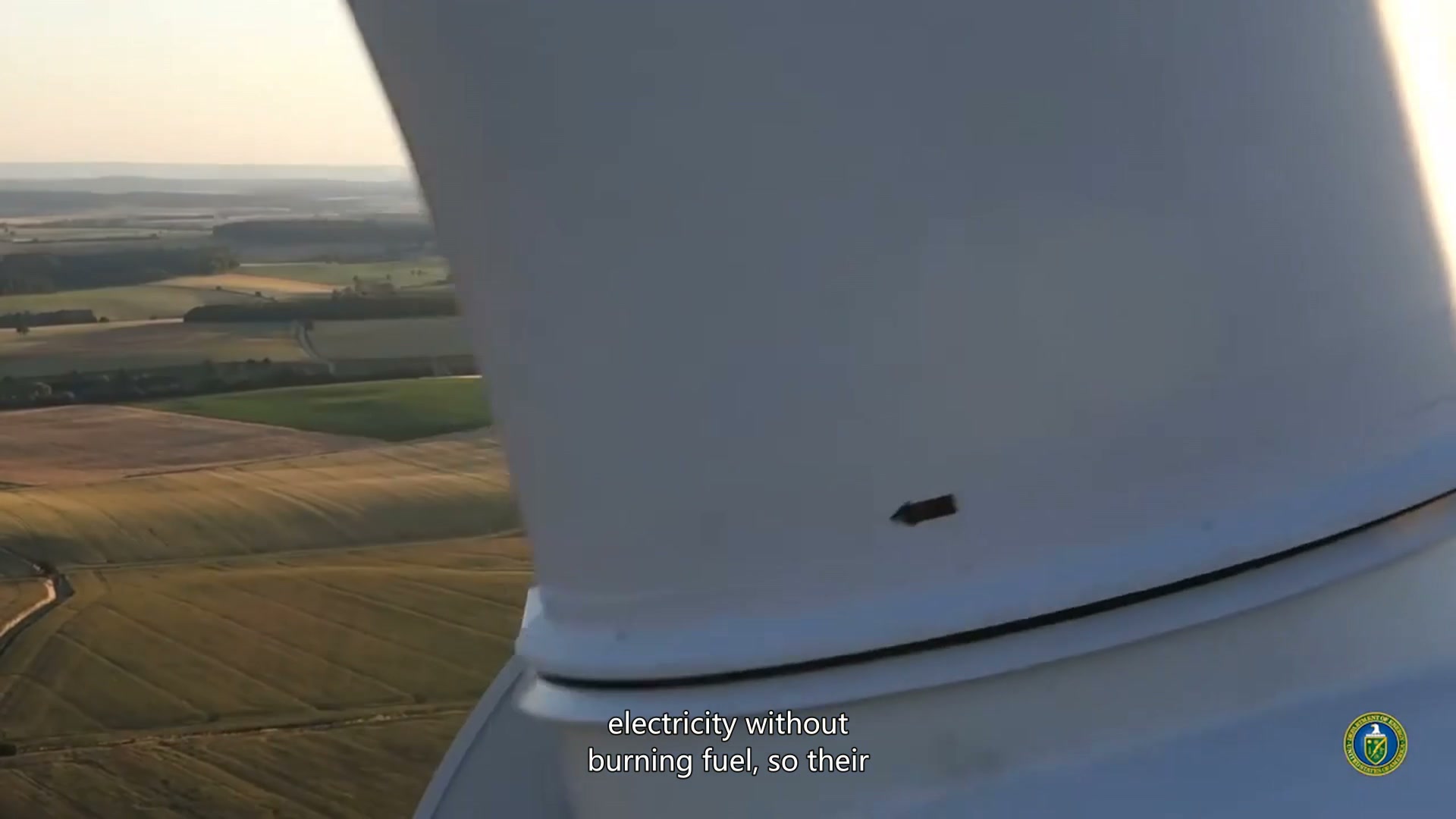}{00:52.3}\hfill%
  \GalleryFrame{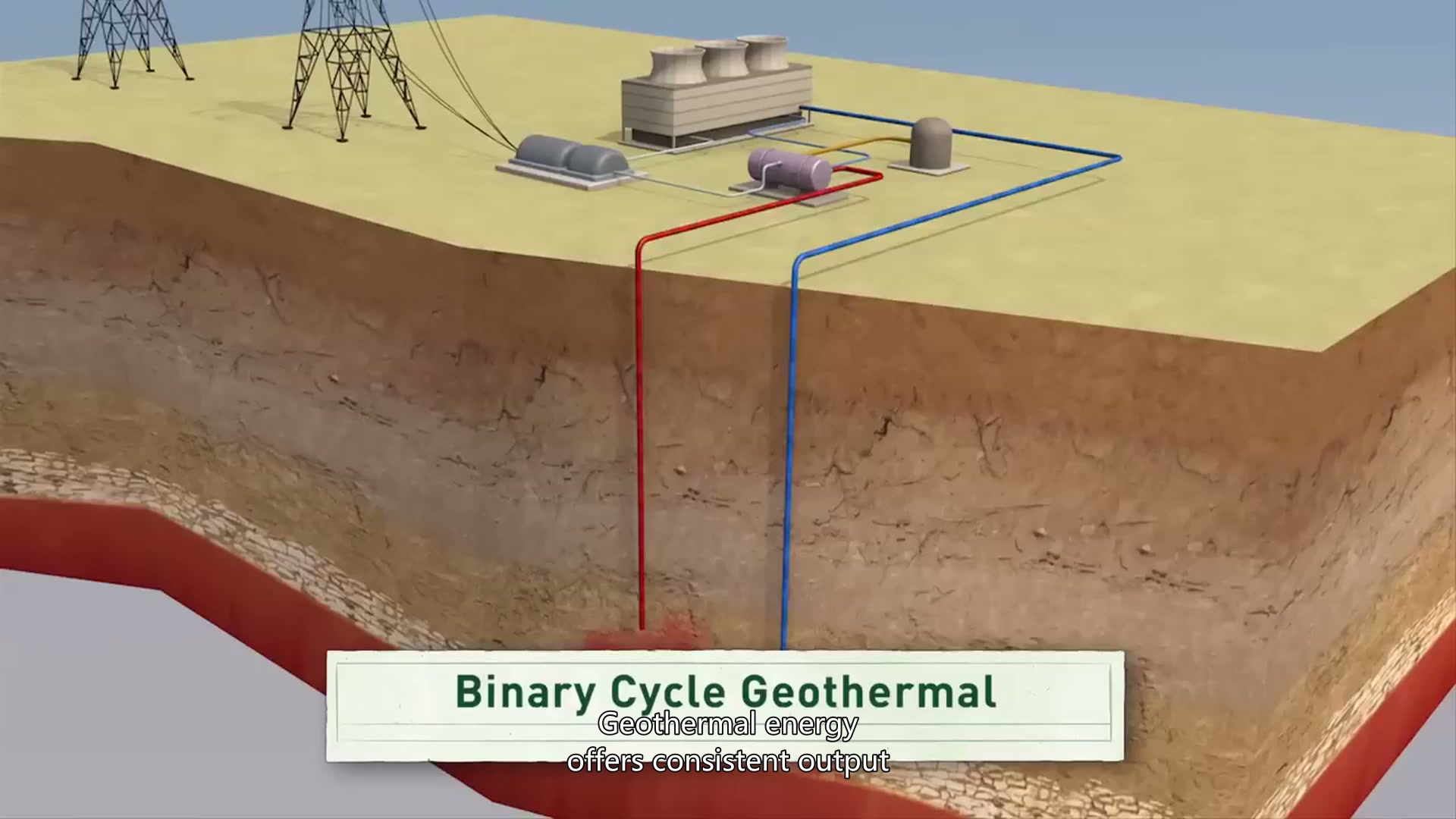}{01:30.4}\hfill%
  \GalleryFrame{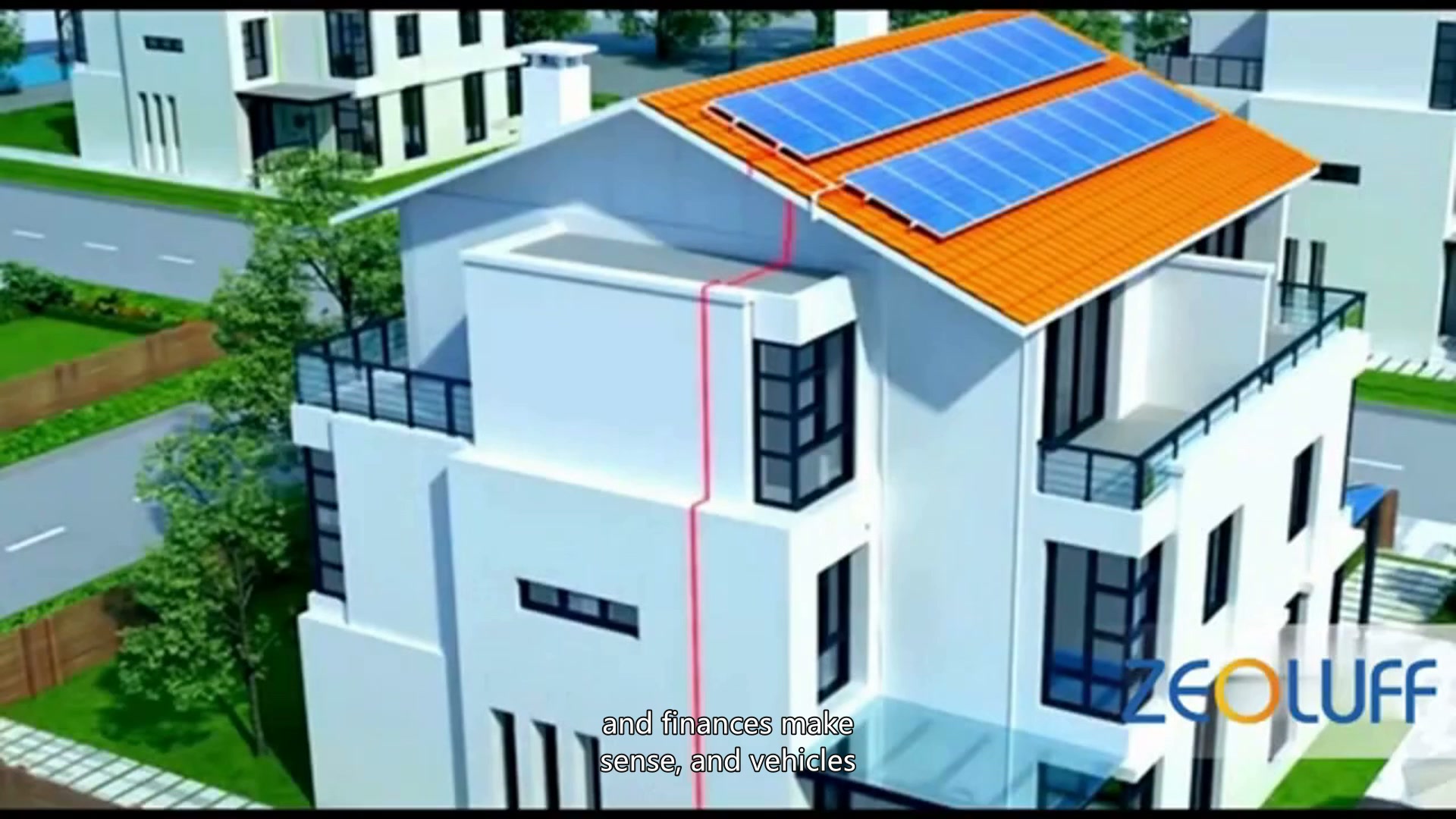}{02:56.0}%
}{fig:video-conventional_and_renewable_energy}
\GalleryExample{Why Advanced Chips Depend on ASML}{3:39}{%
  \GalleryFrame{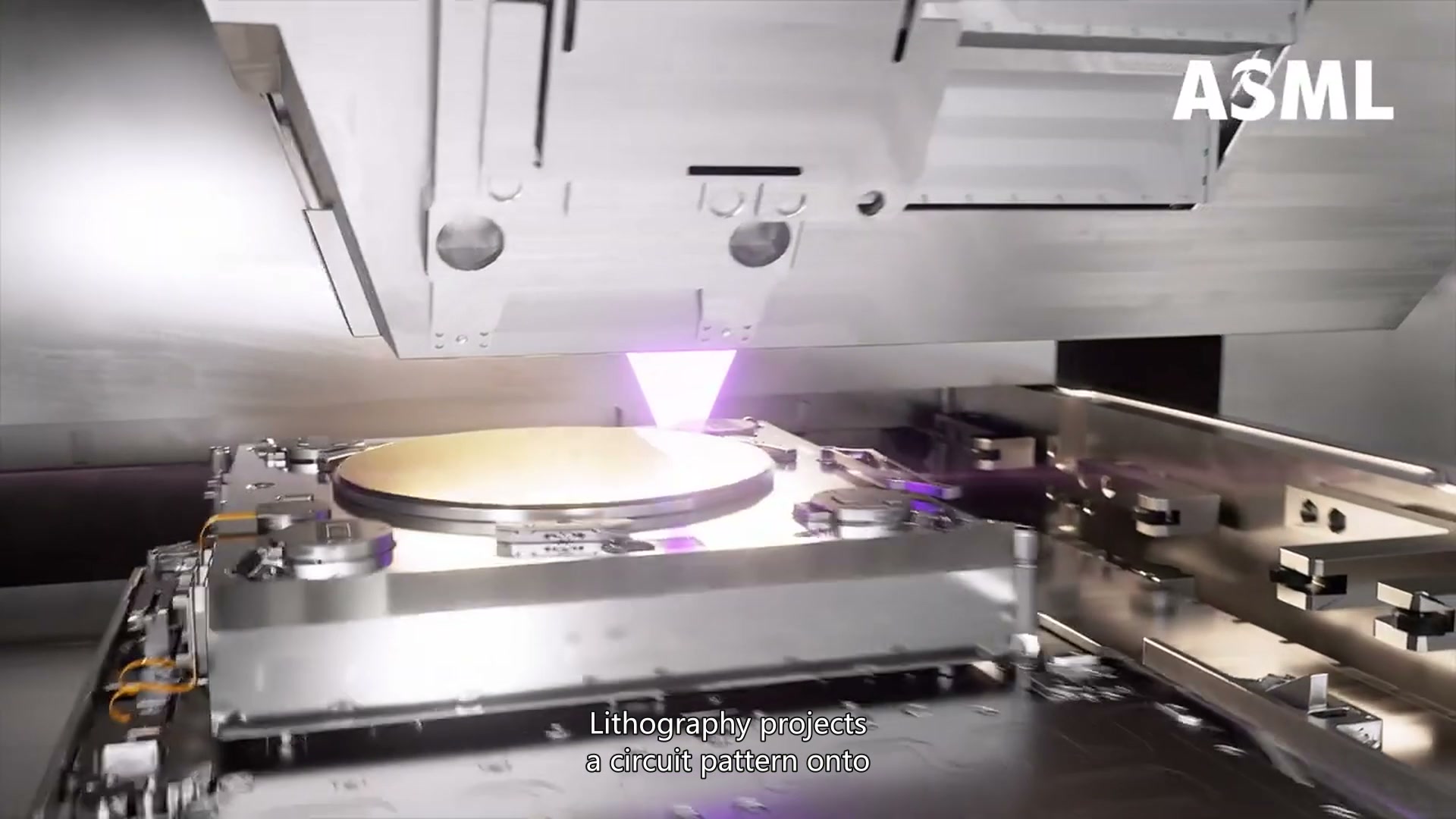}{00:22.9}\hfill%
  \GalleryFrame{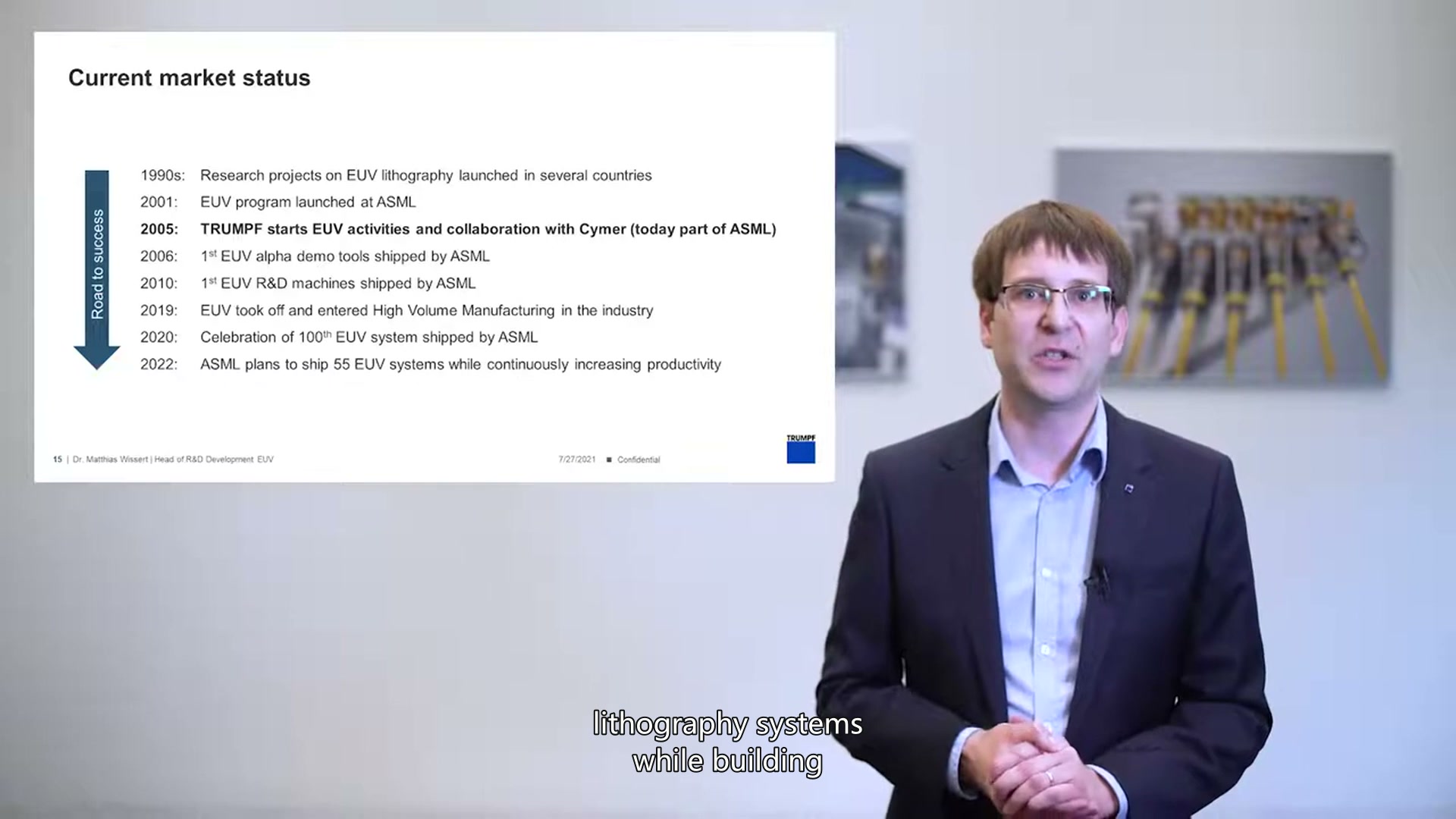}{01:17.8}\hfill%
  \GalleryFrame{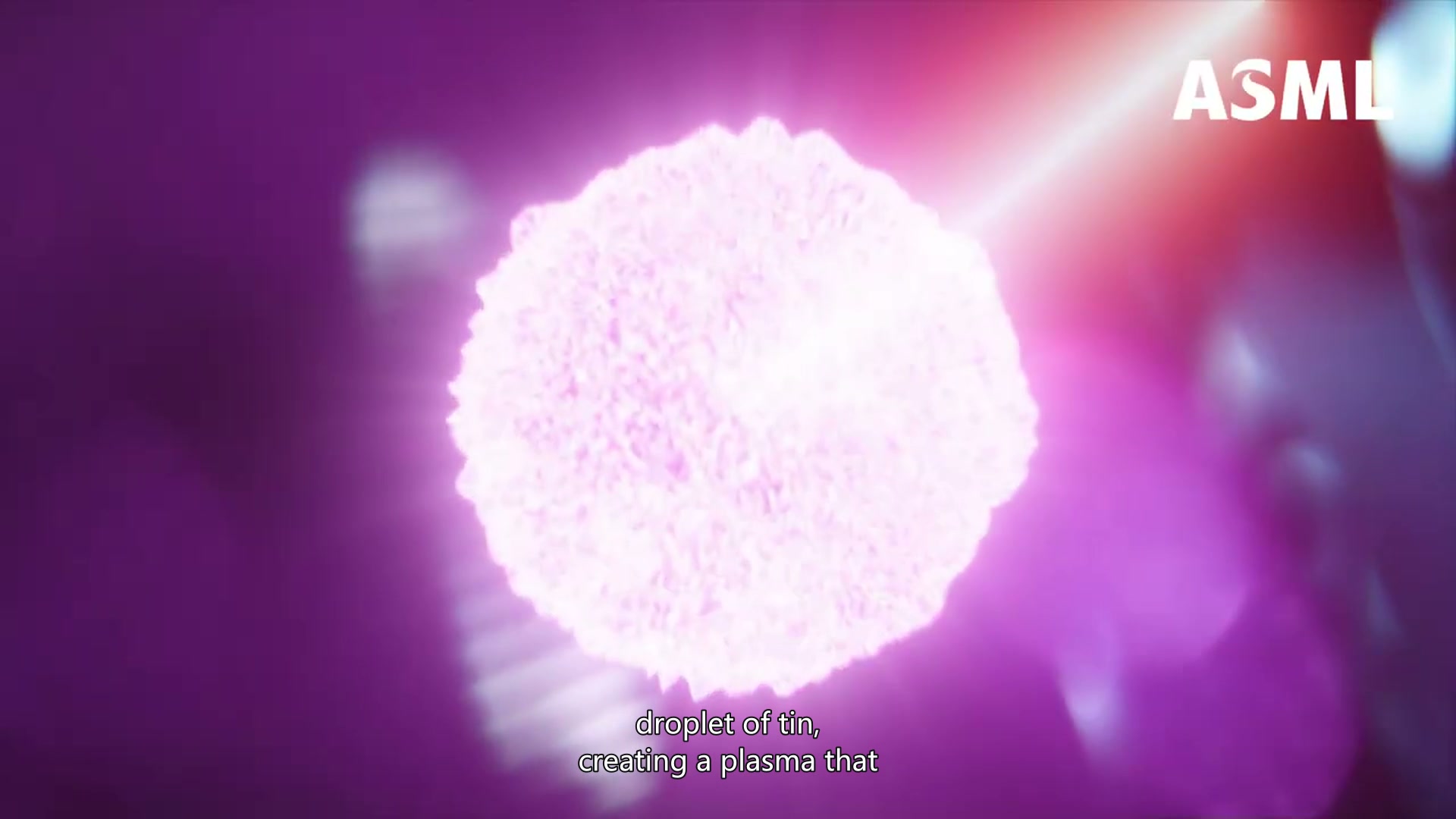}{02:12.7}\hfill%
  \GalleryFrame{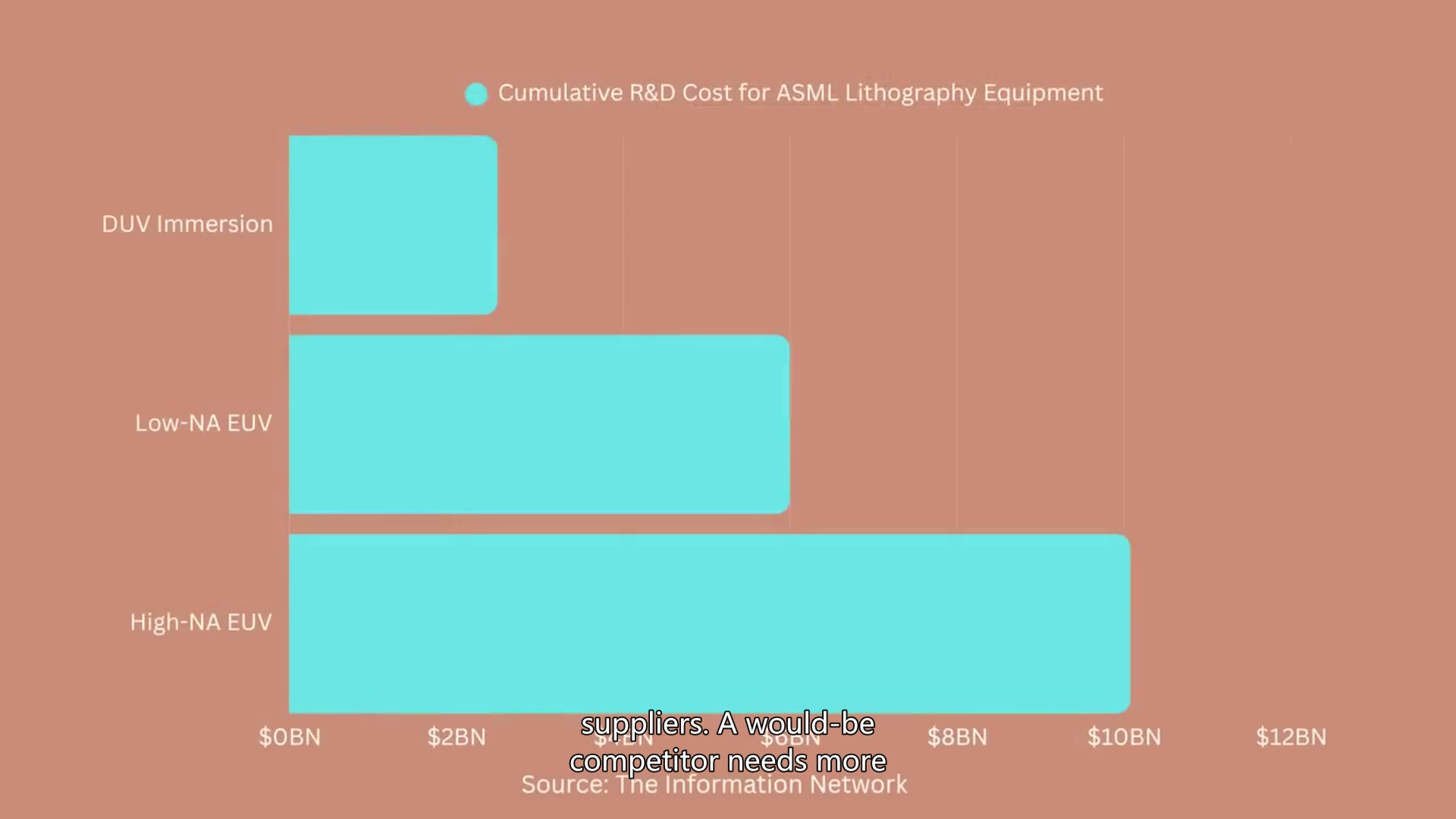}{03:07.6}%
}{fig:video-asml_and_euv_lithography}
\GalleryExample{Five Competitive Video Games}{2:47}{%
  \GalleryFrame{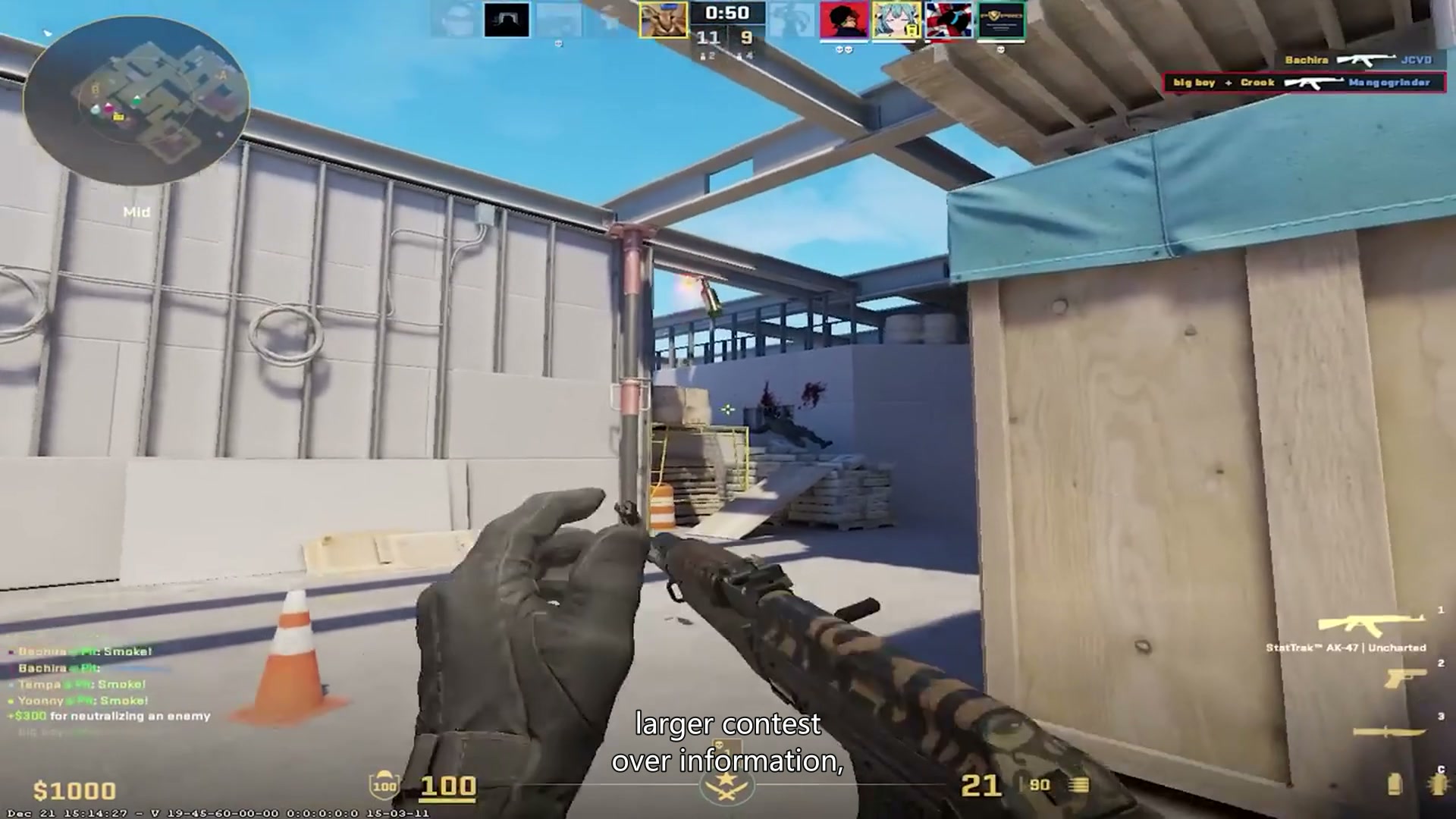}{00:17.4}\hfill%
  \GalleryFrame{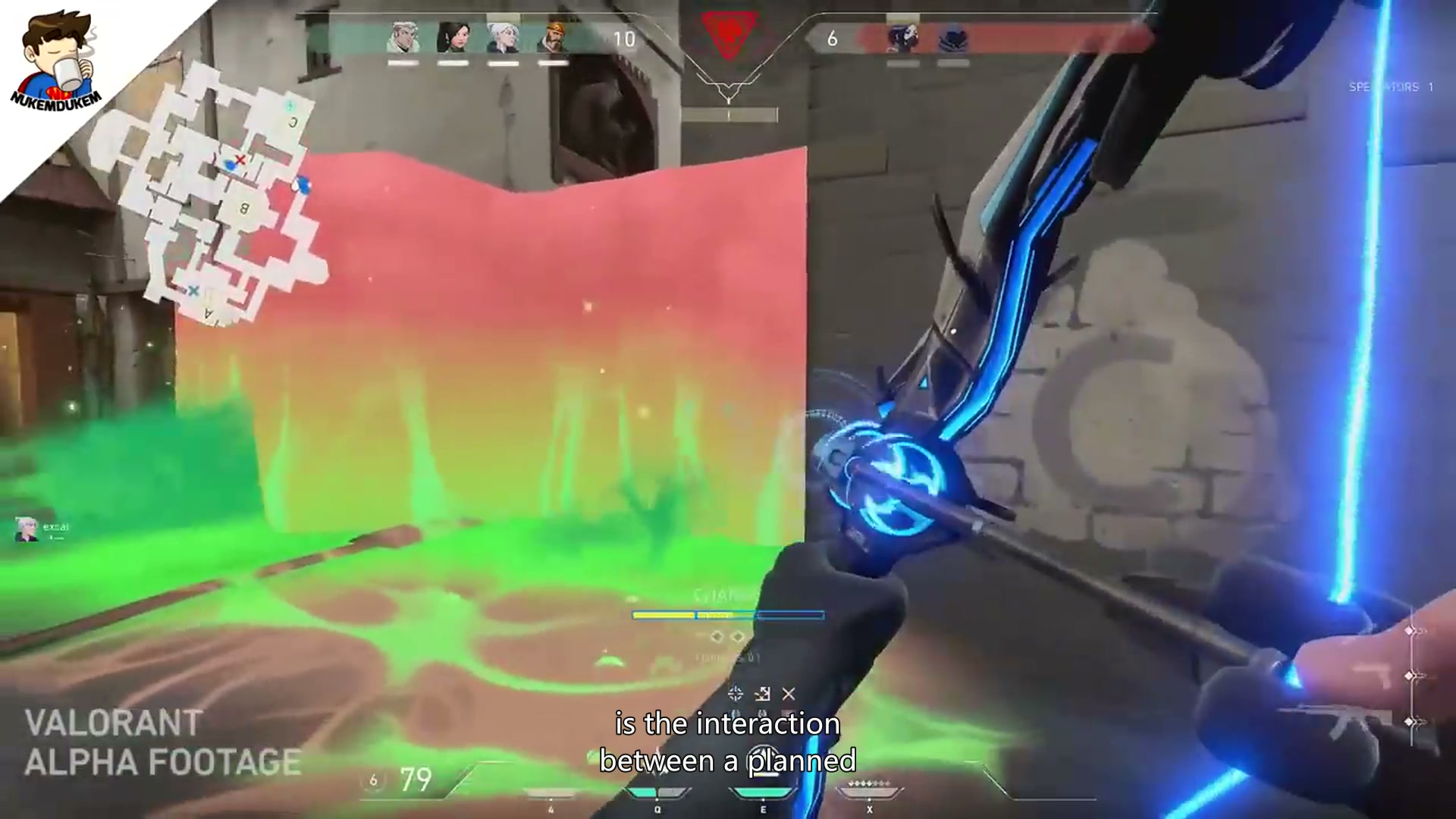}{00:59.2}\hfill%
  \GalleryFrame{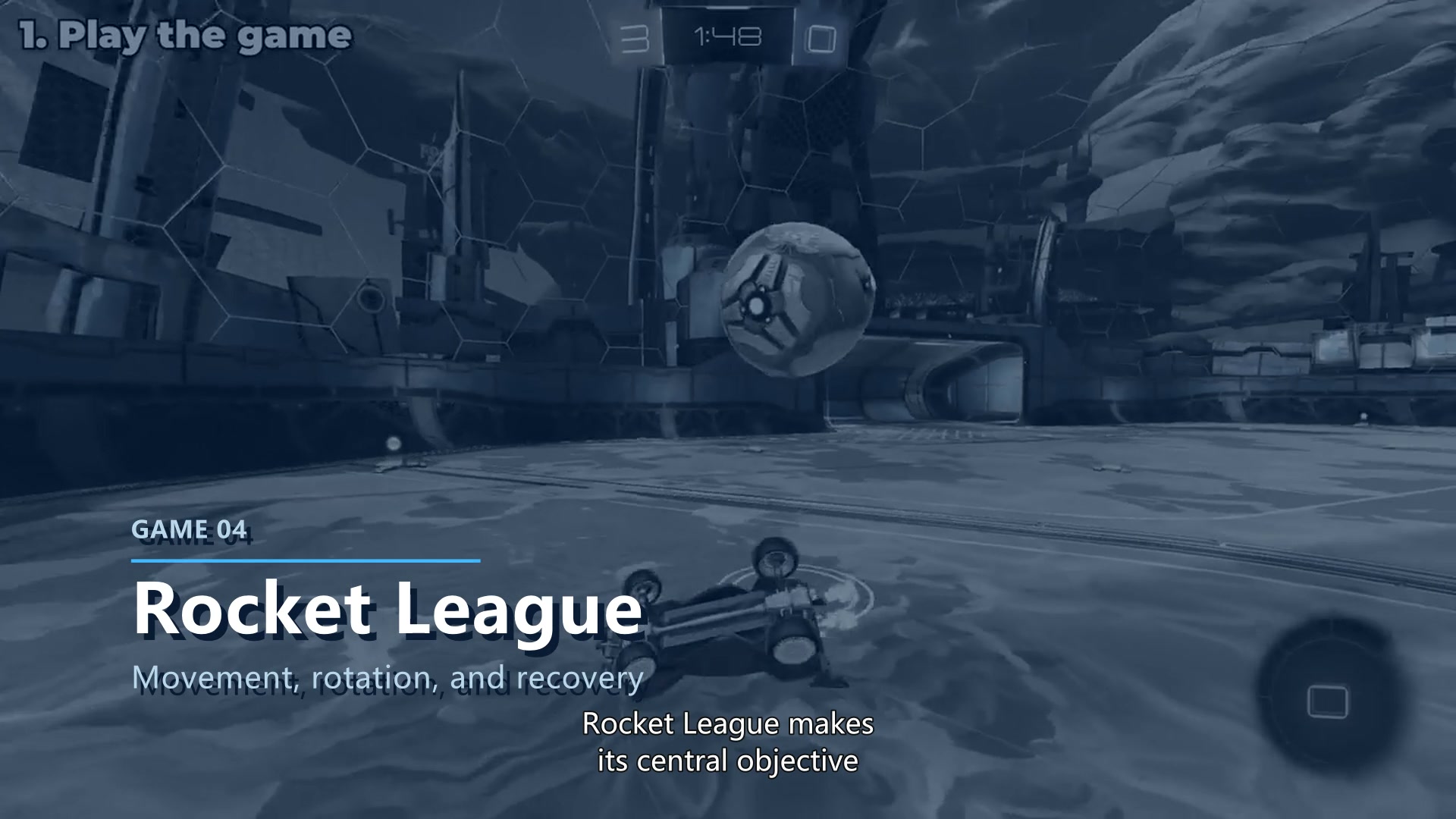}{01:41.0}\hfill%
  \GalleryFrame{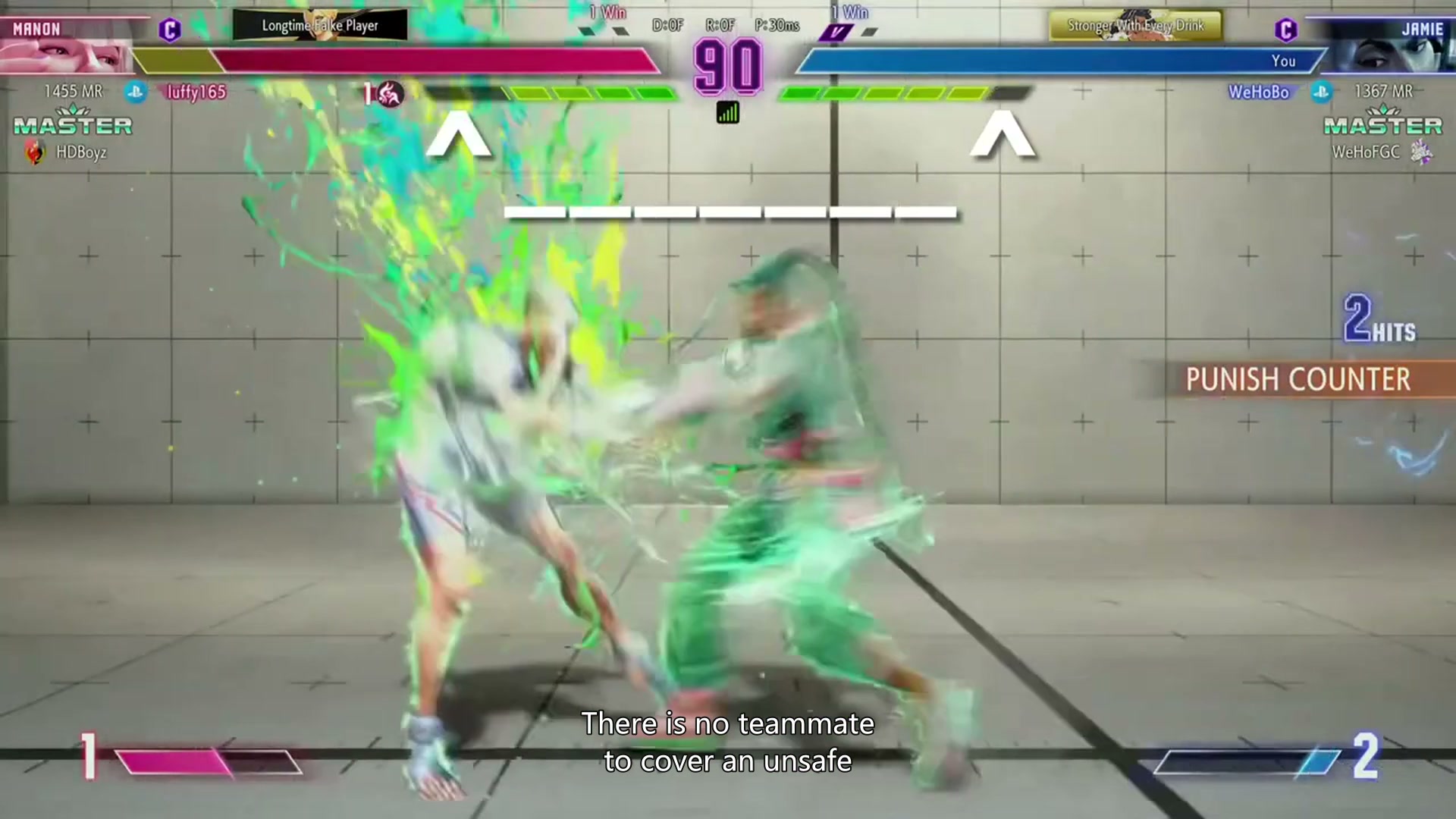}{02:22.8}%
}{fig:video-competitive_video_games}
\GalleryExample{Five Chinese Foods to Try}{3:09}{%
  \GalleryFrame{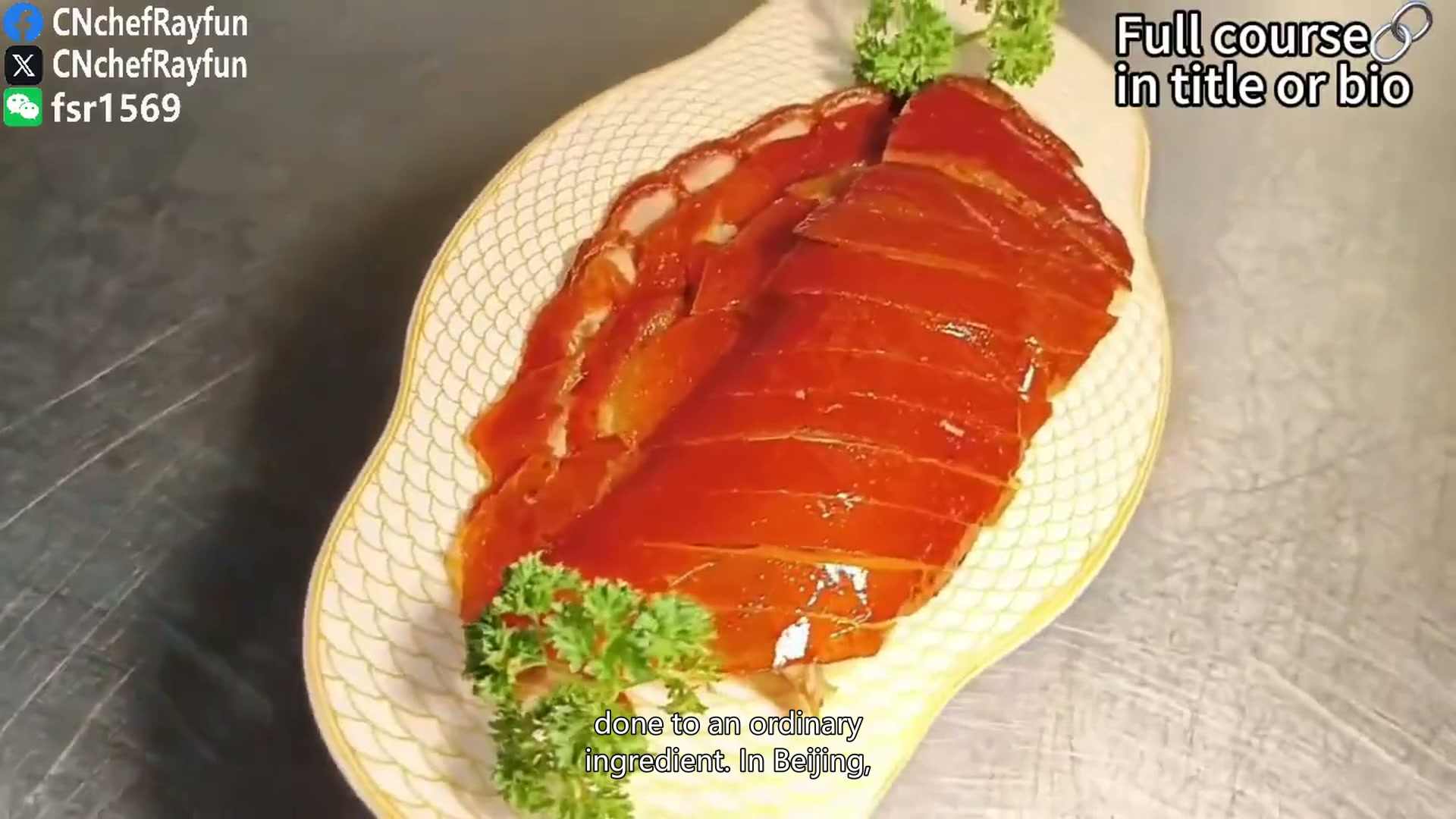}{00:19.7}\hfill%
  \GalleryFrame{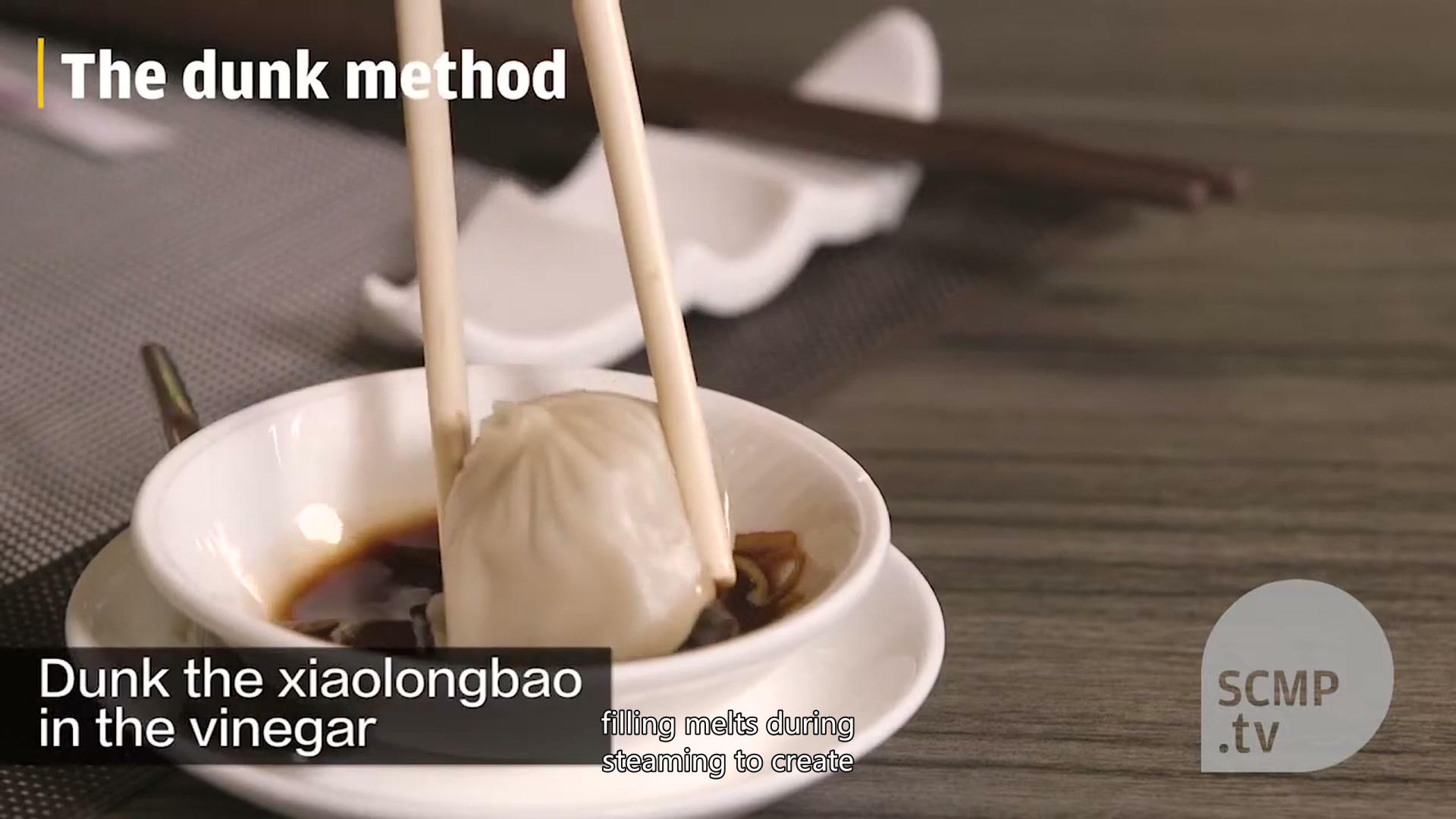}{01:07.1}\hfill%
  \GalleryFrame{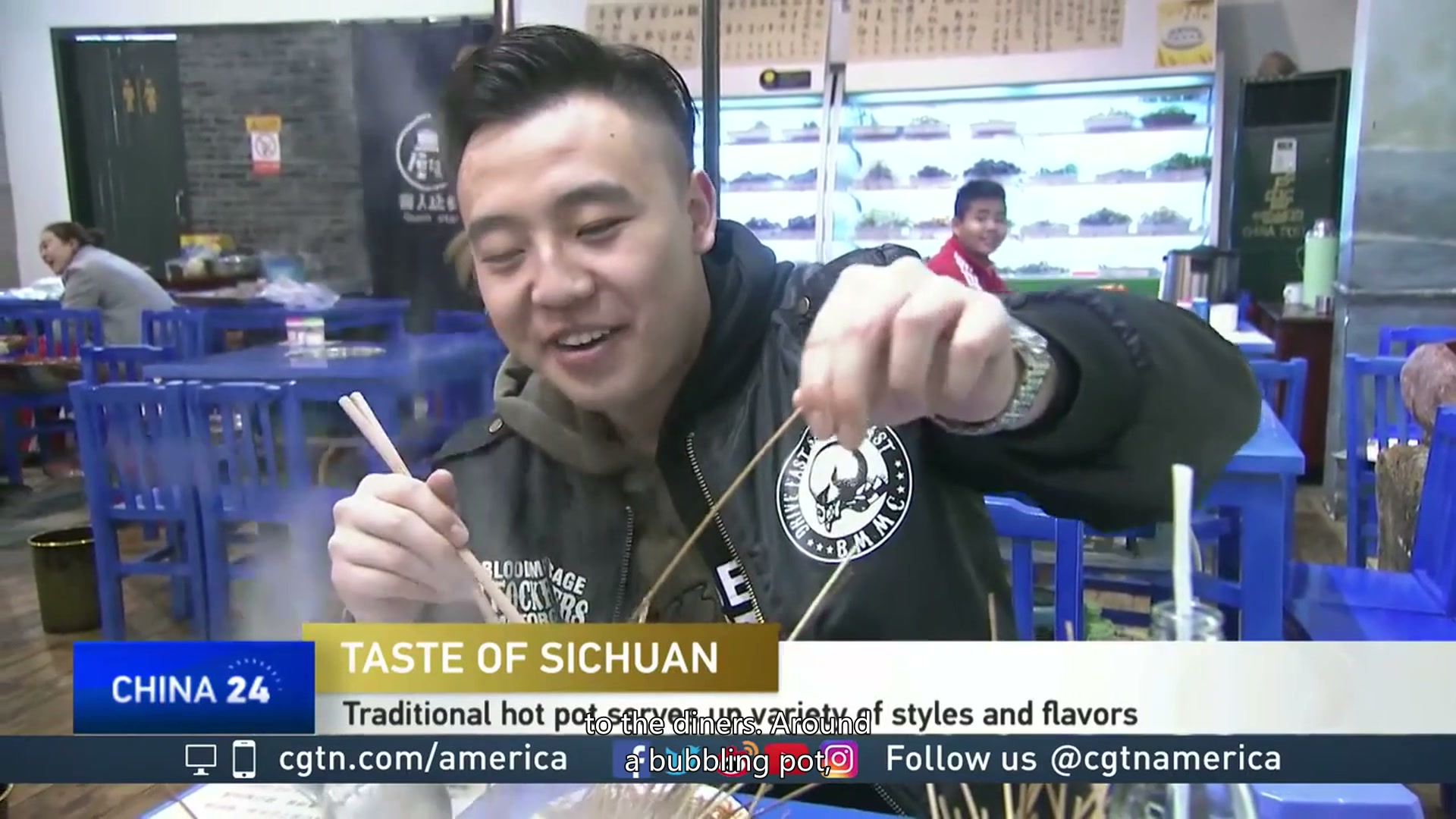}{01:54.4}\hfill%
  \GalleryFrame{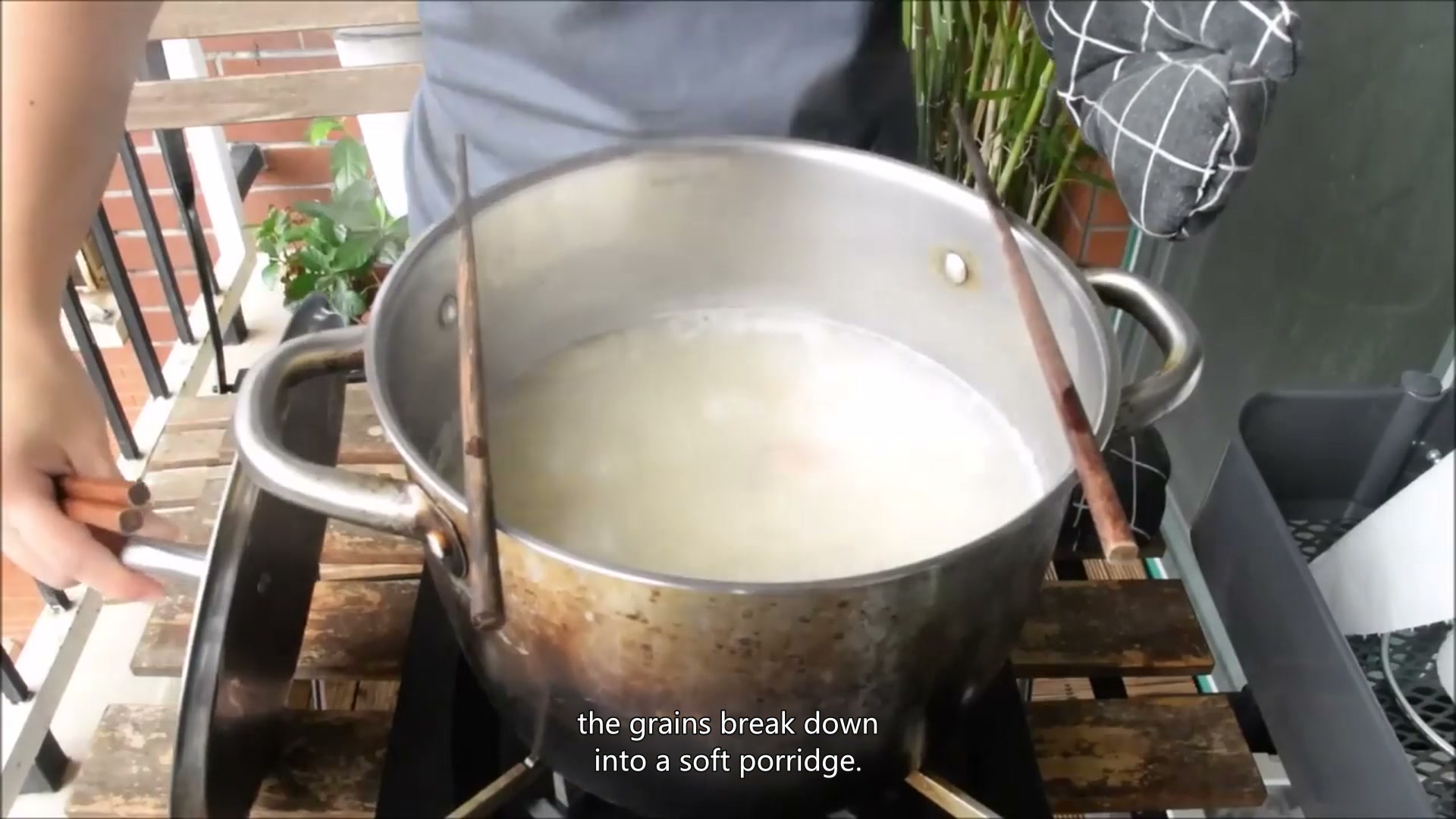}{02:41.8}%
}{fig:video-five_chinese_foods}
\GalleryExample{Why Lithium Prices Boom and Bust}{3:42}{%
  \GalleryFrame{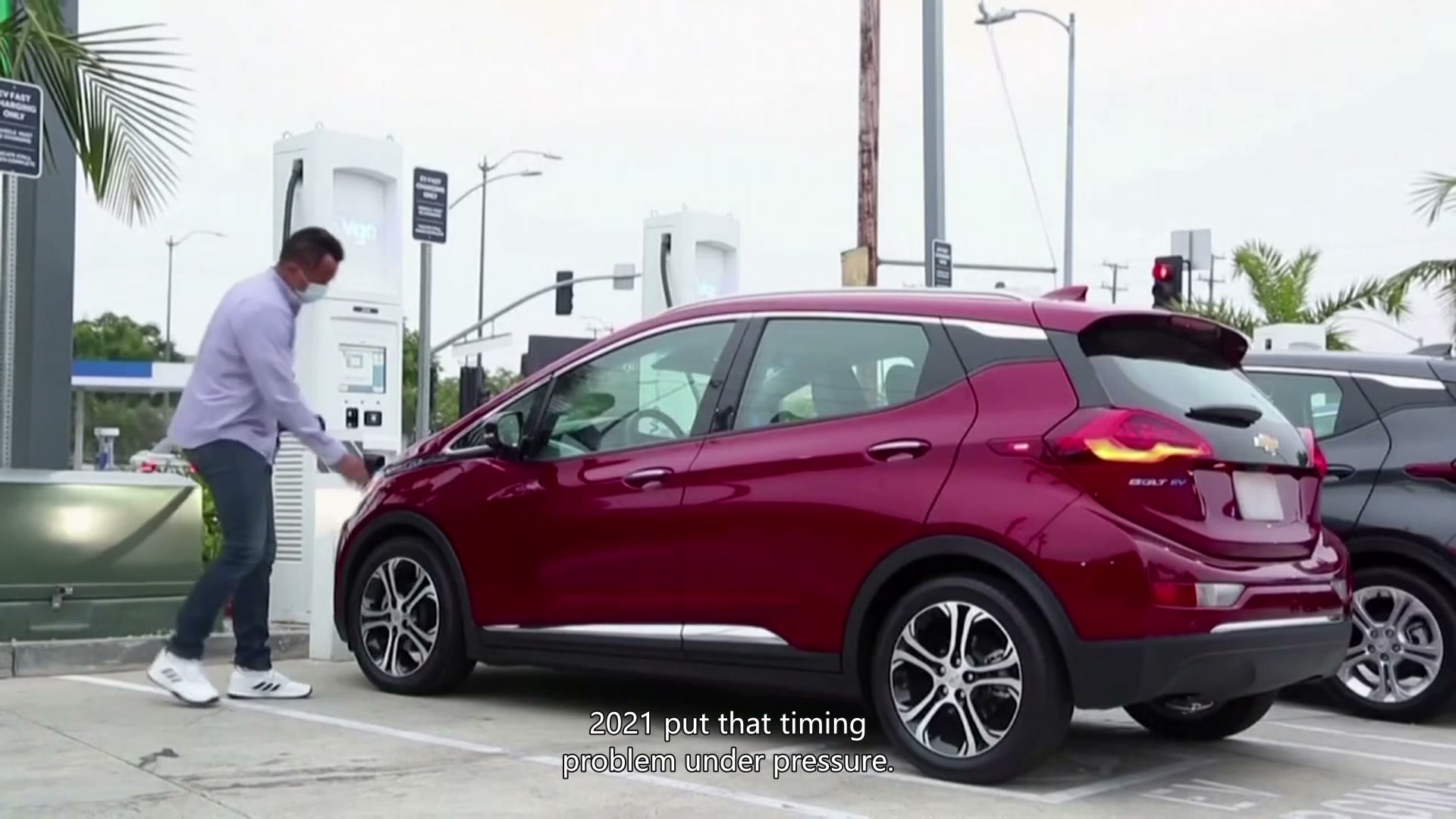}{00:23.1}\hfill%
  \GalleryFrame{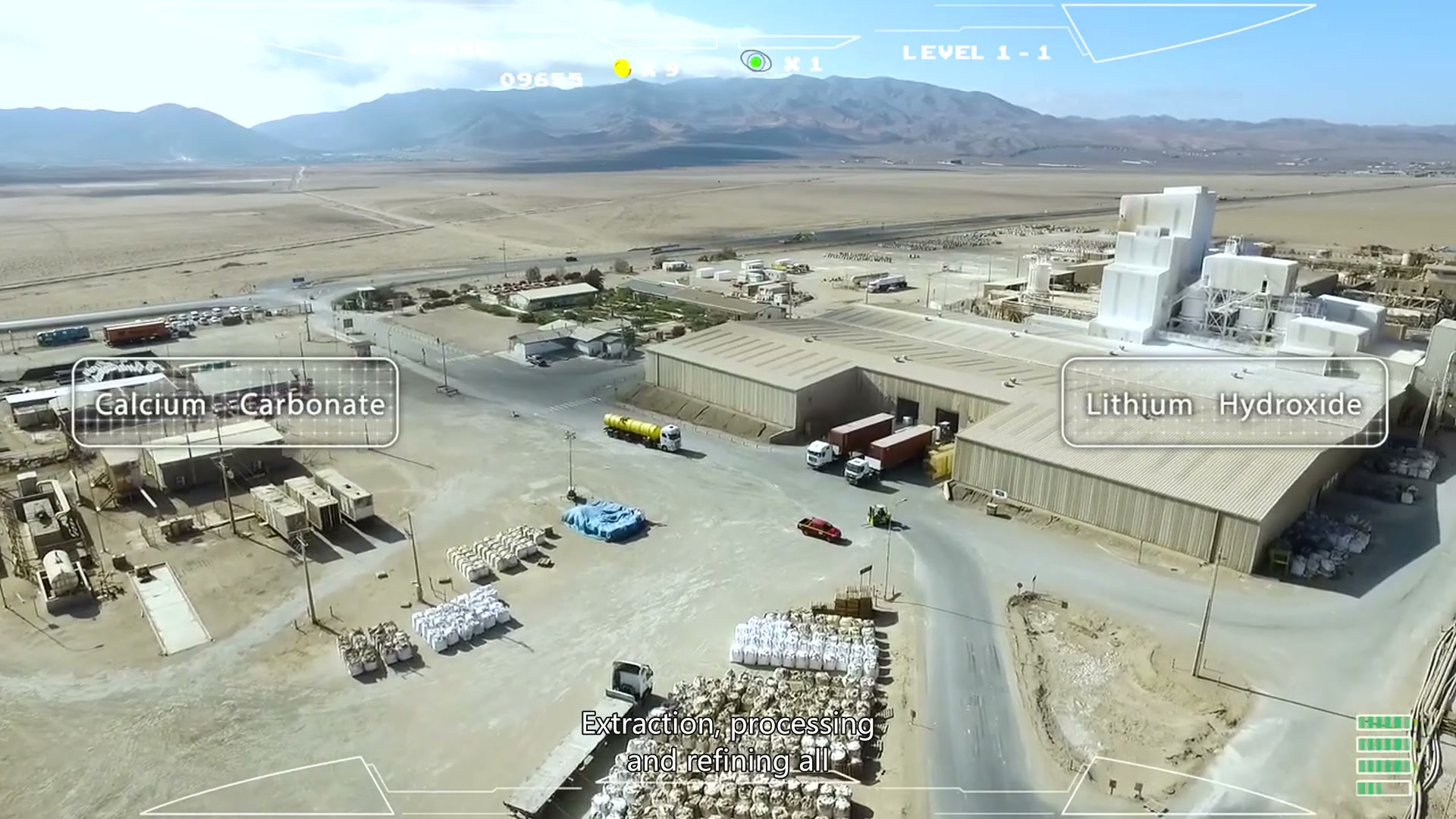}{01:18.6}\hfill%
  \GalleryFrame{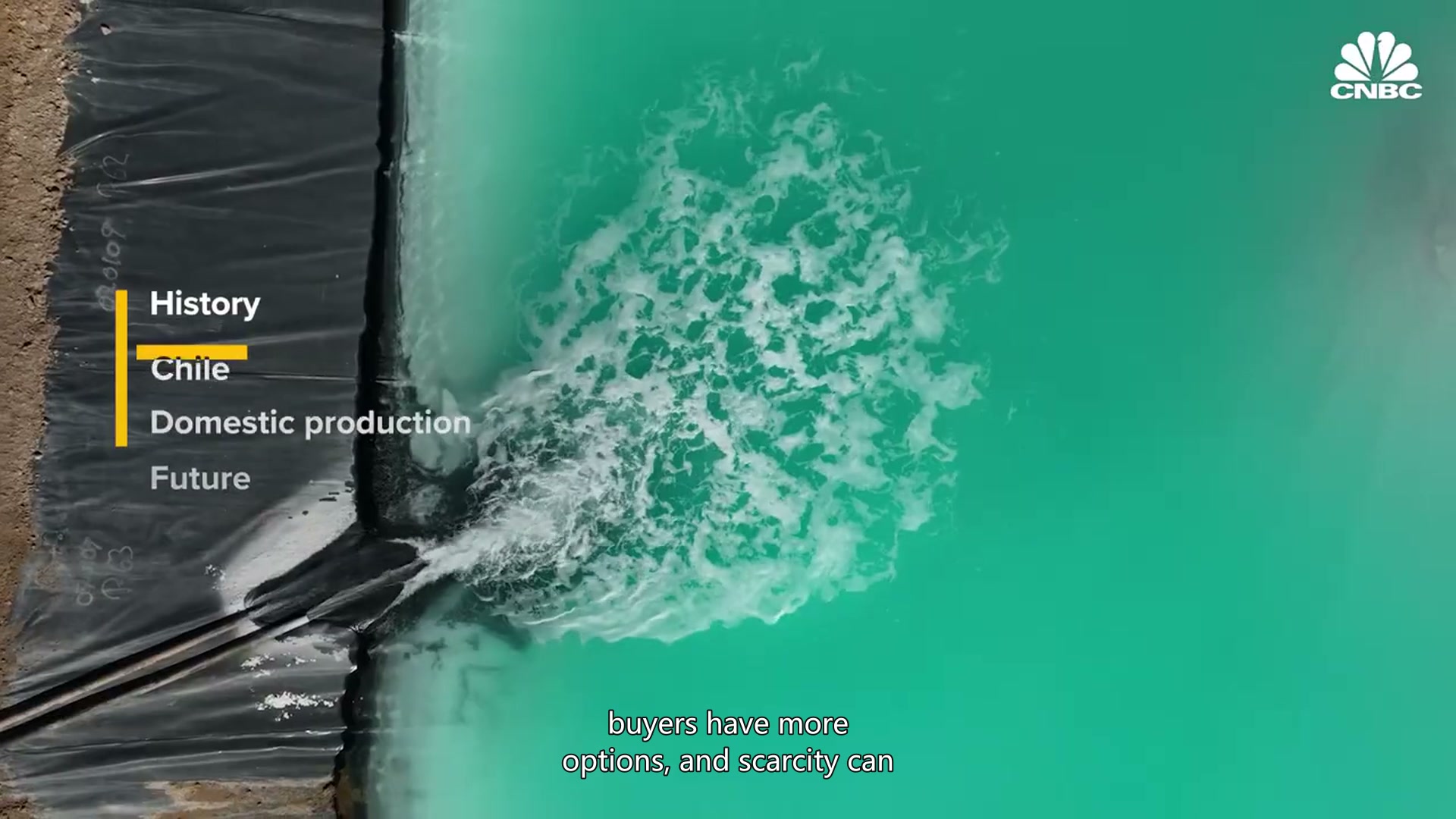}{02:14.1}\hfill%
  \GalleryFrame{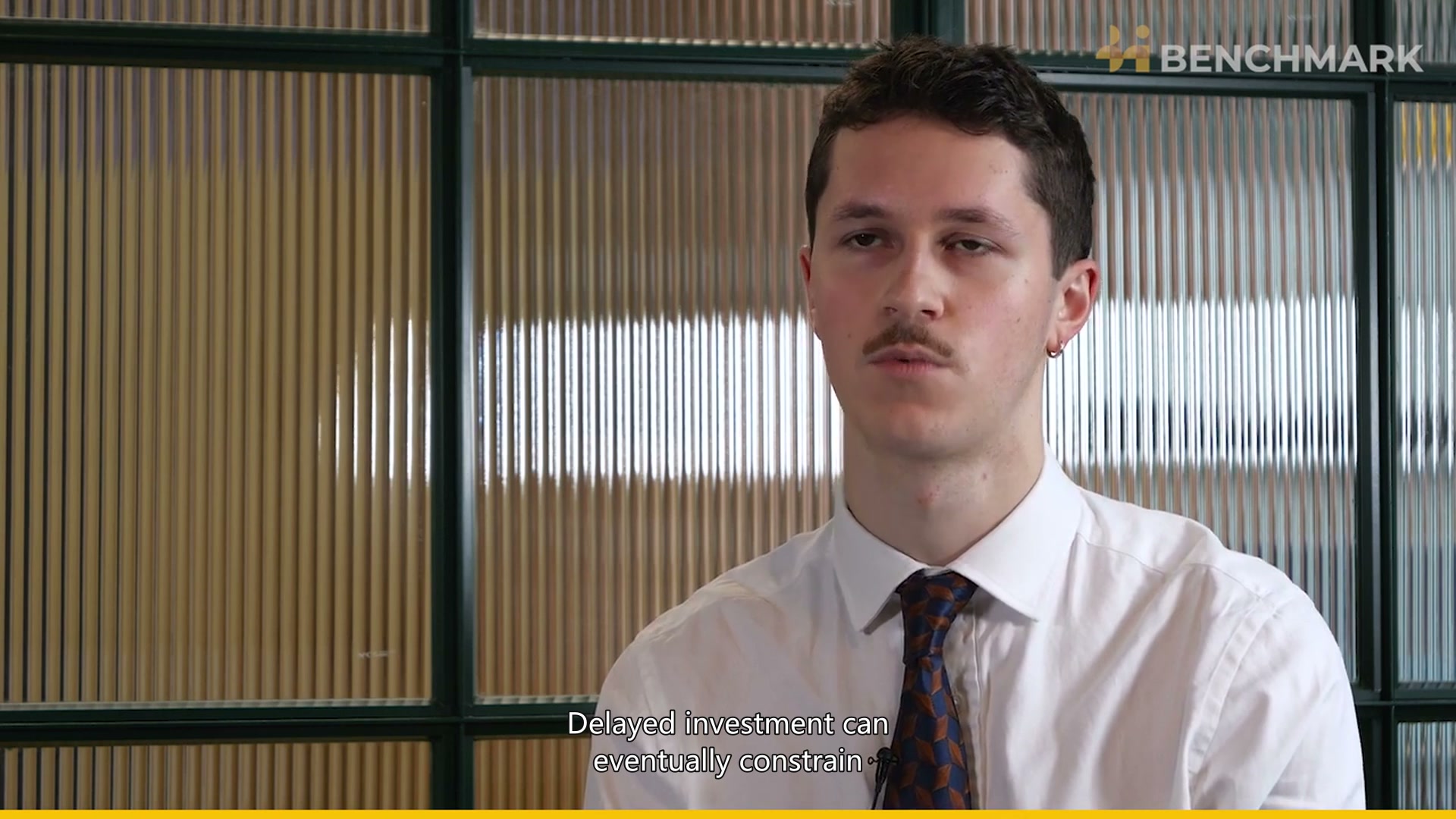}{03:09.6}%
}{fig:video-lithium_price_cycles}
\GalleryExample{Why Drought Causes Shipping Delays at the Panama Canal}{3:29}{%
  \GalleryFrame{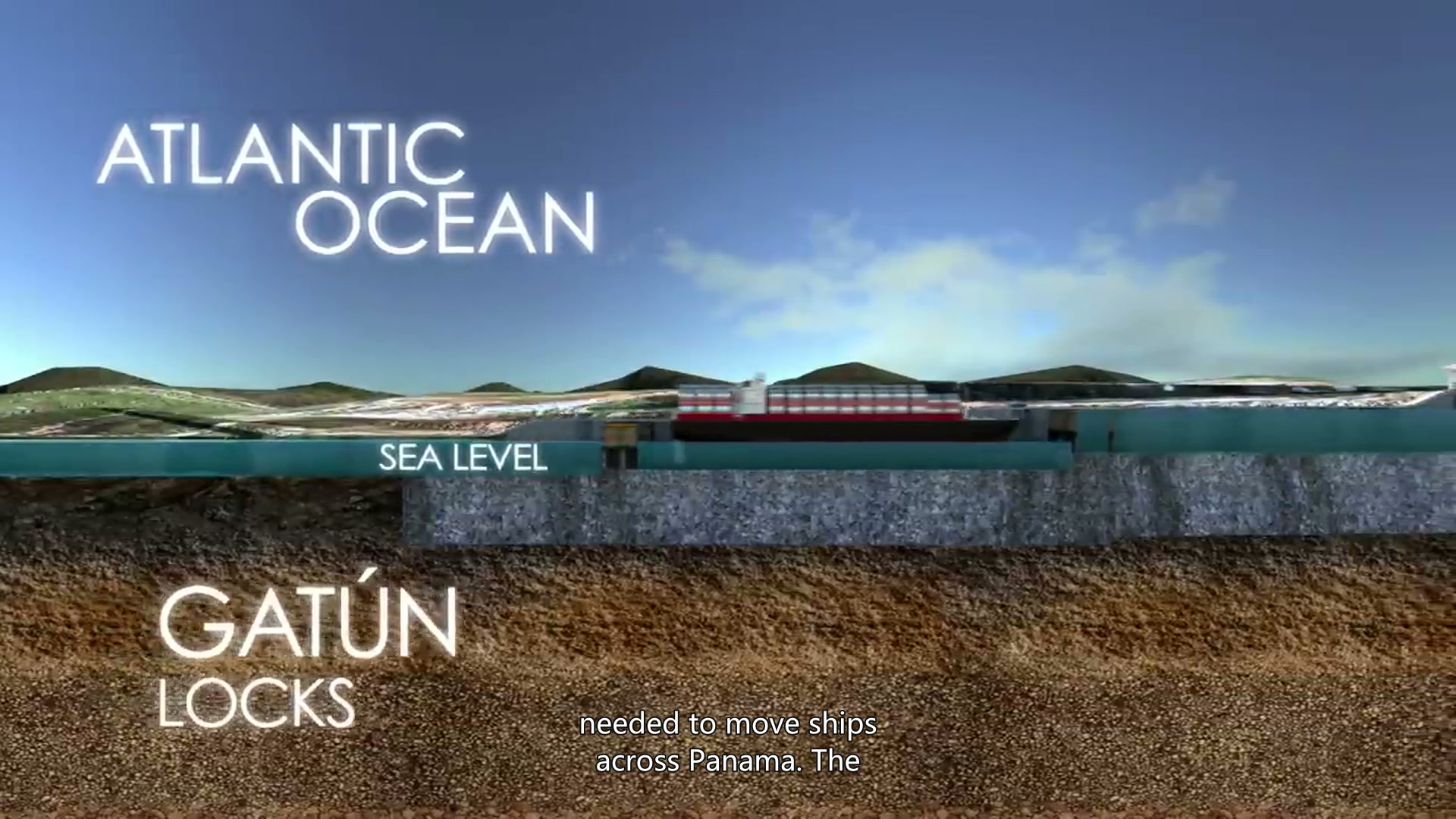}{00:21.8}\hfill%
  \GalleryFrame{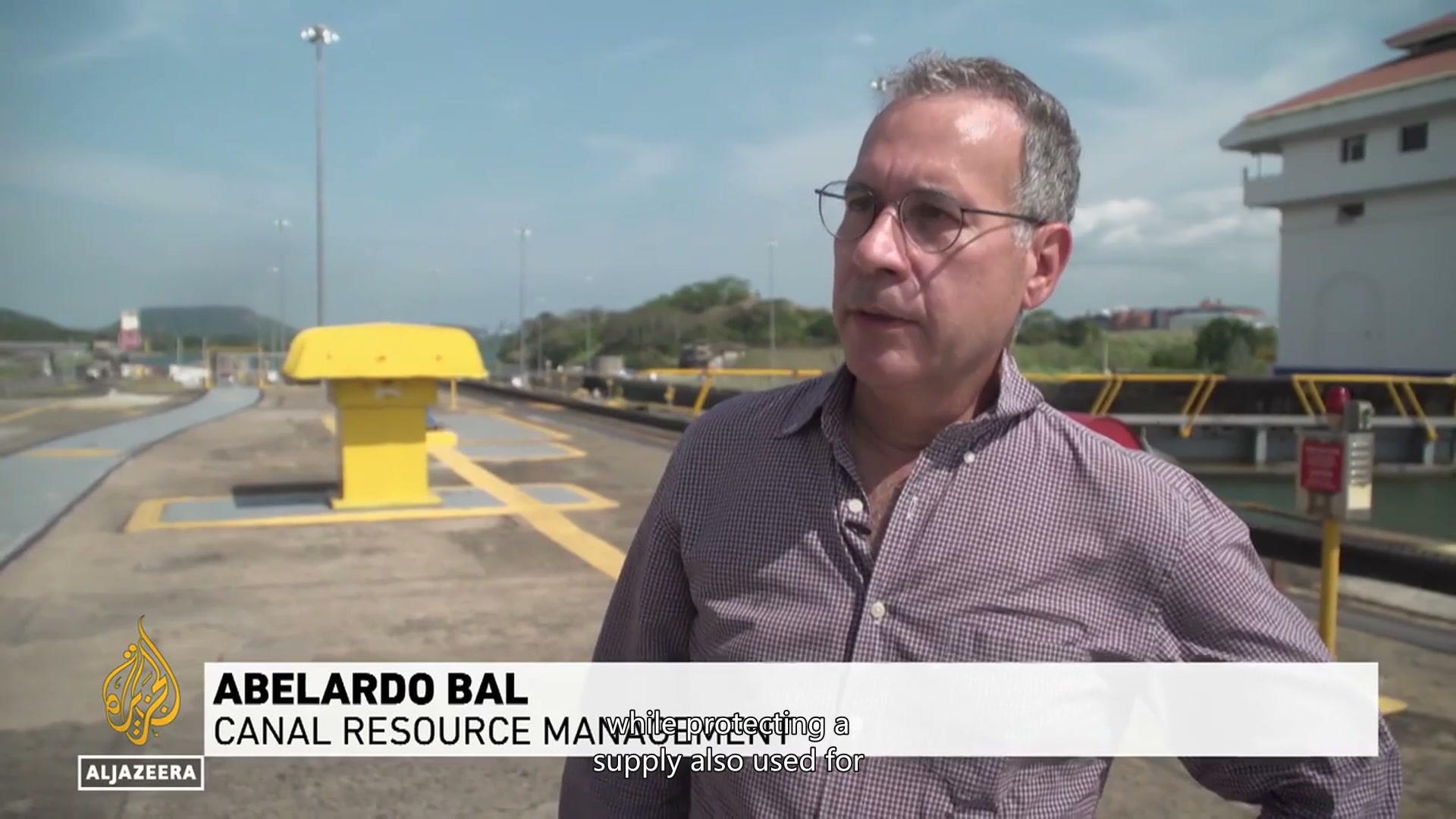}{01:14.2}\hfill%
  \GalleryFrame{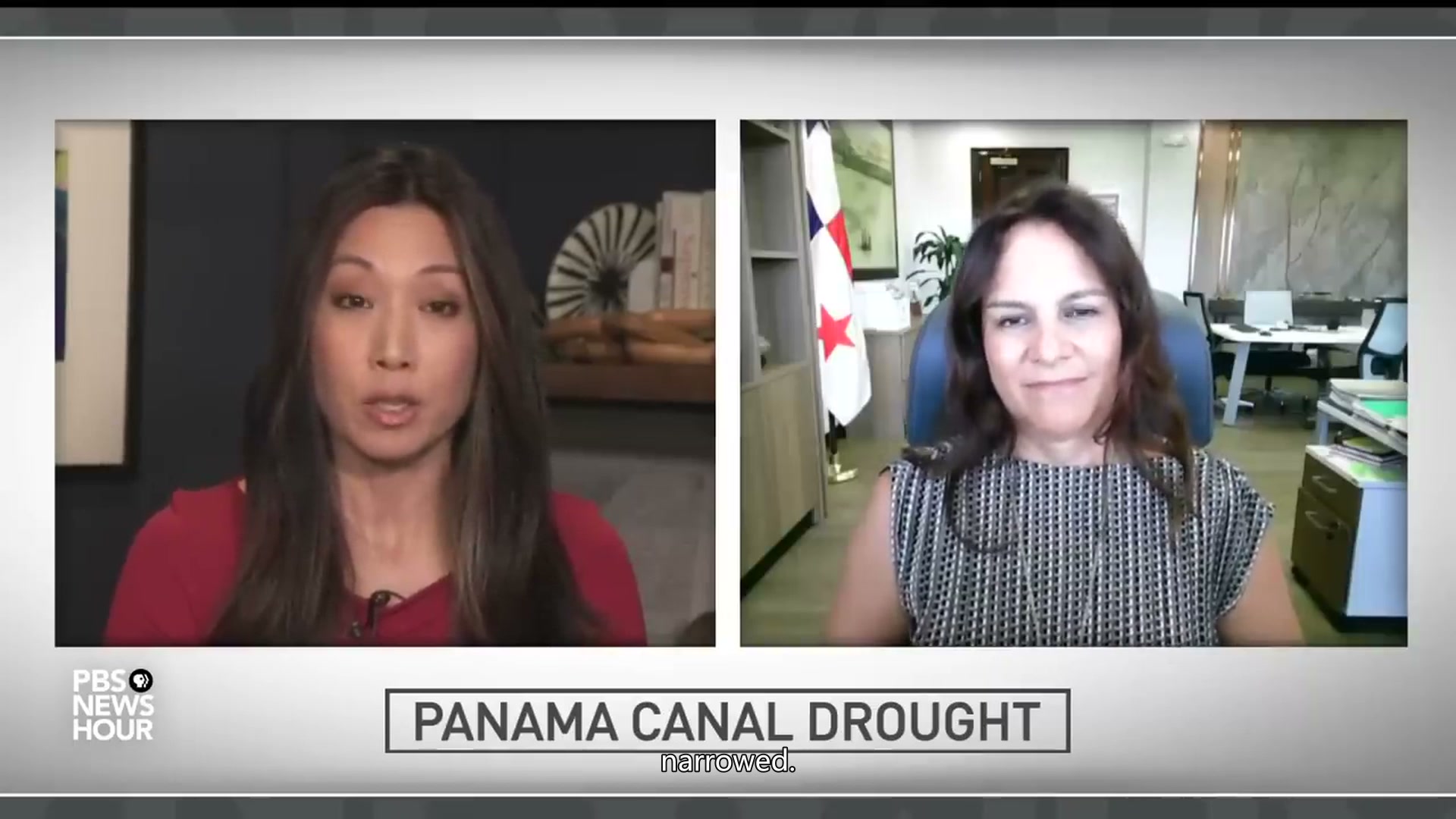}{02:06.5}\hfill%
  \GalleryFrame{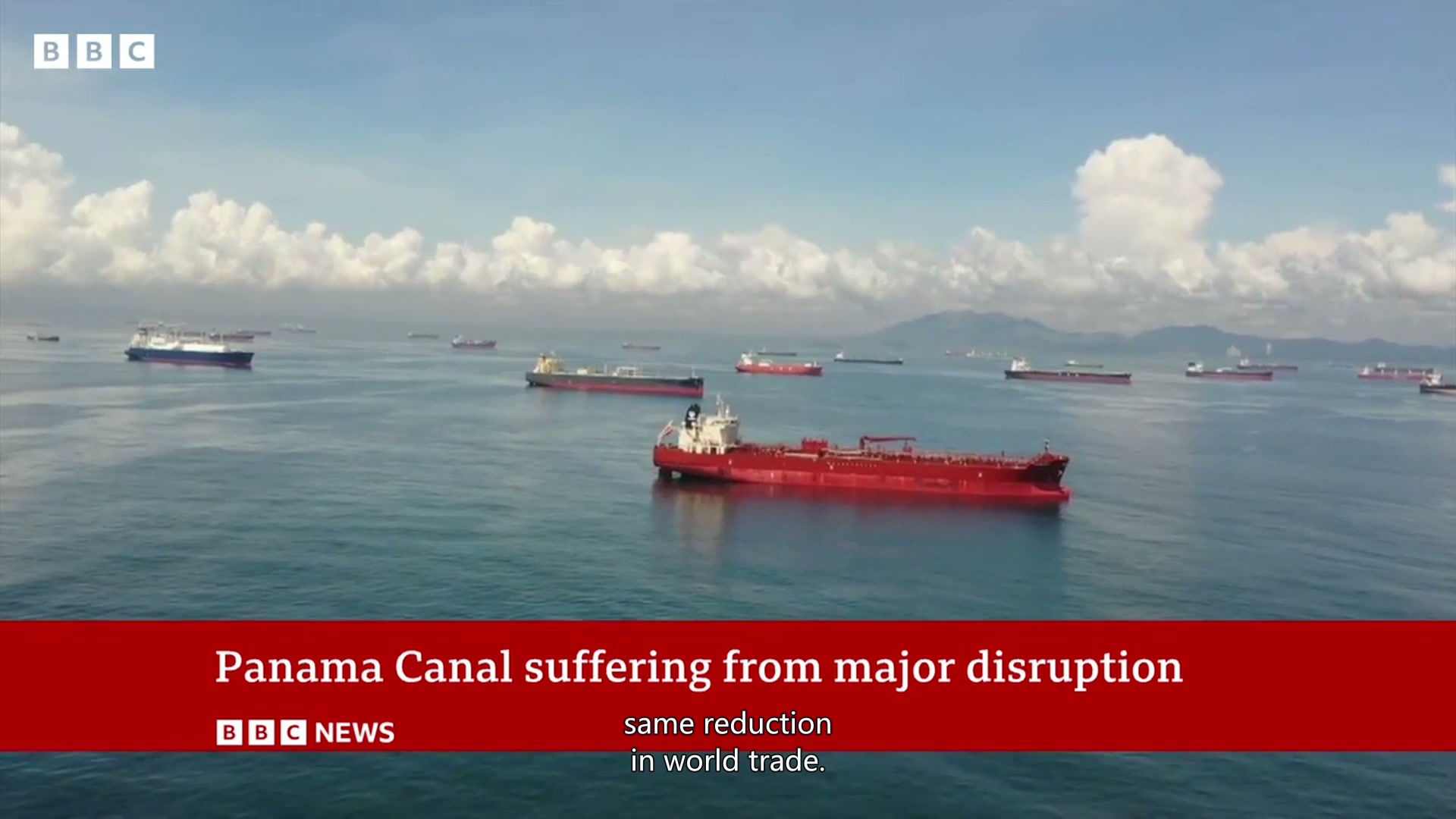}{02:58.9}%
}{fig:video-panama_canal_drought}
\GalleryExample{Badminton Tactics}{3:14}{%
  \GalleryFrame{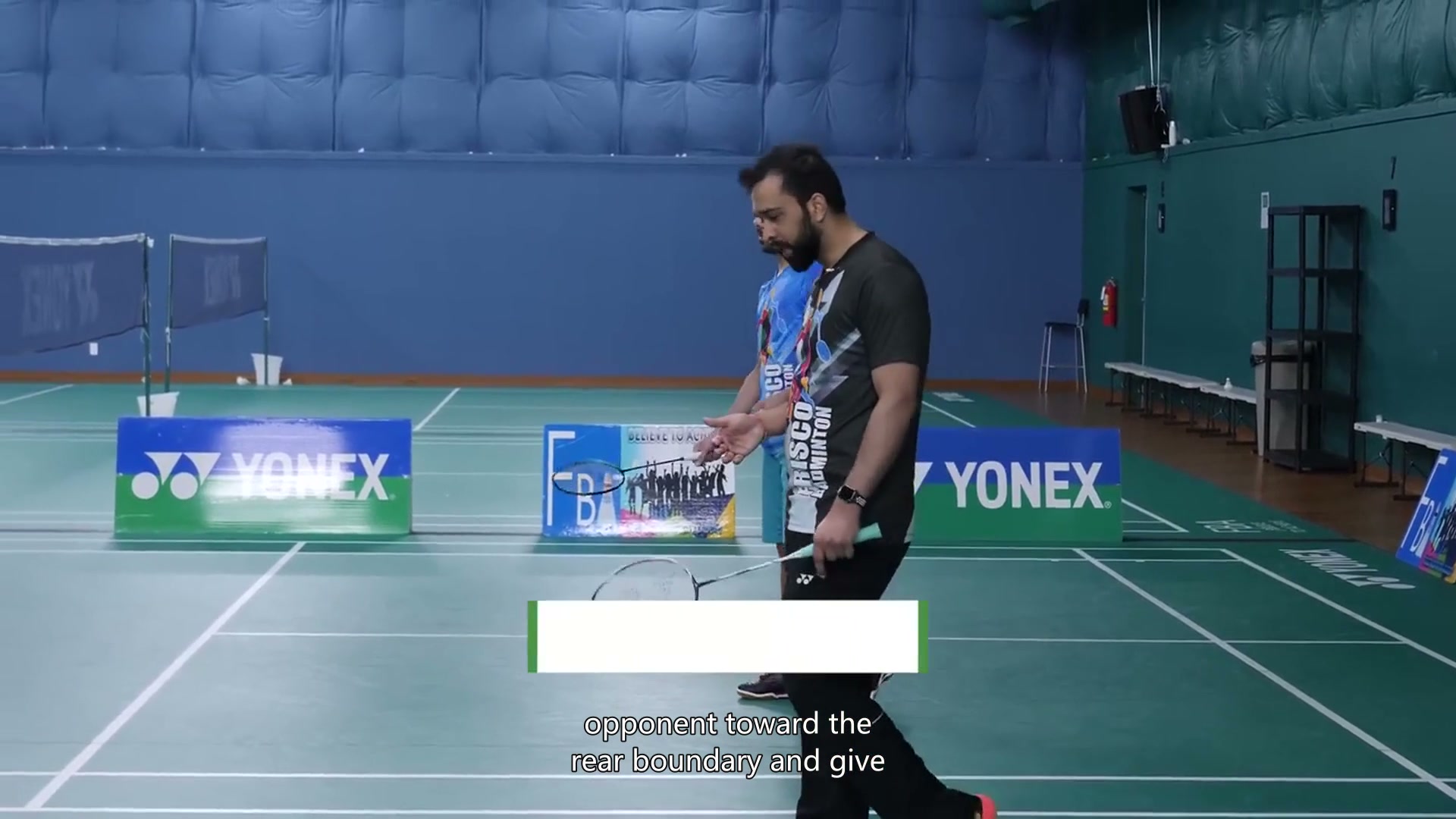}{00:20.2}\hfill%
  \GalleryFrame{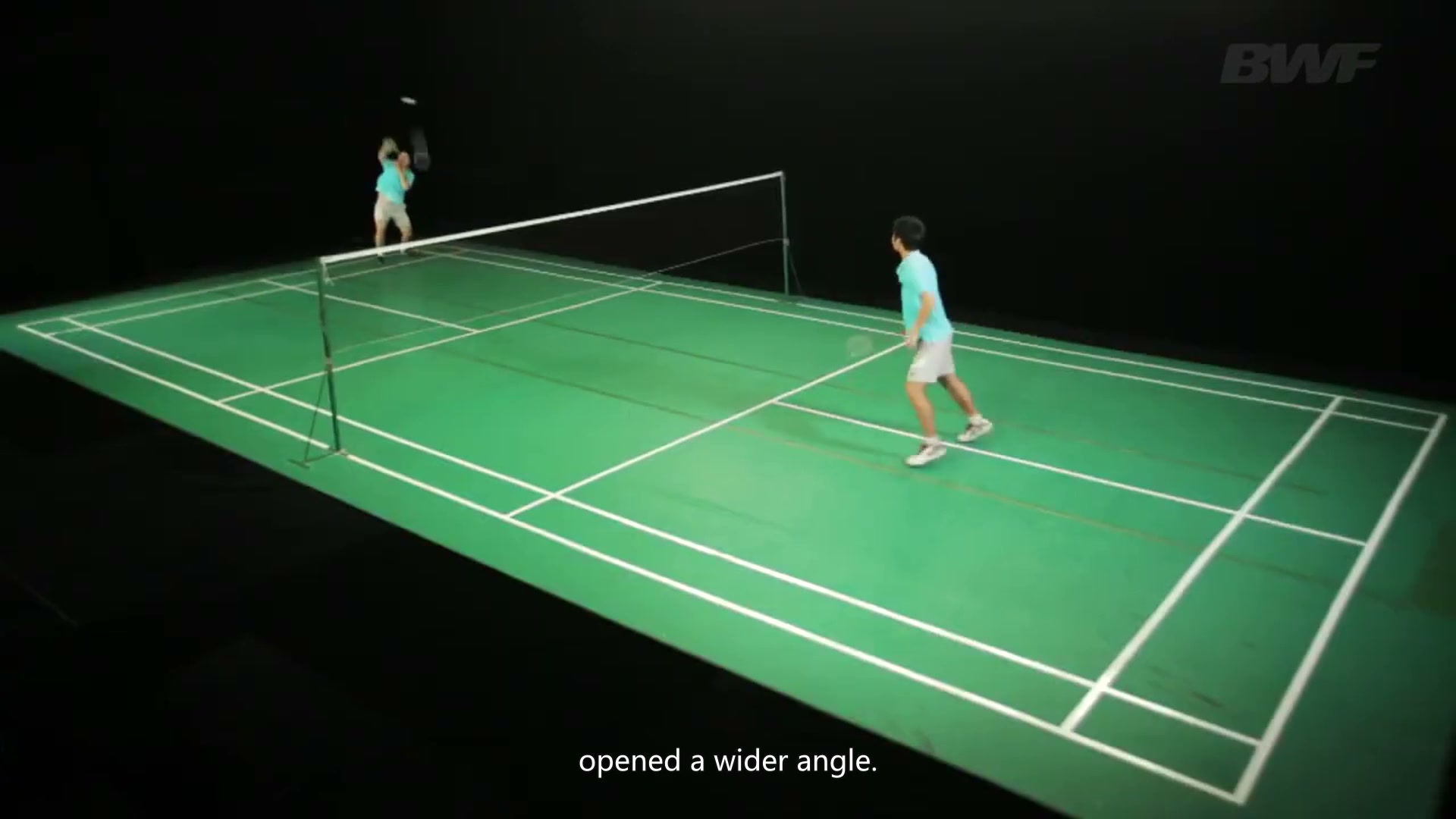}{01:08.7}\hfill%
  \GalleryFrame{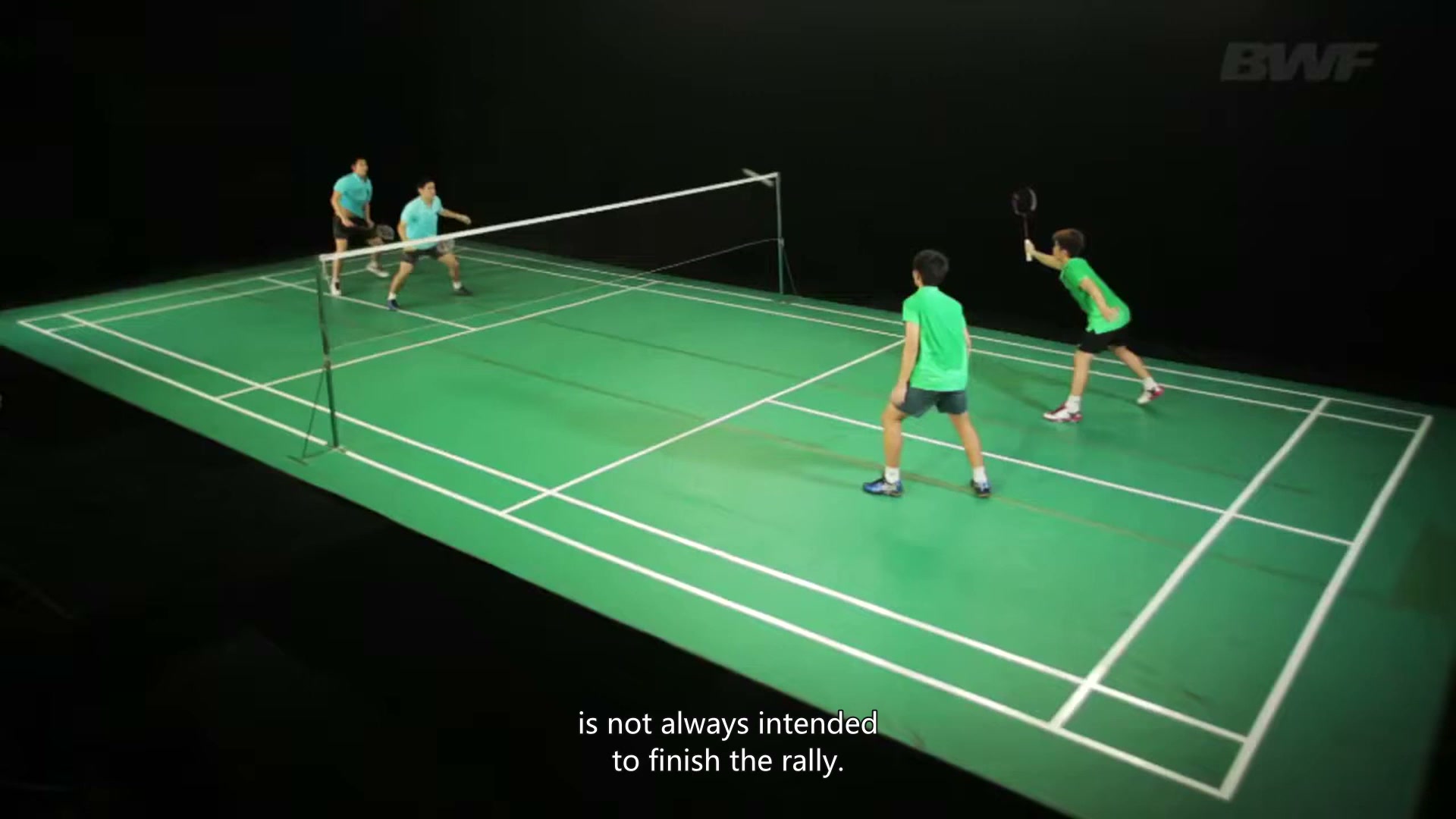}{01:57.3}\hfill%
  \GalleryFrame{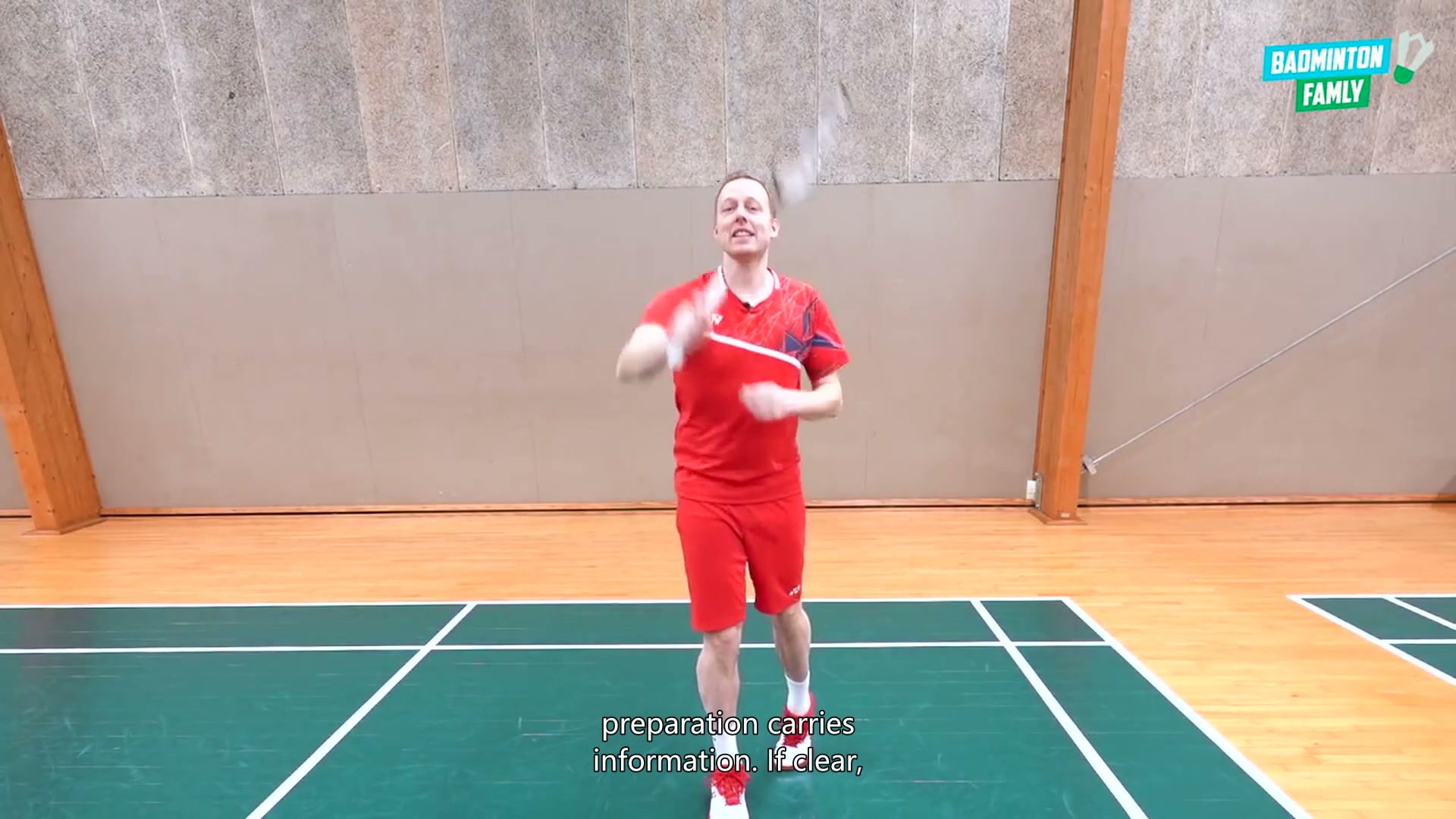}{02:45.8}%
}{fig:video-badminton_tactics}
\GalleryExample{How Rocket Launches Work}{3:17}{%
  \GalleryFrame{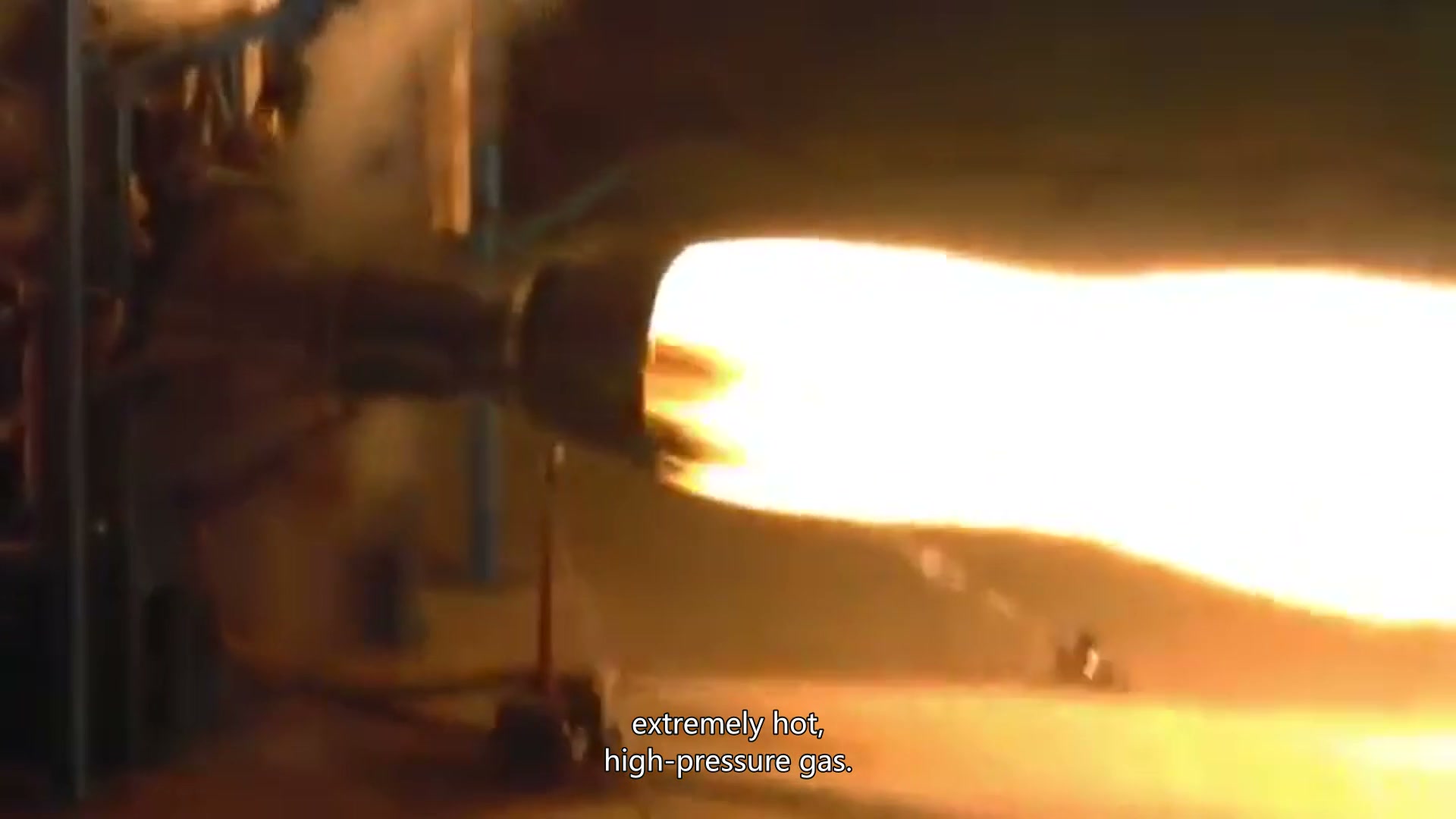}{00:28.8}\hfill%
  \GalleryFrame{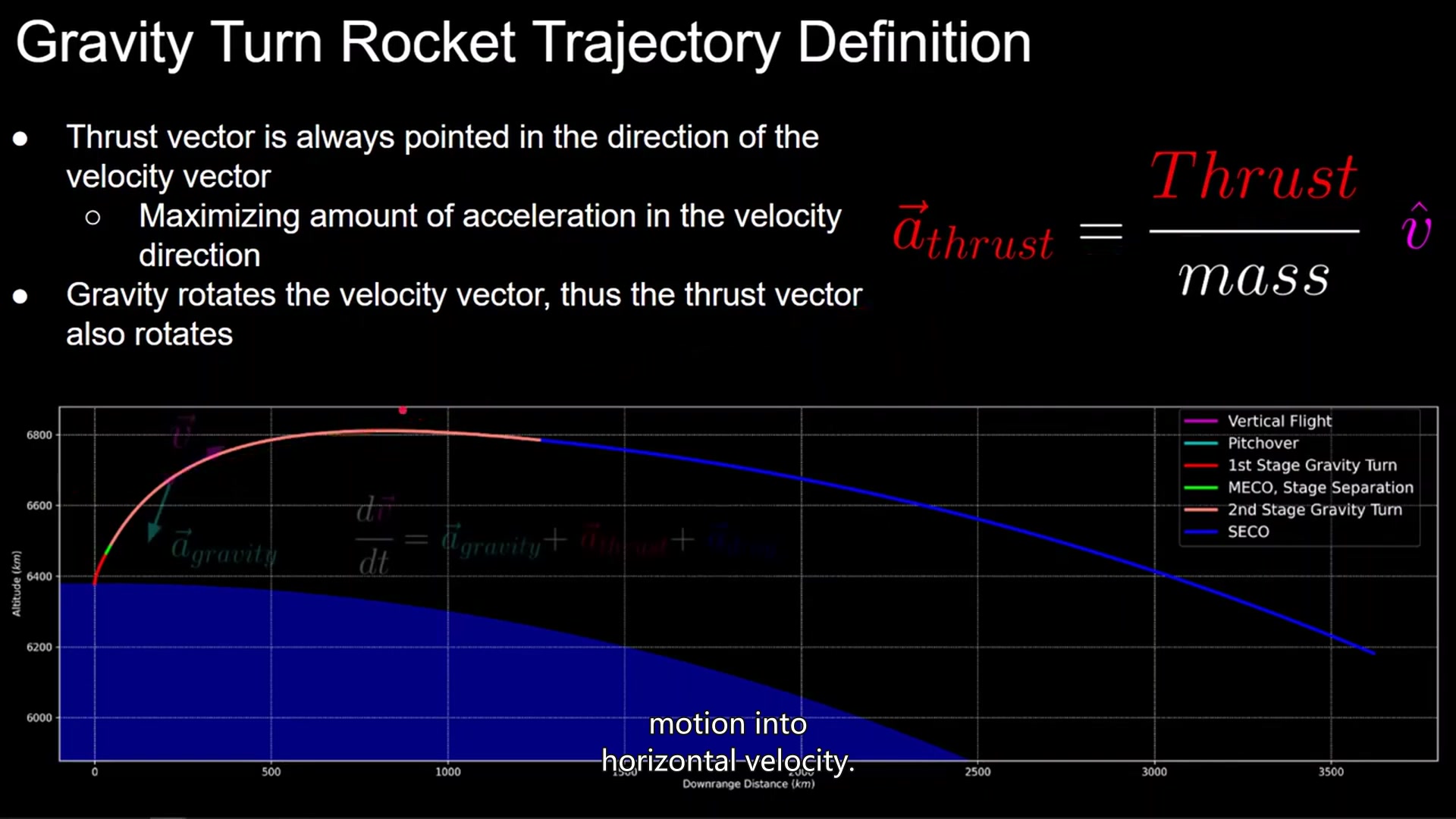}{01:09.9}\hfill%
  \GalleryFrame{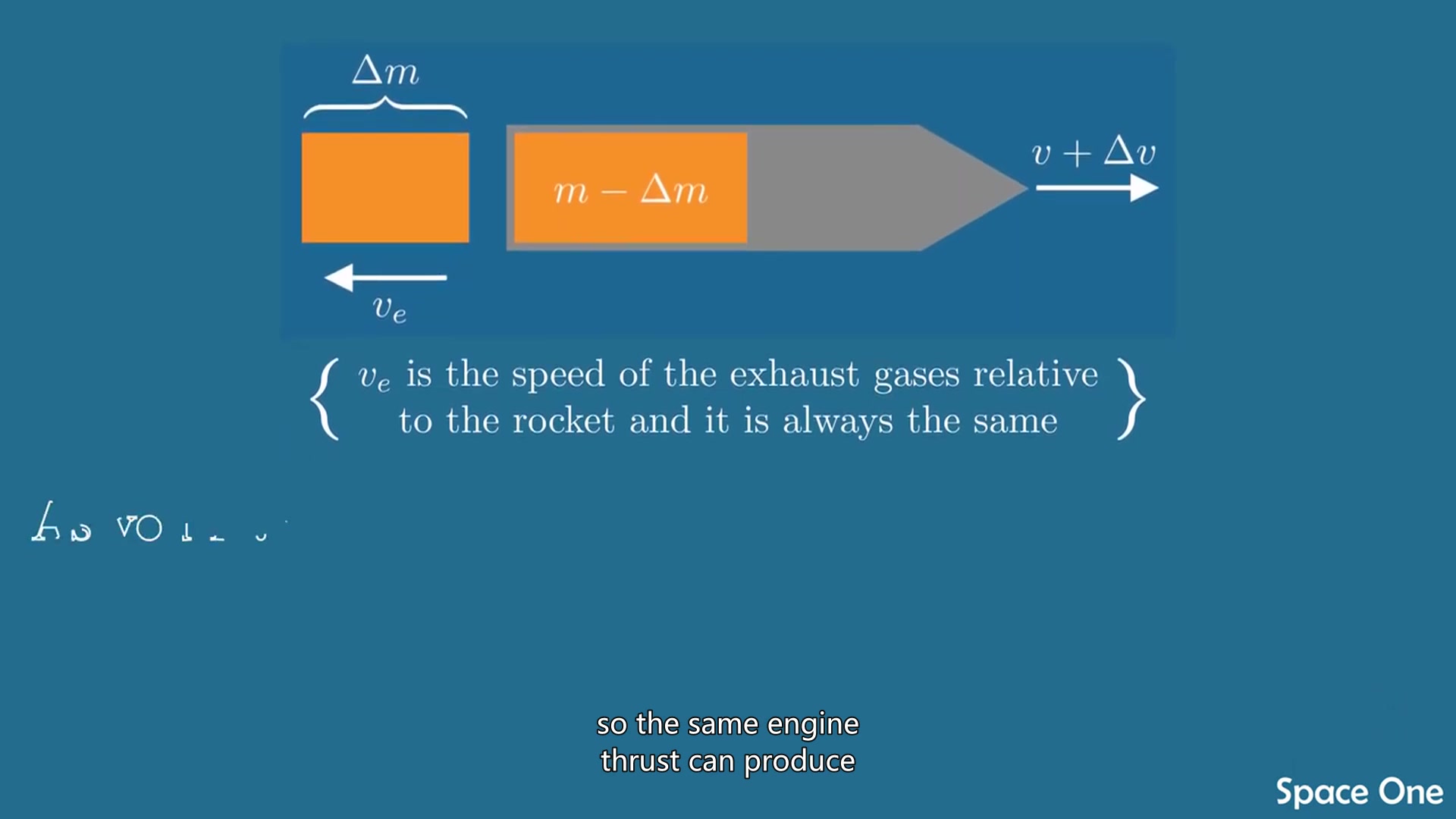}{01:26.4}\hfill%
  \GalleryFrame{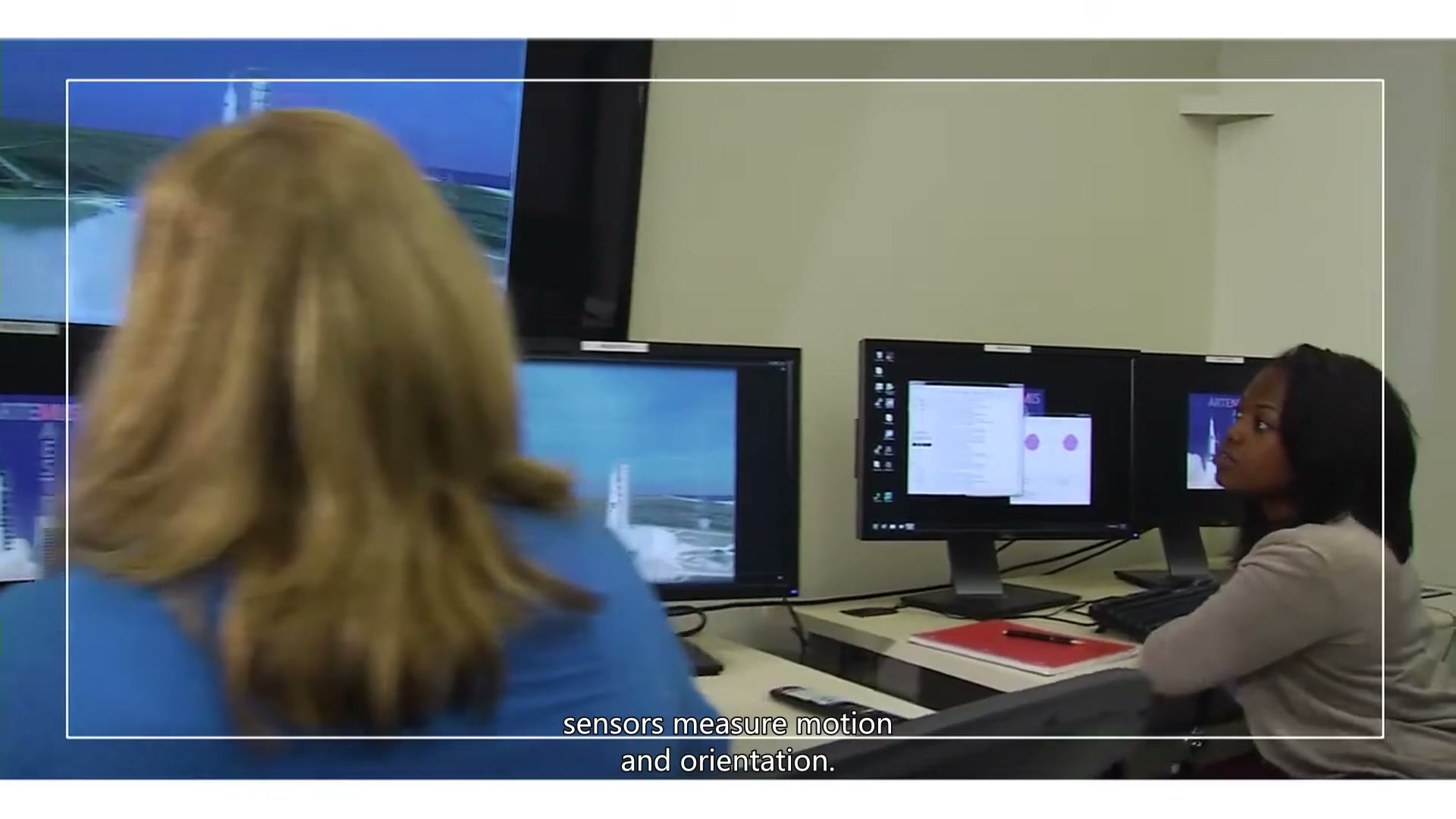}{02:32.2}%
}{fig:video-rocket_launch_principles}
\GalleryExample{Why Vaccines Eradicated Smallpox but Not Every Infectious Disease}{4:05}{%
  \GalleryFrame{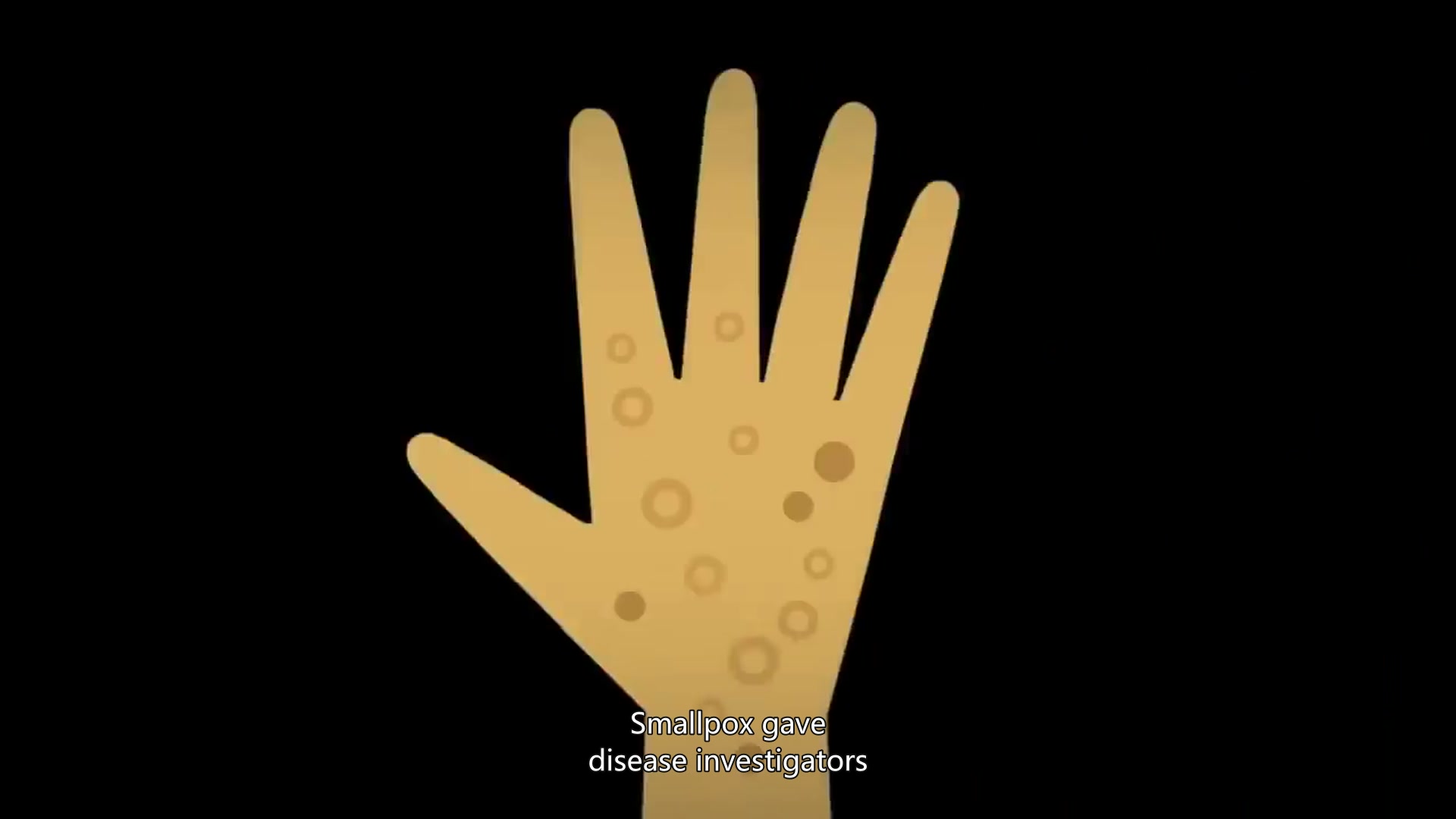}{00:46.1}\hfill%
  \GalleryFrame{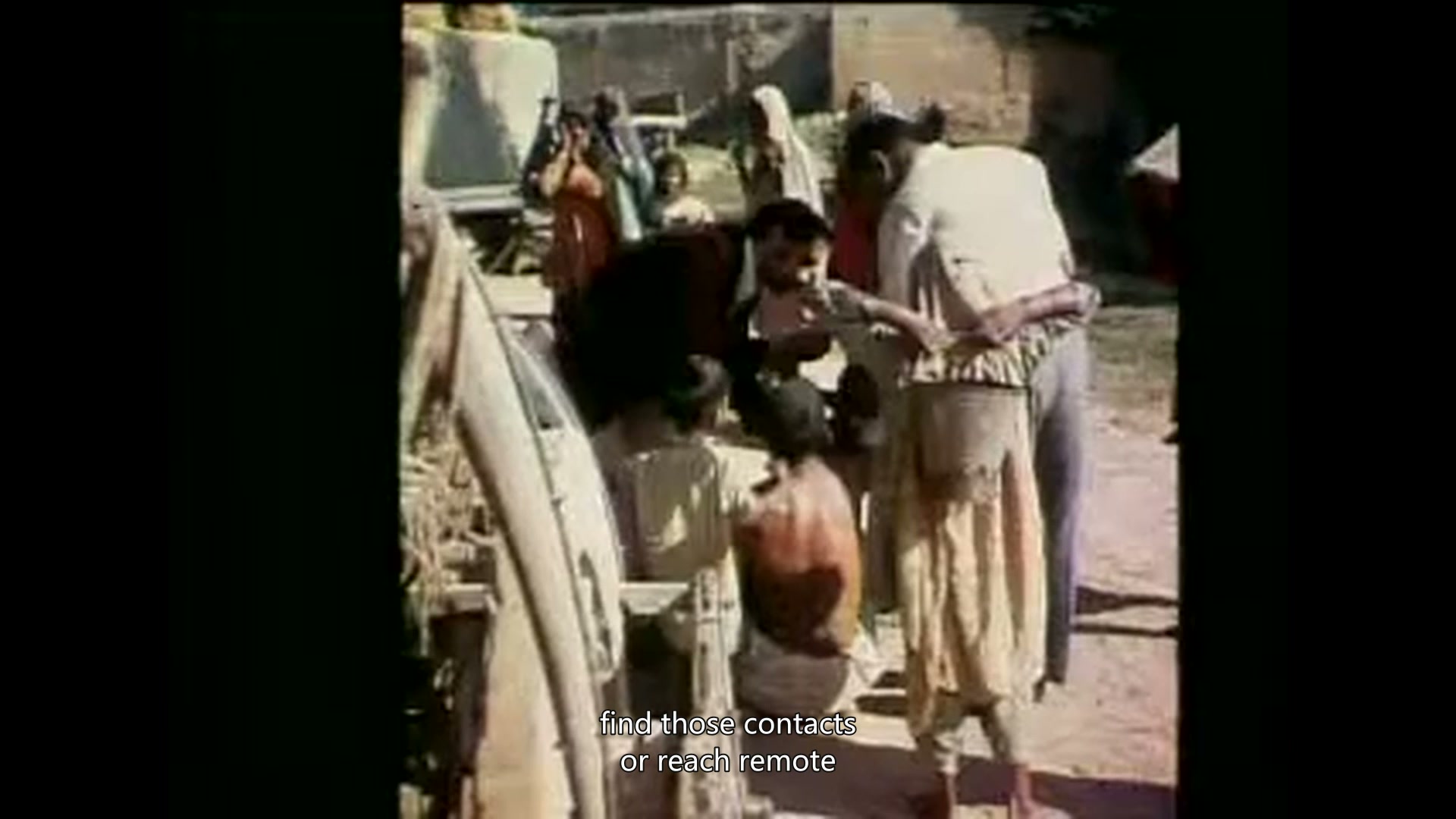}{01:27.1}\hfill%
  \GalleryFrame{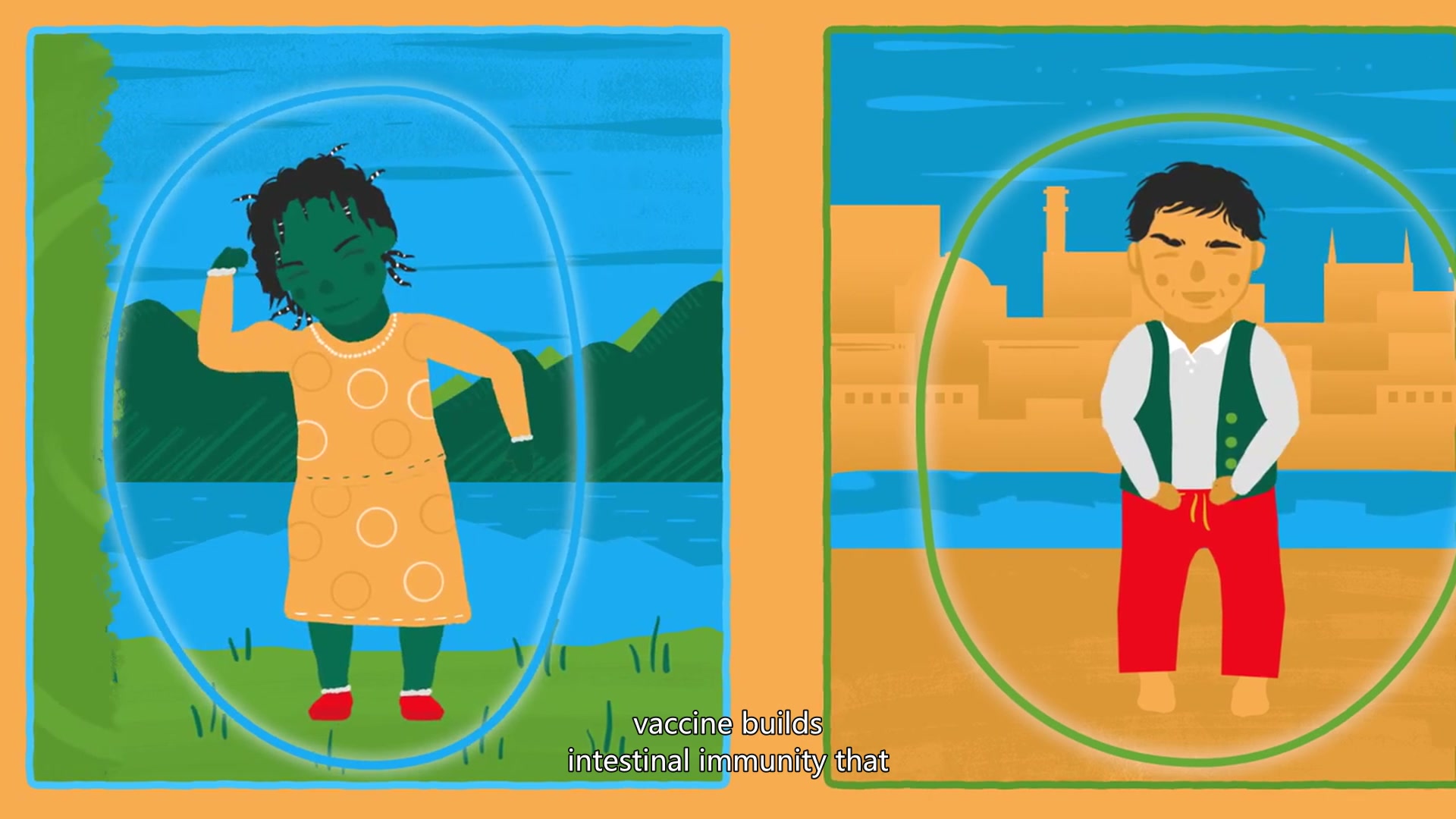}{02:28.6}\hfill%
  \GalleryFrame{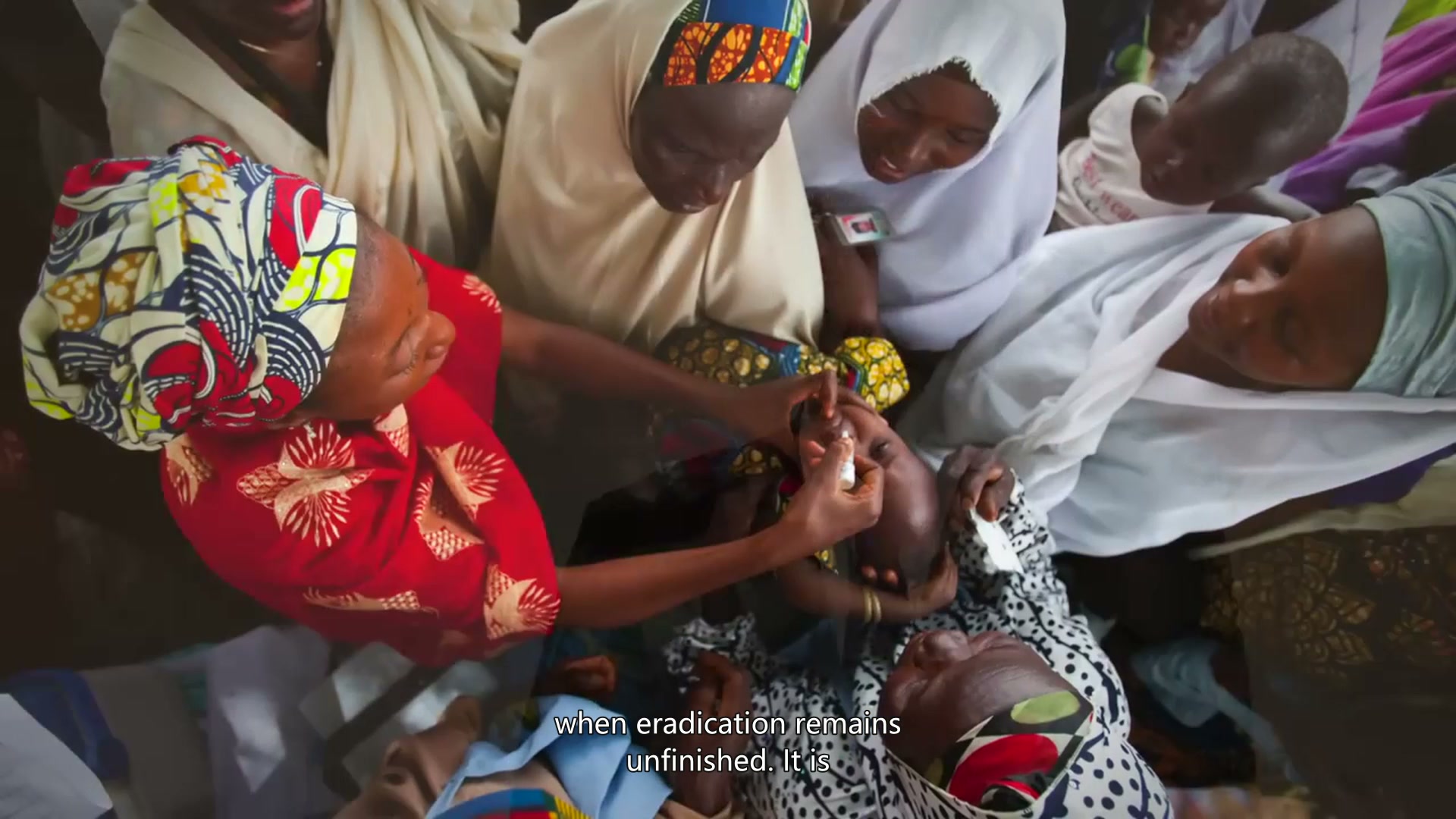}{04:00.8}%
}{fig:video-vaccines_and_eradication}
\clearpage